\documentclass[10pt,twocolumn,letterpaper]{article}

\usepackage[pagenumbers]{cvpr} %
\usepackage{minitoc}

\usepackage[dvipsnames]{xcolor}
\usepackage[export]{adjustbox}
\usepackage{wrapfig}
\usepackage[subrefformat=parens]{subcaption}

\usepackage[dvipsnames]{xcolor}

\newcommand{\ap}[1]{``#1''}
\def\Bezier{B\'{e}zier\xspace}

\definecolor{cvprblue}{rgb}{0.21,0.49,0.74}
\usepackage[pagebackref,breaklinks,colorlinks,citecolor=cvprblue]{hyperref}
\usepackage{tikz,lipsum}
\usepackage[most]{tcolorbox}
\usepackage[capitalize]{cleveref}
\usepackage{multirow}
\usepackage{listings}
\usepackage{xcolor}
\usepackage{tabularx}

\def\confName{CVPR}
\def\confYear{2025}

\title{\vspace{-0.6cm}SketchAgent: Language-Driven Sequential Sketch Generation\vspace{-0.19cm}}

\author{\hspace{-1cm} Yael Vinker$^{1}$ \hspace{-0.8cm} 
\and Tamar Rott Shaham$^{1}$ \hspace{-0.8cm}
\and Kristine Zheng$^{2}$ \hspace{-0.8cm} 
\and Alex Zhao$^{1}$ \hspace{-0.8cm}
\and Judith E Fan$^{2}$ \hspace{-0.8cm} 
\and Antonio Torralba$^{1}$ \hspace{-1cm}
\and \hspace{0.9\linewidth} \and
$^{1}$MIT \\ {\tt\small \{yaelvink,tamarott,alexzhao,torralba\}@mit.edu}\and $^{2}$Stanford University  \\ {\tt\small \{jefan,kxzheng\}@stanford.edu} \and
\small\url{https://sketch-agent.csail.mit.edu/}
}

\begin{document}
\doparttoc %
\faketableofcontents %

\twocolumn[{%
\vspace{-0.5cm}
\maketitle
\renewcommand\twocolumn[1][]{#1}%
\vspace{-0.9cm}
\begin{center}
    \centering
    \includegraphics[width=0.9\linewidth]{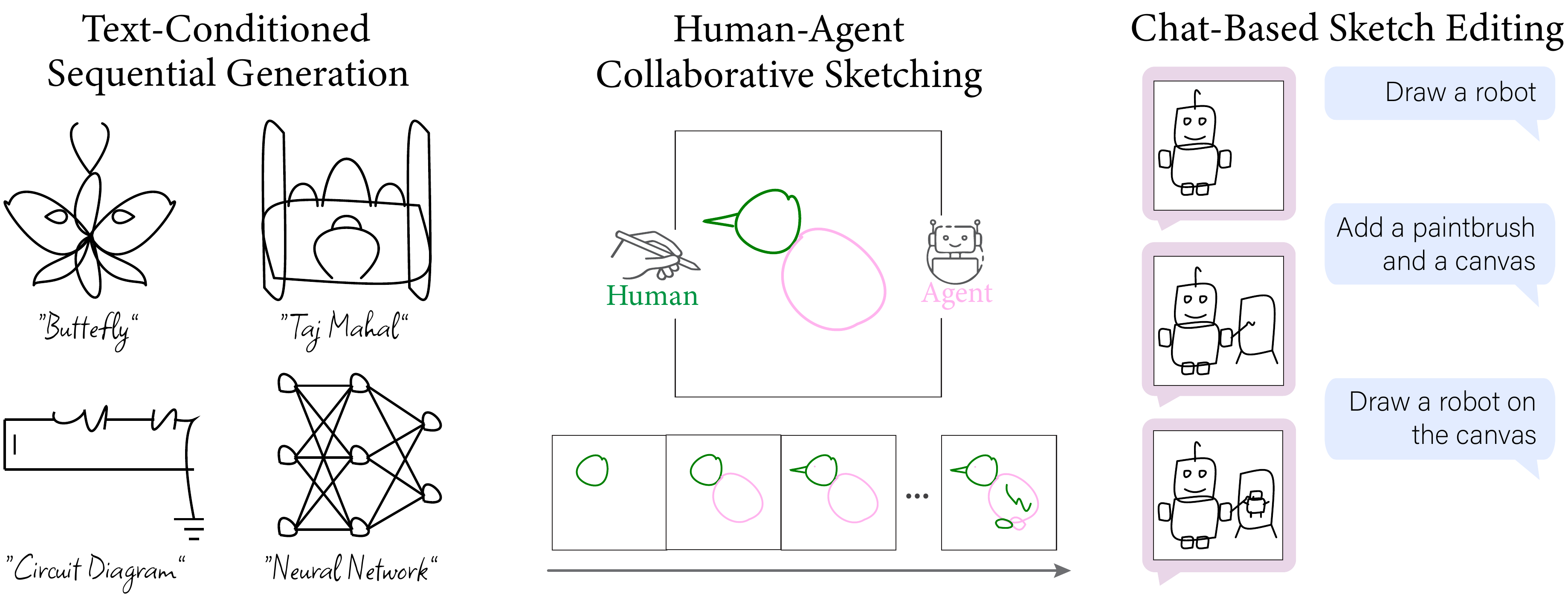}
    \captionsetup{type=figure}\caption{
    SketchAgent leverages an off-the-shelf multimodal LLM to facilitate language-driven, sequential sketch generation through an intuitive sketching language. It can sketch diverse concepts, engage in interactive sketching with humans, and edit content via chat.} 
    \label{fig:teaser}
\end{center}
}]

\begin{abstract}
\vspace{-0.2cm}
Sketching serves as a versatile tool for externalizing ideas, enabling rapid exploration and visual communication that spans various disciplines. While artificial systems have driven substantial advances in content creation and human-computer interaction, capturing the dynamic and abstract nature of human sketching remains challenging. 
In this work, we introduce SketchAgent, a language-driven, sequential sketch generation method that enables users to create, modify, and refine sketches through dynamic, conversational interactions.
Our approach requires no training or fine-tuning. Instead, we leverage the sequential nature and rich prior knowledge of off-the-shelf multimodal large language models (LLMs). We present an intuitive sketching language, introduced to the model through in-context examples, enabling it to \ap{draw} using string-based actions. These are processed into vector graphics and then rendered to create a sketch on a pixel canvas, which can be accessed again for further tasks.
By drawing stroke by stroke, our agent captures the evolving, dynamic qualities intrinsic to sketching. We demonstrate that SketchAgent can generate sketches from diverse prompts, engage in dialogue-driven drawing, and collaborate meaningfully with human users.
\end{abstract}
    
\section{Introduction}
\label{sec:intro}

\begin{figure}
    \centering
    \includegraphics[width=0.87\linewidth]{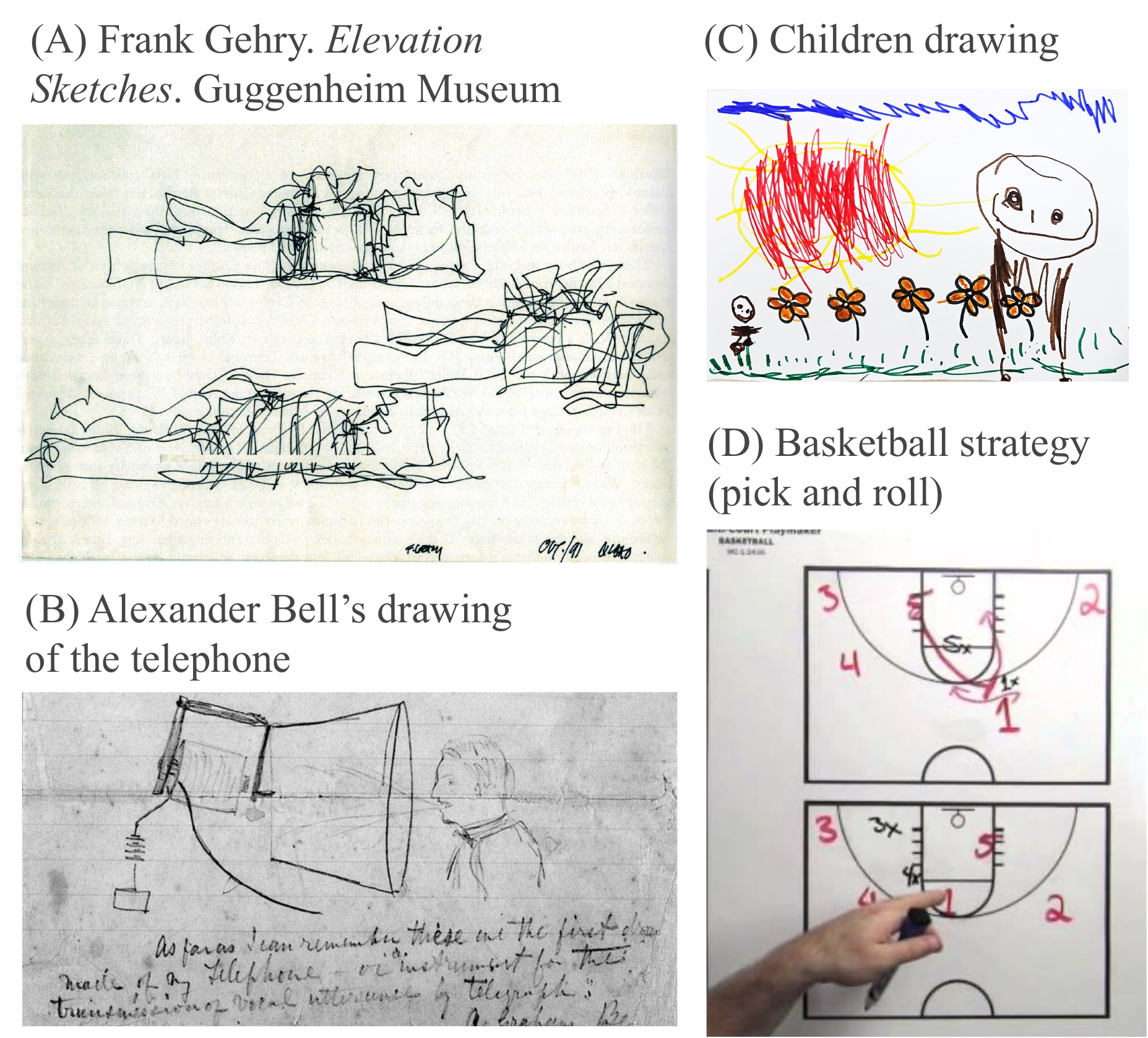}
    \vspace{-0.2cm}
    \caption{Examples of sketches used across disciplines and goals. (A) Ideation and design: \textit{Process Elevation Sketches} by the architect Frank Gehry, Guggenheim Museum. (B) Engineering: Alexander Bell’s telephone drawing. (C) Expressing emotions: Children’s sketches. (D) Visual communication: Planning and communicating game strategy in basketball.
    \vspace{-0.5cm}
}
    \label{fig:sketch-example}
\end{figure}

Sketching is a powerful tool for distilling ideas into their simplest form. Its fluid and spontaneous nature makes sketching a uniquely versatile tool for visualization, rapid ideation, and communication across cultures, generations, and disciplines \cite{fan2023drawing, tversky2013visualizing}.
For example, designers use sketches to explore new ideas \cite{goldschmidt1992serial,tversky2003sketches}, scientists employ them to formulate problems \cite{kaiser2019drawing, nasim2019observing}, and children engage in sketching to learn and express themselves~\cite{forbus2011cogsketch, fiorella2020creating} (see \cref{fig:sketch-example}). 
Artificial systems, in principle, have the potential to support and enhance human creativity, problem-solving, and visual expression through sketching, adapting flexibly to their exploratory nature~\cite{TholanderIdeation2023,Zhang2023GenerativeIA,Epstein2023ArtAT}.

Traditionally, sketch generation methods rely on human-drawn datasets to train generative models \cite{SketchRNN,Bhunia2020PixelorAC,das2020beziersketch,ge2020creative,bhunia2022doodleformer,Deformable_Stroke}. However, fully capturing the diversity of sketches within datasets remains challenging \cite{fan2023drawing}, limiting these methods in both scale and diversity.
Recent advancements in vision-language models, such as CLIP \cite{Radfordclip} and text-to-image diffusion \cite{rombach2022highresolution}, have enabled sketch generation methods that reduce reliance on human-drawn datasets \cite{CLIPdraw, vinker2022clipasso, jain2023vectorfusion}. These methods leverage pretrained model guidance and differentiable rendering~\cite{diffvg} to optimize parametric curves, creating sketches that go beyond predefined styles and categories. 

While representing a significant step toward a general-purpose sketching system, these methods lack a crucial aspect of human drawing: the \textit{process} itself. Current methods, though versatile, optimize all strokes simultaneously, making the intermediate sketching steps meaningless. As a result, the sketch cannot be decomposed into a coherent sequence of strokes that reflects the drawing process.
In contrast, humans draw iteratively, stroke by stroke, incorporating visual feedback and continuously adapting—a dynamic, evolving process that fosters creativity, ideation, and communication \cite{laseau2000graphic, schon1986reflective,Tversky2002WhatDS}.

In this work, we introduce SketchAgent, a sketch generation agent that leverages the prior knowledge and sequential nature of multimodal large language models (LLMs) to enable versatile, progressive, language-driven sketching.
Our agent can generate sketches across a wide range of textual concepts—from animals to engineering principles (\cref{fig:teaser}, left). Its sequential nature facilitates interactive human-agent sketching and supports iterative refinement and editing through a chat-based dialogue (\cref{fig:teaser}, right).

Unlike vision-language models that directly generate images from text \cite{ramesh2022hierarchical, rombach2022highresolution, sdxl2023}, multimodal LLMs \cite{openai2024gpt4technicalreport, geminiteam2024geminifamilyhighlycapable, chu2024visionllamaunifiedllamabackbone, claude,NEURIPS2023_LLAVA,BLIP2022,Flamingo2024} accept text and images as input but only output text. To produce visuals, they either utilize external \ap{tools} (such as calling a text-to-image model) or are prompted to generate executable code (e.g., Python~\cite{Han2023ChartLlamaAM}, SVG~\cite{cai2024delving}) to create charts, diagrams, or graphics.
However, prompting for such representations to directly produce sketches often results in a mechanical appearance with uniform, precise shapes that lack the subtle irregularities and spontaneous qualities characteristic of human sketches (see \cref{fig:related}B).
Additionally, despite their robustness in textual tasks, these models often struggle with fine-grained spatial reasoning \cite{HallusionBenchGuan_2024_CVPR, Yang2023TheDO} as they are primarily optimized for text, making sketch editing more challenging.

To address these limitations, we introduce an intuitive sketching language that enables an off-the-shelf multimodal LLM agent to \ap{draw} sketches on a canvas by providing string-based actions, without additional training or fine-tuning.
We define the canvas as a numbered grid, allowing the agent to reference specific coordinates (e.g., \texttt{x2y8}) to enhance its spatial reasoning capabilities. We represent a sketch as a sequence of semantically meaningful strokes, each defined by a series of such coordinates.
We leverage In-Context Learning (ICL) \cite{NEURIPS2020_1457c0d6} to introduce the agent to the new representation, and Chain of Thought (CoT) \cite{Chain-of-thought24} to enhance its planning capabilities.
Given a sketching task, the agent produces a textual response following our representation, which we process by fitting a smooth \Bezier curve to each coordinate sequence. The curves are then rendered onto the canvas to form the final sketch.
We find this approach useful in emulating a more natural sketch appearance.
For collaborative sketching, the canvas remains accessible to both the user and the agent throughout the session. The agent generates strokes sequentially and pauses according to an adjustable stopping token, allowing the user to add their own strokes directly to the canvas. These strokes are then integrated into the agent's sequence, enabling it to continue drawing, with real-time canvas updates.

We demonstrate SketchAgent's capabil to generate sketches of diverse concepts while capturing the inherently sequential and dynamic nature of sketching. 
We showcase our agent's ability to collaborate effectively with humans in real time to create novel and meaningful sketches.
Our method is the first to leverage pretrained multimodal LLMs for sequential sketching without additional training, paving the way for a general-purpose artificial sketching system that supports iterative, evolving interactivity.

\section{Related Work}
\label{sec:prev_work}
\begin{figure}[t]
\small
    \centering
    \setlength{\tabcolsep}{3pt}
    {\small
    \begin{tabular}{c c c c}
    \small
        \includegraphics[width=0.23\linewidth]{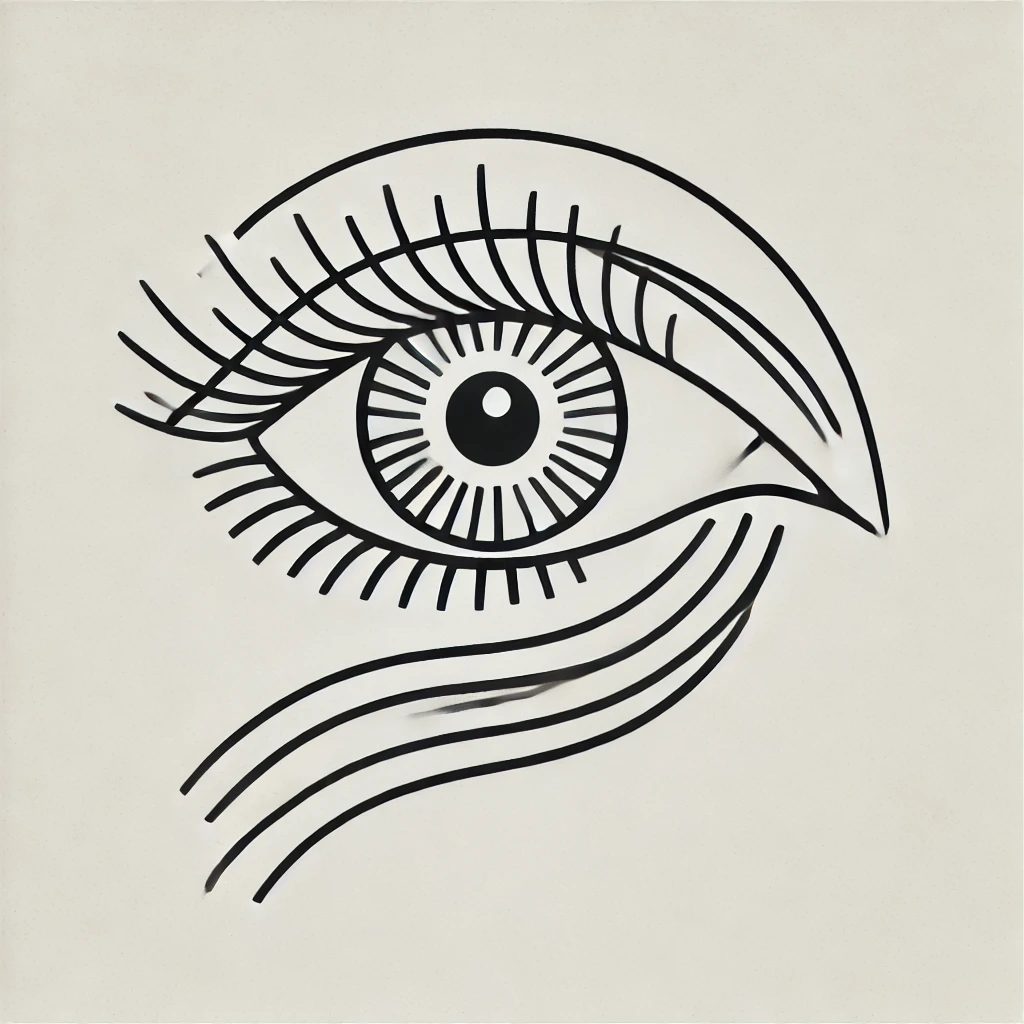} &
        \includegraphics[width=0.23\linewidth]{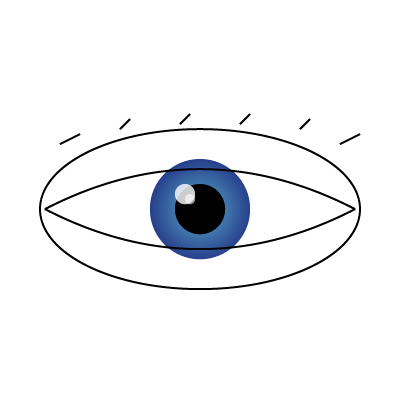} &
        \includegraphics[width=0.23\linewidth]{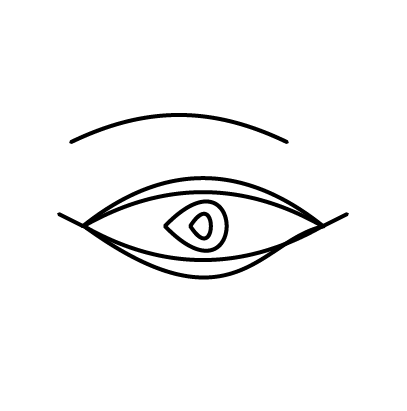} &
        \includegraphics[width=0.23\linewidth]{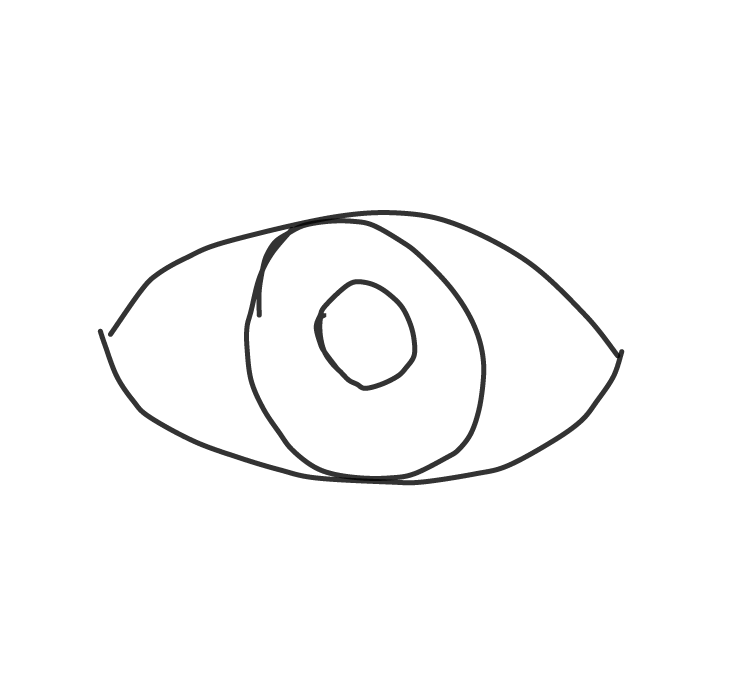} \\
        \footnotesize{(A) DALLE3 \cite{BetkerImprovingIG}} &
        \begin{tabular}[c]{@{}c@{}}\footnotesize{(B) LLMs \cite{claude}} \\ \footnotesize{(SVG)}\end{tabular} &
        \footnotesize{(C) SketchAgent}& 
        \footnotesize{(D) Human \cite{quickDrawData}} \\
    \end{tabular}
    }
    \vspace{-0.3cm}
    \caption{Sketch appearance. (A) Text-to-image diffusion models operate in pixel space, lacking thesequential nature of sketches. (B) Prompting LLMs to produce visuals with SVG results in a uniform, mechanical appearance. (C) Sketches produced by our agent appear less mechanical, more closely resembling the nature of (D) Human sketches, which are often spontaneous and irregular.}
    \vspace{-0.2cm}
    \label{fig:related}
\end{figure}

\paragraph{Sketch Generation}
Early methods approached sketch generation by designing image filters to simulate sketch-like effects \cite{canny1986computational, Winnemller2012XDoGAE}. With the advent of deep learning, data-driven approaches emerged to address a range of sketch-related tasks \cite{xu2020deep}, including category-conditioned sketching \cite{song2018learning,SketchRNN,qi2021sketchlattice}, object sketching \cite{Deformable_Stroke,liu2021neural}, scene-sketching \cite{chan2022learning,li2019photo,xie2015holistically,li2019im2pencil}, sketch completion \cite{bhunia2021doodleformer,su2020sketchhealer,Liu_2019_CVPR,liu2019sketchgan}, portrait drawing \cite{yi2020unpaired,Berger2013,yi2019apdrawinggan}, part-based generation \cite{bhunia2021doodleformer,ge2021creative,SketchRNN,Zhou2018LearningTS}, and more. 
While sketch data collection has been broadly explored \cite{xiao2022differsketching,OpenSketchGryaditskaya2019,EitzTUBerlin2012,Sketchy-Database,gao2020sketchycoco,mukherjee2023SEVA}, the wide variation in sketch styles and their adaptation to specific tasks \cite{Fan2023DrawingAA} makes collecting datasets that encompass this diversity challenging. 
For example, QuickDraw~\cite{quickDrawData}, the largest available sketch dataset with 50 million sketches, covers only 345 object categories and primarily focuses on simple, iconic representations. This limits data-driven methods to the style, abstraction level, and concepts seen during training.
Recently, large pretrained vision-language models \cite{Radfordclip,ramesh2022hierarchical,sdxl2023,rombach2022highresolution,saharia2022photorealistic} have shown remarkable text-to-image generation capabilities by leveraging extensive visual knowledge from billions of training images \cite{schuhmann2022laion5b}. 
While these models can be prompted to generate sketch-like images (see \cref{fig:related}A), they do so in a single step and in pixel space, lacking the sequential, stroke-based process of human sketching.
Subsequent approaches \cite{jain2023vectorfusion, NEURIPS2023_DiffSketcher, svgdreamer_xing_2023, CLIPdraw, Text-to-Vector2024Zhang,vinker2022clipasso, Vinker_2023_ICCV, 3Doodle2024Choi,gal2023breathing} leverage the priors of these models to guide an iterative optimization of parametric curves, with a differentiable rasterizer \cite{diffvg} linking pixel and vector representations. 
This approach reduces reliance on human-drawn datasets, enabling robust sketch generation beyond pre-defined categories. However, optimizing all strokes simultaneously results in sketches that lack temporal and semantic structure, and the process is time-consuming, taking 5 minutes to over an hour per sketch, making it suboptimal for collaborative sketching.

\vspace{-0.3cm}
\paragraph{Sequential and Collaborative Sketching}
Collaborative human-machine sketching holds promise in enhancing creativity,  ideation, communication, and learning, as explored in various fields, including human-computer interaction (HCI) \cite{Davis2015DrawingApprentice, KarimiHCI2020, Ibarrola2022ACI, LawtonHCI2023, Reframer2023,Karimi2020}, computer graphics \cite{ShadowDraw2011,Sketchpad1998}, robotics \cite{schaldenbrand2024cofrida, SchaldenbrandFRIDA2023}, cognitive science \cite{McCarthy2024, collabdraw2019,fan_pragmatic_2020,Garrod2007}, learning sciences \cite{gijlers_collaborative_2013, educsci12010045, Tytler2020}, and more. 
Central to collaborative sketching is its sequential, adaptive, and dynamic process, with each action carrying intent.
Existing methods employ diverse training strategies to account for the discrete nature of sequential sketches, including reinforcement and adversarial learning \cite{mellor2019unsupervised,Ganin2018SynthesizingPF,Zhou2018LearningTS}, multi-agent referential games \cite{qiu2021emergent,mihai2021learning}, transformers \cite{Bhunia2020PixelorAC,Ribeiro2020SketchformerTR,Lin2020SketchBERTLS,bhunia2021doodleformer,carlier2020deepsvg,Wu2023IconShopTV,Ganin2021ComputerAidedDA}, and more. SketchRNN \cite{SketchRNN} is a pioneering work in this area, introducing the QuickDraw dataset \cite{quickDrawData}, a crowd-sourced collection of real-time sketch sequences made by users. They utilize this dataset to train a recurrent neural network for sequential sketch generation, which was later shown \cite{collabdraw2019,Changhoon2018} to have potential for human-machine collaboration.
However, this approach remains constrained by the predefined categories encountered during training.

\vspace{-0.3cm}
\paragraph{Multimodel LLMs for Content Creation}
LLMs~\cite{Touvron2023LLaMAOA,devlin-etal-2019-bert,NEURIPS2020_gpt3,T5Colin2020} and multimodal LLMs~\cite{openai2024gpt4technicalreport, geminiteam2024geminifamilyhighlycapable, chu2024visionllamaunifiedllamabackbone, claude,NEURIPS2023_LLAVA,BLIP2022,Flamingo2024} receive text as input (or text and images for multimodal) and output text. 
To enable visual content generation, these models are often paired with external \ap{tools} that extend their functionality~\cite {wu2023visual,yang2023idea2img, hu2024visual,shaham2024multimodal}. For example, ChatGPT~\cite{openai2024gpt4technicalreport} generates images by internally calling a separate model, DALLE-3 \cite{BetkerImprovingIG}. 
Another approach involves prompting models to produce code in languages like Python~\cite{Han2023ChartLlamaAM}, Processing~\cite{sharma2024vision}, SVG~\cite{cai2024delving}, or TikZ~\cite{bubeck2023sparks} that can be rendered into visuals such as graphs, charts, and vector graphics.
However, such code-generated content often looks rigid, with uniform and overly precise shapes that lack the subtle irregularities and spontaneous qualities characteristic of freehand sketches (see Fig.~\ref{fig:related}B). 
In contrast, we propose a sketching language grounded in spatial information that encourages the model to produce a more natural sketch appearance, which we then process into vector graphics.
Common strategies for enhancing LLMs capabilities include Chain-of-Thought prompting~\cite{rose2023visual, shao2024visual, chen2024visual, mitra2024compositional, zhang2023multimodal}, which breaks down tasks into smaller, logical steps to mimic human reasoning, and In-Context Learning (ICL)~\cite{zhang2023makes, sun2023exploring, zhang2024instruct, doveh2024towards}, where examples of input-output pairs are provided to help the model infer task patterns.

\begin{wrapfigure}[7]{R}{0.3\linewidth}
  \begin{center}
  \vspace{-1.7cm}
    \includegraphics[width=1\linewidth]{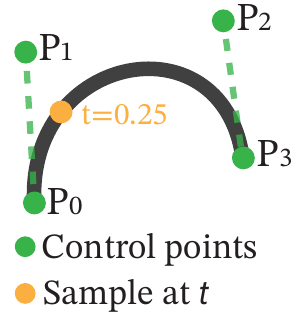}
  \end{center}
  \vspace{-0.6cm}
  \caption{Cubic \Bezier curve.}
  \label{fig:bezier}
\end{wrapfigure}

\begin{figure*}[h]
    \centering
    \includegraphics[width=1\linewidth]{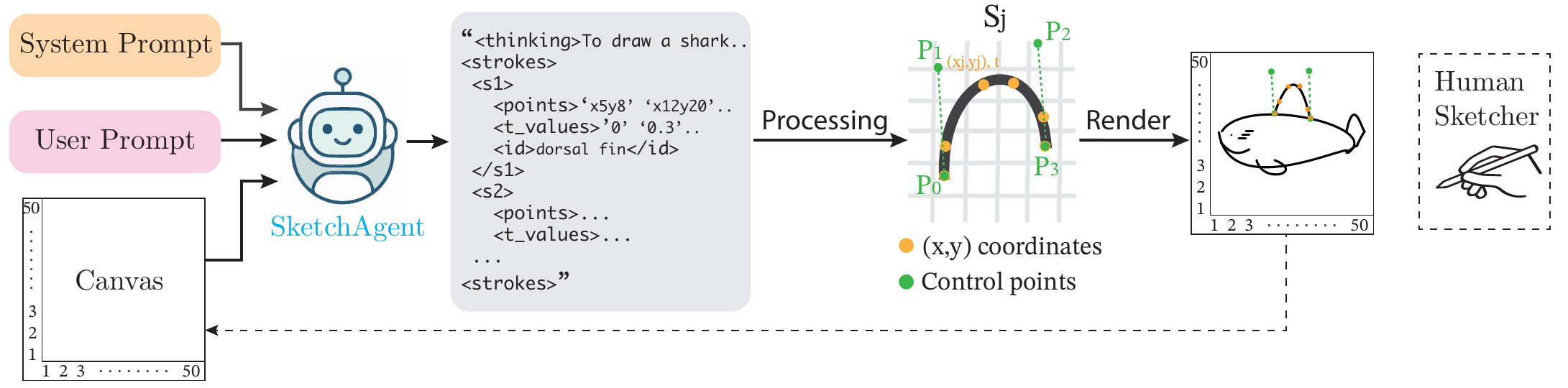}
    \vspace{-0.7cm}
    \caption{Method Overview. SketchAgent (blue) receives drawing instructions and generates a string representing the intended sketch. Inputs include: (1) a system prompt (orange) introducing the sketching language and canvas, (2) a user prompt (pink) specifying the task (e.g., \ap{draw a shark}), and (3) a numbered canvas. The agent's response outlines a sketching strategy (in thinking tags) and a sequence of strokes defined by coordinates, which are processed into Bézier curves and rendered onto the canvas.}
    \vspace{-0.2cm}
    \label{fig:high-level-pipe}
\end{figure*}

\section{Preliminaries}
\label{sec:preliminaries}
\paragraph{Vector Graphics and \Bezier Curves}
Vector graphics allow us to create visual images directly from geometric shapes such as points, lines, curves, and polygons. Unlike raster images (represented with pixels), vector graphics are resolution-free, more compact, and editable.
SVG \cite{svg_spec} is an XML-based format for storing vector graphics, popular for its scalability and compatibility with modern web browsers.
The process of transferring vector graphics into pixel images is called rasterization or rendering.
Cubic \Bezier curves are commonly used to represent sketches in vector graphics. A cubic \Bezier curve (\cref{fig:bezier}) is a smooth parametric curve defined by four points: a start point $P_0$, an end point $P_3$, and two control points $P_1$ and $P_2$ that shape the curvature . The set $P=\{P_0, P_1, P_2, P_3\}$ is often referred to as the curve's control points. 
The curve is described by the following polynomial equation:
\begin{equation}
    B(t) = (1 - t)^3P_0 + 3(1-t)^2tP_1 + 3(1-t)t^2P_2 + t^3P_3,
    \label{eq:cubic_bezier}
\end{equation}
where $t\in[0,1]$ is a parameter that moves the point along the curve from $P_0$ at $t=0$ to $P_3$ at $t=1$.

\section{Method}
Our goal is to enable an off-the-shelf pretrained multimodal LLM to draw sketches based on natural language instructions.
An overview of our pipeline is illustrated in \cref{fig:high-level-pipe}. 
We utilize a frozen multimodal LLM (\ap{SketchAgent} shown in blue), which receives three inputs: (1) a system prompt containing guidelines for using our new sketching language, (2) a user prompt with additional task-specific instructions (e.g., \ap{\textit{Draw a shark}}), and (3) a blank canvas on which the agent can draw.
Based on the given task, the agent generates a textual response, representing the sequence of strokes to be drawn, which we then process into vector graphics and render onto the canvas. The canvas can then be reused in two ways: it can be fed back into the model with an updated user prompt for additional tasks and editing, or it can be accessed by a human user who can draw directly on it to facilitate collaborative sketching.
Next, we describe each component of the pipeline.

\begin{figure}[t]
    \centering
    \includegraphics[width=1\linewidth]{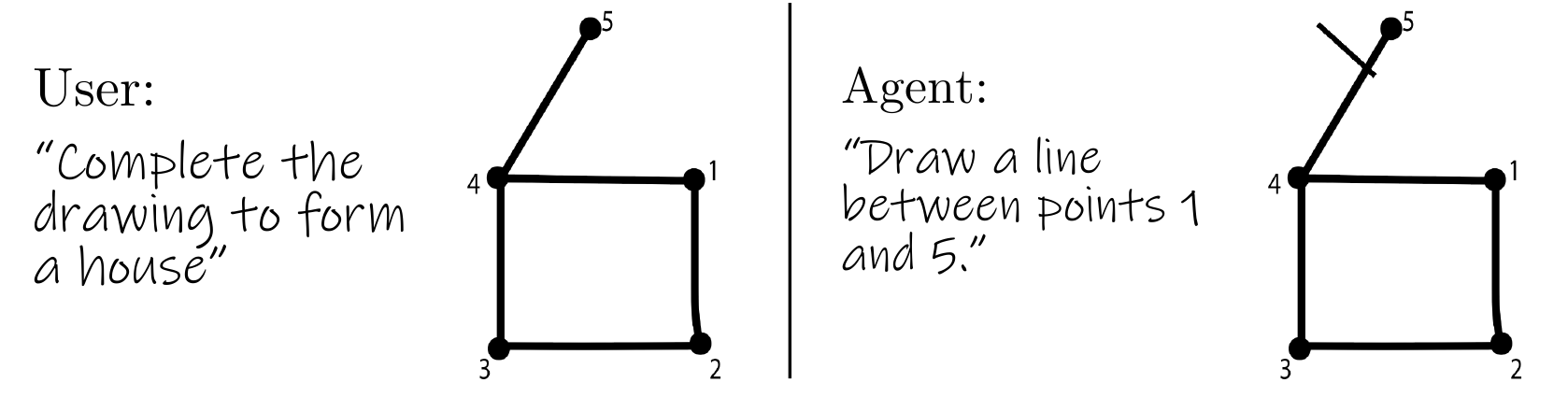}
    \vspace{-0.6cm}
    \caption{Although excelling in visual reasoning, multimodal LLMs often struggle to translate these abilities into spatial actions. In this example, GPT-4o \cite{openai2024gpt4technicalreport} intends to draw a line between points 1 and 5 but fails to execute this with a  \texttt{draw\_line} function that accepts pixel coordinates.}
    \vspace{-0.2cm}
    \label{fig:spatial_limitation}
\end{figure}

\vspace{-0.5cm}
\paragraph{The Canvas}
Although multimodal LLMs demonstrate remarkable reasoning abilities, they often struggle with spatial reasoning tasks \cite{fu2024blink, OpenEQA2023, tong2024eyeswideshutexploring}. 
We present a simple example (see \cref{fig:spatial_limitation}) to illustrate how this limitation affects the naive use of these models for sketch generation and interactive sketching.
We provide GPT-4o \cite{openai2024gpt4technicalreport} with an image depicting a simple line drawing of a partial house featuring five numbered points (from 1 to 5), and ask it to identify which points should be connected to complete the house. 
While the model correctly identifies the pair of points, it fails to select the correct pixel coordinates when given a basic \texttt{draw\_line} tool that connects two points, even after multiple attempts.
To enhance the model's spatial reasoning ability, we utilize a numbered canvas that forms a grid.
This grid features numbers (1 to 50) along the x-axis and the y-axis (\cref{fig:high-level-pipe}, left). Each cell is uniquely identified by a combination of the corresponding x-axis and y-axis numbers (e.g., the bottom-left cell is \texttt{x1y1}). The agent interacts with the canvas by specifying desired \texttt{(x,y)} coordinates.

\paragraph{Sketch Representation}
We define a sketch as a sequence of $n$ ordered strokes $S=\{S_1,S_2, ... S_n\}$.
Each stroke $S_i$ is defined by a sequence of $m$ cell coordinates on the grid: $S_i = \{(x_j,y_j)\}_{j=1}^m$, represented in string format as: \texttt{<points>x1y1, x15y20, ...</points>}.

\begin{figure}[t]
    \centering
    \includegraphics[width=0.8\linewidth]{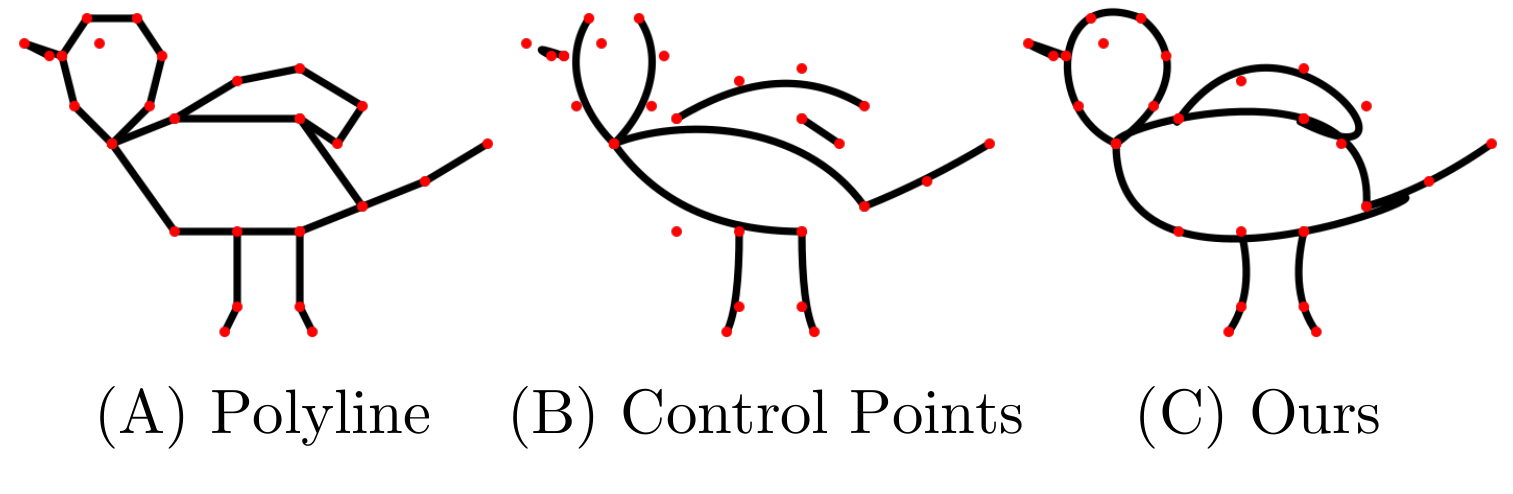}
    \vspace{-0.4cm}
    \caption{Methods for processing the agent's coordinate sequence (in red): (A) Polyline results in an unnatural appearance. (B) Directly using coordinates as Bézier control points is challenging as they do not lie on the curve. (C) Fitting a Bézier curve to sampled coordinates provides smoother results.}
    \vspace{-0.2cm}
    \label{fig:naive-stroke}
\end{figure}

A naive approach to processing the textual sequence of coordinates would be to use a polyline, connecting consecutive points with line segments. However, our grid-based representation sparsifies the canvas, resulting in a non-smooth and unnatural appearance when using polylines (see \cref{fig:naive-stroke}, left). To achieve a smoother appearance, an alternative approach is to treat the coordinates as a sequence of control points defining smooth curves. However, as illustrated in \cref{fig:bezier}, the control points often do not lie directly on the curve. Consequently, if the agent aims for a stroke that passes through specific coordinates, it must derive the control points that define this stroke, which is challenging. 

We propose an alternative approach: we treat the specified \texttt{(x,y)} coordinates as a set of desired points sampled \textbf{along} the curve, and fit a smooth \Bezier curve to them (\cref{fig:naive-stroke}, right).
To accommodate curves with complex curvature, we also task the model with determining \textbf{when} each point on the curve should be passed through, corresponding to the $t$ value described in \cref{eq:cubic_bezier}.
Thus, for each stroke $S_i$, the agent provides a set of $m$ sampled points $S_i = \{(x_j, y_j)\}_{j=1}^m$, along with a corresponding set of $t$ values: $T_i = \{t_j\}_{j=1}^m$. Based on these, we fit a cubic \Bezier curve to the sampled points by solving a system of linear equations using least squares, where the unknowns are the control points $P = \{P_0, P_1, P_2, P_3\}$:
\begin{equation}
    P = \text{argmin}_P ||AP - B||,
\end{equation}

where $A \in \mathbb{R}^{m \times 4}$ contains the cubic Bézier basis functions evaluated at specific $t_j$ values (as described in \cref{eq:cubic_bezier}), and $B \in \mathbb{R}^{m \times 2}$ contains the $m$ sampled points $\{(x_j, y_j)\}_{j=1}^m$. The least squares solution minimizes the error between the fitted Bézier curve and the sampled points.
For long sequences resulting in a large fitting error, we recursively split the curve. Additionally, we account for Bézier curves of lower degrees, including quadratic curves, linear lines, and points.
Upon completing this process, we render the parametric curves onto the canvas. 

\vspace{-0.2cm}
\paragraph{Drawing Instructions}
We provide the model with a system prompt and a user prompt (marked in orange and pink in \cref{fig:high-level-pipe}).
In the system prompt, we supply the agent with context about its expertise (\ap{\textit{You are an expert artist specializing in drawing sketches}}) and introduce it to the grid canvas along with examples of how to use our sketching language for drawing single-stroke primitives (full prompts are provided in the Appendix). The system prompt is fixed and can be applied to a variety of sketching tasks.
The user prompt includes a description of the desired task and an example of a simple sketch of a house drawn with our sketching language.
We find this to be crucial in assisting the agent with preserving the correct format that could be parsed directly \cite{NEURIPS2020_1457c0d6}.
The agent is tasked with responding in the format shown in the gray text box in \cref{fig:high-level-pipe}.
In the \texttt{<thinking>} tags, the agent is tasked to outline the overall sketching strategy \cite{Chain-of-thought24}. This typically includes describing the different components of the sketch, the intended sketching order, and the overall placement of each part.
The agent is also tasked with providing an ID tag following each stroke, which is useful for further analysis and for producing annotated sketches in scale.

\subsection{In-Chat Editing and Collaborative Sketching}
The above process can be repeated iteratively to support multiple sketching tasks and interactions.
Text-based sketch editing in a chat dialogue is enabled by feeding the rendered canvas back to the agent (see dashed arrow in \cref{fig:high-level-pipe}) and updating the user prompt with the desired edits.
To support collaborative human-agent sketching, the canvas remains accessible to both the human user and the agent throughout the entire sketching session. We define an adjustable stopping token, \texttt{</s\{j\}>}, which instructs the agent to pause generating the sequence at stroke number $j$. We then process and render the generated strokes onto the canvas up to that point, then the user can add strokes directly to the canvas to continue the sketch. The user-drawn strokes are processed and converted into the agent’s format by reversing our fitting process, i.e., sampling each stroke at multiple $t$ values (as shown in \cref{eq:cubic_bezier}), and selecting the points closest to each cell's center on the grid. The converted user strokes are then chained with the agent’s sequence, after which the agent resumes sketching until the next stopping token.

\section{Results}
We evaluate the performance of our method qualitatively and quantitatively across a selected set of sketching tasks.
Additional tasks, evaluations, and examples are provided in the Appendix.
All results presented in the paper were generated using Claude3.5-Sonnet \cite{claude} as our backbone model, unless stated otherwise.

\begin{figure}[t]
\small
    \centering
    \setlength{\tabcolsep}{2pt}
    {
    \begin{tabular}{c c c c}
    \small
        \includegraphics[width=0.23\linewidth]{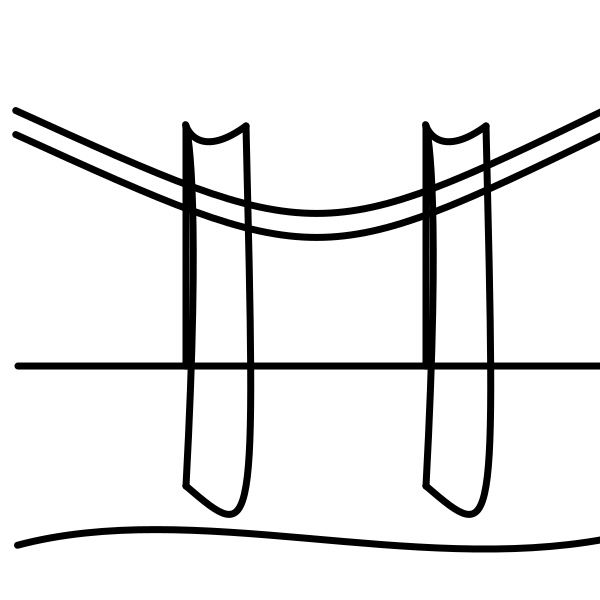} &
        \includegraphics[width=0.23\linewidth]{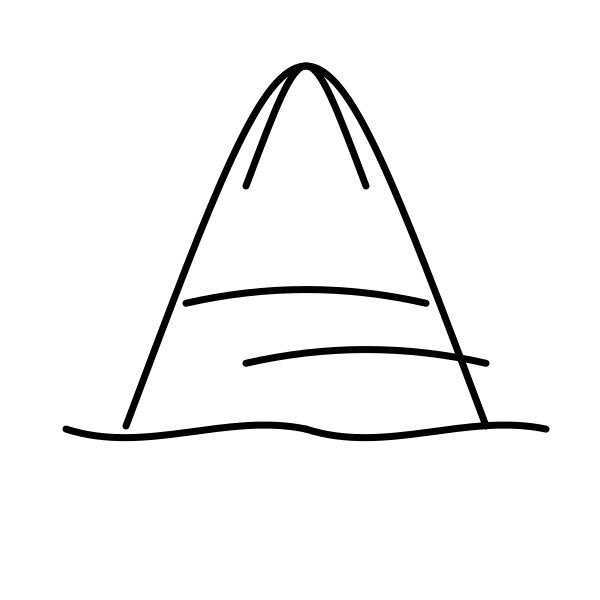} &
        \includegraphics[width=0.23\linewidth]{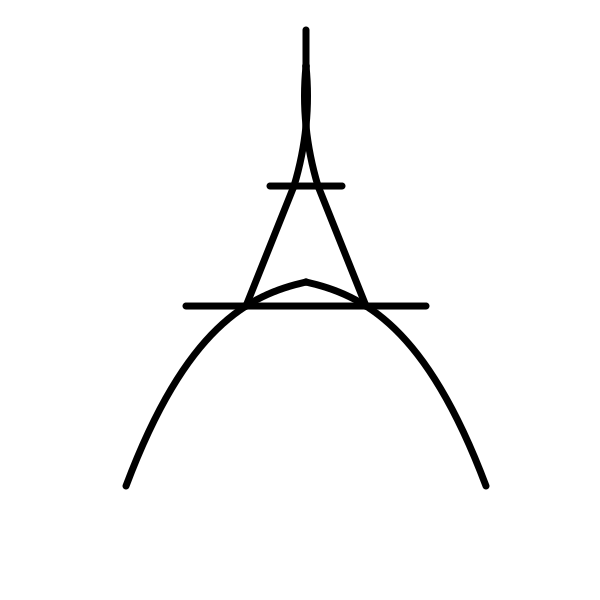} &
        \includegraphics[width=0.23\linewidth]{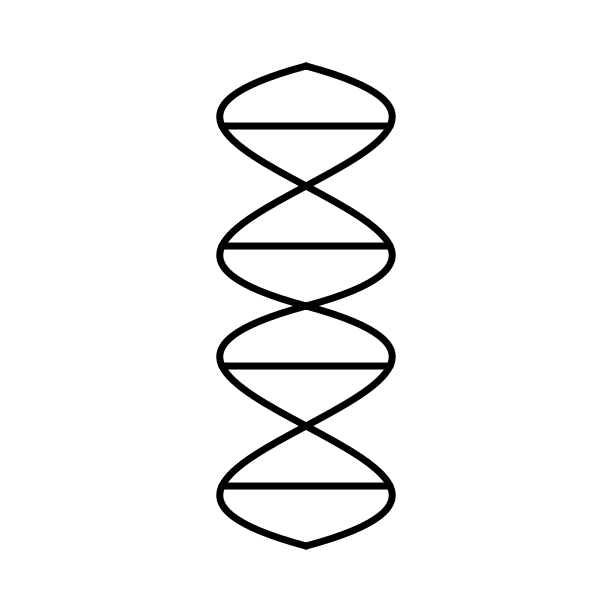} \\
        
        \begin{tabular}[c]{@{}c@{}}\small{Golden Gate} \\ Bridge\end{tabular} & \begin{tabular}[c]{@{}c@{}}Mount  \\ Fuji\end{tabular} & 
        \begin{tabular}[c]{@{}c@{}}Eiffel\\Tower\end{tabular} & 
        \begin{tabular}[c]{@{}c@{}}DNA Dou-  \\ ble Helix\end{tabular} \\

        \includegraphics[width=0.23\linewidth]{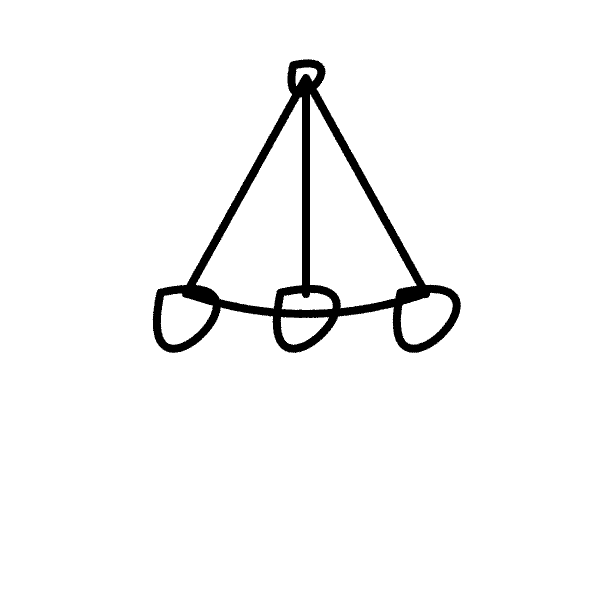} &
        \includegraphics[width=0.23\linewidth]{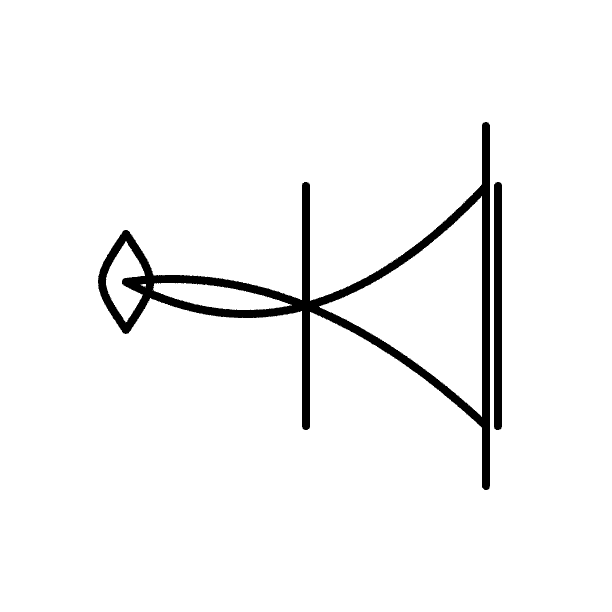} &
        \includegraphics[width=0.23\linewidth]{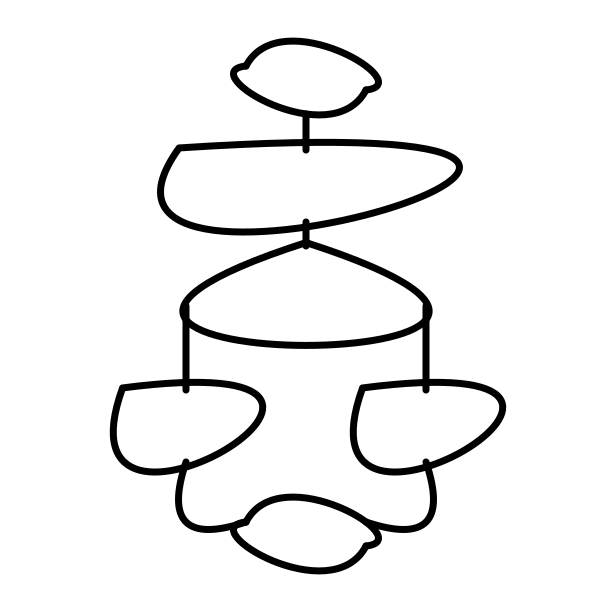} &
        \includegraphics[width=0.23\linewidth]{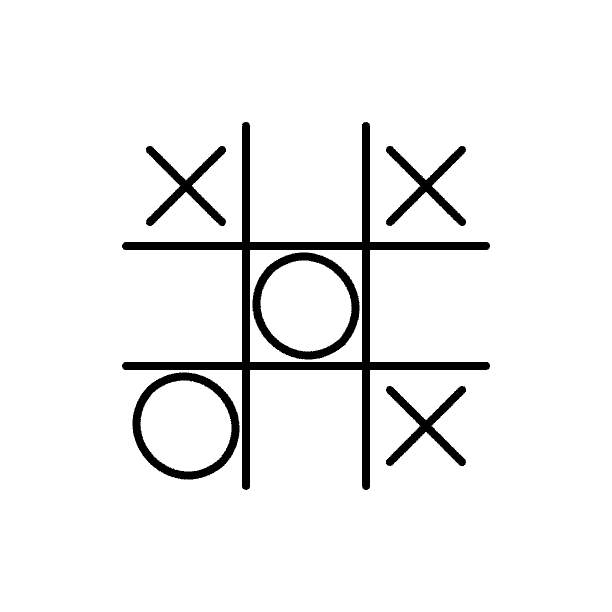} \\
        
        \begin{tabular}[c]{@{}c@{}}Pendulum  \\ Motion\end{tabular} & \begin{tabular}[c]{@{}c@{}}Double-Slit  \\ Experiment\end{tabular} &
        \begin{tabular}[c]{@{}c@{}}Flowchart  \\ \end{tabular} &
        \begin{tabular}[c]{@{}c@{}}Tic-tac \\ toe\end{tabular}\\
        
    \end{tabular}
    }
    \vspace{-0.1cm}
    \caption{Sketches produced by SketchAgent for concepts beyond pre-defined categories. The textual input describing the desired concept shown below each image.}
    \label{fig:versatile}
\end{figure}

\begin{table}[t]
\resizebox{\columnwidth}{!}{%
\setlength{\tabcolsep}{3pt}
\begin{tabular}{l| c c c c |c c}
\toprule
         & \begin{tabular}[c]{@{}c@{}}GPT-4o\\ \end{tabular} & \begin{tabular}[c]{@{}c@{}}GPT-4o\\-mini\end{tabular} & \begin{tabular}[c]{@{}c@{}}Claude3  \\ Opus\end{tabular}& \begin{tabular}[c]{@{}c@{}}Claude3.5 \\ -Sonnet*\end{tabular} & \begin{tabular}[c]{@{}c@{}}Claude3.5 \\ -Sonnet \\(SVG)\end{tabular} & \begin{tabular}[c]{@{}c@{}}Human\\(QD \cite{quickDrawData})\end{tabular} \\
       \midrule
      
        \multirow{2}{20pt}{Top1}  & 0.15      & 0.04       &0.13&  0.23      &  0.23      & $0.27$      \\
                                & $\pm 0.04$ & $\pm 0.03$ &$\pm 0.04$&  $\pm0.05$ &  $\pm0.04$ & $\pm0.07$   \\
        \hline
        \multirow{2}{20pt}{Top5}  & 0.30       & 0.10       &0.27& 0.44       & 0.43       & 0.49       \\
                                & $\pm 0.06$  & $\pm 0.04$ &$\pm 0.05$& $\pm0.03$  & $\pm0.06$  & $\pm0.06$  \\
        \hline
        
\raisebox{2\height}{Vis.} & 
\includegraphics[width=0.15\linewidth]{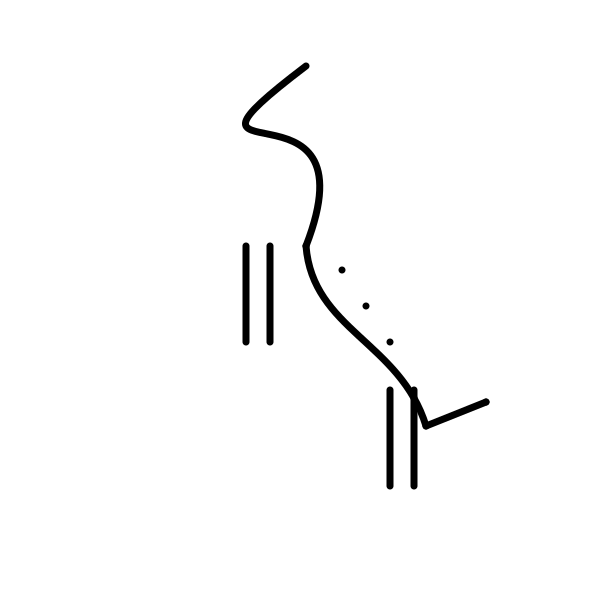} &
\includegraphics[width=0.15\linewidth]{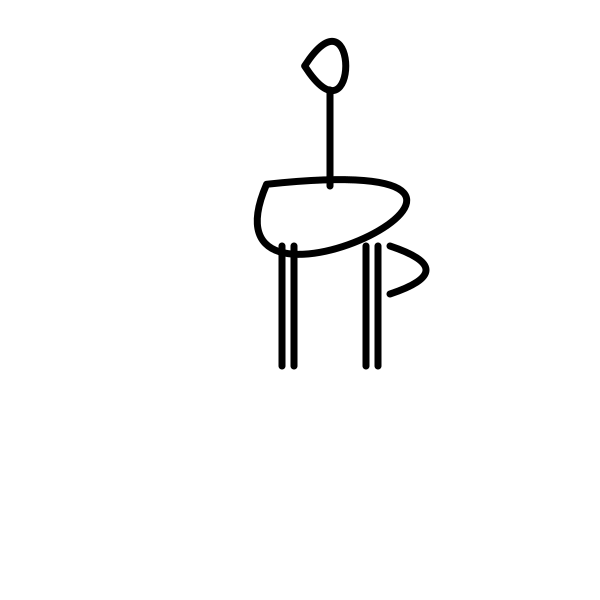} & 
\includegraphics[width=0.15\linewidth]{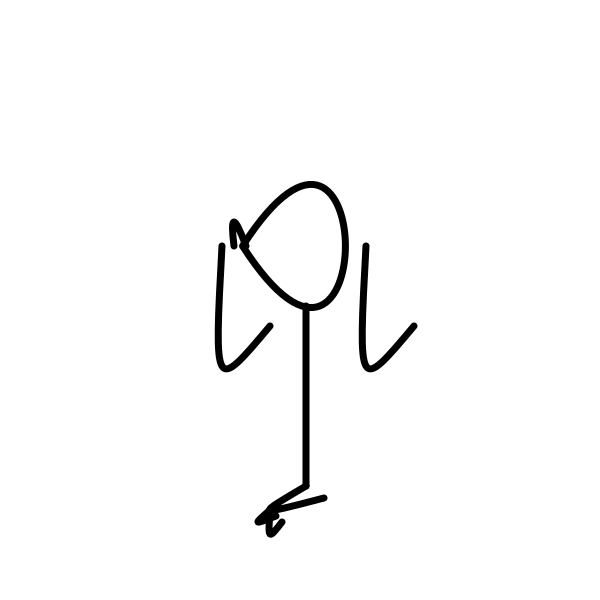} & 
\includegraphics[width=0.15\linewidth]{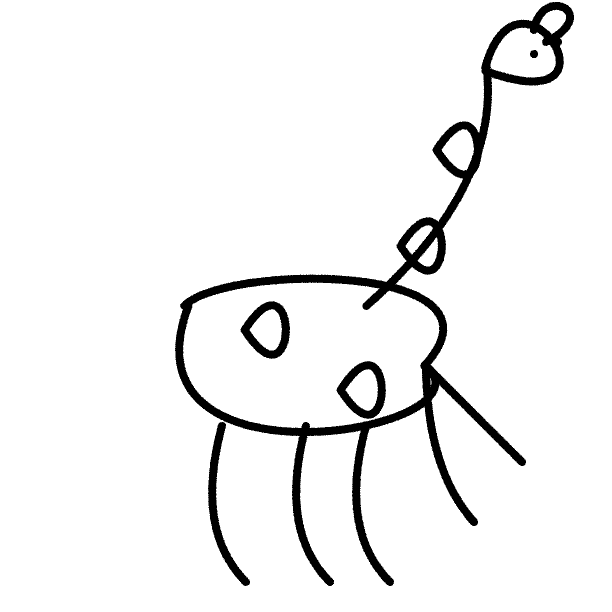} & 
\includegraphics[width=0.15\linewidth]{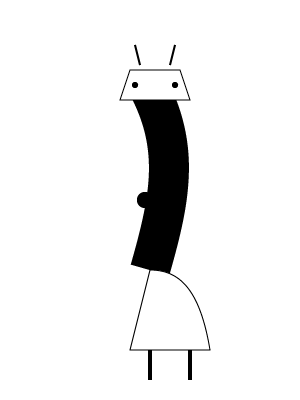} & 
\includegraphics[width=0.15\linewidth]{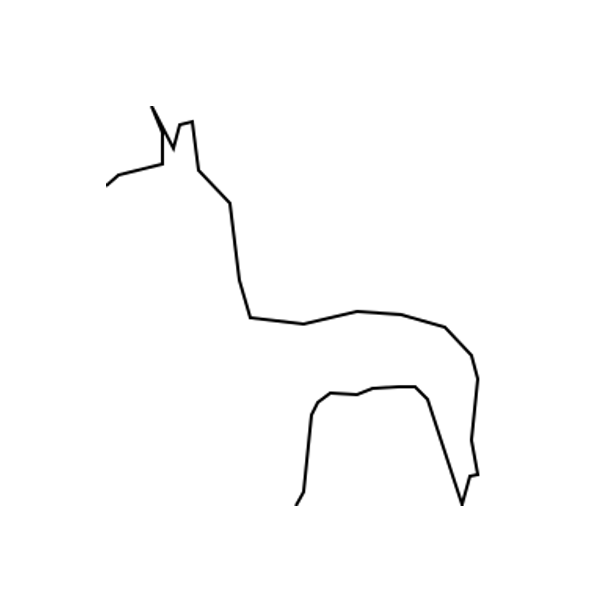}\\
\bottomrule
\end{tabular}
}
\vspace{-0.2cm}
\caption{Sketch recognition evaluation. Average Top-1 and Top-5 sketch recognition accuracy computed with CLIP zero-shot classifier on 500 sketches from 50 categories. The last row visualizes one sample from each experiment. *Indicates our default settings, which receives the highest accuracy among all models.}
\vspace{-0.2cm}
\label{tab:CLIPscore}
\end{table}

\subsection{Text-Conditioned Sketch Generation}
\label{subsec:class-cond}
\Cref{fig:teaser,fig:versatile} demonstrate SketchAgent's capability to generate sketches of various concepts that extend beyond standard categories, which includes scientific concepts (e.g., \ap{the double-slit experiment}, \ap{pendulum motion}), diagrams (e.g.,\ap{circuit diagram}, \ap{a flowchart}), and notable landmarks (e.g., \ap{Taj Mahal}, \ap{Eiffel Tower}). More examples are provided in the Appendix.
To quantitatively evaluate text-conditioned generation we utilize the QuickDraw dataset \cite{quickDrawData}. We randomly sample 50 categories (out of 345), and apply our method to generate 10 sketch instances per category, resulting in 500 sketches in total. 
Following common practice \cite{vinker2022clipasso,Vinker_2023_ICCV,NEURIPS2023_DiffSketcher,svgdreamer_xing_2023}, we utilize a CLIP zero-shot classifier \cite{Radfordclip} to evaluate how well the generated sketches depict the intended category.
We compare the performance of different multimodal LLMs by repeating the same process with GPT-4o-mini \cite{openai2024gpt4technicalreport}, GPT-4o \cite{openai2024gpt4technicalreport}, and Claude3-Opus \cite{claude} as our backbone model (in addition to Claude3.5-Sonnet \cite{claude}, our default backbone). 
As a baseline, we include human-drawn sketches sampled from the QuickDraw dataset~\cite{quickDrawData}.
The average Top-1 and Top-5 sketch classification accuracy are presented in \Cref{tab:CLIPscore}. 
As can be seen, human sketches achieve the highest recognition accuracy, with Claude3.5-Sonnet performing best among all models, approaching human-level rates under the CLIP-score metric. 
More evaluation of confusion patterns and visualization of the data are provided in the Appendix.

We additionally compare to prompting Claude3.5-Sonnet to directly generate SVGs using the following prompt:
\textit{\ap{Write SVG string that draws a sketch of a $<$concept$>$. Use only black and white colors}}. The corresponding scores are shown in the fifth column of \cref{tab:CLIPscore}.
While this approach achieves recognition scores comparable to those of SketchAgent, the outputs are often characterized by uniform and precise shapes, failing to replicate the fluidity and natural irregularity of free-hand human sketches (e.g., Fig.~\ref{fig:related}). 
To evaluate how ``human-like'' our agent’s sketches appear, we conduct a two alternative forced choice (2AFC) user study with 150 participants. Each participant was presented with pairs of sketches depicting the same object class produced by different methods, and asked to choose the sketch they believed was human-drawn.
150 sketches across 50 object classes were tested, comparing three methods: direct prompting, SketchAgent, and human sketches from QuickDraw (see Appendix for details).
Results indicate SketchAgent’s drawings appeared more human-like, being chosen as human-drawn in $74.90\pm3.35\%$ of cases when compared with direct prompting. When compared to human drawings, users slightly preferred human sketches ($54.68\pm4.61\%$) over SketchAgent's, while direct prompting was chosen only $38.9\pm5.55\%$ of the time.

\begin{figure}[t]
\centering
    \setlength{\tabcolsep}{0pt}
    {\small
    \begin{tabular}{c c}
\includegraphics[width=0.45\linewidth, valign=t]{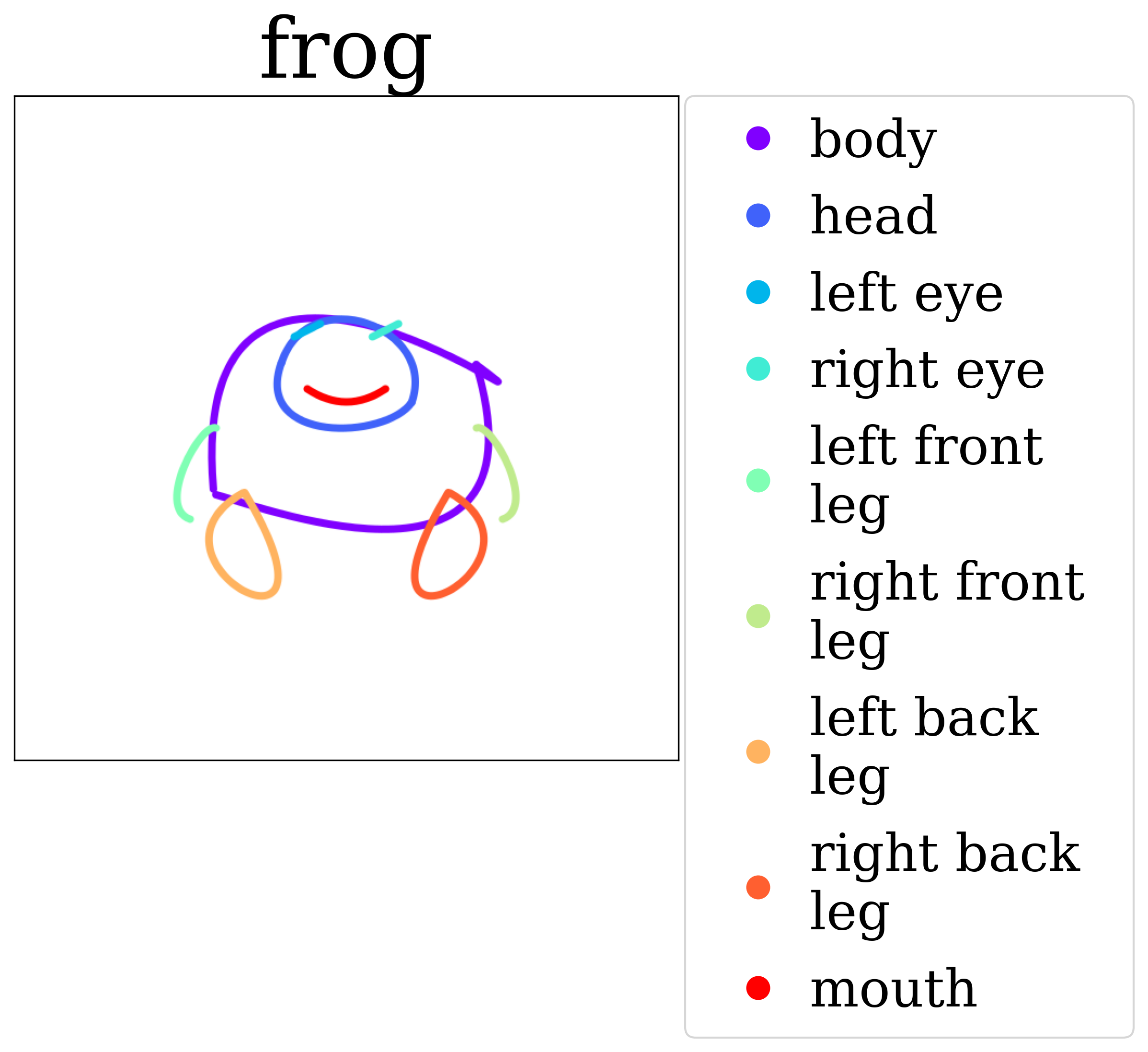} & 
\includegraphics[width=0.45\linewidth, valign=t]{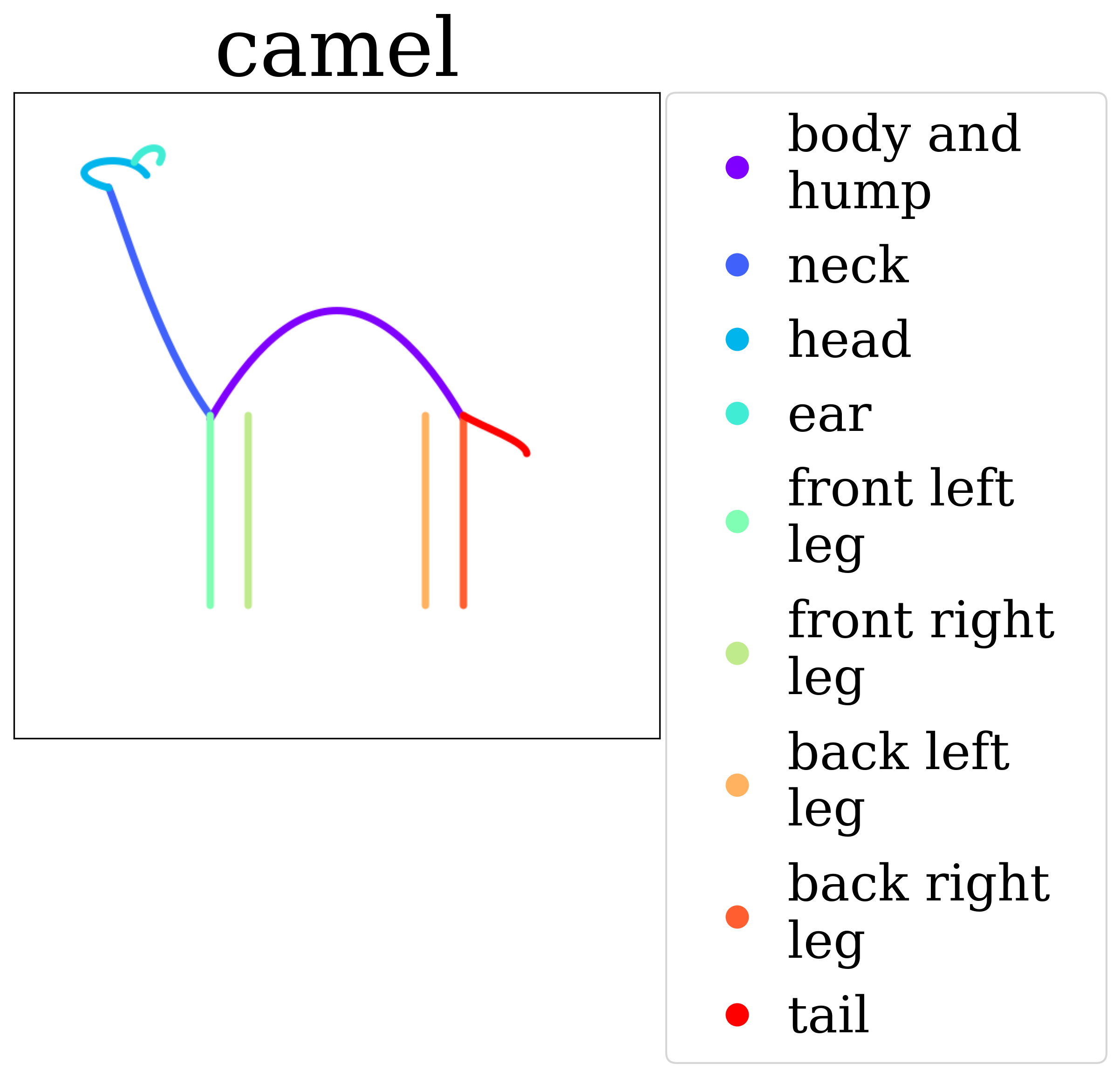}
    \end{tabular}
    }
    \vspace{-0.4cm}
    \caption{SketchAgent gradually draws stroke-by-stroke, each stroke is annotated by the agent with a semantic meaning.}
    \vspace{-0.2cm}
    \label{fig:seqlabeled}
\end{figure}

\subsection{Sequential Sketching}
\begin{figure}[t]
    \centering
    \setlength{\tabcolsep}{0pt}
    {\small
    \begin{tabular}{c c c c c c c c}
        \toprule
        \multicolumn{8}{c}{SVGDreamer \cite{svgdreamer_xing_2023}. $\approx 1.6$ hours} \\
        \midrule
        \multicolumn{2}{c}{\includegraphics[width=0.17\linewidth]{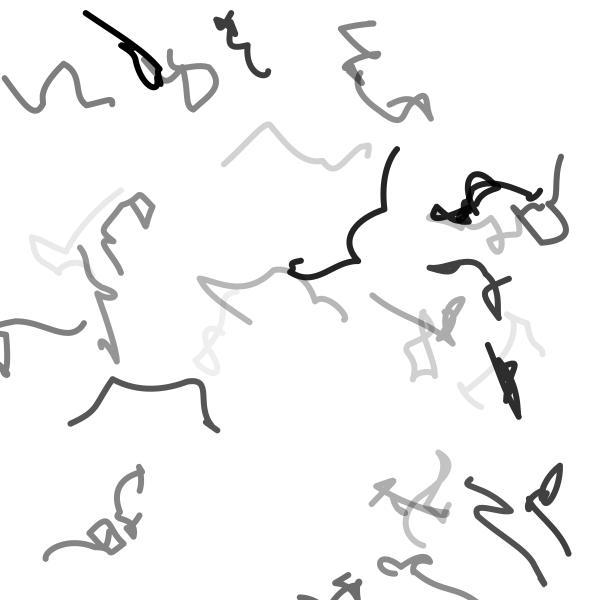}} &
        \multicolumn{2}{c}{\includegraphics[width=0.17\linewidth]{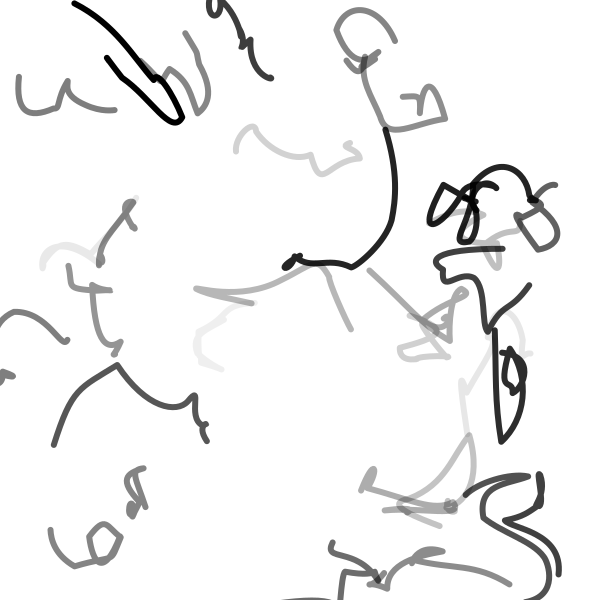}} &
        \multicolumn{2}{c}{\includegraphics[width=0.17\linewidth]{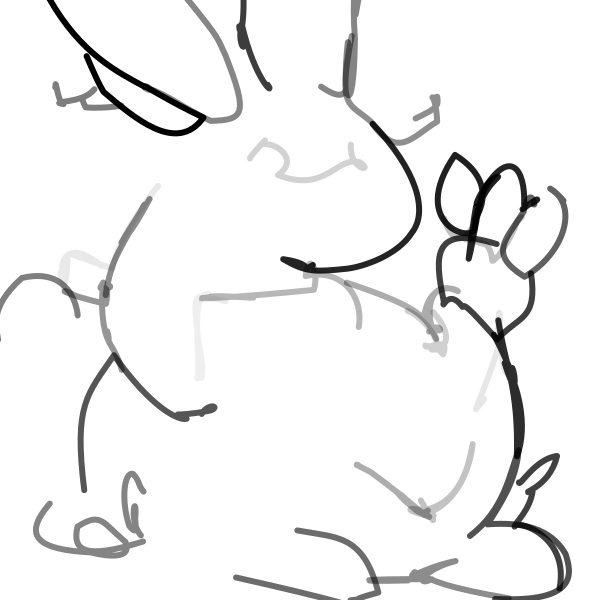}} &
        \multicolumn{2}{c}{\includegraphics[width=0.17\linewidth]{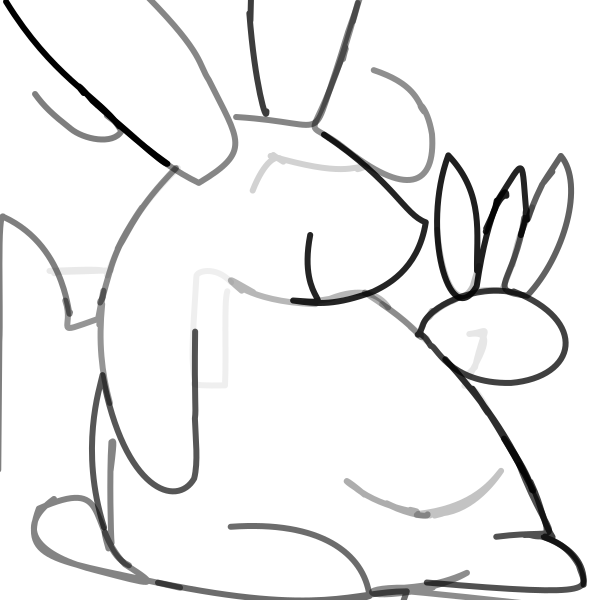}} \\

        \midrule
        \multicolumn{8}{c}{SketchRNN \cite{SketchRNN}. $\approx 4$ seconds} \\
        \midrule
        \includegraphics[trim={5cm 1cm 5cm 1cm},clip,width=0.12\linewidth]{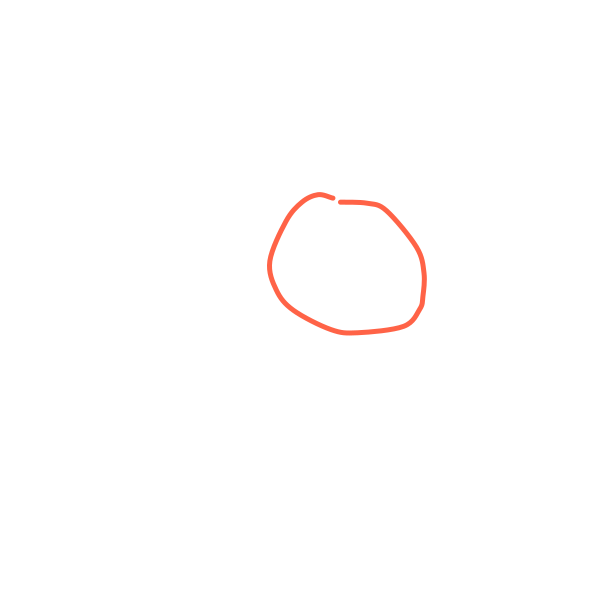} &
        \includegraphics[trim={5cm 1cm 5cm 1cm},clip,width=0.12\linewidth]{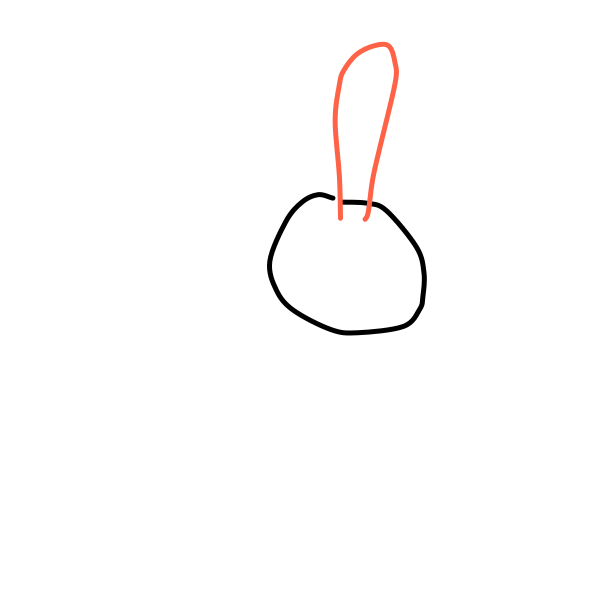} &
        \includegraphics[trim={5cm 1cm 5cm 1cm},clip,width=0.12\linewidth]{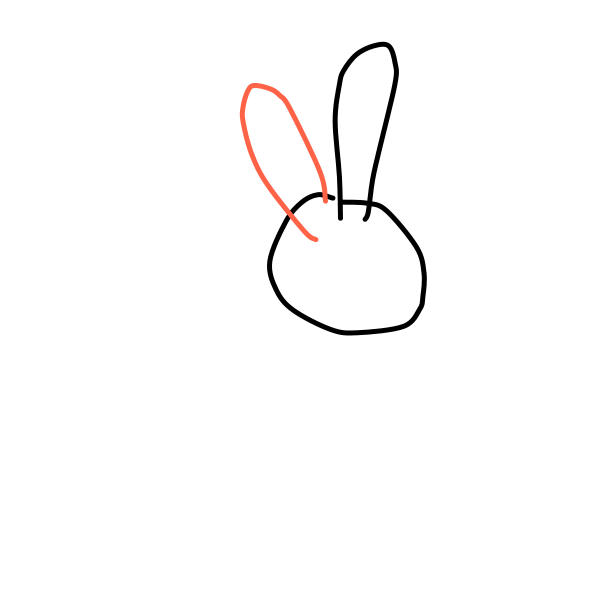} &
        \includegraphics[trim={5cm 1cm 5cm 1cm},clip,width=0.12\linewidth]{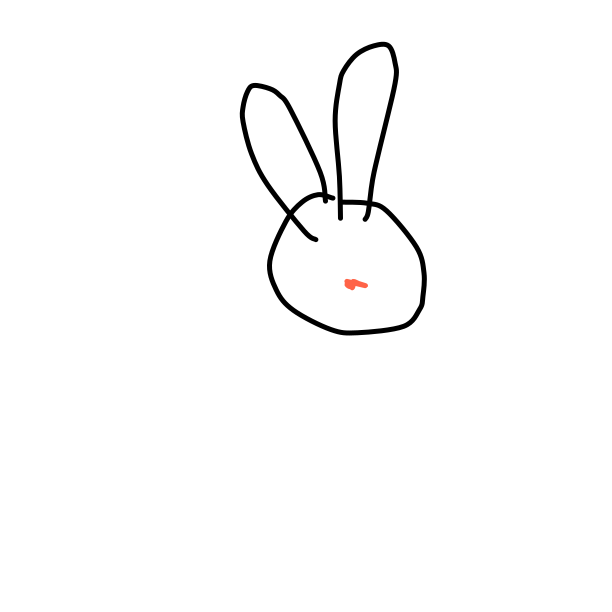} &        
        \includegraphics[trim={5cm 1cm 5cm 1cm},clip,width=0.12\linewidth]{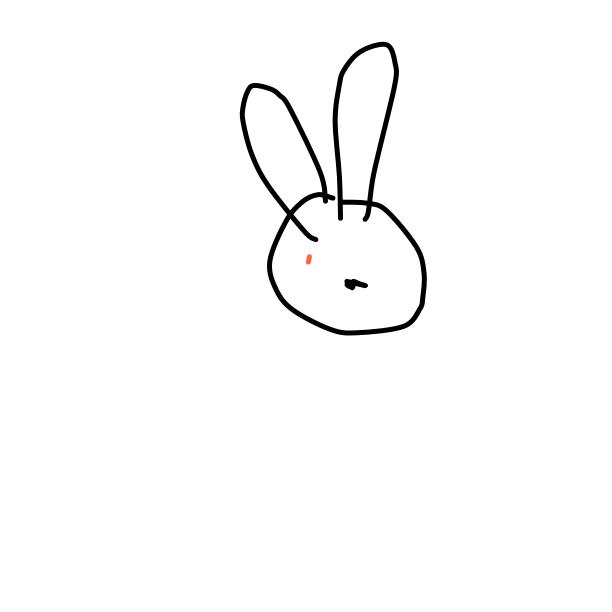} &
        \includegraphics[trim={5cm 1cm 5cm 1cm},clip,width=0.12\linewidth]{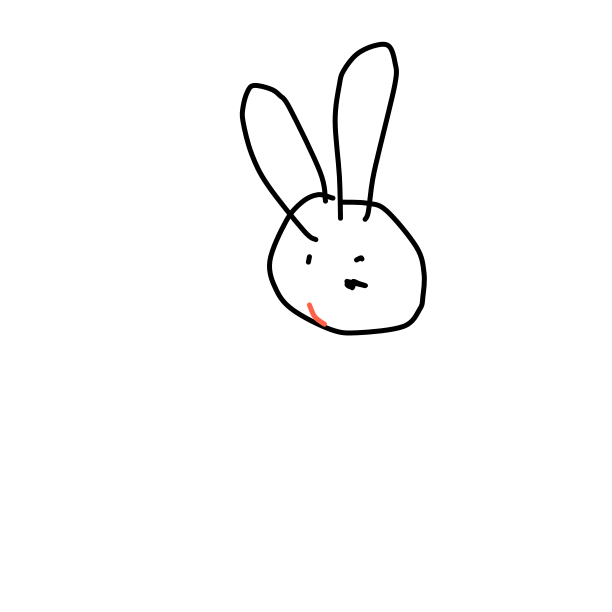} &    
        \includegraphics[trim={5cm 1cm 5cm 1cm},clip,width=0.12\linewidth]{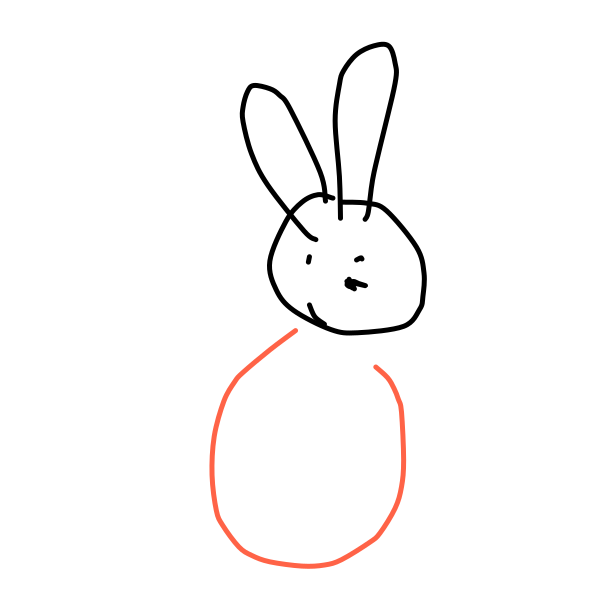} &    
        \includegraphics[trim={5cm 1cm 5cm 1cm},clip,width=0.12\linewidth]{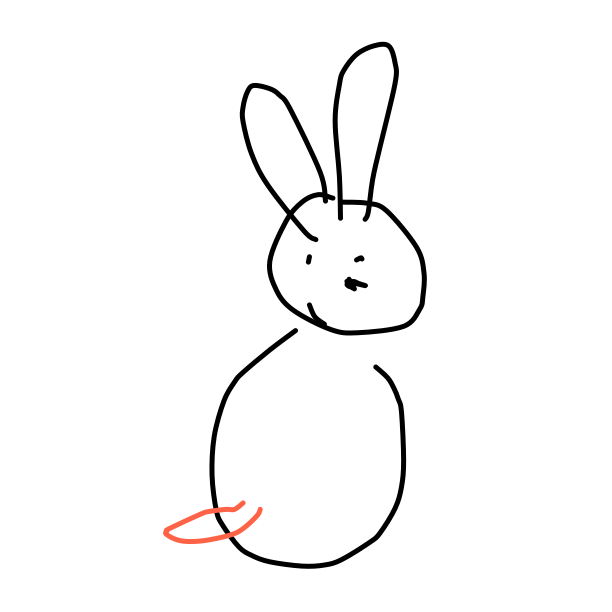}  \\
        
        \midrule
        \multicolumn{8}{c}{SketchAgent (Ours). $\approx 20$ seconds} \\
        \midrule
        \includegraphics[trim={5cm 1cm 5cm 1cm},clip,width=0.11\linewidth]{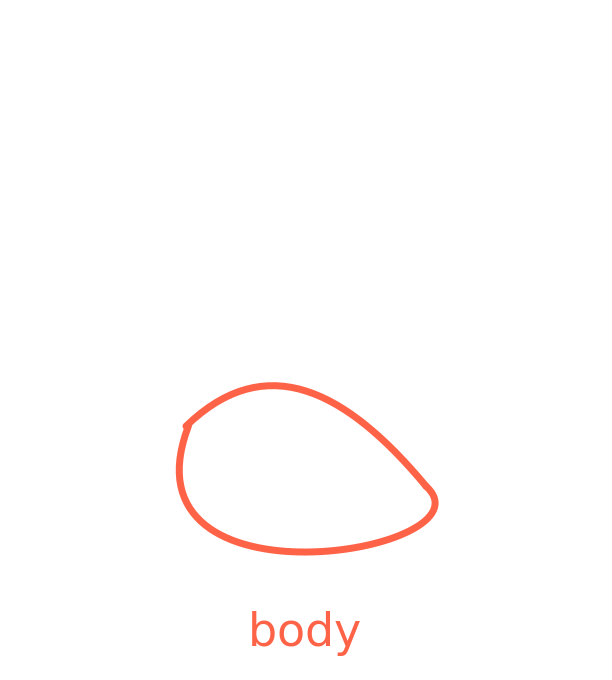} &
        \includegraphics[trim={5cm 1cm 5cm 1cm},clip,width=0.11\linewidth]{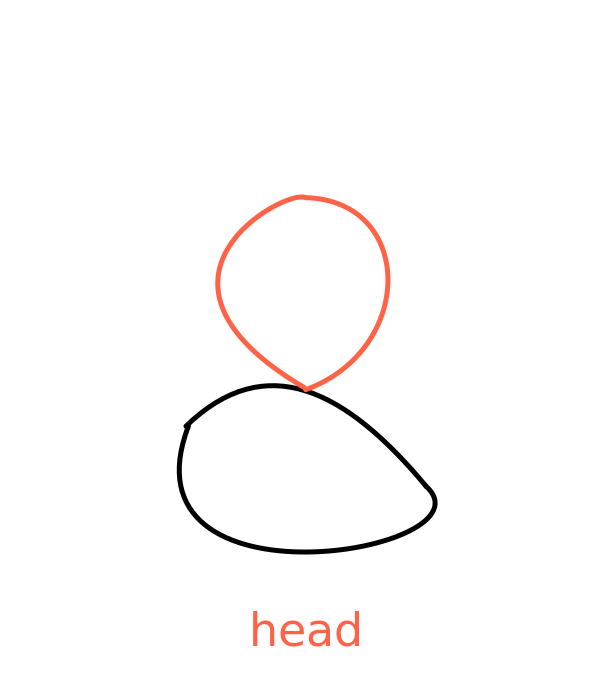} &
        \includegraphics[trim={5cm 1cm 5cm 1cm},clip,width=0.11\linewidth]{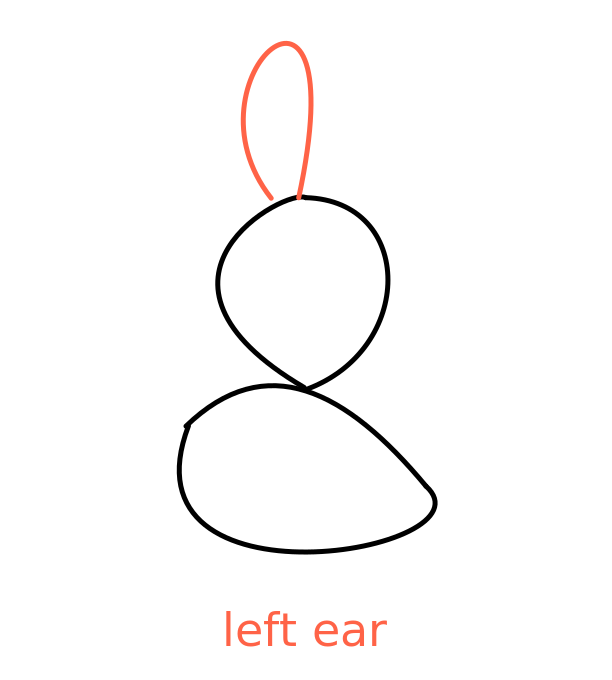} &
        \includegraphics[trim={5cm 1cm 5cm 1cm},clip,width=0.11\linewidth]{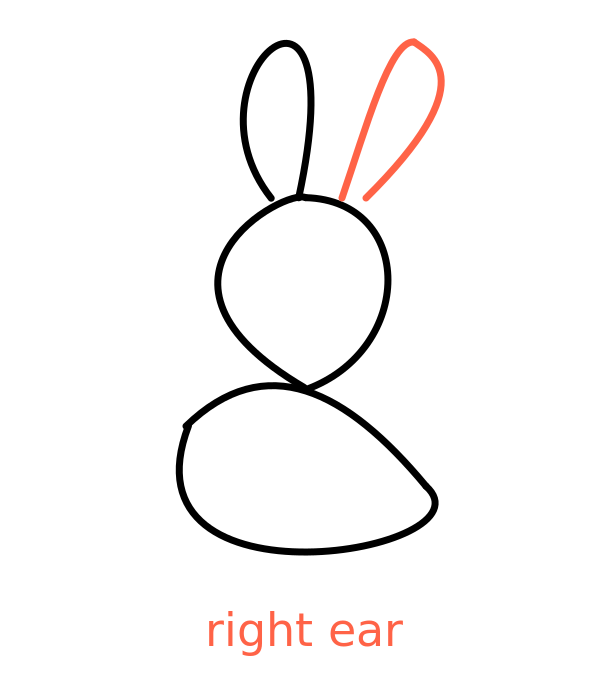} &
        \includegraphics[trim={5cm 1cm 5cm 1cm},clip,width=0.11\linewidth]{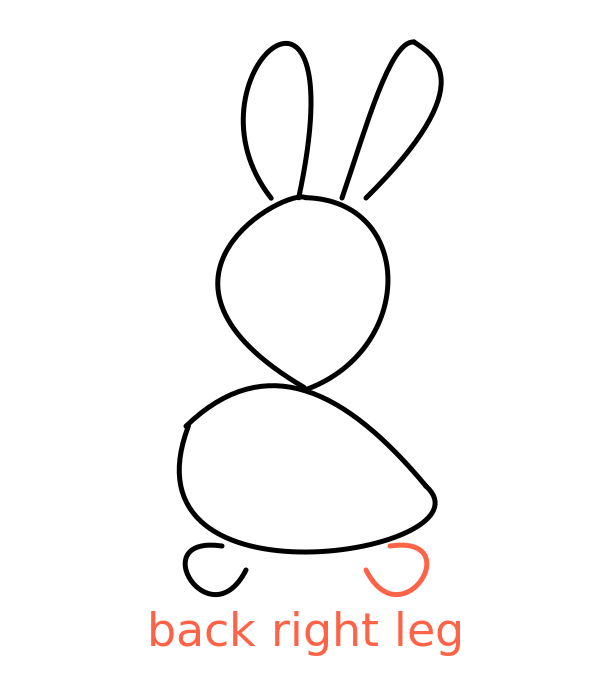} &
        \includegraphics[trim={5cm 1cm 5cm 1cm},clip,width=0.11\linewidth]{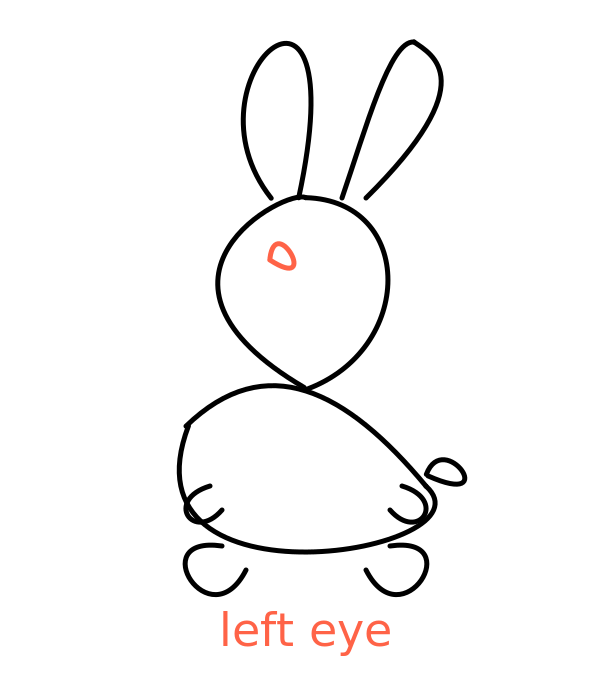} &
        \includegraphics[trim={5cm 1cm 5cm 1cm},clip,width=0.11\linewidth]{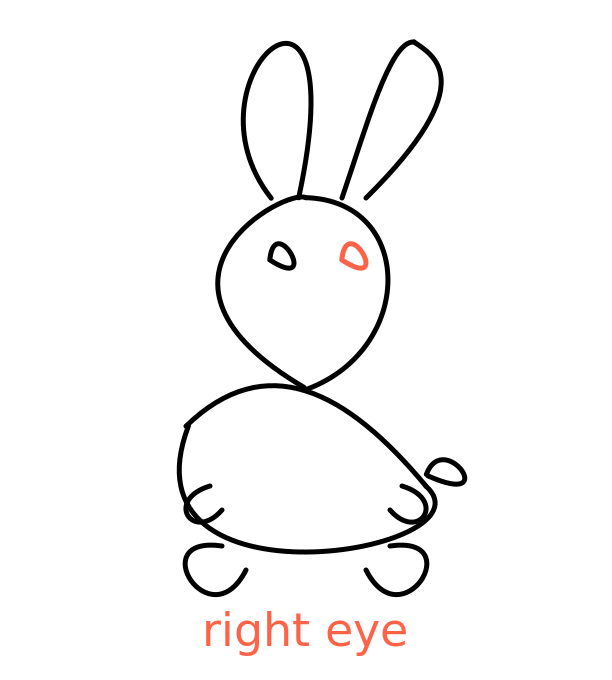} &
        \includegraphics[trim={5cm 1cm 5cm 1cm},clip,width=0.11\linewidth]{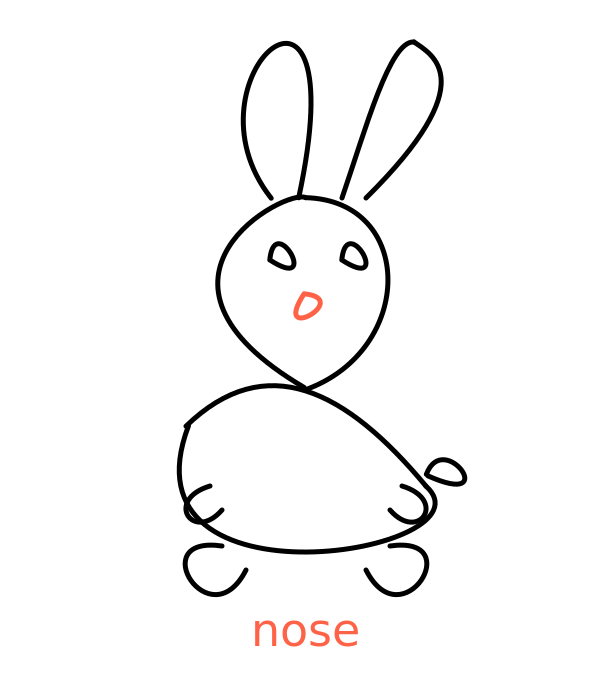}  \\

        \midrule
        \multicolumn{8}{c}{Human \cite{quickDrawData}. $\leq 20$ seconds} \\
        \midrule
        \includegraphics[trim={5cm 4cm 5cm 3cm},clip,width=0.12\linewidth]{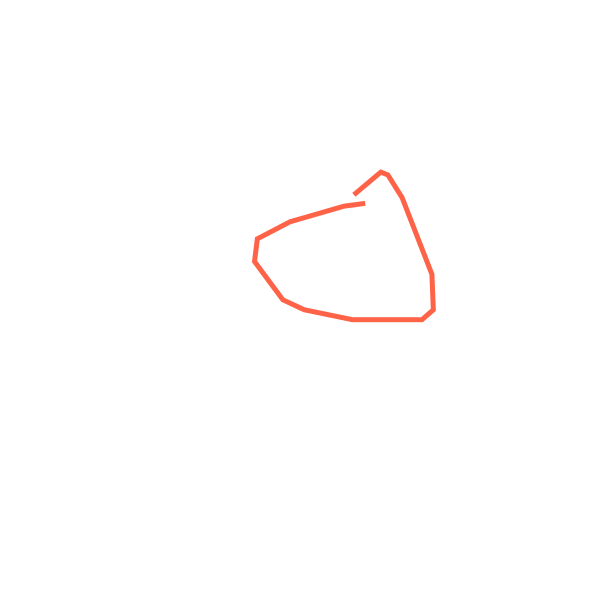} &
        \includegraphics[trim={5cm 4cm 5cm 3cm},clip,width=0.12\linewidth]{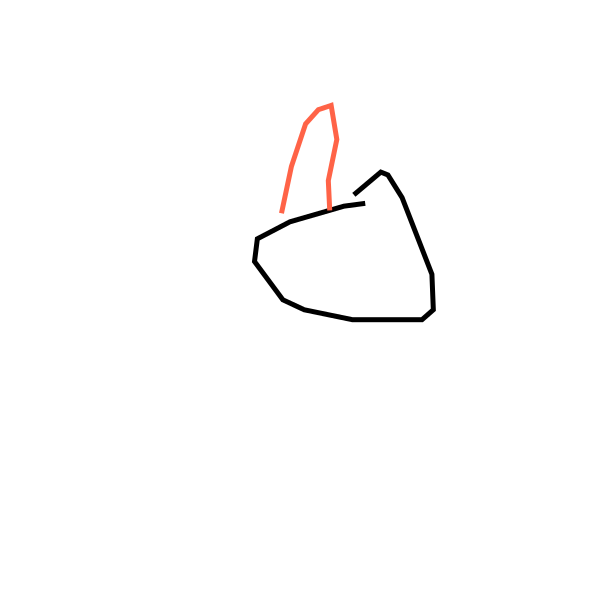} &
        \includegraphics[trim={5cm 4cm 5cm 3cm},clip,width=0.12\linewidth]{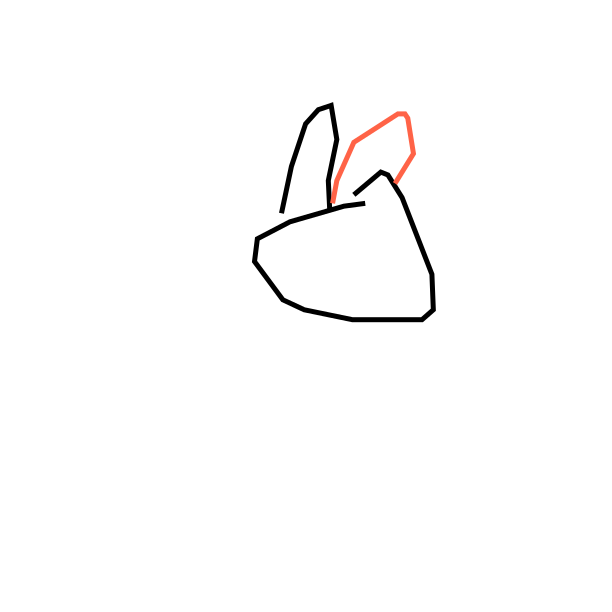} &
        \includegraphics[trim={5cm 4cm 5cm 3cm},clip,width=0.12\linewidth]{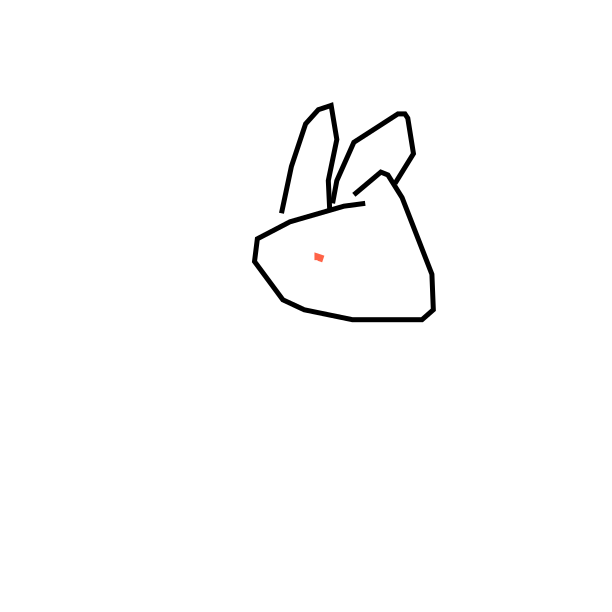} &
        \includegraphics[trim={5cm 4cm 5cm 3cm},clip,width=0.12\linewidth]{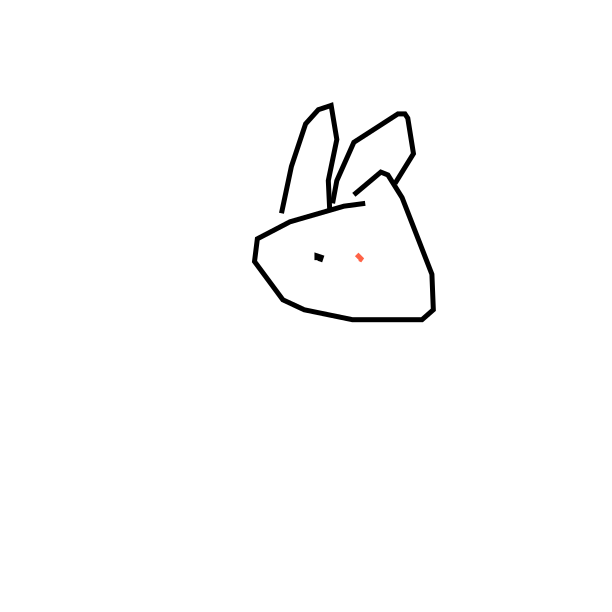} &
        \includegraphics[trim={5cm 4cm 5cm 3cm},clip,width=0.12\linewidth]{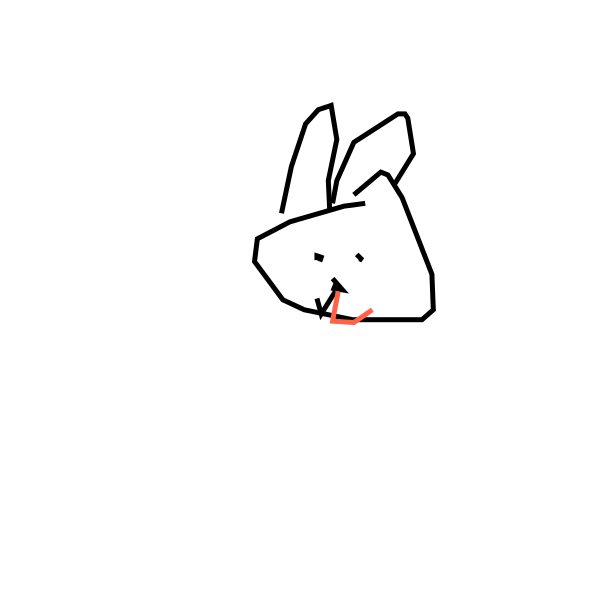} &
        \includegraphics[trim={5cm 4cm 5cm 3cm},clip,width=0.12\linewidth]{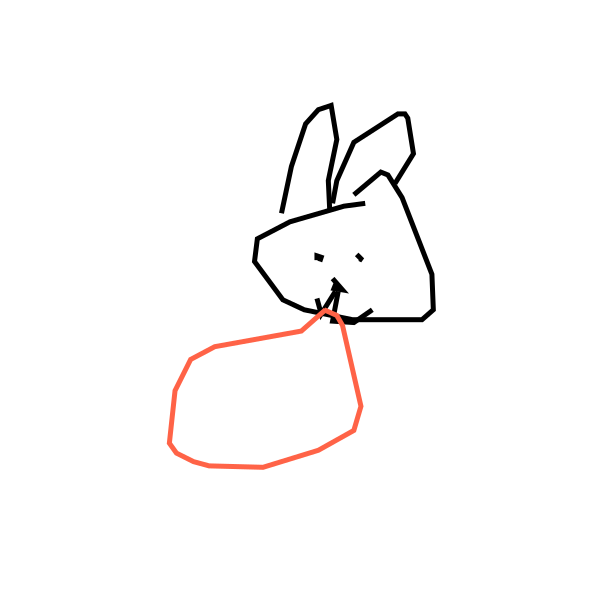} &
        \includegraphics[trim={5cm 4cm 5cm 3cm},clip,width=0.12\linewidth]{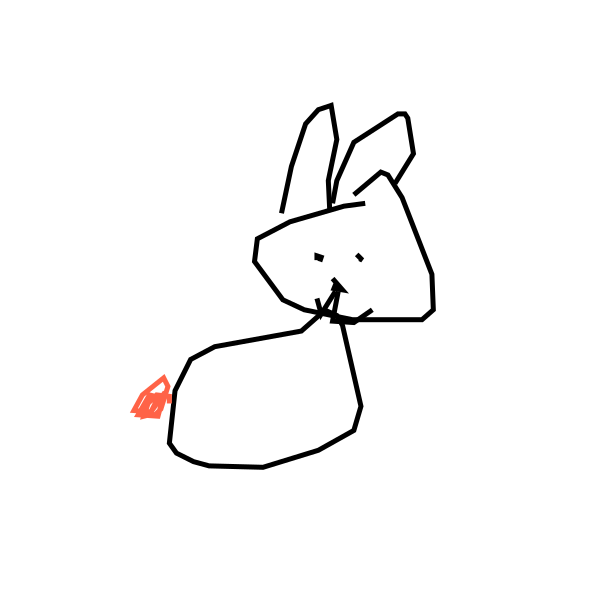} \\
        
        \bottomrule
        
    \end{tabular}
    }
    \vspace{-0.2cm}
    \caption{Sequential sketching process. SVGDreamer \cite{svgdreamer_xing_2023} requires 2000 iterations (1.6 hours) with intermediate steps lacking semantic meaning. SketchRNN \cite{SketchRNN} operates in real-time with coherent steps but is limited to QuickDraw categories. SketchAgent creates sketches gradually with meaningful strokes and no category restrictions. Human sketches also evolve through gradual, meaningful steps.}
    \vspace{-0.5cm}
    \label{fig:visual-comp}
\end{figure}

\Cref{fig:seqlabeled} shows stroke-by-stroke sketch generation by SketchAgent, with the labels on the right indicating the sketching order and the meaning our agent associates with each generated stroke (see Appendix for more examples). Stroke annotation during generation is enabled by utilizing the prior of the backbone LLM, providing a valuable feature for analysis and data collection~\cite{CreativeSeg2024, ge2021creative,Zhangsegmentation2022,Long2024,wang2023contextseg}.
In \cref{fig:visual-comp}, we illustrate why accounting for the sequential nature of sketching more closely emulates the process of human drawing.
We present the sketch creation process of SketchAgent alongside SVGDreamer \cite{svgdreamer_xing_2023}, SketchRNN \cite{SketchRNN}, and a human sketch sampled from QuickDraw \cite{quickDrawData}.
SVGDreamer (first row), is an optimization-based method, where a set of randomly initialized parametric curves (leftmost column) are iteratively refined to form a sketch, guided by a pretrained text-to-image diffusion model \cite{rombach2022highresolution}. This process is time-consuming, taking 2000 iterations (1.6 hours), which makes it unsuitable for interactive sketching.
While the final sketch (rightmost column) appears detailed and artistic due to the powerful vision backbone, the intermediate sketching and individual strokes lack clear semantic meaning.
In contrast, SketchRNN (second row) is a sequential generative model trained on human-drawn dataset, producing sketches in real-time with strokes added progressively, emulating closer a human-like sketching process (as shown in the last row). 
Similarly, SketchAgent (third row) produces sketches gradually, with each stroke carrying a semantic meaning, by utilizing the sequential nature of its backbone model.
While SketchRNN is restricted to generating sketches only within the 345 categories it was trained on, SketchAgent leverages the extensive prior knowledge of its backbone multimodal LLM, enabling it to create sketches of general visual concepts.

\begin{figure}
    \centering
    \includegraphics[width=1\linewidth]{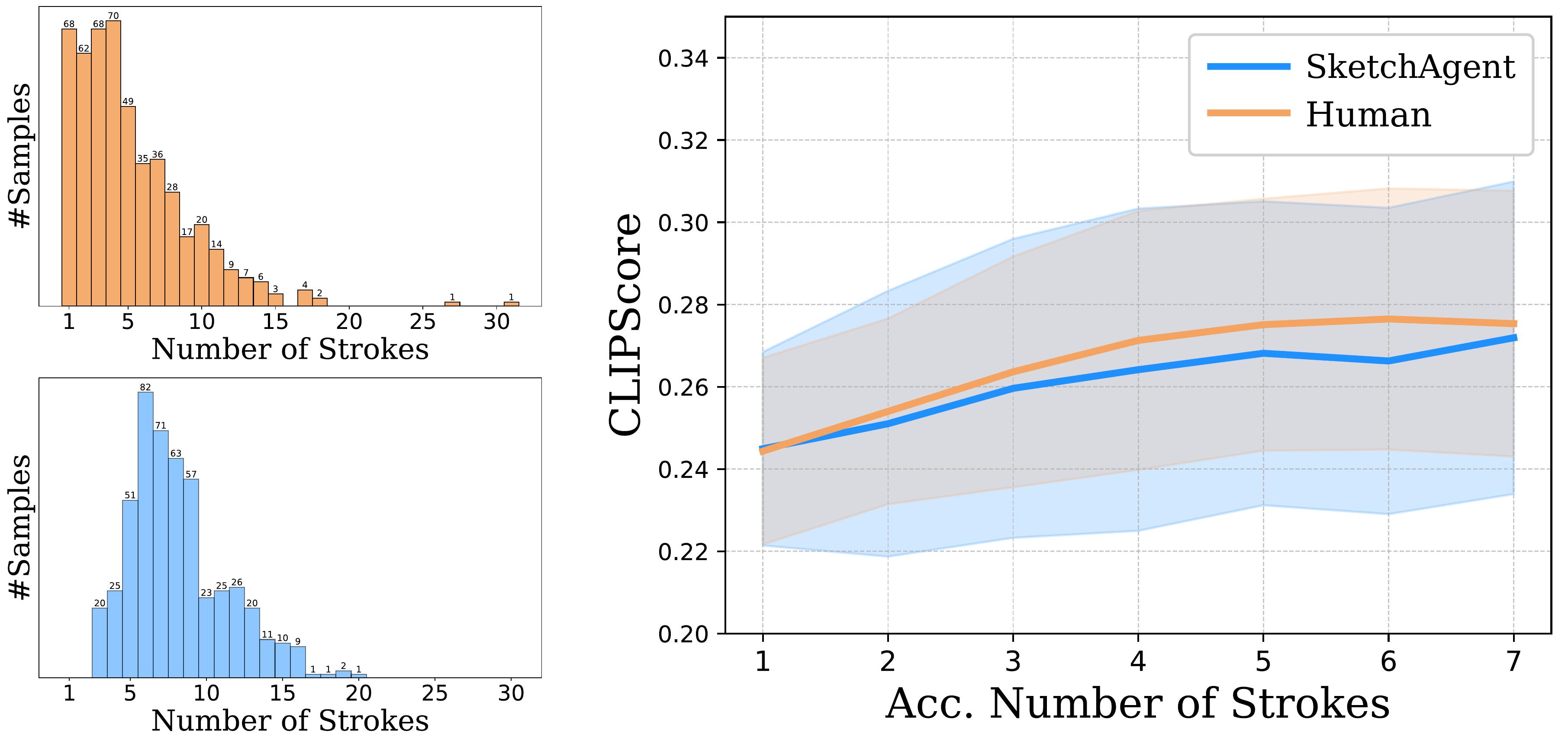}
    \vspace{-0.6cm}
    \caption{Sequential sketching analysis of SketchAgent (blue) and Humans \cite{quickDrawData} (orange). Left: Histograms of stroke distribution per sketch, showing QuickDraw sketches are more abstract on average. Right: CLIPScore as a function of the accumulated number of strokes for sketches containing 4-7 strokes, showing a similar recognition pattern over time.}
    \label{fig:clipscore-temporal}
\end{figure}

We use the set of 500 samples described in \cref{subsec:class-cond} to quantitatively analyze the sequential nature of our agent's sketches compared to human drawings. On the left of \cref{fig:clipscore-temporal}, we present histograms comparing the number of strokes in QuickDraw sketches (orange) and our sketches (blue). Most QuickDraw sketches contain 1 to 6 strokes, while our sketches show a broader distribution, peaking between 5 to 10 strokes. This suggests that, on average, QuickDraw sketches appear more abstract. To ensure a balanced comparison of sketches with similar levels of abstraction, we select sketches from both groups with a similar number of strokes (the largest intersection is found in sketches with 4-7 strokes, comprising 204 of our sketches and 120 from QuickDraw) and measure the change in CLIPScore as a function of the accumulated number of strokes (\cref{fig:clipscore-temporal}, right). Both QuickDraw and our sketches exhibit a generally similar pattern, with CLIPScore increasing as more strokes are added, suggesting that sketches become progressively more recognizable as they evolve.

\subsection{Human-Agent Collaborative Sketching}
\begin{figure}[t]
    \centering
    \setlength{\tabcolsep}{4pt}
    {\small
    \begin{tabular}{c c}
        \includegraphics[width=0.34\linewidth]{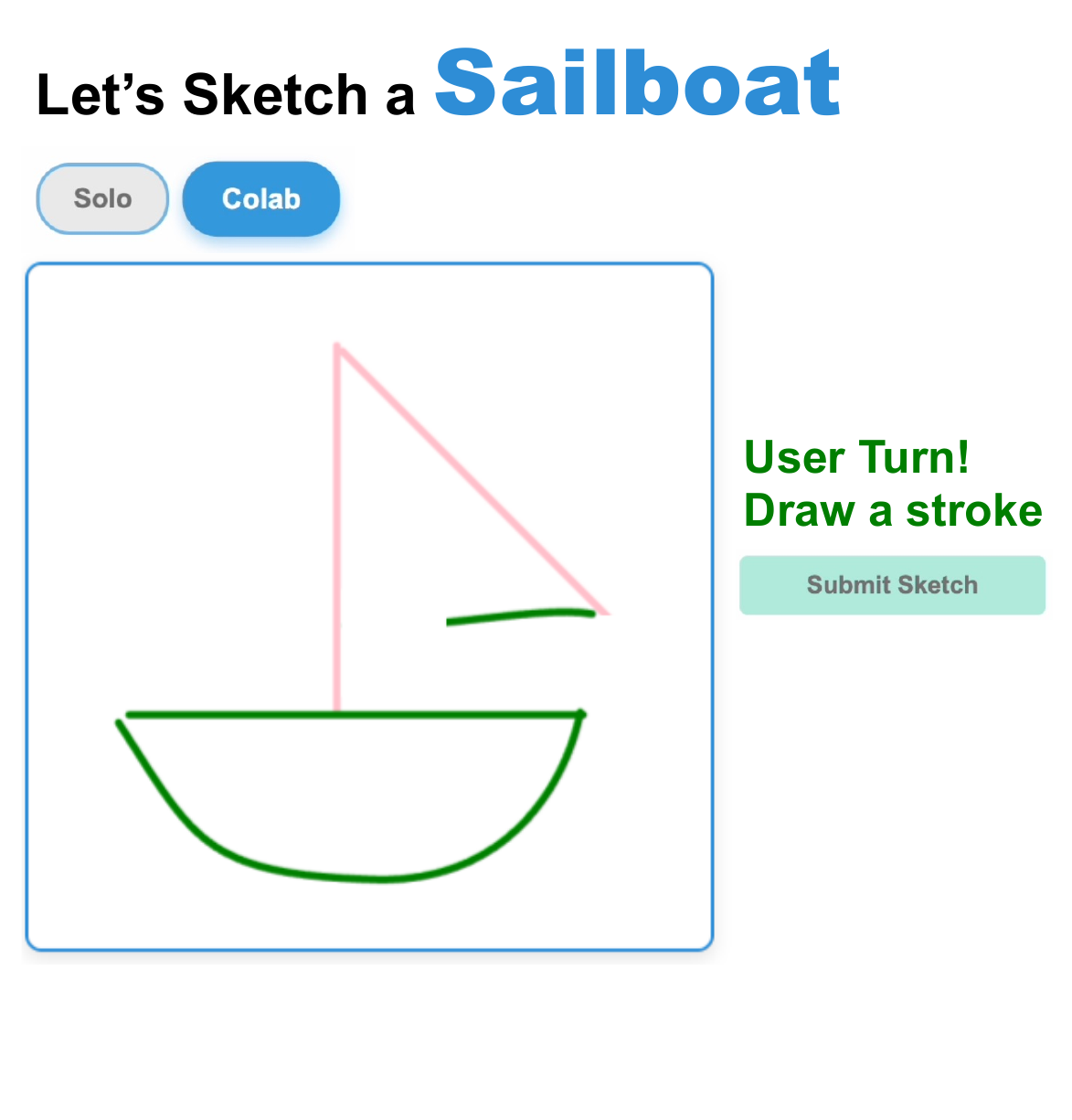}
        &
        \includegraphics[trim={0cm 0cm 0cm 2.5cm},clip,width=0.5\linewidth]{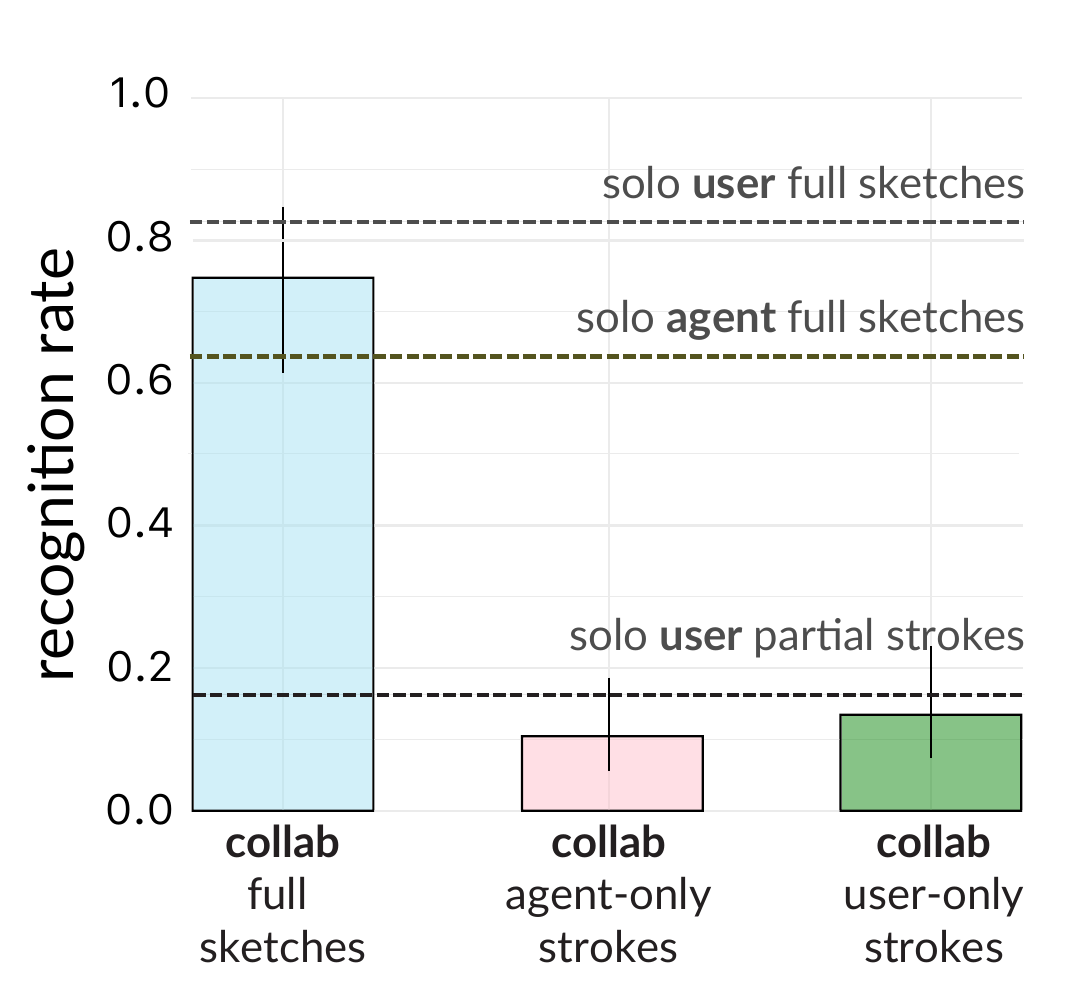}
        \\ 
        \footnotesize{(A) Sketching interface} & \footnotesize{(B) Collaborative user study results}
    \end{tabular}
    }
    \vspace{-0.2cm}
    \caption{Collaborative sketching evaluation measured using CLIP classification.
    Sketches created collaboratively (blue) approaching those made solely by users (dashed lines). In collaborative sketches, keeping agent-only strokes (pink) or user-only strokes (green) significantly reduces recognizability.}
    \label{fig:collab_res}
\end{figure}

We demonstrate the potential of our system for facilitating interactive human-agent collaboration, resulting in semantically meaningful and recognizable sketches. We design a web-based collaborative sketching environment (\cref{fig:collab_res}A) where users and SketchAgent take turns drawing on a shared canvas to create a recognizable sketch from a given textual concept.
Following the evaluation protocol in collabdraw \cite{collabdraw2019}, we select 8 simple concepts, based on the agent's demonstrated ability to draw them independently, to focus evaluations on assessing the impact of \textit{collaboration}. 
Participants sketched concepts in two modes: \textit{solo}, where users drew independently, and \textit{collab}, where users and SketchAgent collaborated, adding one stroke at a time until either was satisfied with the drawing.
We collect sketches from 30 participants, resulting in 480 sketches in total.
Average CLIP recognition rates are shown in \Cref{fig:collab_res}B. Collaboratively produced sketches (blue) achieve recognition levels close to those made solely by users and higher than those produced by the agent alone (dashed lines).
To assess the contribution of each party in collaborative mode, we analyze partial sketches with only agent-made strokes (pink) or user-made strokes (green), resulting in a significant reduction in recognizability. This suggests that both user and agent contribute meaningfully to the recognizability of the complete sketch.

\subsection{Chat-Based Sketch Editing}
We next demonstrate the effectiveness of our method in performing interactive text-based sketch editing within a chat dialogue, where the input to the agent combines both text and images. 
Inspired by~\cite{sharma2024vision}, we explore edits that involve spatial reasoning and object relations.
We focus on three object categories: outdoor, indoor, and animals, with three objects each, and design editing prompts to add objects to the input sketches.
For outdoor and indoor objects, we specify relative locations of added concepts, e.g., \ap{left to}, \ap{on top of} (see \cref{fig:editing} left). 
For the animals category, we tasked the agent with adding accessories to each animal without guidance on their exact placement, testing its ability to infer placement based on semantics (e.g., placing a hat on a head (see \cref{fig:editing} right).
The full list of object and editing instructions is provided in the Appendix. 
We produced a total of 54 sketches. Evaluating the edited sketches reveals that SketchAgent correctly follows instructions 92\% of the time, with 94\% accuracy for specified relations and 88\% accuracy for inferred semantic relations. 

\begin{figure}
    \centering
    \small
    \setlength{\tabcolsep}{1pt}
    \renewcommand{\arraystretch}{0.83}
    {\small
\begin{tabular}{>{\centering\arraybackslash}p{.33\linewidth}>{\centering\arraybackslash}p{.33\linewidth}>{\centering\arraybackslash}p{.33\linewidth}}

        \footnotesize\ap{Building} & \footnotesize\ap{Nightstand} & \footnotesize\ap{Cat} \\
        \includegraphics[trim={3cm 1.8cm 0cm 0cm},clip,width=.7\linewidth]{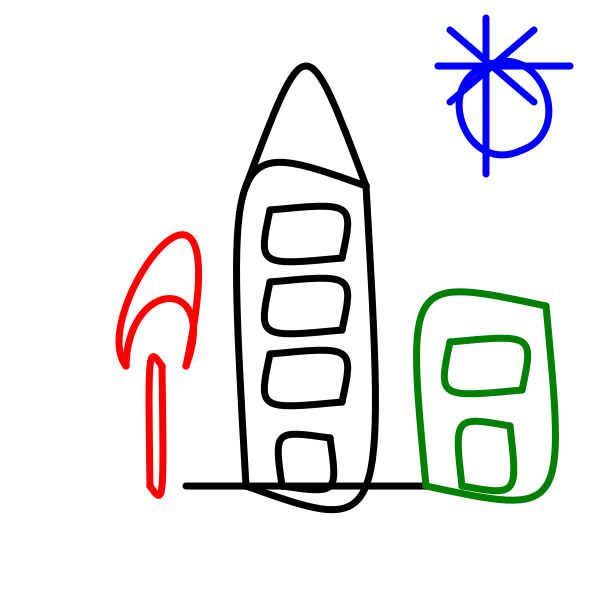} &
        \includegraphics[trim={0cm 1cm 2cm 0cm},clip,width=.7\linewidth]{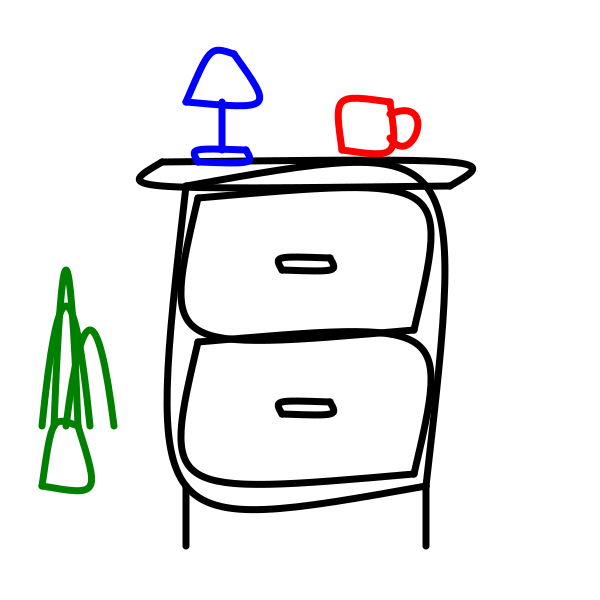} &
        \includegraphics[trim={2cm 2cm 2cm 0cm},clip,width=.65\linewidth]{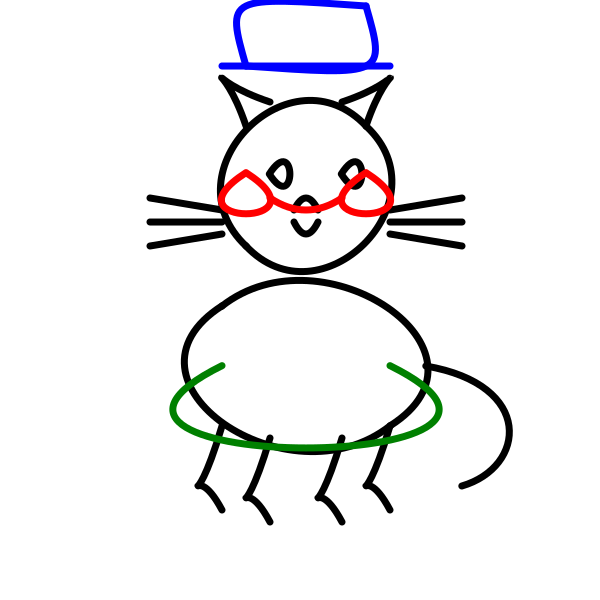}\\
        
        \begin{tabular}[c]{@{}c@{}}\textcolor{red}{\footnotesize{\ap{Tree to the left}}} \\ \textcolor{blue}{\footnotesize{\ap{Sun on top right}}} \\ \textcolor{ForestGreen}{\footnotesize{``Building to the right''}}
        \end{tabular} & 
        \begin{tabular}[c]{@{}c@{}}\textcolor{red}{\footnotesize{\ap{Coffee mug on top}}} \\ \textcolor{blue}{\footnotesize{\ap{Lamp on top}}} \\ \textcolor{ForestGreen}{\footnotesize{\ap{Plant to the left}}} \end{tabular} &
        \begin{tabular}[c]{@{}c@{}}\textcolor{red}{\footnotesize{\ap{Add glasses}}} \\ \textcolor{blue}{\footnotesize{\ap{Add a hat}}} \\ \textcolor{ForestGreen}{\footnotesize{\ap{Add a skirt}}}\end{tabular} \\
        
    \end{tabular}
    }
    \vspace{-0.3cm}
    \caption{Chat-based sketch editing. We iteratively prompt SketchAgent to add objects to sketches through chat dialogues.}
    \vspace{-0.4cm}
    \label{fig:editing}
\end{figure}

\section{Ablation}
We evaluate the impact of each component of our method by systematically removing them and measuring sketch recognition rates as detailed in~\ref{subsec:class-cond}. We assess the effects of removing the system prompt, omitting the CoT process (i.e., excluding thinking tags and 'think step-by-step' instructions), and modifying ICL (the complete sketch example provided in the user prompt). When modifying ICL, we use a correctly formatted single-stroke example instead of the complete sketch, as fully removing ICL results in outputs that do not follow the expected format making them unparsable.
The results in Table~\ref{tab:ablation} show that the full SketchAgent pipeline achieves the highest performance, highlighting the importance of each component. Interestingly, not providing a complete sketch example significantly reduces performance. Additional visualizations and analyses are provided in the Appendix.

\begin{table}[]
\centering
\resizebox{\columnwidth}{!}{
\begin{tabular}{l| c c c c}
\toprule
       & \begin{tabular}[c]{@{}c@{}} w/o System \\ Prompt \end{tabular} 
       &w/o CoT
       &\begin{tabular}[c]{@{}c@{}} Modified \\ ICL \end{tabular}
       & \begin{tabular}[c]{@{}c@{}} SketchAgent \\ (full) \end{tabular}
       \\
       \midrule
        Top1 & $0.20\pm0.04$ & 0.14  $\pm0.02$    & $0.07\pm0.02$    & $0.23\pm0.04$   \\
        \hline
        Top5 & $0.42\pm0.03$    & $0.29\pm0.04$     & $0.16\pm0.03$     & $0.43\pm0.06$ \\   
\bottomrule
\end{tabular}
}
\vspace{-0.2cm}
\caption{Ablation study. Average Top-1 and Top-5 CLIP recognition accuracy. We systematically remove each component in our pipeline, showcasing all components contribute to the agent's full performance.}
\label{tab:ablation}
\end{table}

\section{Limitations and Future Work} 
\begin{figure}[h]
    \centering
    \includegraphics[width=1\linewidth]{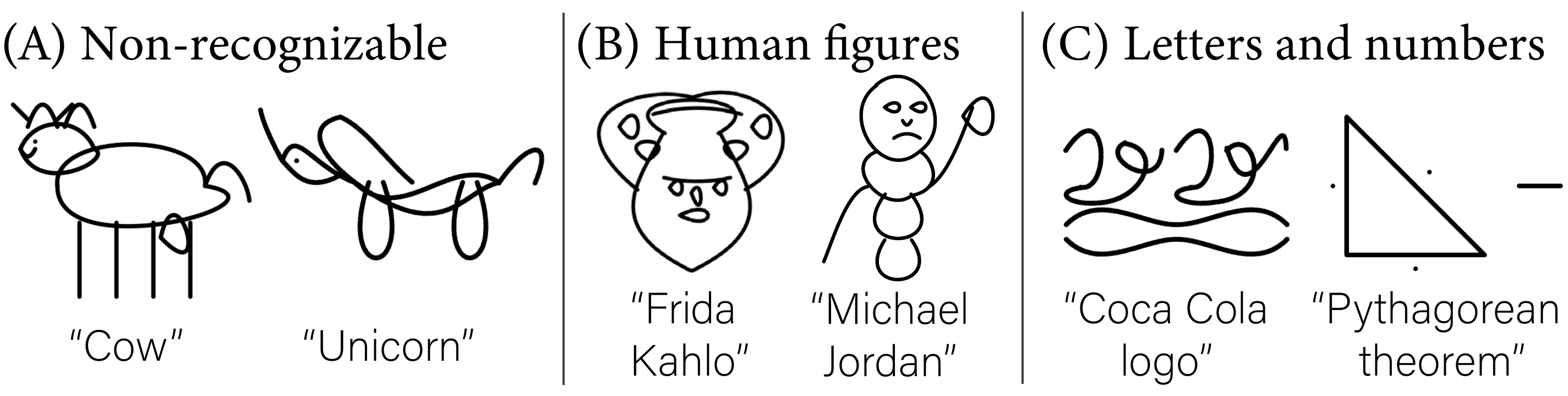}
    \vspace{-0.6cm}
    \caption{Limitations. Sketches of complex concepts (A) and human figures (B) appear too abstract and unrecognizable with non-professional style. (C) Fail to depict letters and numbers.}
    \label{fig:limitations}
\end{figure}

SketchAgent has several limitations. 
First, it is constrained by the priors of the backbone model, primarily optimized for text rather than visual content. As a result, the agent often produces rich textual descriptions of object parts but struggles to convert these into effective sketching actions, resulting in overly abstract and unrecognizable outputs. For example, in \cref{fig:limitations}A, the agent effectively describes key parts of a unicorn (e.g., the horn), but the sketch is unrecognizable.
This constraint also impacts the depiction of human figures (\cref{fig:limitations}B). While distinctive features (e.g., Frida Kahlo's eyebrows or Michael Jordan's dunk) may be captured well in language, the resulting sketches are overly simple, with an amateur style, lacking expressivity. We expect this issue to improve as future models advance in vision capabilities. Lastly, the agent may struggle with drawing letters and numbers. This could be improved in future work by providing relevant in-context examples.

\section{Conclusions}
We presented a method for language-driven, sequential sketch generation, that can produce versatile sketches in real-time and meaningfully engage in collaborative sketching sessions with humans. We show that the prior knowledge embedded in pretrained multimodal LLMs can be effectively leveraged for sketch generation through an intuitive sketching language and a grid canvas, without requiring additional training or fine-tuning. 
We hope our work represents a meaningful step toward developing general-purpose sketching systems with the potential to enhance human-computer communication and computer-aided ideation.

\section{Acknowledgements}
We thank Yuval Alaluf, Hila Chefer, Assaf Ben Kish, Joanna Materzynska, Rinon Gal, Elad Richardson, Arnav Verma, and Ellie Arar for providing feedback on early versions of our manuscript. We are especially grateful to Yarden Frenkel for his insights, early explorations, and engaging discussions.
This work was partially supported by NSF CAREER $\#2047191$, NSF DRL $\#2400471$, Stanford Human Centered AI Institute Hoffman-Yee Grant, Hyundai Motor Company, ARL grant W911NF-18-2-021, the Zuckerman STEM Leadership Program, and the Viterbi Fellowship. The funders had no role in the experimental design or analysis, the decision to publish, or manuscript preparation. The authors have no competing interests to report.

{
    \small
    \bibliographystyle{ieeenat_fullname}
    \bibliography{main}
}

\clearpage
\appendix
\setcounter{page}{1}

\newpage
\twocolumn[
\centering
\Large
\textbf{\thetitle}\\
\vspace{2em}Supplementary Material \\
\vspace{1.0em}
] %

\part{}
\vspace{-20pt}
\parttoc

\small{\paragraph{IRB Disclosure} We received IRB approvals for all user studies, from all of the institutions involved. Accordingly, we took measures to ensure participant anonymity and refrained from showing them potentially offensive content.}

\section{Technical Details}
\begin{figure}
    \centering
    \includegraphics[width=1\linewidth]{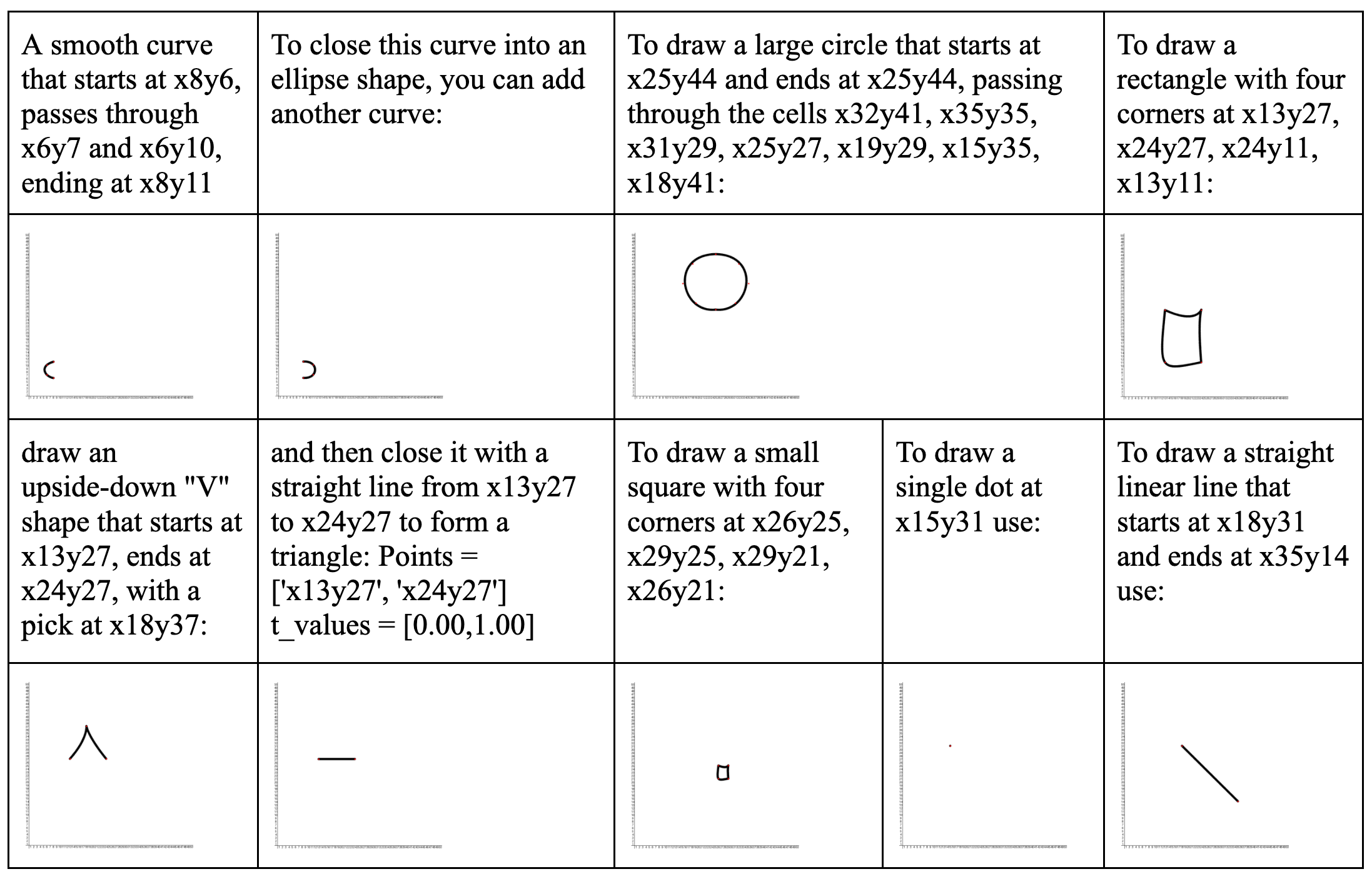}
    \vspace{-0.5cm}
    \caption{Visualization of single-stroke primitives used in the system prompt to introduce the grid and sketching language to the agent.}
    \label{fig:agent-examples1}
\end{figure}

\begin{figure}
    \centering
    \frame{\includegraphics[width=0.4\linewidth]{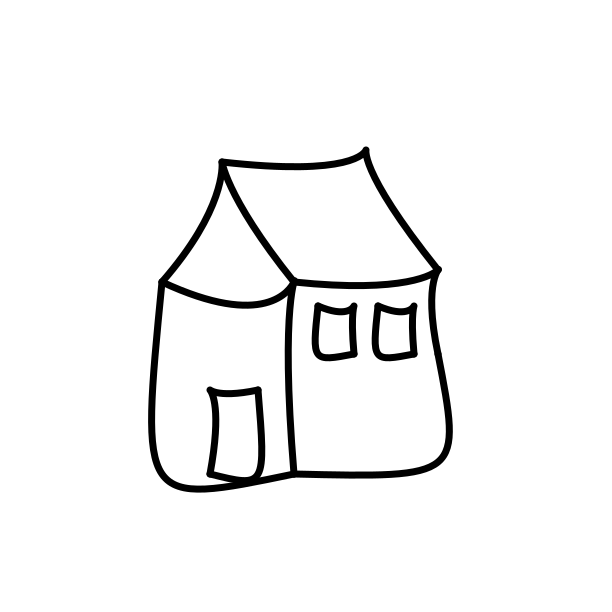}}
    \caption{Visualization of the simple sketch of a house provided as an in-context example, represented with our sketching language through the user prompt.}
    \label{fig:agent-examples2}
\end{figure}

We will publicly release the full source code, including our interactive platform. 
Our default backbone model is Claude3.5-Sonnet (version 20240620) \cite{claude}. We use the official API of Anthropic, with an average cost of \$0.05 per sketch. We employ CairoSVG \cite{cairosvg} for rendering the SVG onto the canvas. Our output sketches are also provided in SVG format to facilitate further editing if needed. SketchAgent generates a complete sketch in approximately 20 seconds, with individual strokes in collaborative mode taking about 8 seconds each.
For the CLIP zero-shot classification, we use the clip-vit-large-patch14 model from Hugging Face \cite{clip-vit-large-patch14}.
Our canvas is defined as a $50\times50$ grid with numbers labeled on the bottom and left edges. Each cell corresponds to a patch of size $12\times12$ pixels, chosen to ensure a clear display of the grid numbers along the edges. This configuration results in a $612\times612$ pixels grid, with the drawing area confined to a $600\times600$ pixel range.
All prompts used in our method are provided in \cref{fig:system-prompt,fig:user-prompt,fig:ICL-prompt}.
The examples provided to the agent in the system and user prompts are visualized in \cref{fig:agent-examples1} and \cref{fig:agent-examples2} respectively.

\begin{figure}[h]
    \centering
    \includegraphics[width=1\linewidth]{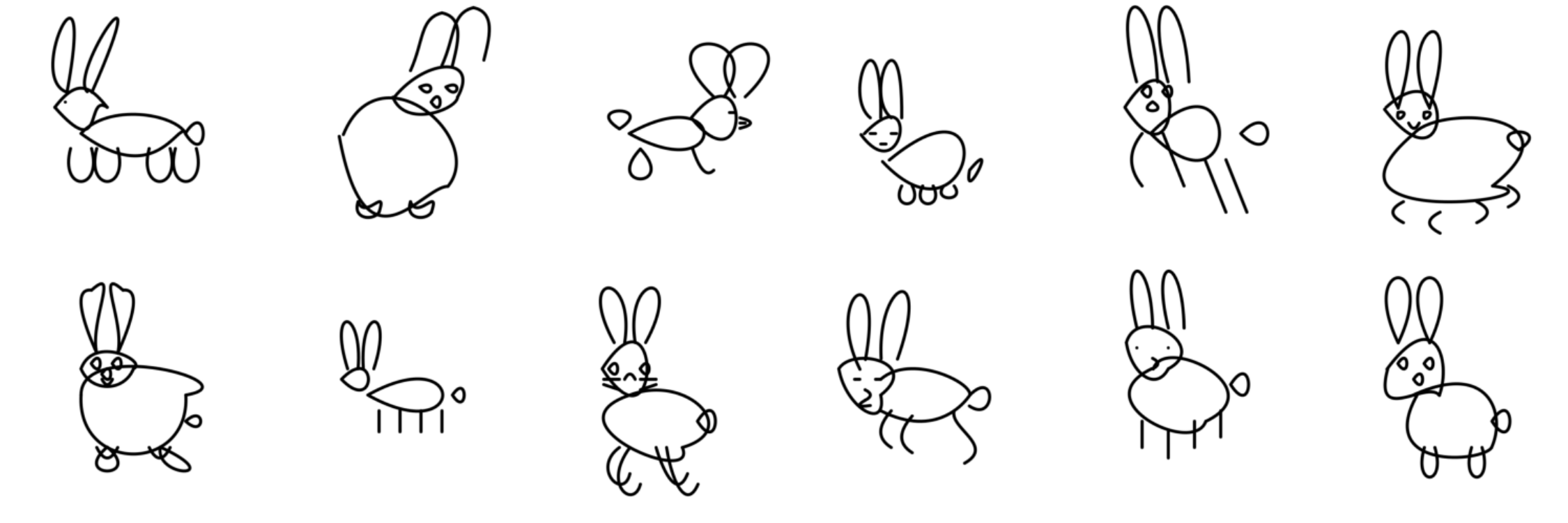}
    \caption{Sketch variability. Example of twelve different sketches produced for the concept \ap{rabbit} by SketchAgent, with the same settings.}
    \label{fig:rabbit-diverse}
\end{figure}

We use Claude3.5-Sonnet in its default settings, which results in significant variability in results, given the highly diverse nature of LLMs. For example, in \cref{fig:rabbit-diverse}, we present 12 sketches produced by our method for the concept \ap{rabbit}, demonstrating high diversity in pose, structure, and quality. To generate variations in the experiments described in Section 5.1 of the main paper (where we applied our method 10 times per category), we use the default settings of Claude3.5-Sonnet. However, during controlled experimental conditions, we reduce variability by setting the temperature to 0 and top\_k to 1, ensuring deterministic outputs. For general use, we recommend the stochastic version to encourage more varied and creative outputs.

\begin{figure}[t]
    \centering
    \includegraphics[trim={1.5cm 6cm 2.2cm 6cm},clip,width=1\linewidth]{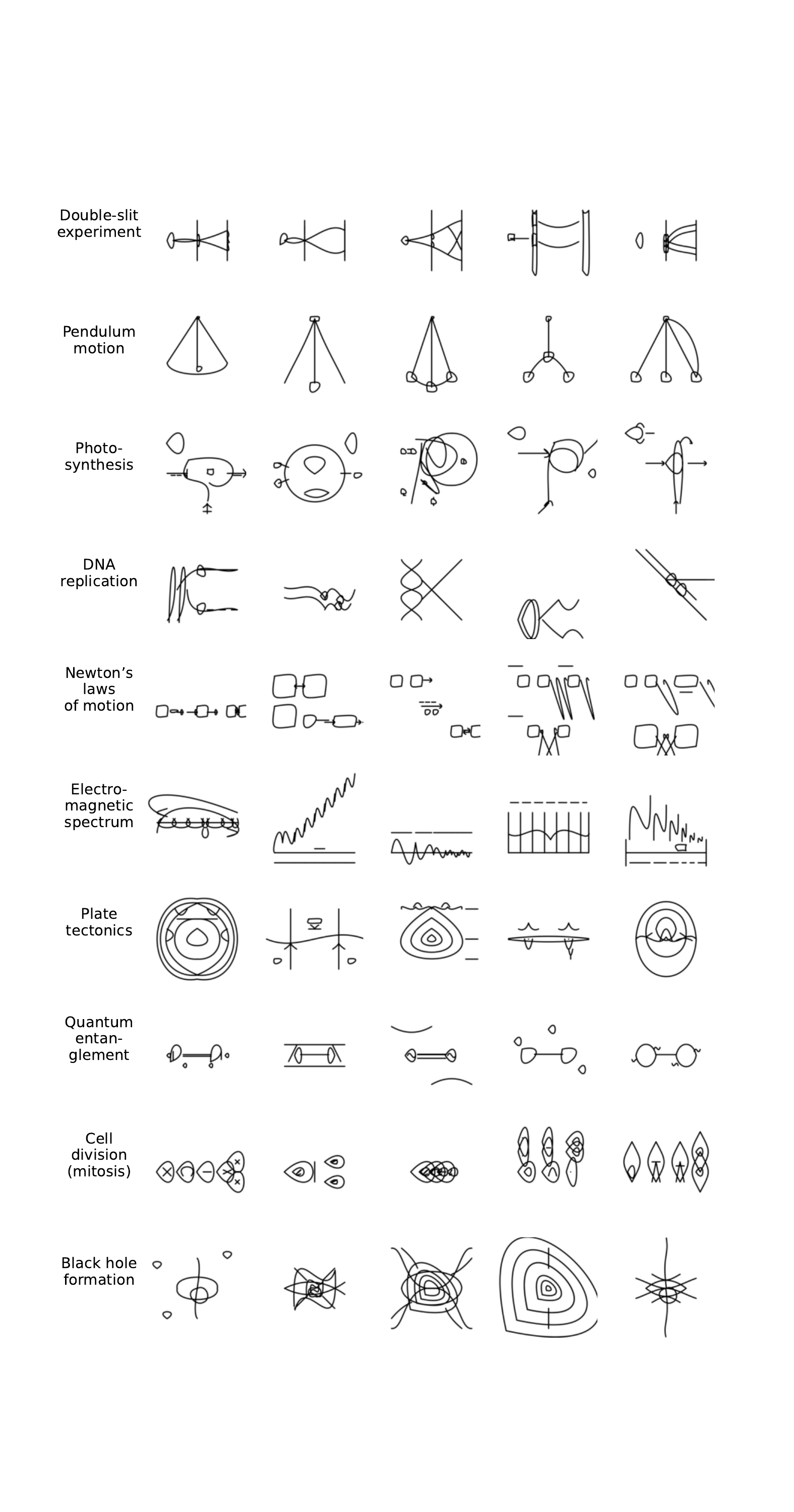}
    \caption{Randomly selected sketches of scientific concepts. Ten textual concepts were randomly selected using GPT-4o. Five sketches were generated per concept, showcasing the variability and diversity of the outputs.}
    \label{fig:unique-scientific}
\end{figure}

\begin{figure}[t]
    \centering
    \includegraphics[trim={1.5cm 6cm 2.2cm 6cm},clip,width=1\linewidth]{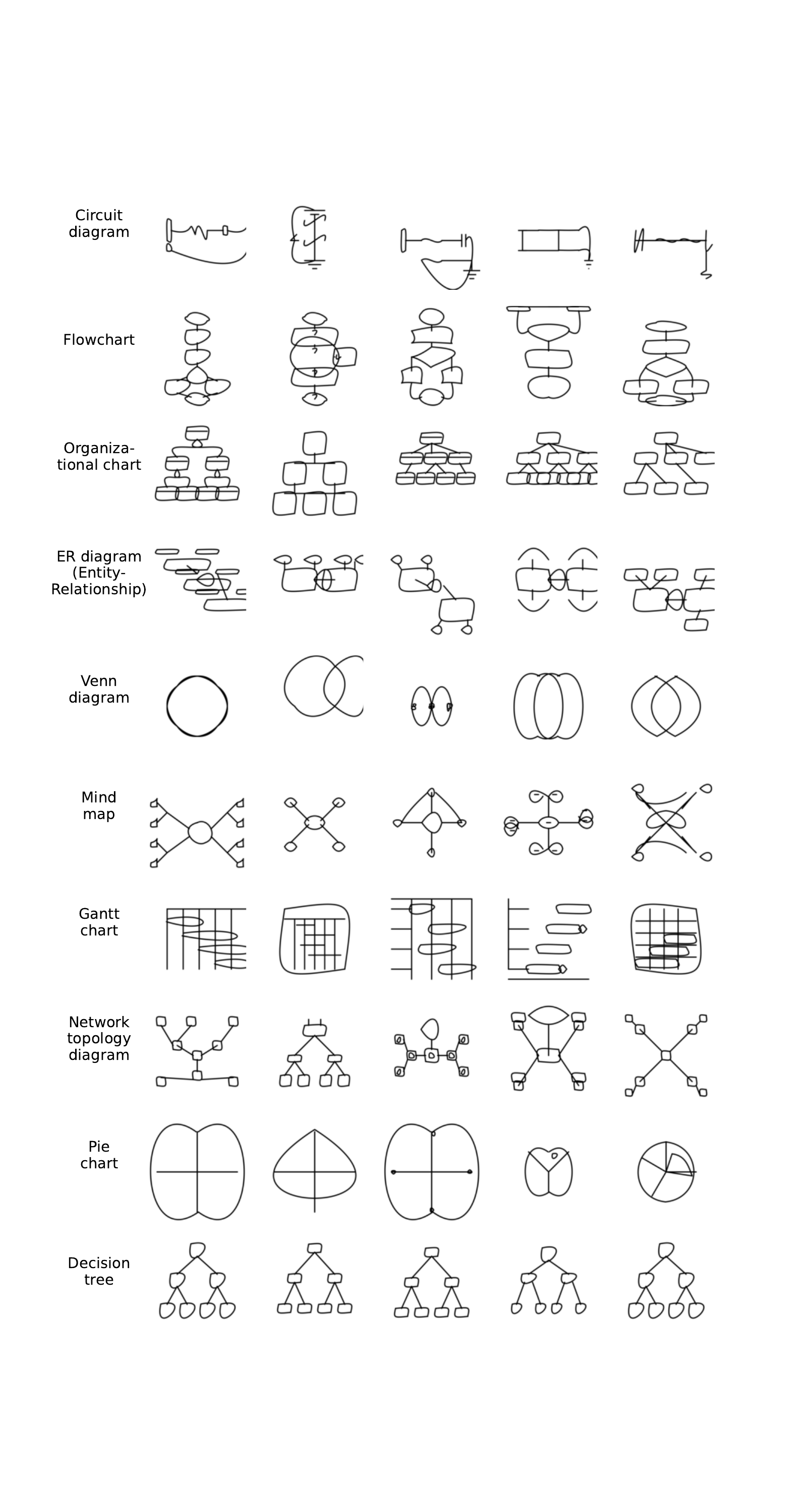}
    \caption{Randomly selected sketches of diagrams across fields. Ten textual concepts were randomly selected using GPT-4o. Five sketches were generated per concept, showcasing the variability and diversity of the outputs.}
    \label{fig:unique-diagrams}
\end{figure}

\section{More Results and Analysis} 
\begin{figure}[t]
    \centering
    \includegraphics[trim={1.5cm 6cm 2.2cm 6cm},clip,width=1\linewidth]{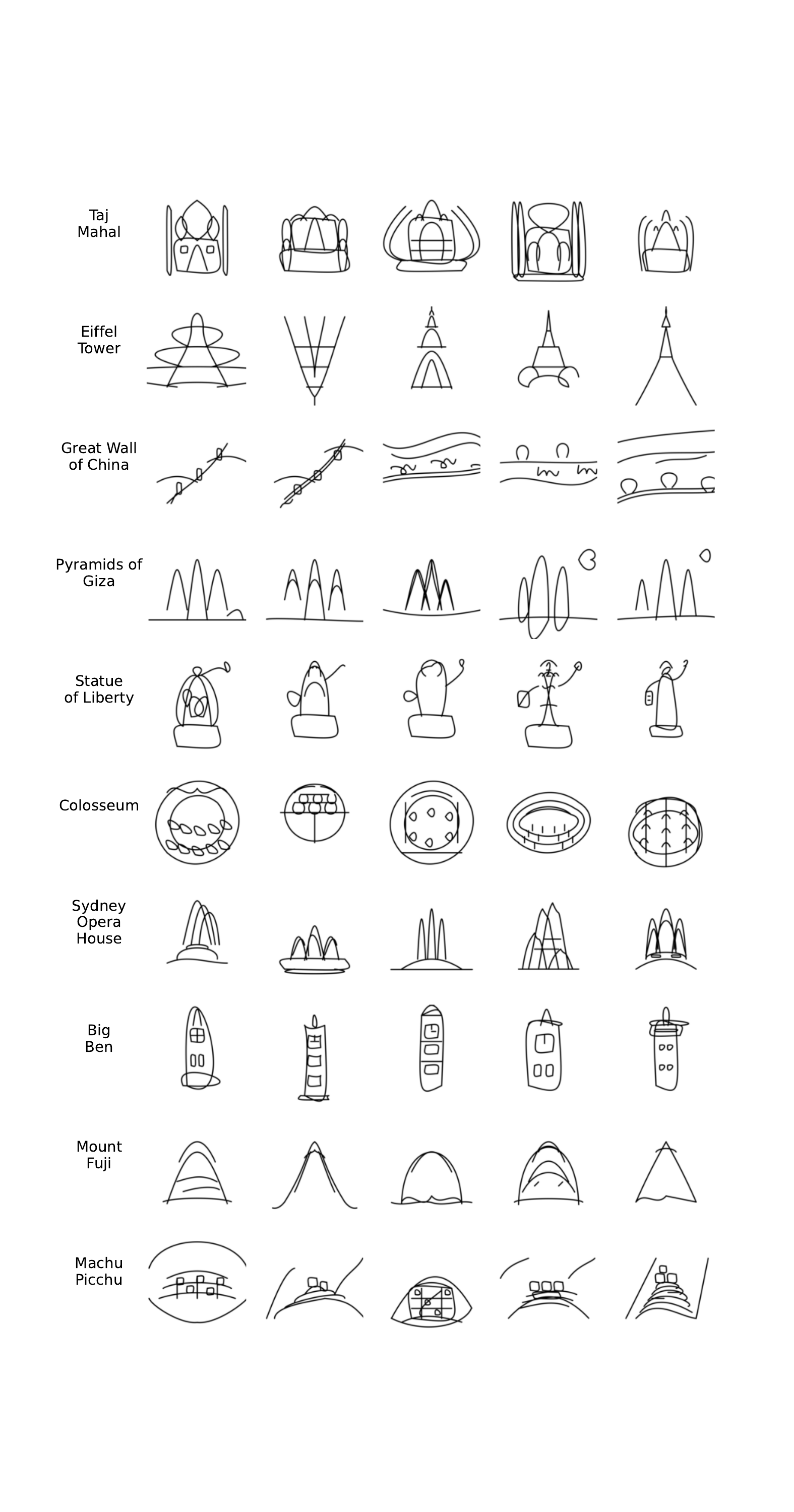}
    \caption{Randomly selected sketches of notable landmarks. Ten textual concepts were randomly selected using GPT-4o. Five sketches were generated per concept, showcasing the variability and diversity of the outputs.}
    \label{fig:unique-landmarks}
\end{figure}

As described in Section 5.1 of the main paper, SketchAgent is capable of generating sketches for a wide range of concepts that extend beyond standard categories. Here we provide additional results to support this claim. We define three unique categories that require general knowledge: Scientific Concepts, Diagrams, and Notable Landmarks, and utilize ChatGPT-4o to produce 10 random textual concepts for each category, resulting in the following random concepts: 
\begin{itemize}
    \item \textbf{Scientific Concepts:} \textit{Double-slit experiment, Pendulum motion, Photosynthesis, DNA replication, Newton’s laws of motion, Electromagnetic spectrum, Plate tectonics, Quantum entanglement, Cell division (mitosis), Black hole formation}.
    \item \textbf{Diagrams:} \textit{Circuit diagram, Flowchart, Organizational chart, ER diagram (Entity-Relationship), Venn diagram, Mind map, Gantt chart, Network topology diagram, Pie chart, Decision tree}.
    \item \textbf{Notable Landmarks:} \textit{Taj Mahal, Eiffel Tower, Great Wall of China, Pyramids of Giza, Statue of Liberty, Colosseum, Sydney Opera House, Big Ben, Mount Fuji, Machu Picchu}.
\end{itemize}
We generate five sketches for each concept (producing 50 sketches per category) by applying our method five times using its default settings. \Cref{fig:unique-scientific,fig:unique-diagrams,fig:unique-landmarks} present the results for Scientific Concepts, Diagrams, and Notable Landmarks, respectively. The resulting sketches generally depict the concepts well, demonstrating diversity in the outputs. As can be seen, our method can generate a diverse set of different types and instances per concept (see \textit{double-slit experiment, pendulum motion, Electromagnetic spectrum} in \cref{fig:unique-scientific} and \textit{Flowchart, Network typology diagram} in \cref{fig:unique-diagrams}).
Naturally, within each set, some concepts were depicted very successfully, while some outputs were less successful (e.g., Statue of Liberty, photosynthesis, pie chart).

\begin{figure}
    \centering
    \includegraphics[width=1\linewidth]{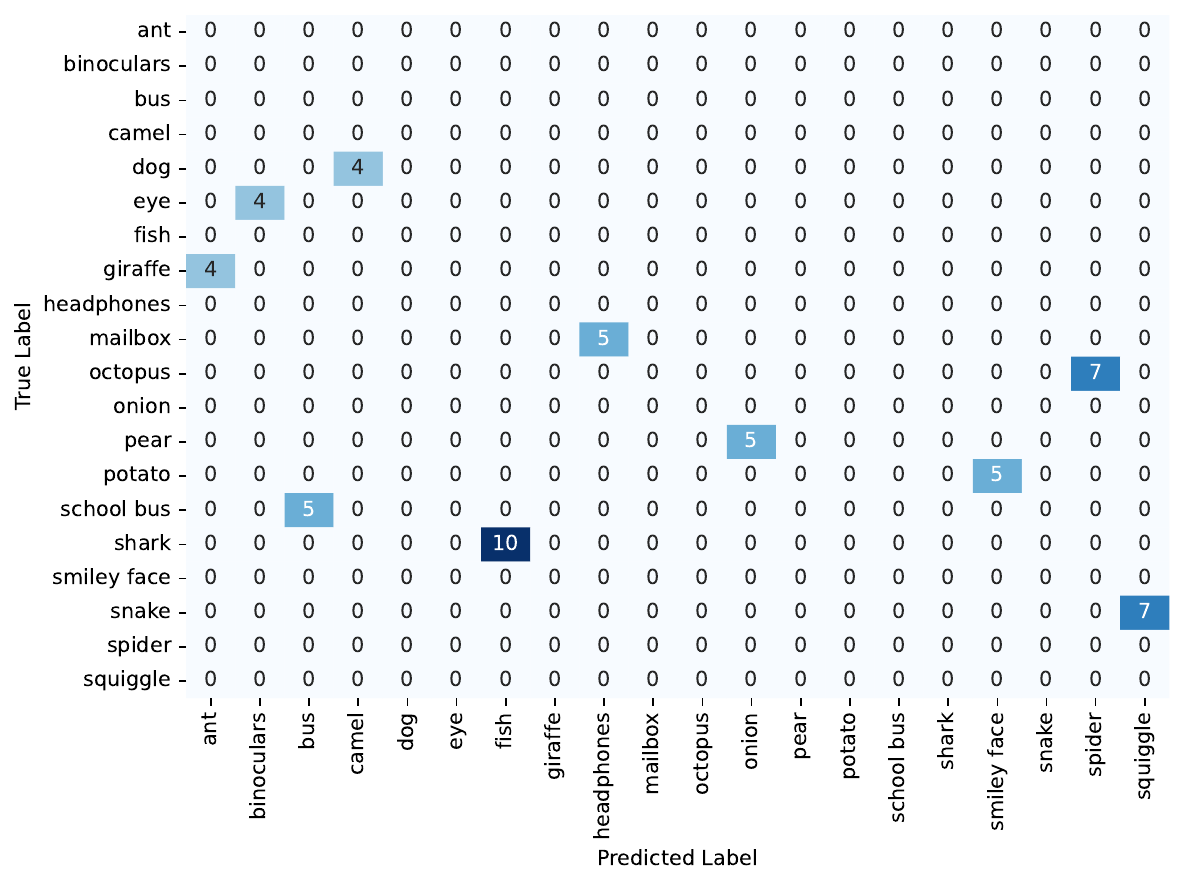}
    \caption{Confusion matrix (showing top 10 confused classes) for the set of 500 sketches generated with SketchAgent default settings (Claude3.5-Sonnet) across 50 categories}
    \label{fig:confuzed-classes}
\end{figure}

\begin{figure}
    \centering
    \includegraphics[width=1\linewidth]{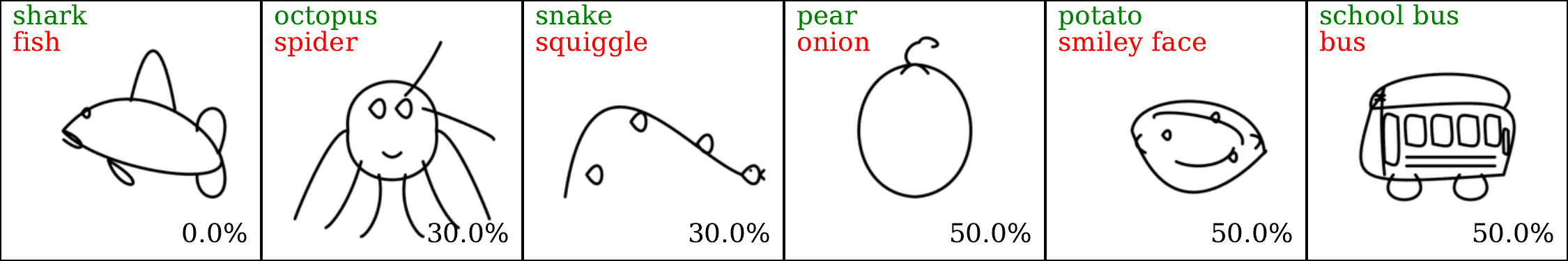}
    \caption{Visualization of sketches from the six most confused classes. The correct category is highlighted in green, while the misclassified category is highlighted in red.}
    \label{fig:top-confuzed-classes}
\end{figure}

\begin{figure}
    \centering
    \includegraphics[width=1\linewidth]{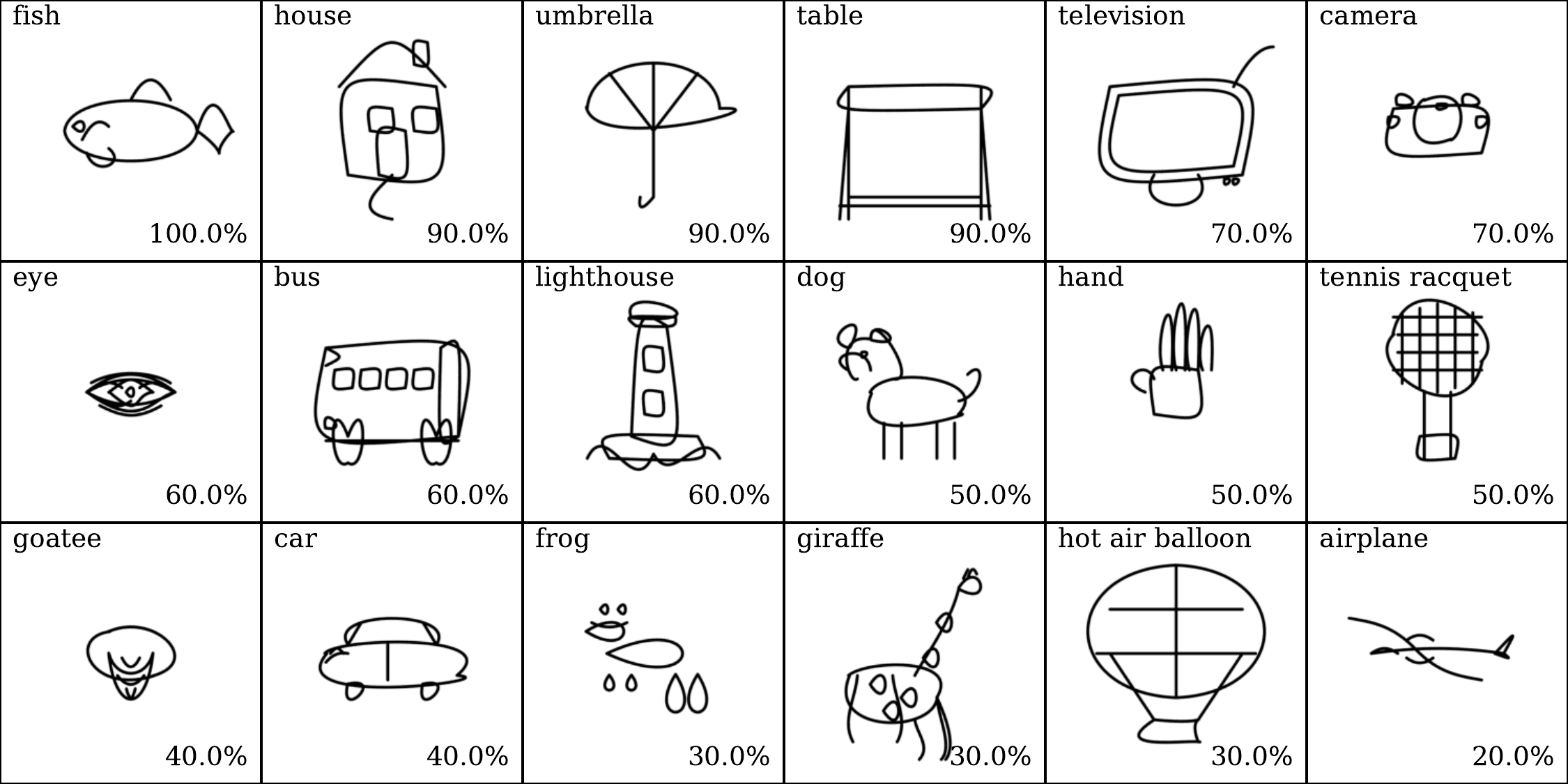}
    \caption{Visualization of the top recognized classes for the set of 500 sketches generated with our default settings (Claude3.5-Sonnet) across 50 categories.}
    \label{fig:top-recognized}
\end{figure}

\begin{figure}
    \centering
    \includegraphics[width=0.8\linewidth]{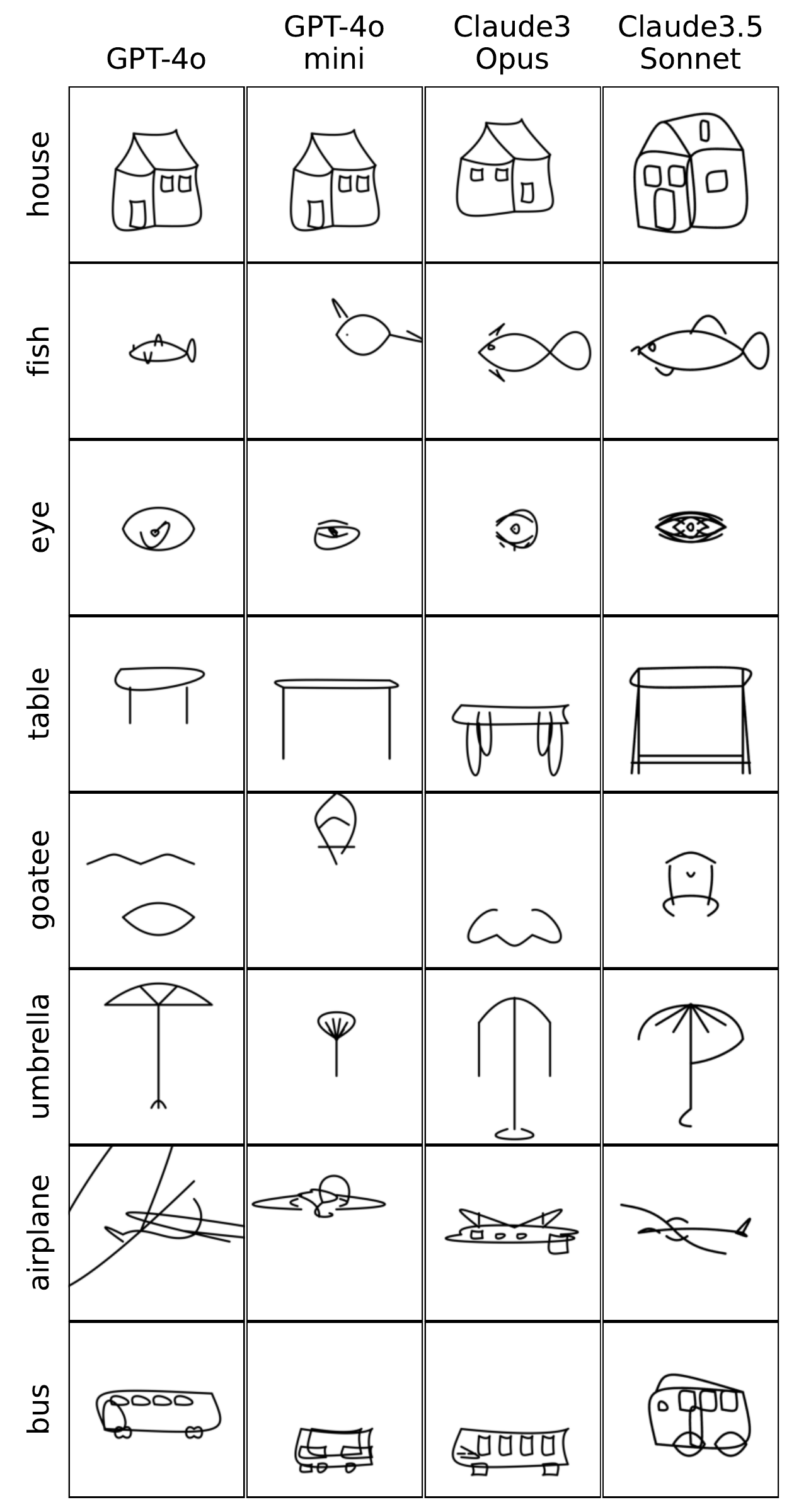}
    \vspace{-0.2cm}
    \caption{Visualization of sketches from the most recognized classes across all backbone models. The classes selected based on the two most recognizable classes in each model.}
    \label{fig:most-recognized-backbones}
\end{figure}

\begin{figure}
    \centering
    \includegraphics[width=0.8\linewidth]{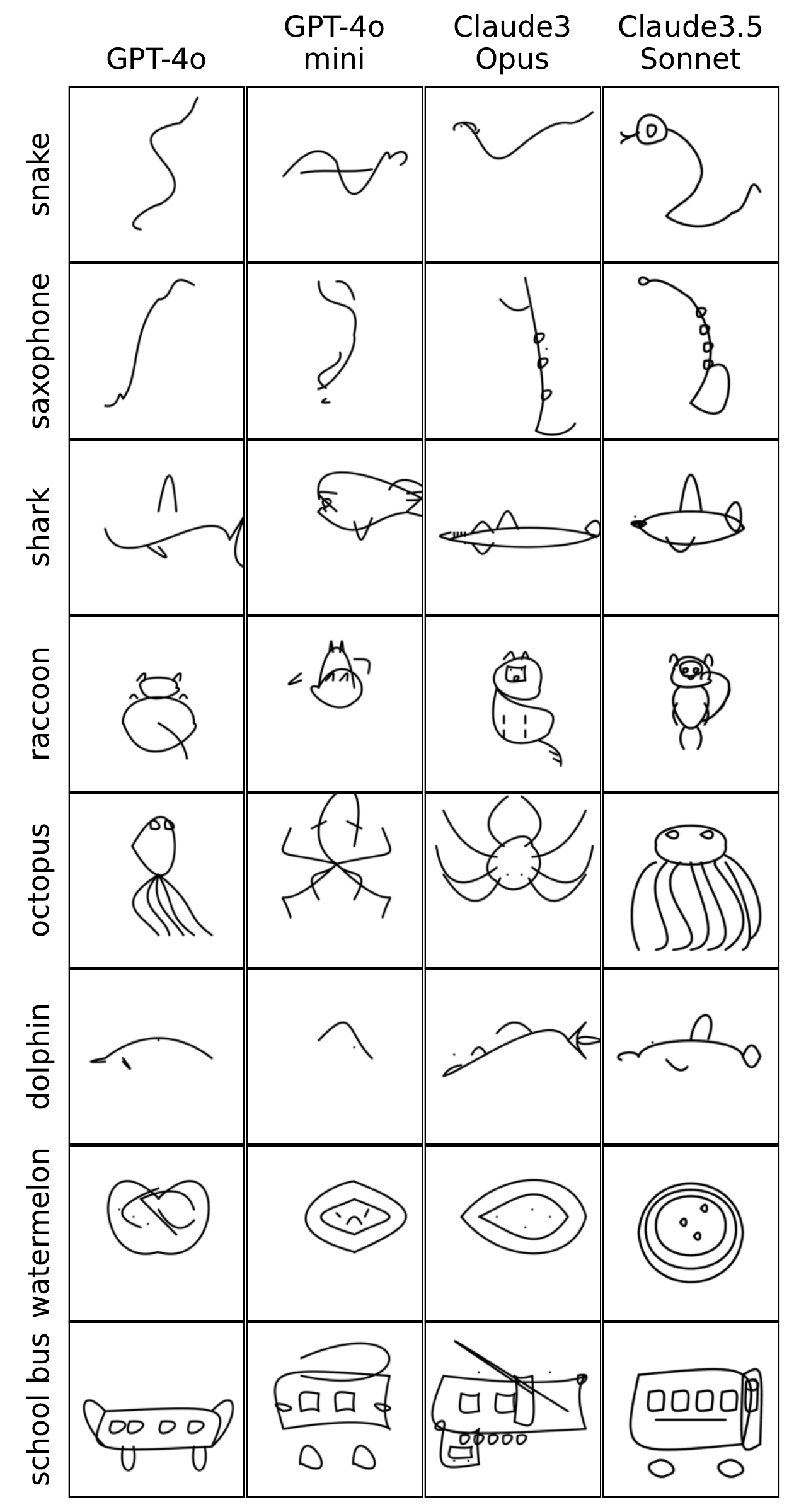}
    \caption{Visualization of sketches from the least recognized classes across all backbone models. The classes selected based on the two least recognizable classes in each model.}
    \label{fig:most-confused-backbones}
\end{figure}

\subsection{Quantitative Text-Conditioned Analysis} In Section 5.1 of the main paper, we presented a quantitative analysis of text-conditioned sketch generation across 50 selected categories from the QuickDraw dataset \cite{quickDrawData}. Here, we provide additional details, visual examples, and further analysis of the experiment.
We begin by providing further analysis of the CLIP classification rates of our default settings (Claude3.5-Sonnet) to explore recognition patterns.
\Cref{fig:confuzed-classes} shows the confusion matrix (top 10 confused categories out of 50) for our set of 500 sketches. The most commonly confused classes are: \ap{shark}, which was often misclassified as a \ap{fish}, \ap{octopus}, which was frequently identified as a \ap{spider}; and \ap{snake}, which was misclassified as a \ap{squiggle}. These confused classes often fall within highly related classes (such as a fish and a shark, or a school bus and a bus), suggesting that our method struggles with emphasizing distinctive features, likely due to its inherently abstract style. In \cref{fig:top-confuzed-classes}, we visualize sketches from the six most confused classes with the correct class shown in green and the misclassified class shown in red.
\Cref{fig:top-recognized} visualize the 10 top recognized classes. The class \ap{fish} was correctly identified across all seeds, followed by \ap{house}, \ap{umbrella}, and \ap{table}, which were correctly recognized in $90\%$ of trials (9 out of 10). Recognition rates for other classes ranged from $70\%$ to $20\%$.
In Section 5.1 of the main paper, we compared the performance of different multimodal LLMs (GPT-4o-mini, GPT-4o, and Claude3-Opus) using our default prompts and settings. \Cref{fig:most-recognized-backbones} visualizes the eight most recognized classes across all backbone models. For this analysis, we select the top two recognized categories from each model and display the sketches with the highest classification probability for each. Note that some categories were at the top two of multiple models (such as house, fish, and eye), in that case, we select the next top recognized category. The chosen top two categories for each model are: GPT-4o: house and eye, GPT-4o-mini: table and goatee, Claude3-Opus: fish and airplane, Claude3.5-Sonnet: umbrella and bus.
Similarly, \Cref{fig:most-confused-backbones} highlights the least recognized categories, chosen using the same selection criteria. The chosen worst two categories for each model are: GPT-4o: snake and school bus, GPT-4o-mini: saxophone and raccoon, Claude3-Opus: octopus and dolphin, Claude3.5-Sonnet: shark and watermelon.
Note that snake, octopus, and shark, were all confused under at least three of the four backbones.
The visualizations align well with the quantitative results presented in Table 1 of the main paper. Among the Anthropic models, Claude3.5-Sonnet produces better sketches than Claude3-Opus, and among the GPT models, GPT-4o outperforms GPT-4o-mini. Overall, the two best-performing backbone models are Claude3.5-Sonnet and GPT-4o.
Interestingly, the sketching style differs between GPT-4o and Claude3.5-Sonnet. Although Claude3.5-Sonnet (our default backbone model) seems to yield the best results, this may be due to the fact that our method was primarily developed using this model. Consequently, the prompts we use were optimized for Claude3.5-Sonnet, and improved results for other models might be achievable with additional prompt engineering. We leave this exploration for future work.

\paragraph{SketchAgent using an open-source model} While open-source models currently lag behind commercial closed-source models, they are rapidly advancing in size and capability, showing significant potential for facilitating sketch generation.

\begin{figure}
    \centering
    \includegraphics[width=1\linewidth]{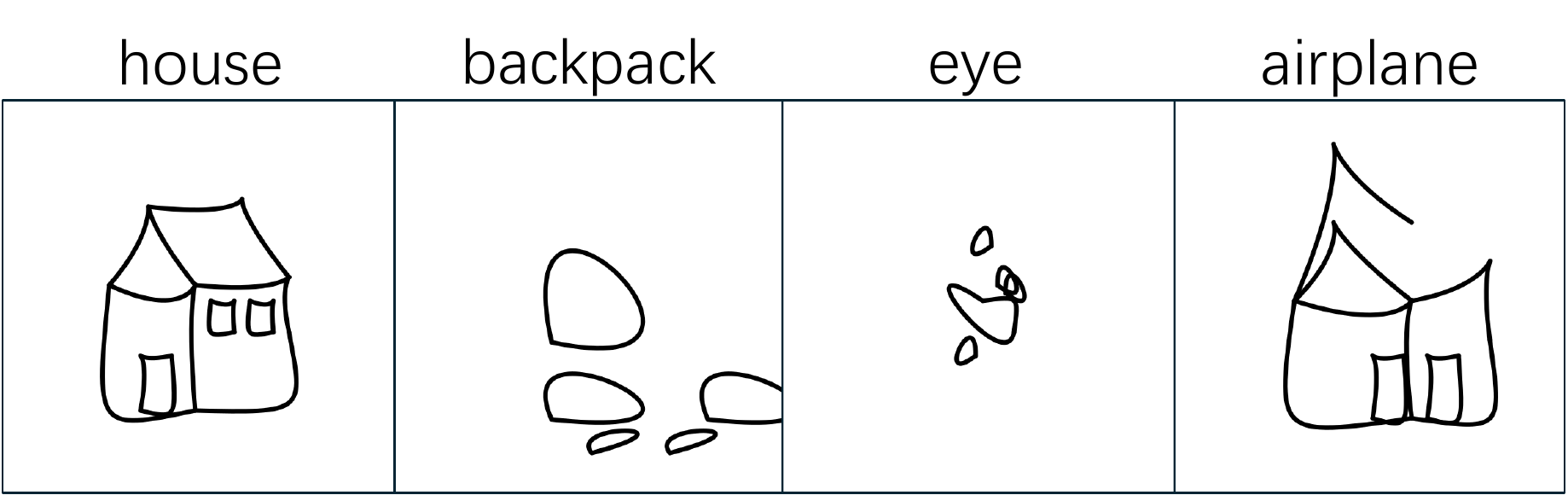}
    \caption{Sketches generated using Llama-3.2-11B-Vision as our backbone models. The model frequently replicates the in-context example of a house provided in the user prompt.}
    \label{fig:llama_failures}
\end{figure}

We begin by experimenting with Llama-3.2-11B-Vision~\cite{dubey2024llama3}, a multimodal large language model developed by Meta AI, as SketchAgent's backbone model. When used with our default prompts and framework, the model fails to generate meaningful sketches, frequently replicating the in-context example of a house provided in the user prompt (examples are shown in \cref{fig:llama_failures}).

\begin{figure}
    \centering
    \includegraphics[width=1\linewidth]{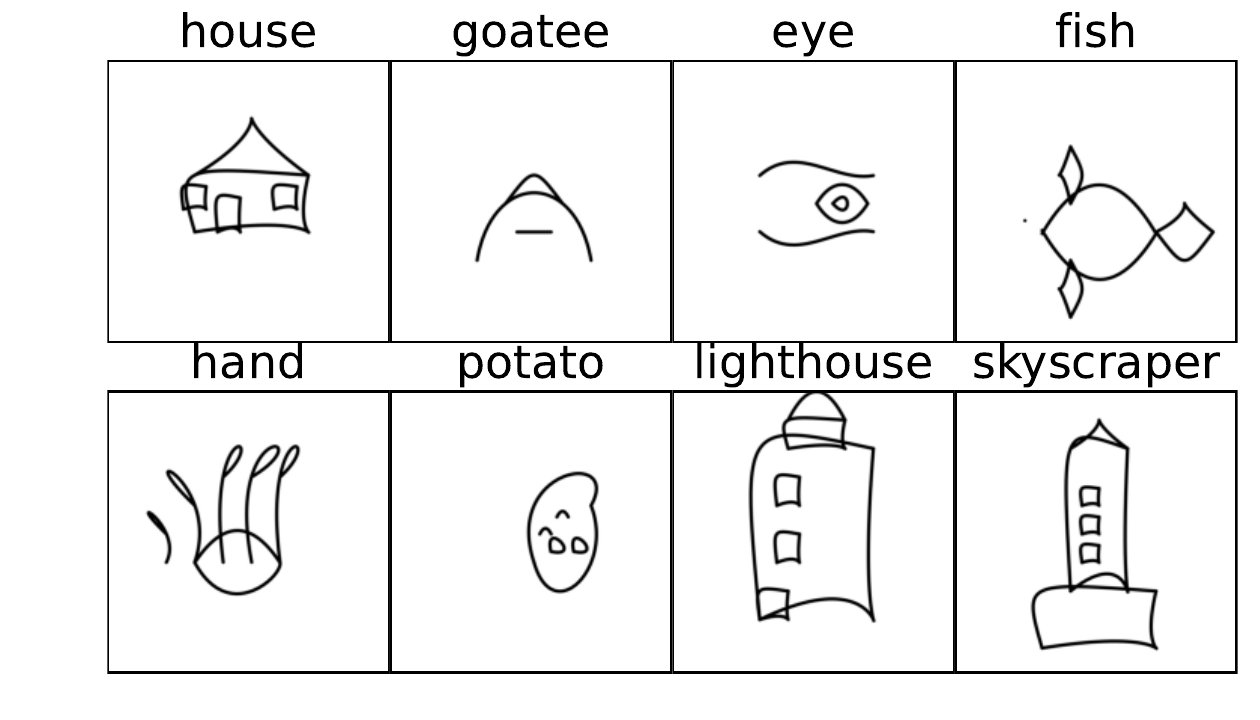}
    \caption{Visualization of the eight top recognized classes for the set of 500 sketches generated with Llama-3.1-405B-Instruct as our backbone model.}
    \label{fig:llama_best_8}
\end{figure}

\begin{figure}
    \centering
    \includegraphics[width=1\linewidth]{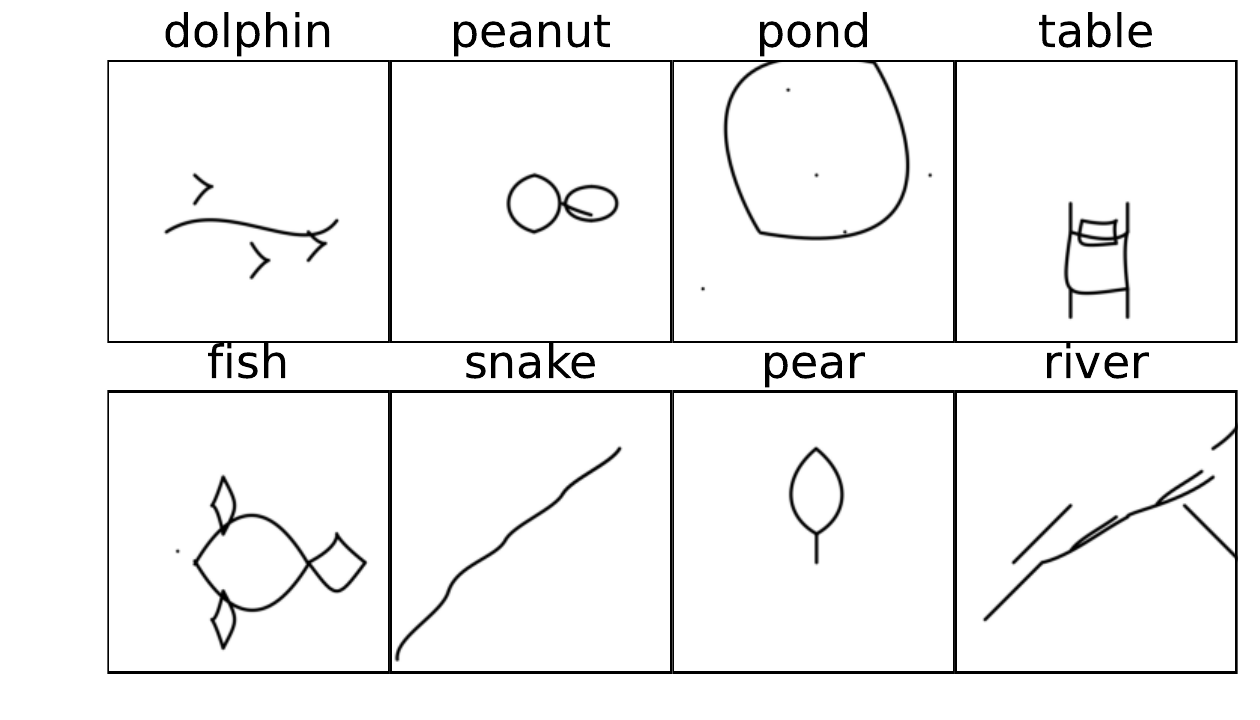}
    \caption{Visualization of sketches from the eight least recognized classes for the set of 500 sketches generated with Llama-3.1-405B-Instruct as our backbone model.}
    \label{fig:llama_worst_8}
\end{figure}

We therefore turn into exploring a larger available open-source model, Llama-3.1-405B-Instruct~\cite{dubey2024llama3}.
This model resulted in better sketches that manage to generalize well beyond the in-context example. We generated 500 random sketches and computed their classification rates using CLIP, as described in Section 5.1 of the main paper. The results yielded lower scores compared to commercial models, with an average Top-1 recognition accuracy of $0.052 \pm 0.03$ and a Top-5 recognition accuracy of $0.1 \pm 0.03$.
Visualizations of the top eight correctly classified classes are shown in \cref{fig:llama_best_8}, and the top eight most confused classes are presented in \cref{fig:llama_worst_8}. Despite the lower recognition rates, the generated sketches are reasonable and visually coherent, showing promise as open-source models continue to improve. This experiment demonstrates the potential for SketchAgent to be implemented using publicly available models. While its performance does not match that of our default backbone, SketchAgent can still function effectively with open-source models, albeit with a slight compromise in performance.

\begin{figure}[h]
    \centering
    \setlength{\tabcolsep}{3pt}
    \small{
    \begin{tabular}{c c c c c}
    \toprule
    & 
    \begin{tabular}[c]{@{}c@{}}GPT-4o\\ \end{tabular} & 
    \begin{tabular}[c]{@{}c@{}}GPT-4o\\-mini\end{tabular} & 
    \begin{tabular}[c]{@{}c@{}}Claude3  \\ Opus\end{tabular}& 
    \begin{tabular}[c]{@{}c@{}}Claude3.5 \\ -Sonnet\end{tabular} \\
    \hline
    \raisebox{2\height}{\begin{tabular}[c]{@{}c@{}}SVG \\ \ap{giraffe}\end{tabular}} & 
    \includegraphics[trim={0cm 0cm 0cm 2cm},clip,width=0.15\linewidth]{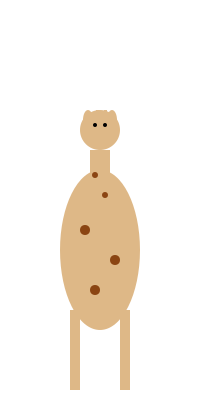} &
    \includegraphics[trim={0cm 0cm 0cm 3cm},clip,width=0.15\linewidth]{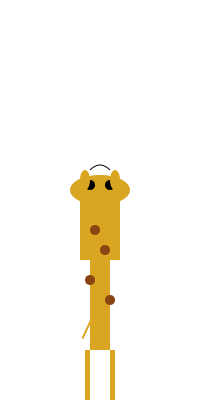} &
    \includegraphics[trim={0cm 0cm 0cm 3cm},clip,width=0.15\linewidth]{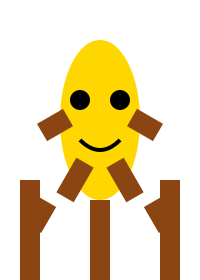} &
    \includegraphics[trim={0cm 0cm 0cm 1cm},clip,width=0.15\linewidth]{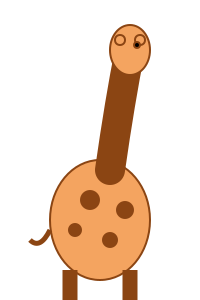} \\
    
    \hline
    \raisebox{1.8\height}{\begin{tabular}[c]{@{}c@{}}Sketch-like \\ SVG \\ \ap{giraffe}\end{tabular}} & 
    \includegraphics[width=0.15\linewidth]{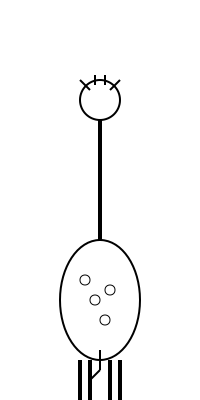} &
    \includegraphics[width=0.15\linewidth]{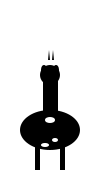} &
    \includegraphics[width=0.15\linewidth]{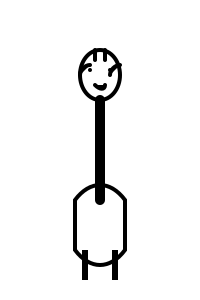} &
    \includegraphics[width=0.15\linewidth]{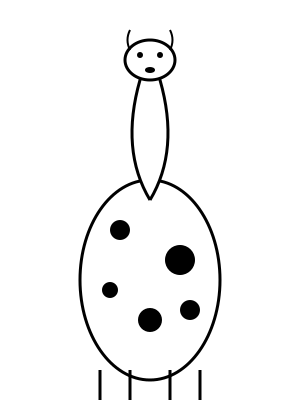} \\
    
    \hline
    \raisebox{2\height}{\begin{tabular}[c]{@{}c@{}}SVG \\ \ap{lighthouse}\end{tabular}} & 
    \includegraphics[width=0.15\linewidth]{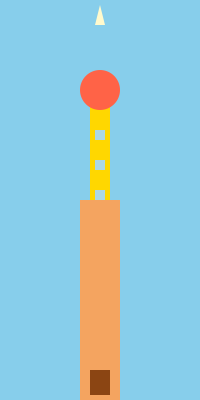} &
    \includegraphics[width=0.15\linewidth]{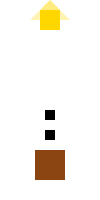} &
    \includegraphics[width=0.15\linewidth]{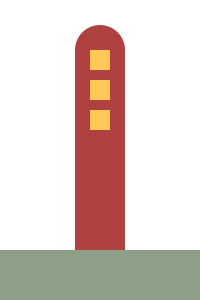} &
    \includegraphics[width=0.15\linewidth]{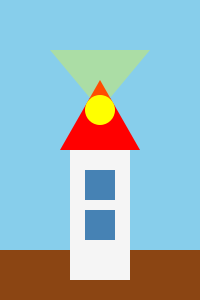} \\
    
    \hline
    \raisebox{2\height}{\begin{tabular}[c]{@{}c@{}}Sketch-like \\ SVG \\ \ap{lighthouse}\end{tabular}} & 
    \includegraphics[width=0.15\linewidth]{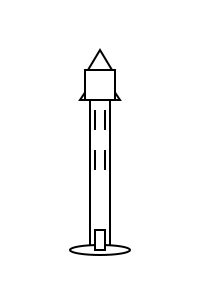} &
    \includegraphics[width=0.15\linewidth]{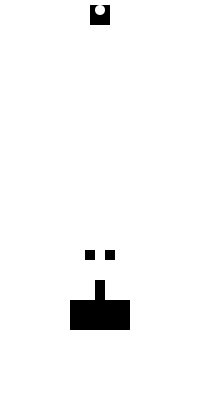} &
    \includegraphics[width=0.15\linewidth]{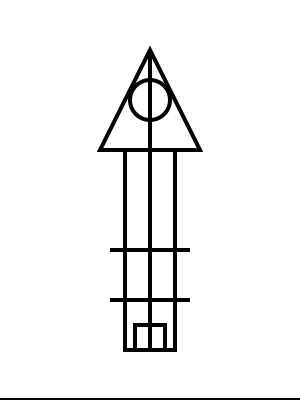} &
    \includegraphics[width=0.15\linewidth]{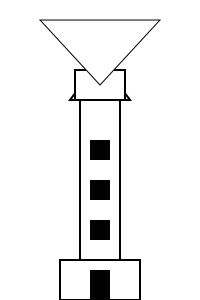} \\
    \bottomrule
    \end{tabular}
    }
    \caption{Direct prompting for SVG generation across different backbone models. The SVGs generated by GPT-4o and Claude3.5-Sonnet appear more expressive and visually appealing compared to those produced by GPT-4o-mini and Claude3-Opus, aligning well with the performance differences observed in sketch generation.}
    \label{fig:direct-prompt-extend}
\end{figure}

\paragraph{Direct Prompting Analysis} In Section 5.1 of the main paper, we compared our method to directly prompting Claude3.5-Sonnet for generating SVGs with a sketch-like appearance. 
In \cref{fig:direct-prompt-extend}, we extend this analysis by visualizing the results of direct prompting with the other backbone models used in the quantitative experiment.
This demonstrates how different models respond to direct SVG generation prompts. We present examples for the concepts \ap{giraffe} and \ap{lighthouse}, using the following SVG generation prompt: \textit{\ap{Write an SVG string of a $<$concept$>$.}}. For sketch-like SVGs, we used the same prompt as in the main paper (\textit{\ap{Write an SVG string that draws a sketch of a $<$concept$>$. Use only black and white colors}}).
As shown, the outputs across all methods often feature uniform and precise geometric shapes (e.g., ellipses, triangles), which diverge from the natural variability and expressiveness characteristic of hand-drawn sketches.
Interestingly, the SVGs generated by GPT-4o and Claude3.5-Sonnet appear more expressive and visually appealing compared to those produced by GPT-4o-mini and Claude3-Opus, aligning well with the performance differences observed in sketch generation.

\begin{figure}[h]
    \centering
\includegraphics[width=01\linewidth]{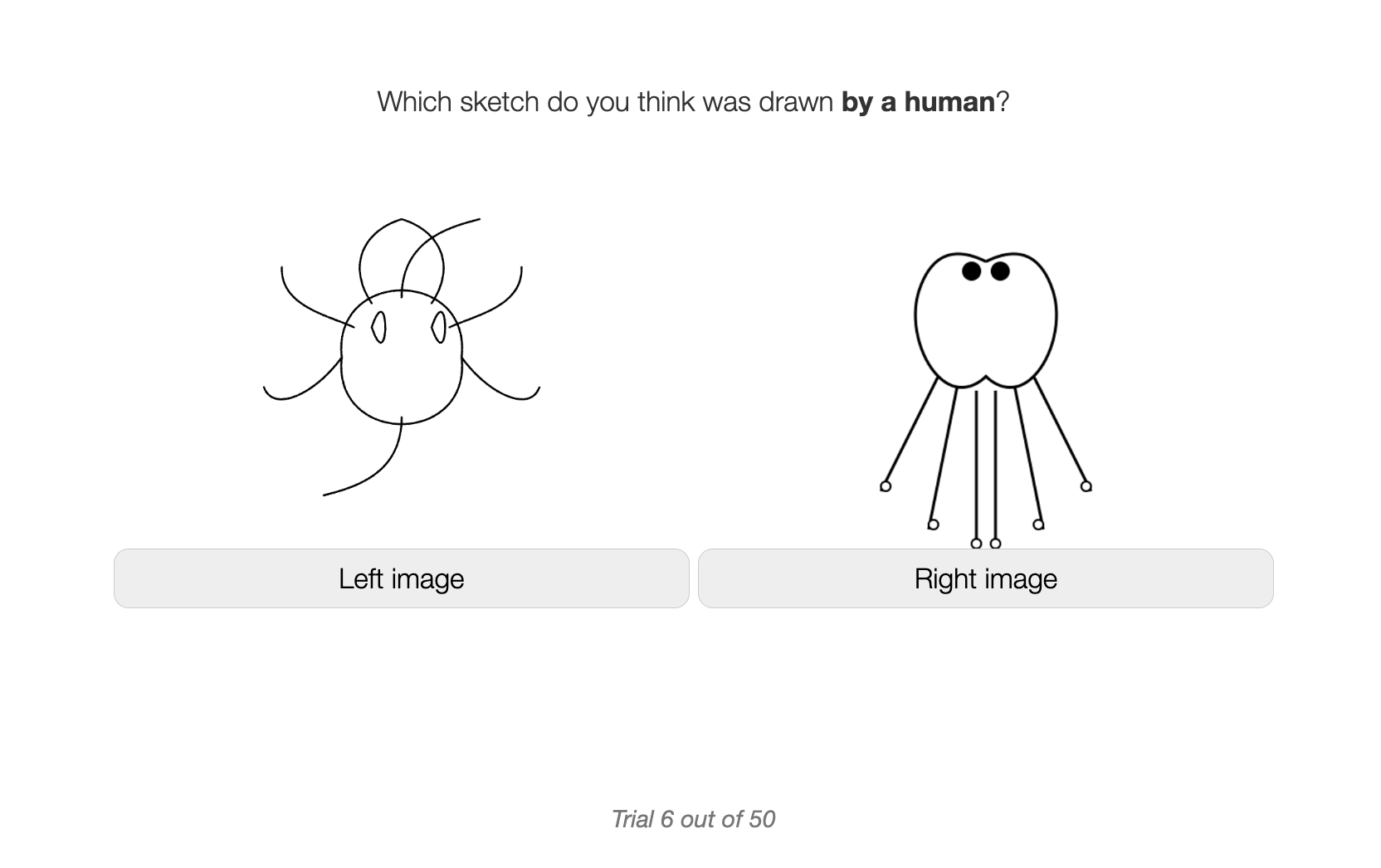}
    \caption{An example of our 2AFC session.}
    \label{fig:2afc-interface}
\end{figure}

\begin{figure}[h]
    \centering
    \includegraphics[width=01\linewidth]{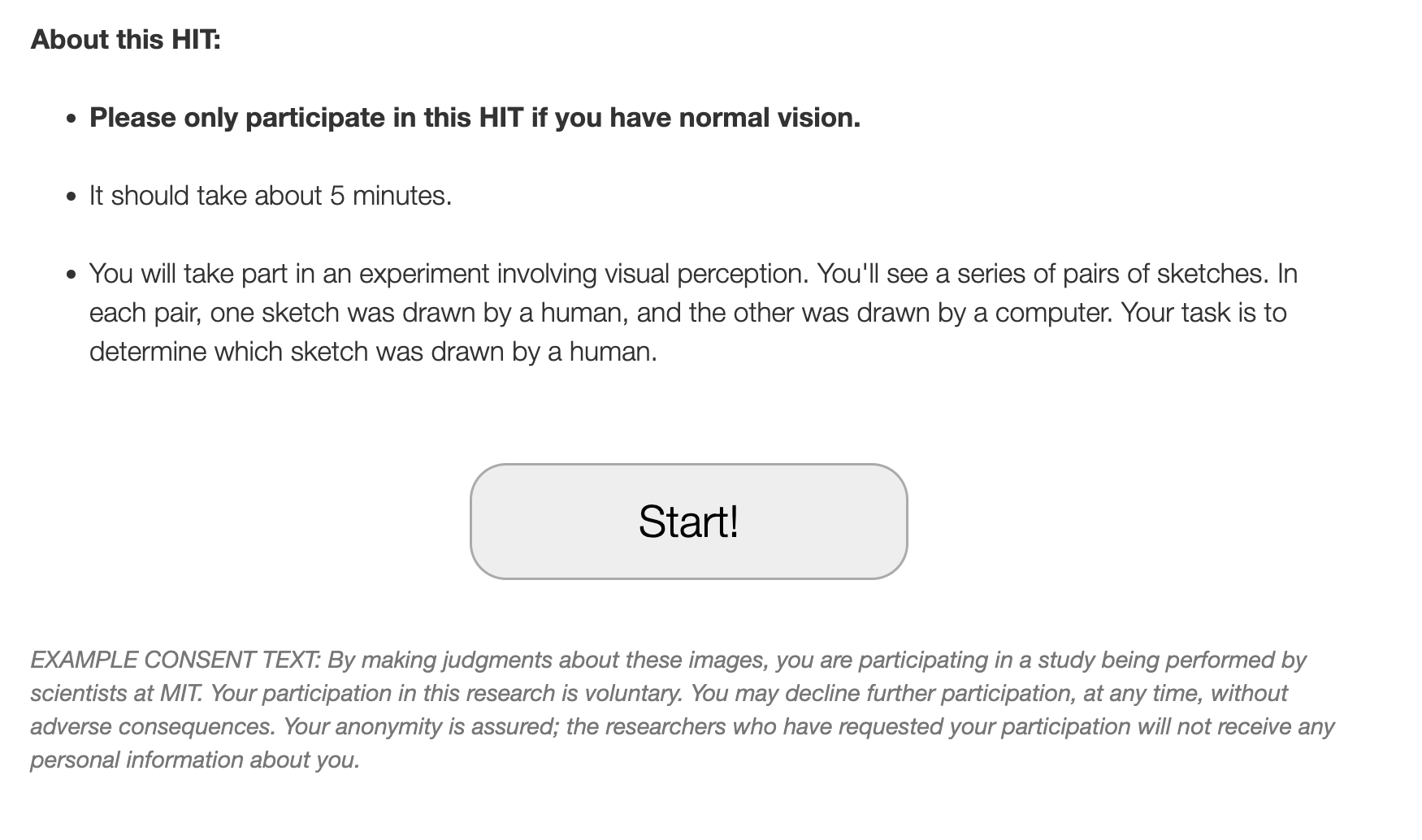}
    \caption{2AFC instructions to users.}
    \label{fig:2afc-instructions}
\end{figure}

\paragraph{2AFC experiment} In section 5.1 of the main paper, we also presented a 2AFC experiment to evaluate how ``human-like'' our agent’s sketches appear compared to sketch-like SVGs generated with direct prompting and human sketches from the QuickDraw dataset. We utilize 50 sketches from 50 classes per method. We recruited a total of 150 workers through Amazon Mechanical Turk, each participating in 50 test sessions, as presented in Fig. \ref{fig:2afc-interface}.
Before starting the test, workers were presented with instructions (Fig.~\ref{fig:2afc-instructions}). We filtered participants with a Mturk approval rate of 99.9\% or higher and with a record of more than 1,000 surveys. Workers were paid \$0.5 for completing the full test.


\clearpage
\newpage

\begin{figure}[t]
    \centering
    \includegraphics[width=1\linewidth]{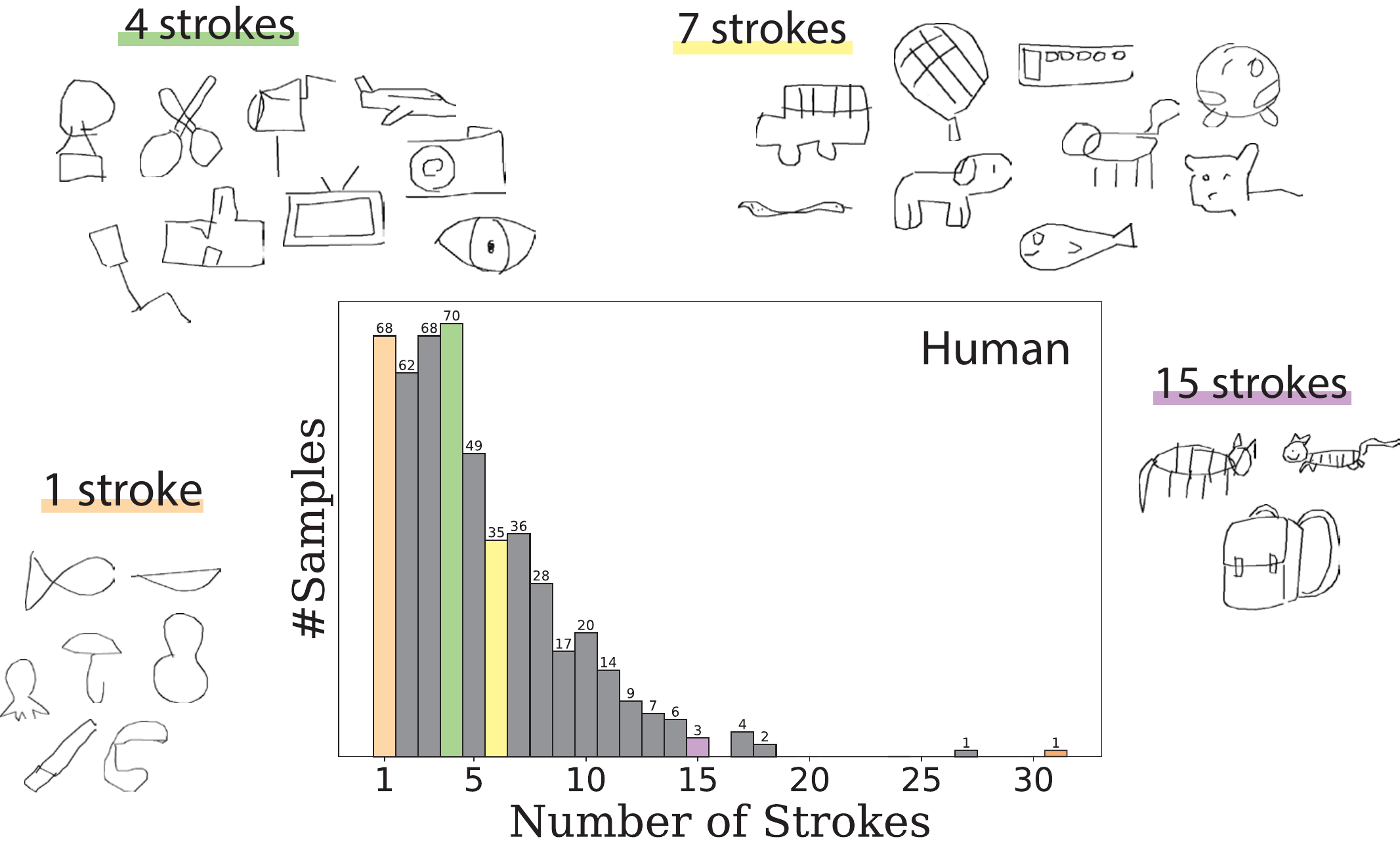}
    \includegraphics[width=0.8\linewidth]{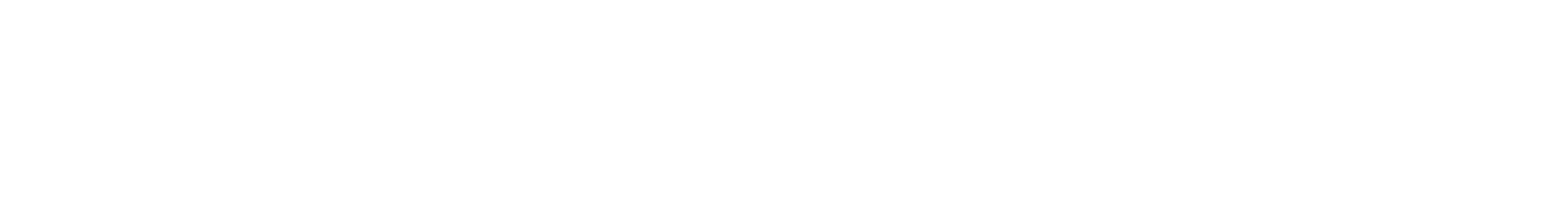}
    \includegraphics[width=1\linewidth]{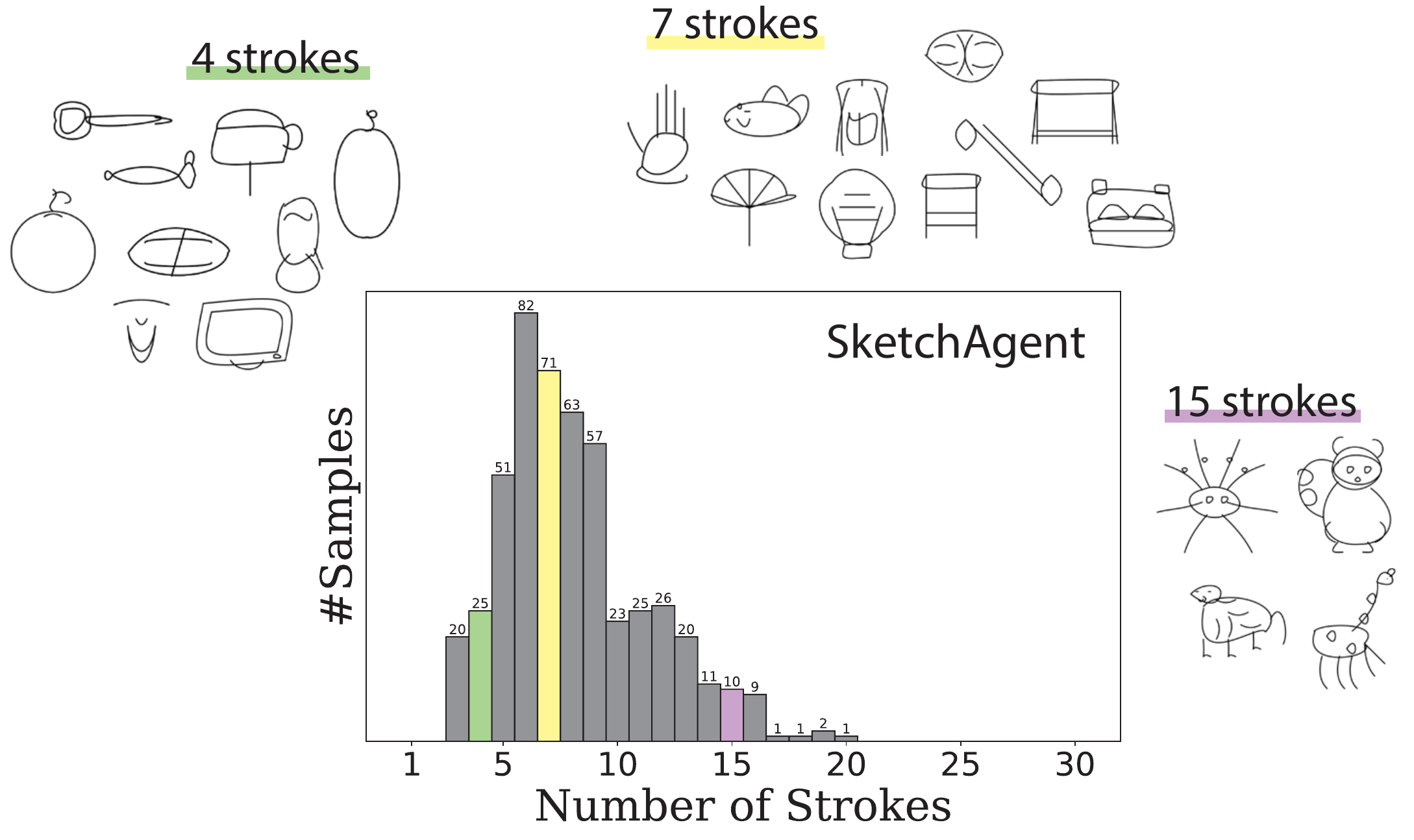}
    \caption{Distribution of human sketches \cite{quickDrawData} (top) and SketchAgent's sketches (bottom) based on the number of strokes per sketch. Representative examples are shown for sketches drawn with 1, 4, 7, and 15 strokes. Notably, in the QuickDraw dataset, single-stroke sketches often consist of a single long continuous line.}
    \label{fig:stroke-distribution}
\end{figure}
\subsection{Sequential sketching}
\begin{figure}[h]
    \centering
    \setlength{\tabcolsep}{1pt}
    \renewcommand{\arraystretch}{0.5}
    \small{
    \begin{tabular}{l c}
    \raisebox{0.6\height}{\rotatebox[origin=t]{90}{\footnotesize{Human}}} & 
    \includegraphics[trim={2.55cm 0.5cm 2cm 0.6cm},clip,width=0.6\linewidth]{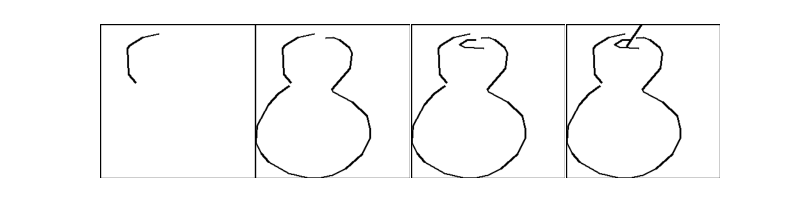} \\
    \raisebox{0.75\height}{\rotatebox[origin=t]{90}{\footnotesize{Ours}}} & 
    \includegraphics[width=0.6\linewidth]{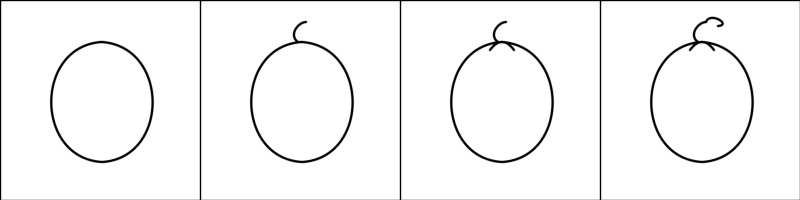} \\
    \hline
    \raisebox{0.6\height}{\rotatebox[origin=t]{90}{\footnotesize{Human}}} & 
    \includegraphics[trim={2.55cm 0.5cm 2cm 0.6cm},clip,width=0.6\linewidth]{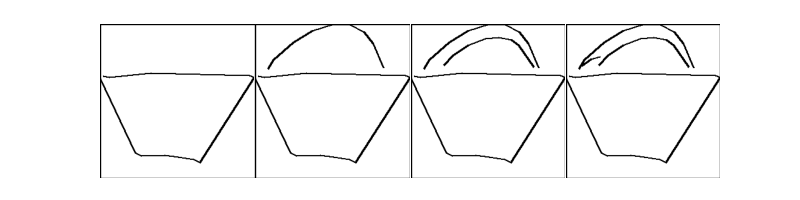} \\
    \raisebox{0.75\height}{\rotatebox[origin=t]{90}{\footnotesize{Ours}}} & 
    \includegraphics[width=0.6\linewidth]{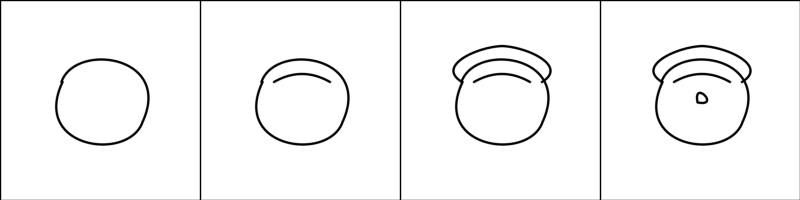} \\
    \hline
    \raisebox{0.6\height}{\rotatebox[origin=t]{90}{\footnotesize{Human}}} & 
    \includegraphics[trim={2.55cm 0.5cm 2cm 0.6cm},clip,width=0.6\linewidth]{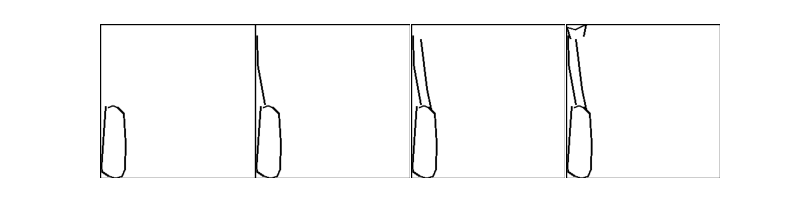} \\
    \raisebox{0.75\height}{\rotatebox[origin=t]{90}{\footnotesize{Ours}}} & 
    \includegraphics[width=0.6\linewidth]{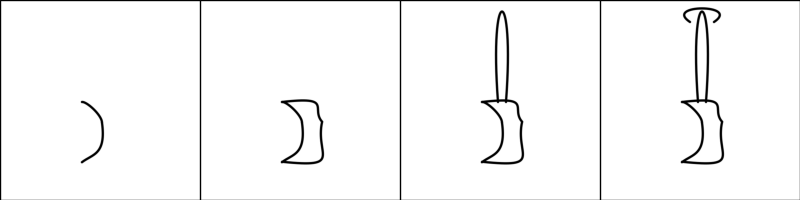} \\
    \end{tabular}
    }
    \caption{Sequential four-stroke sketches of a pear, purse, and screwdriver, created by humans \cite{quickDrawData} and by SketchAgent.}
    \label{fig:4-strokes-set}
\end{figure}

\begin{figure}[h]
    \centering
    \setlength{\tabcolsep}{1pt}
    \renewcommand{\arraystretch}{0.5}
    \small{
    \begin{tabular}{l c}
    \raisebox{0.6\height}{\rotatebox[origin=t]{90}{\footnotesize{Human}}} & 
    \includegraphics[trim={3.2cm 0.5cm 2.5cm 0.6cm},clip,width=0.75\linewidth]{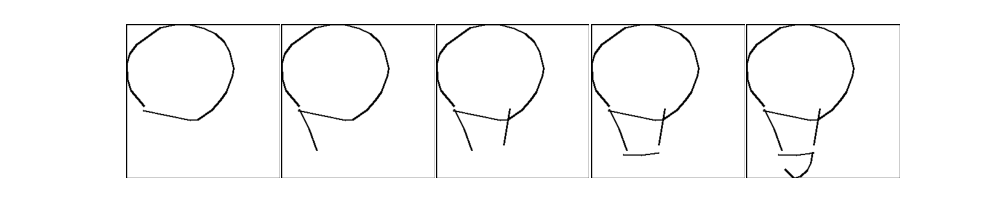} \\
    \raisebox{0.77\height}{\rotatebox[origin=t]{90}{\footnotesize{Ours}}} & 
    \includegraphics[width=0.75\linewidth]{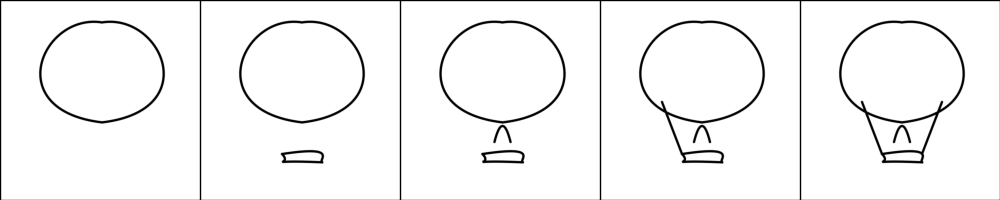} \\
    \hline
    \raisebox{0.6\height}{\rotatebox[origin=t]{90}{\footnotesize{Human}}} & 
    \includegraphics[trim={3.2cm 0.5cm 2.5cm 0.6cm},clip,width=0.75\linewidth]{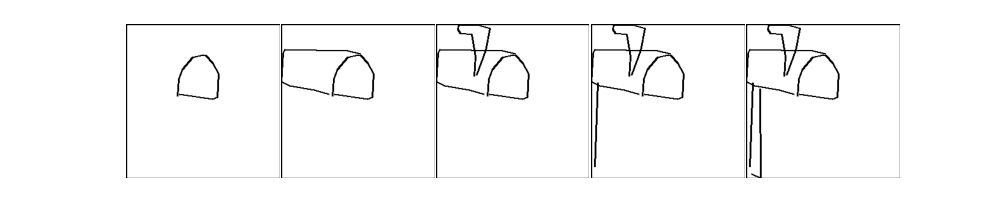} \\
    \raisebox{0.77\height}{\rotatebox[origin=t]{90}{\footnotesize{Ours}}} & 
    \includegraphics[width=0.75\linewidth]{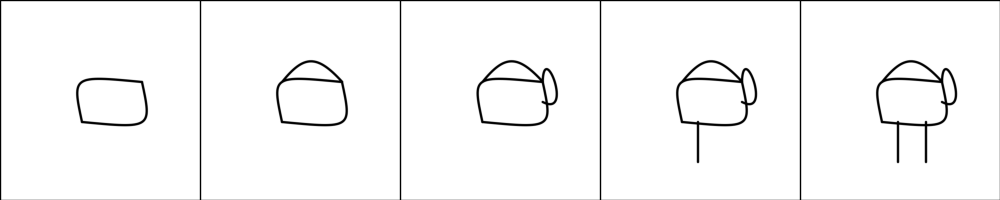} \\
    \hline
    \raisebox{0.6\height}{\rotatebox[origin=t]{90}{\footnotesize{Human}}} & 
    \includegraphics[trim={3.2cm 0.5cm 2.5cm 0.6cm},clip,width=0.75\linewidth]{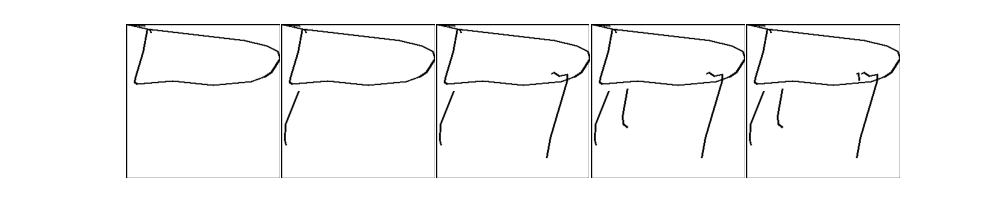} \\
    \raisebox{0.77\height}{\rotatebox[origin=t]{90}{\footnotesize{Ours}}} & 
    \includegraphics[width=0.75\linewidth]{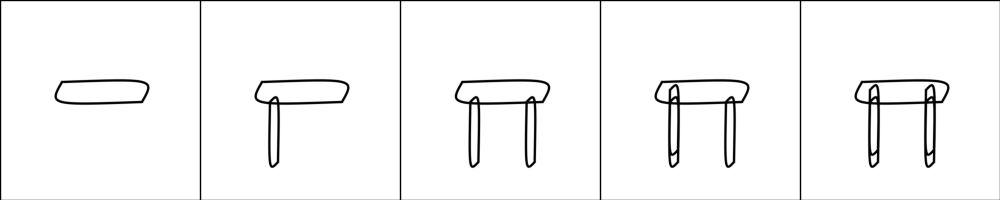} \\
    \end{tabular}
    }
    \caption{Sequential five-stroke sketches of a pear, purse, and screwdriver, created by humans \cite{quickDrawData} and by SketchAgent.}
    \label{fig:5-strokes-set}
\end{figure}

\begin{figure}[h]
    \centering
    \setlength{\tabcolsep}{1pt}
    \renewcommand{\arraystretch}{0.5}
    \small{
    \begin{tabular}{l c}
    \raisebox{0.65\height}{\rotatebox[origin=t]{90}{\footnotesize{Human}}} & 
    \includegraphics[trim={3.8cm 0.46cm 3cm 0.57cm},clip,width=0.9\linewidth]{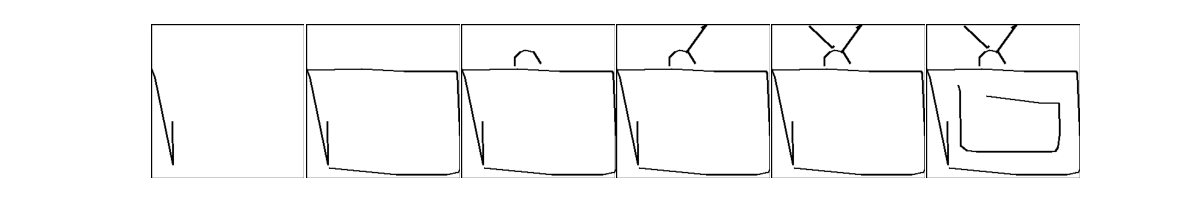} \\
    \raisebox{0.83\height}{\rotatebox[origin=t]{90}{\footnotesize{Ours}}} & 
    \includegraphics[width=0.9\linewidth]{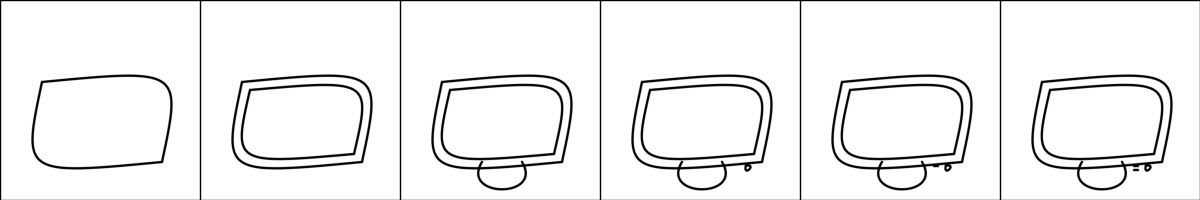} \\
    \hline
    \raisebox{0.65\height}{\rotatebox[origin=t]{90}{\footnotesize{Human}}} & 
    \includegraphics[trim={3.8cm 0.46cm 3cm 0.57cm},clip,width=0.9\linewidth]{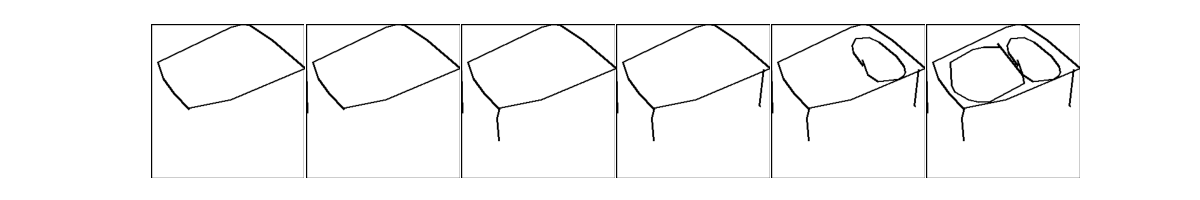} \\
    \raisebox{0.83\height}{\rotatebox[origin=t]{90}{\footnotesize{Ours}}} & 
    \includegraphics[width=0.9\linewidth]{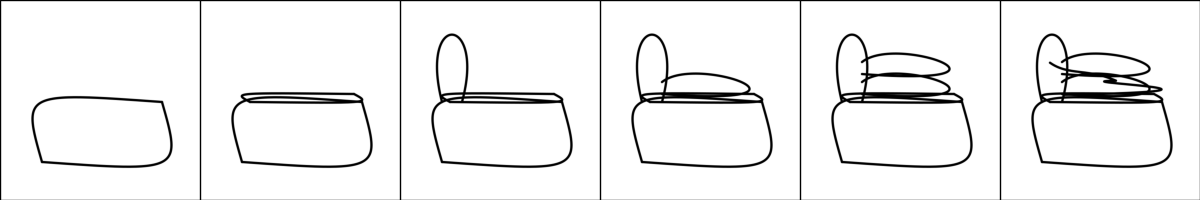} \\
    \hline
    \raisebox{0.65\height}{\rotatebox[origin=t]{90}{\footnotesize{Human}}} & 
    \includegraphics[trim={3.8cm 0.46cm 3cm 0.57cm},clip,width=0.9\linewidth]{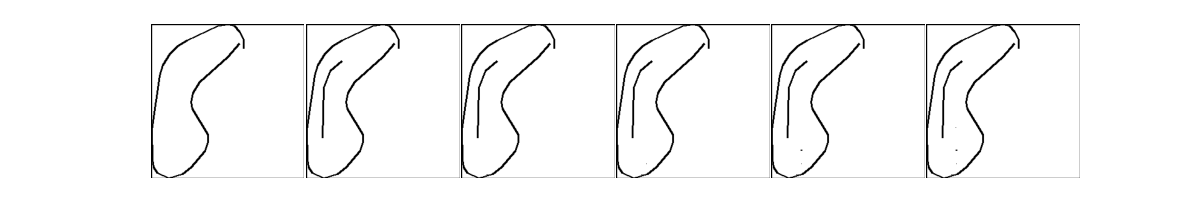} \\
    \raisebox{0.83\height}{\rotatebox[origin=t]{90}{\footnotesize{Ours}}} & 
    \includegraphics[width=0.9\linewidth]{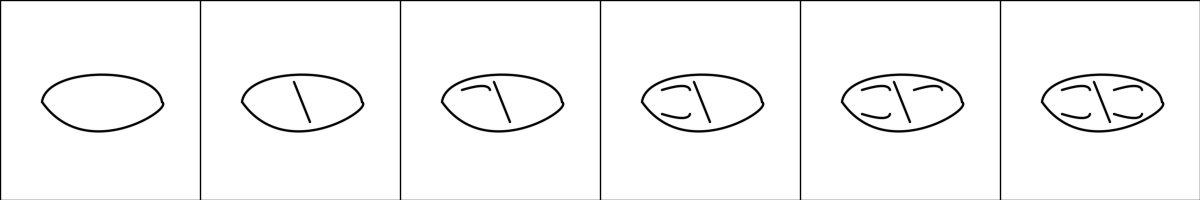} \\  
    \end{tabular}
    }
    \caption{Sequential six-stroke sketches of a television, bed, and peanut, created by humans \cite{quickDrawData} and by SketchAgent.}
    \label{fig:6-strokes-set}
\end{figure}

\begin{figure}[h]
    \centering
    \setlength{\tabcolsep}{1pt}
    \renewcommand{\arraystretch}{0.5}
    \small{
    \begin{tabular}{l c}
    \raisebox{0.55\height}{\rotatebox[origin=t]{90}{\footnotesize{Human}}} & 
    \includegraphics[width=1\linewidth]{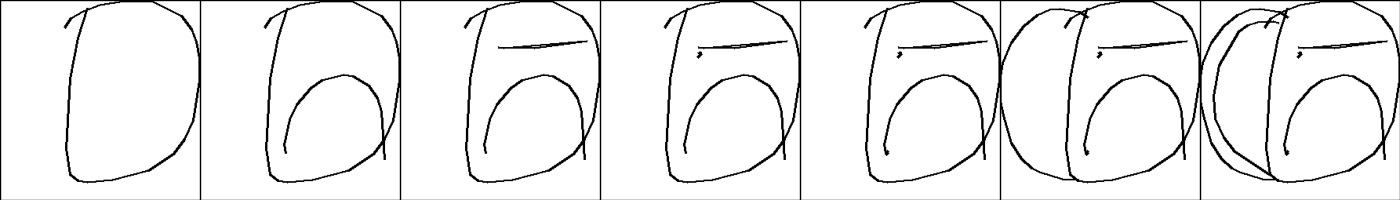} \\
    \raisebox{0.7\height}{\rotatebox[origin=t]{90}{\footnotesize{Ours}}} & 
    \includegraphics[width=1\linewidth]{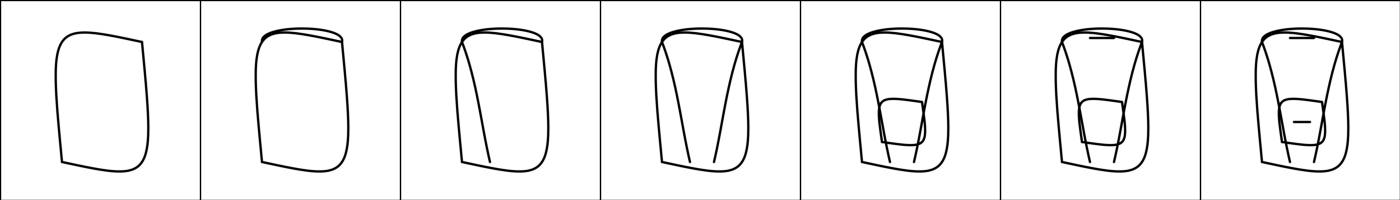} \\
    \hline
    \raisebox{0.55\height}{\rotatebox[origin=t]{90}{\footnotesize{Human}}} & 
    \includegraphics[width=1\linewidth]{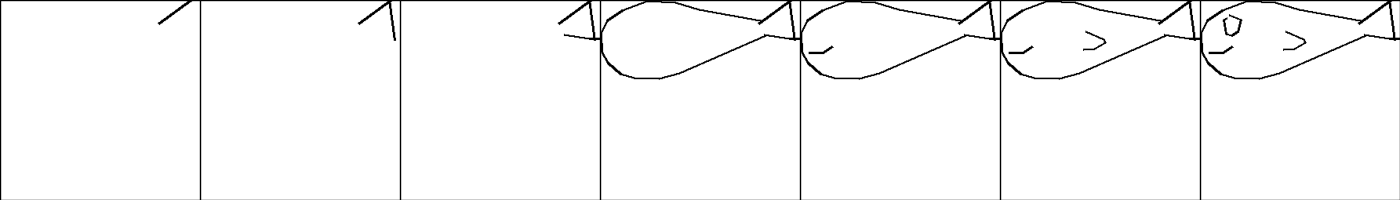} \\
    \raisebox{0.7\height}{\rotatebox[origin=t]{90}{\footnotesize{Ours}}} & 
    \includegraphics[width=1\linewidth]{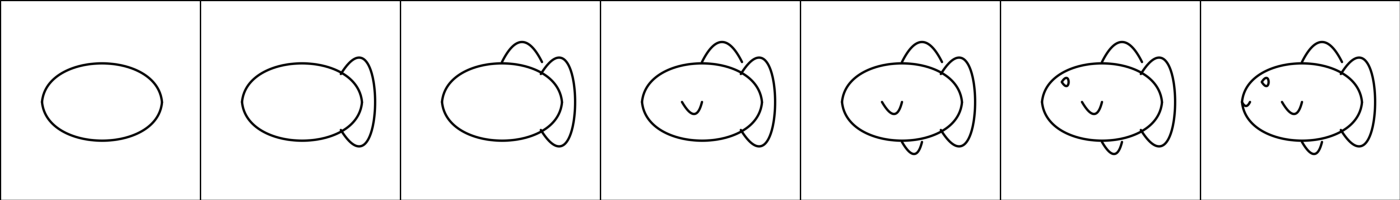} \\
    \hline
    \raisebox{0.55\height}{\rotatebox[origin=t]{90}{\footnotesize{Human}}} & 
    \includegraphics[width=1\linewidth]{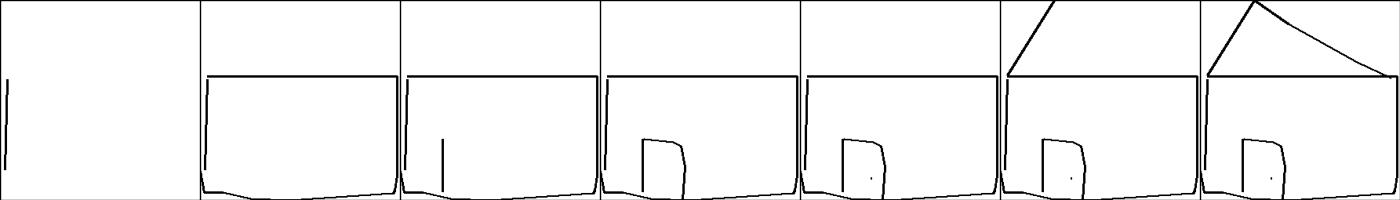} \\
    \raisebox{0.7\height}{\rotatebox[origin=t]{90}{\footnotesize{Ours}}} & 
    \includegraphics[width=1\linewidth]{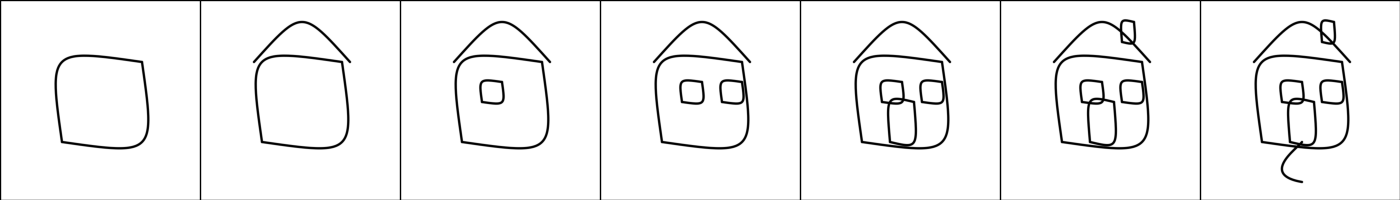} \\
    \end{tabular}
    }
    \caption{Sequential seven-stroke sketches of a backpack, fish, and house, created by humans \cite{quickDrawData} and by SketchAgent.}
    \label{fig:7-strokes-set}
\end{figure}

\begin{figure*}[t]
    \centering
    \setlength{\tabcolsep}{0pt}
    {\small
    \begin{tabular}{c c c c}
    \includegraphics[width=0.25\linewidth, valign=t]{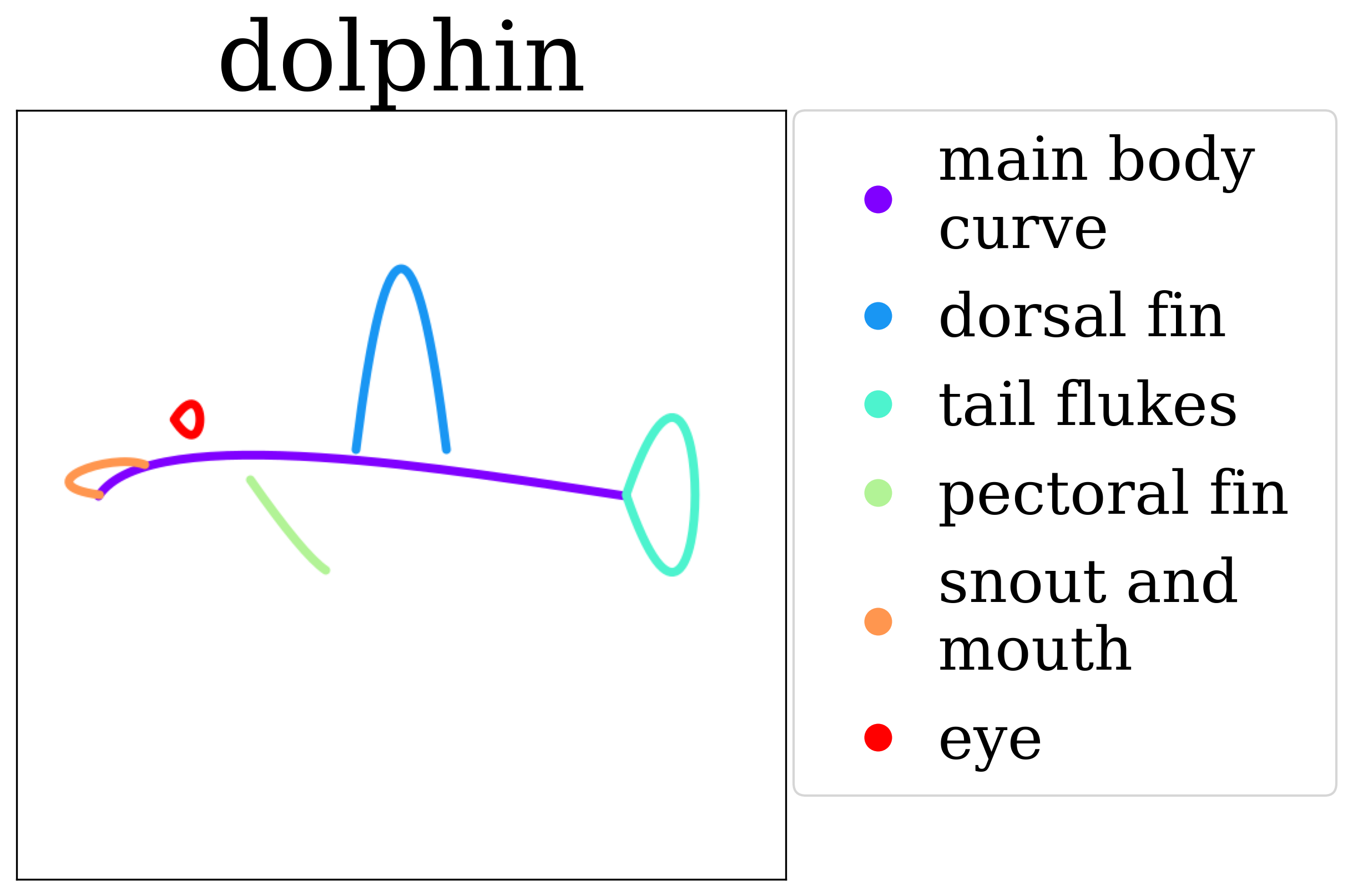} & 
    \includegraphics[width=0.25\linewidth, valign=t]{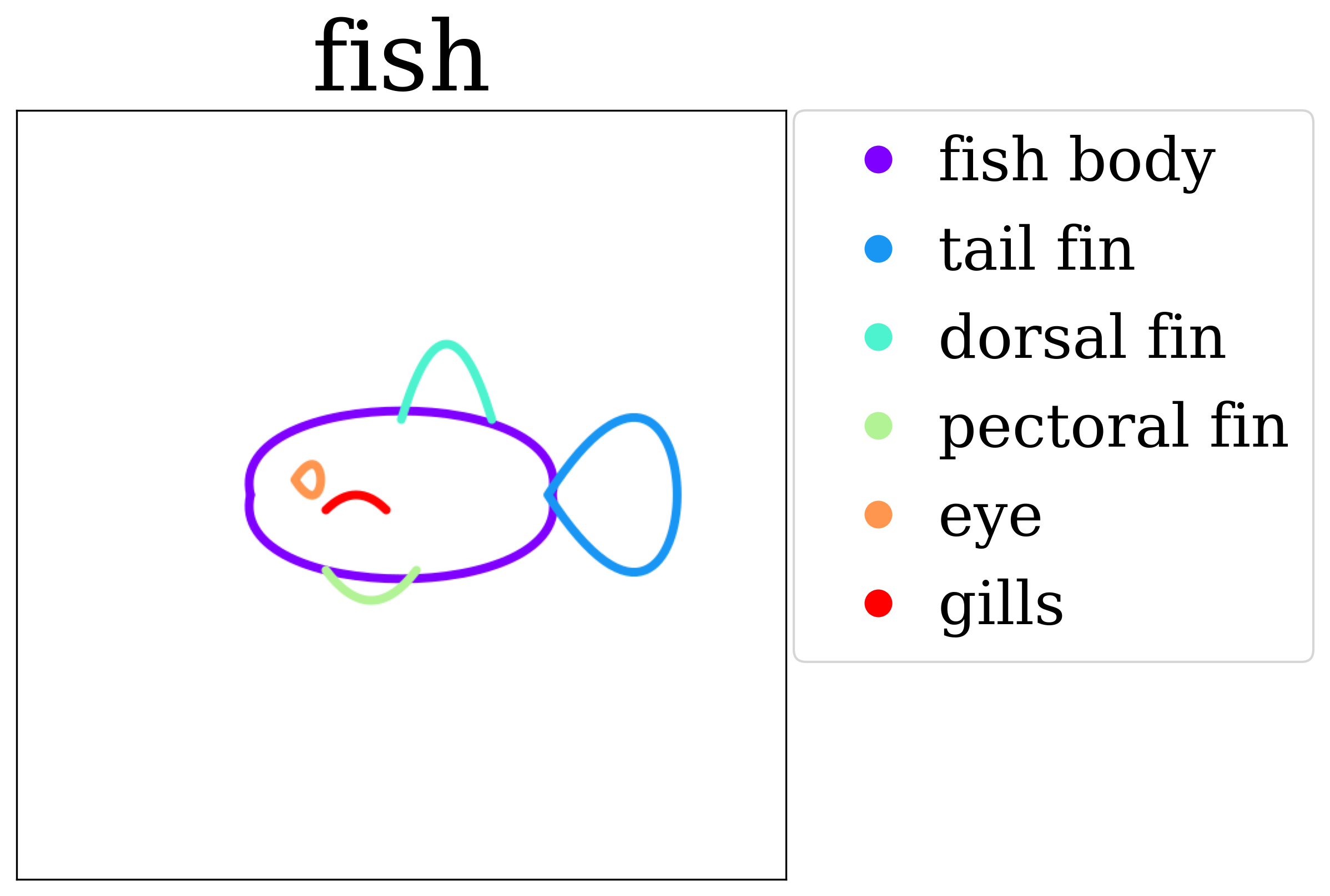} & 
    \includegraphics[width=0.25\linewidth, valign=t]{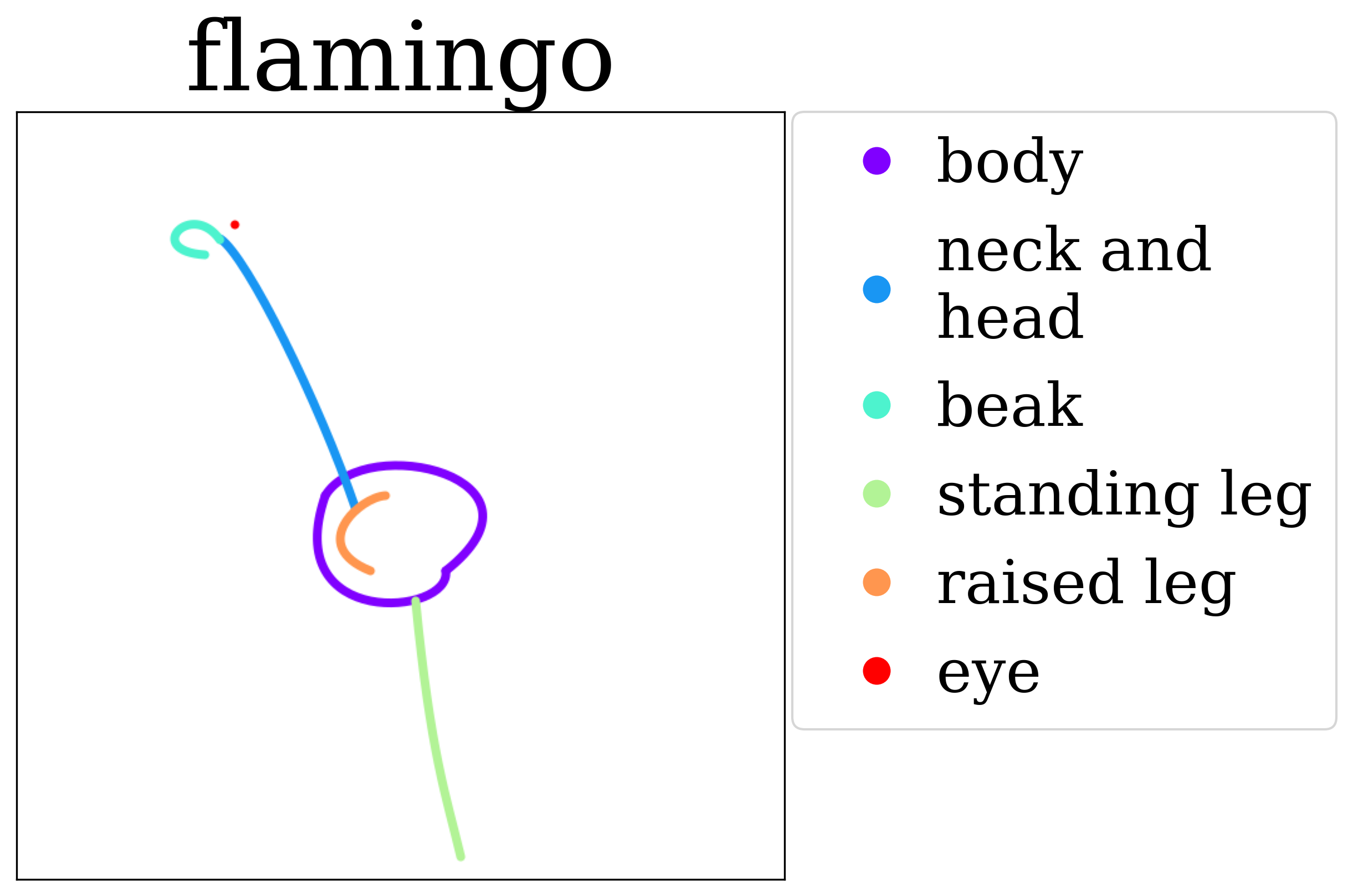} & 
    \includegraphics[width=0.25\linewidth, valign=t]{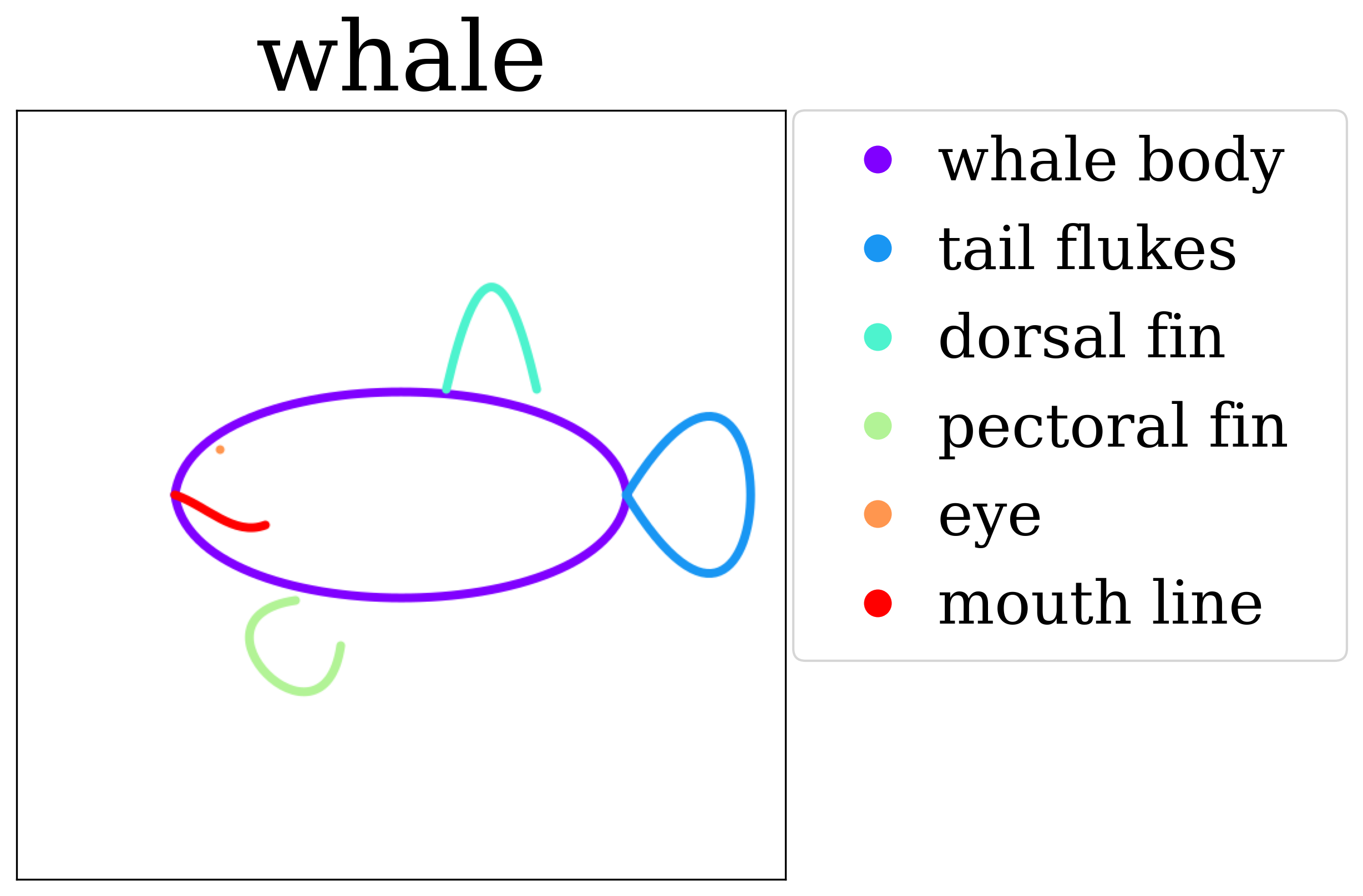} \\ 
    \includegraphics[width=0.25\linewidth, valign=t]{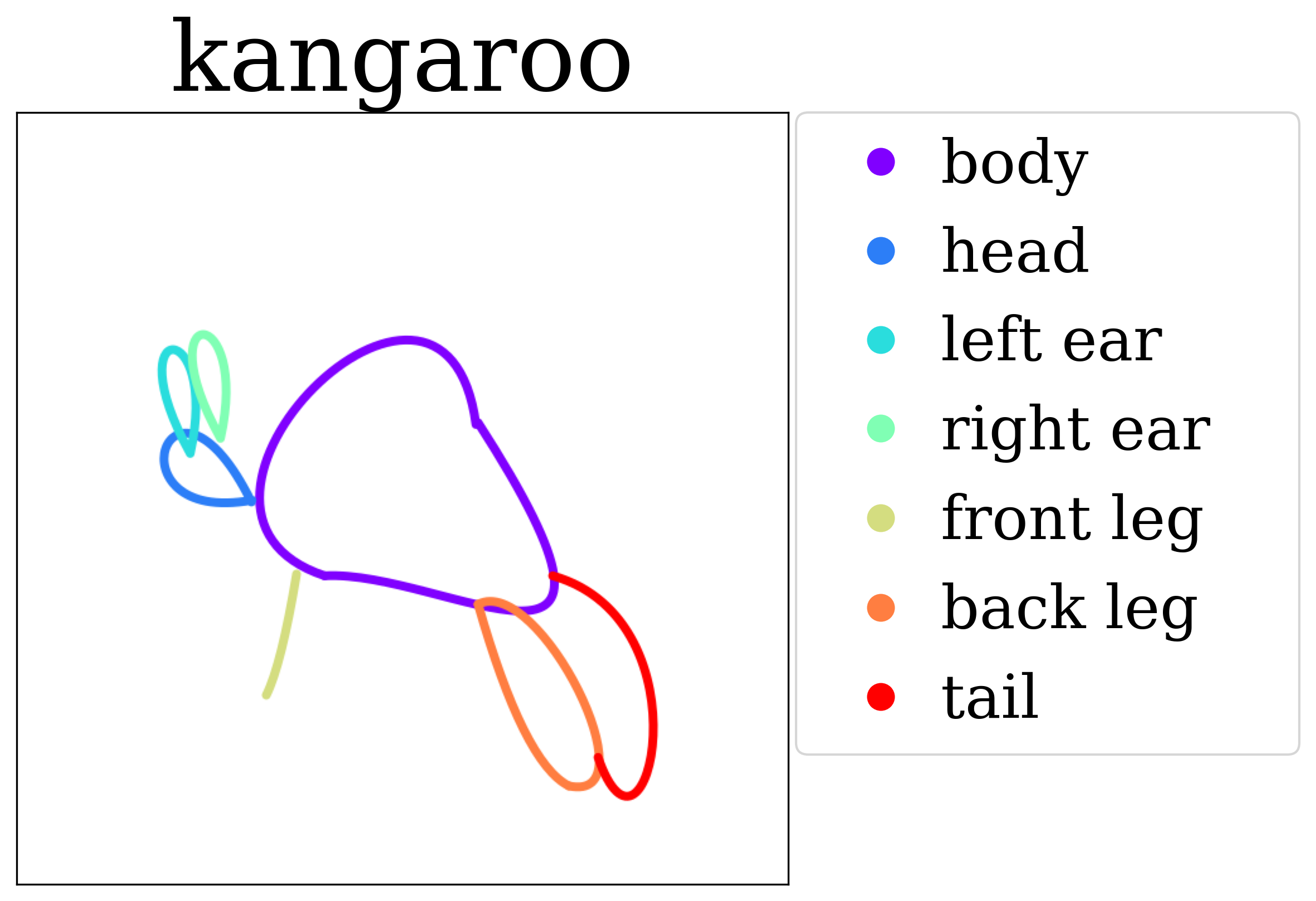} & 
    \includegraphics[width=0.25\linewidth, valign=t]{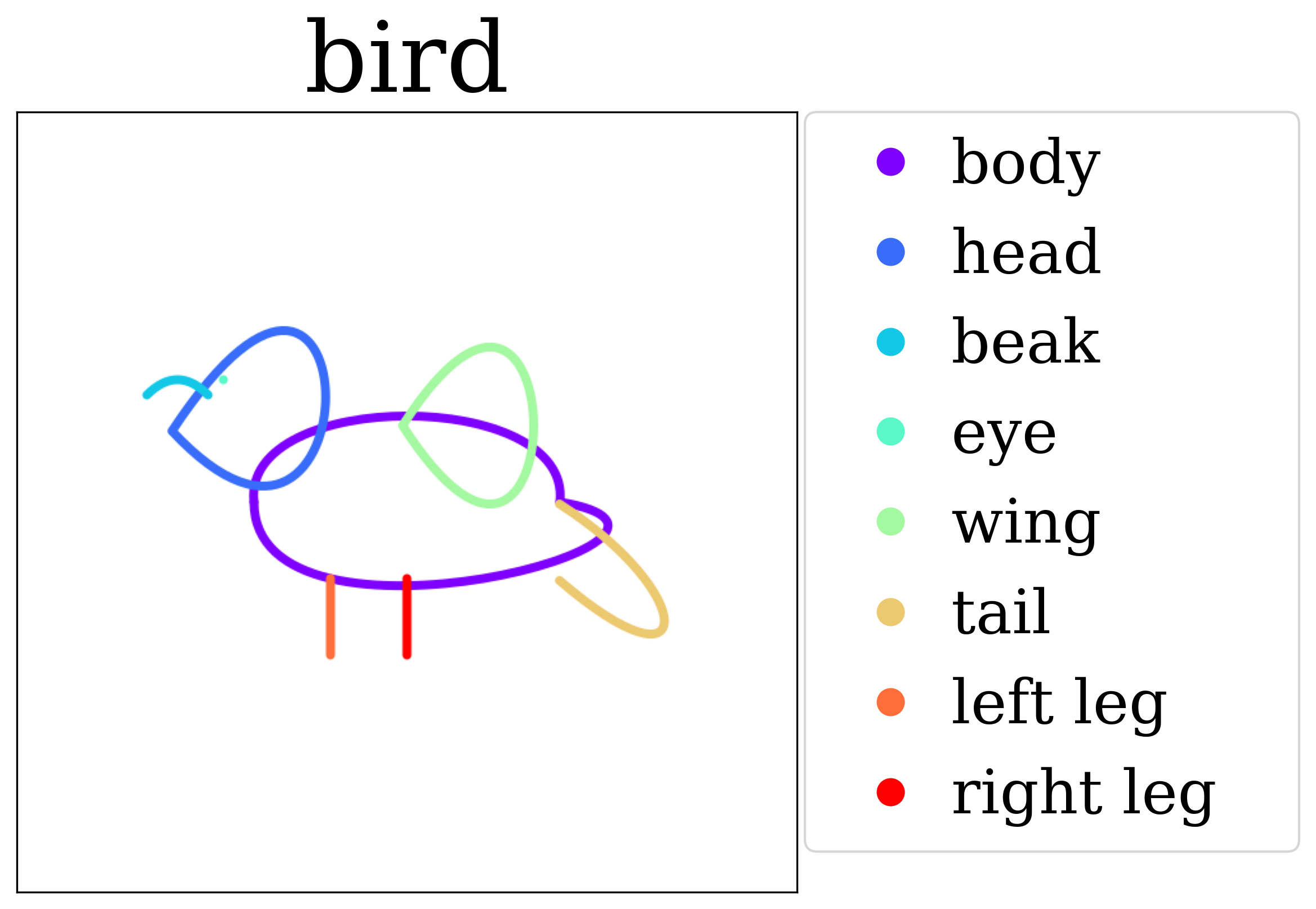} &
    \includegraphics[width=0.25\linewidth, valign=t]{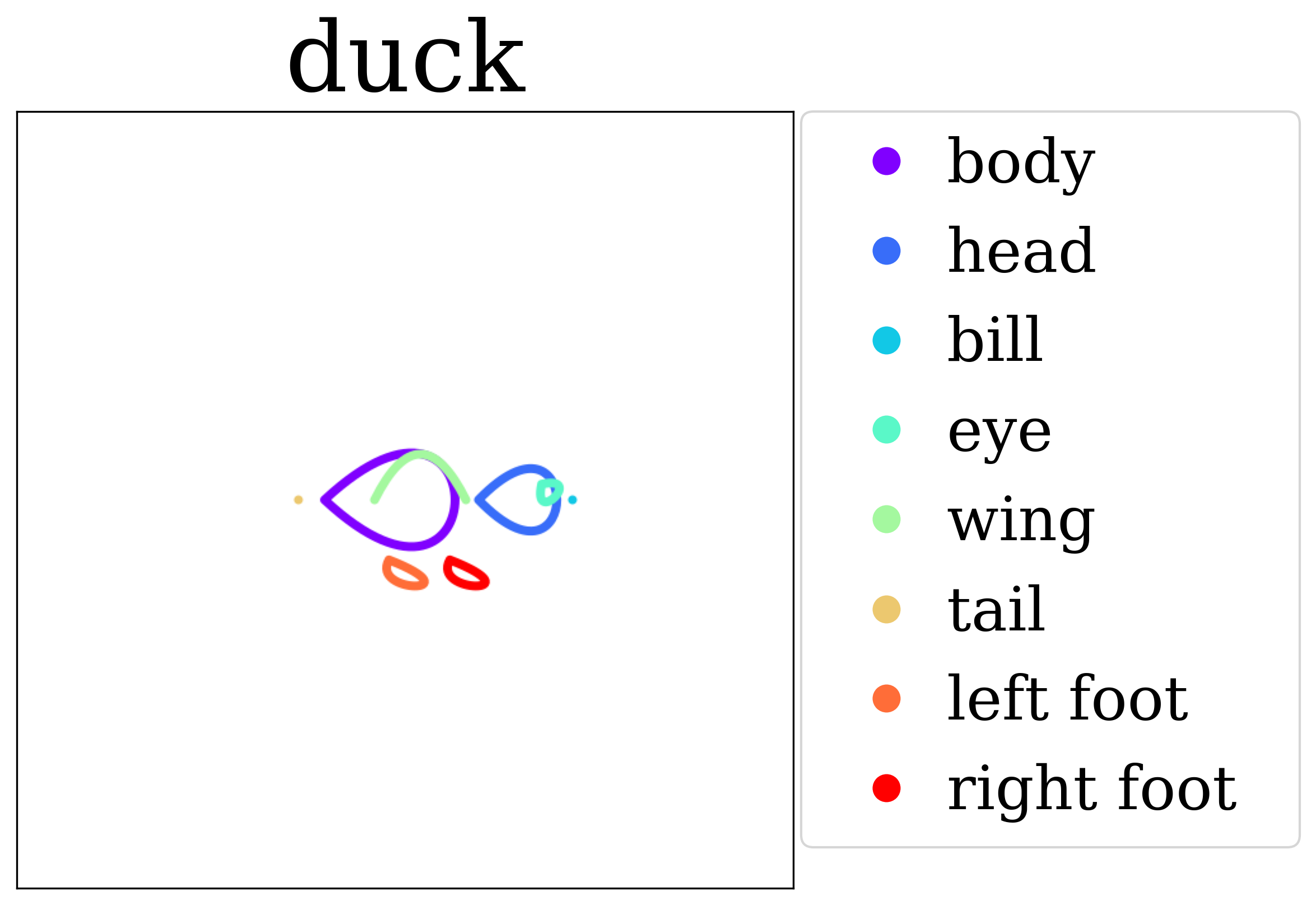} & 
    \includegraphics[width=0.25\linewidth, valign=t]{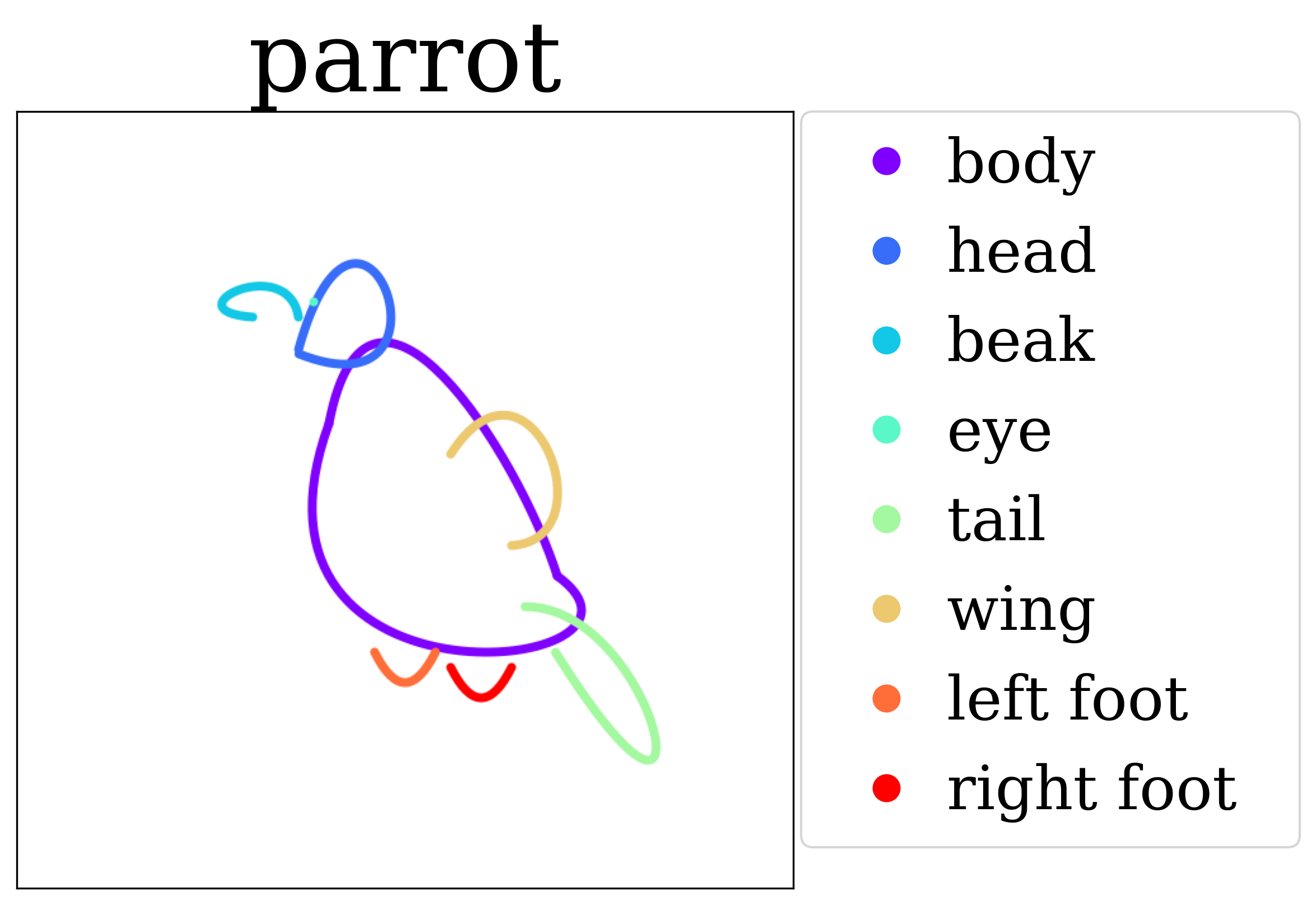} \\
    \includegraphics[width=0.25\linewidth, valign=t]{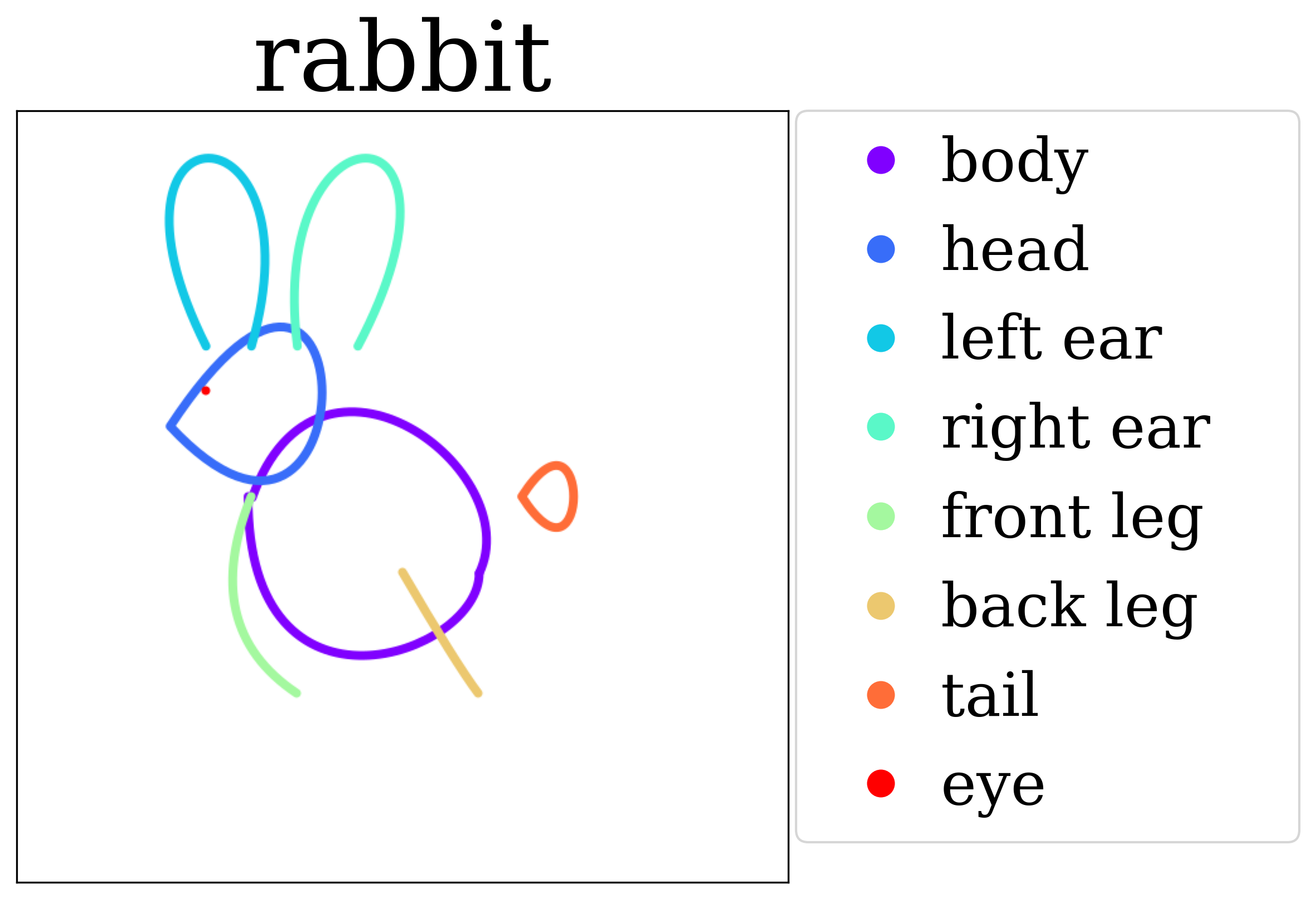} & 
    \includegraphics[width=0.25\linewidth, valign=t]{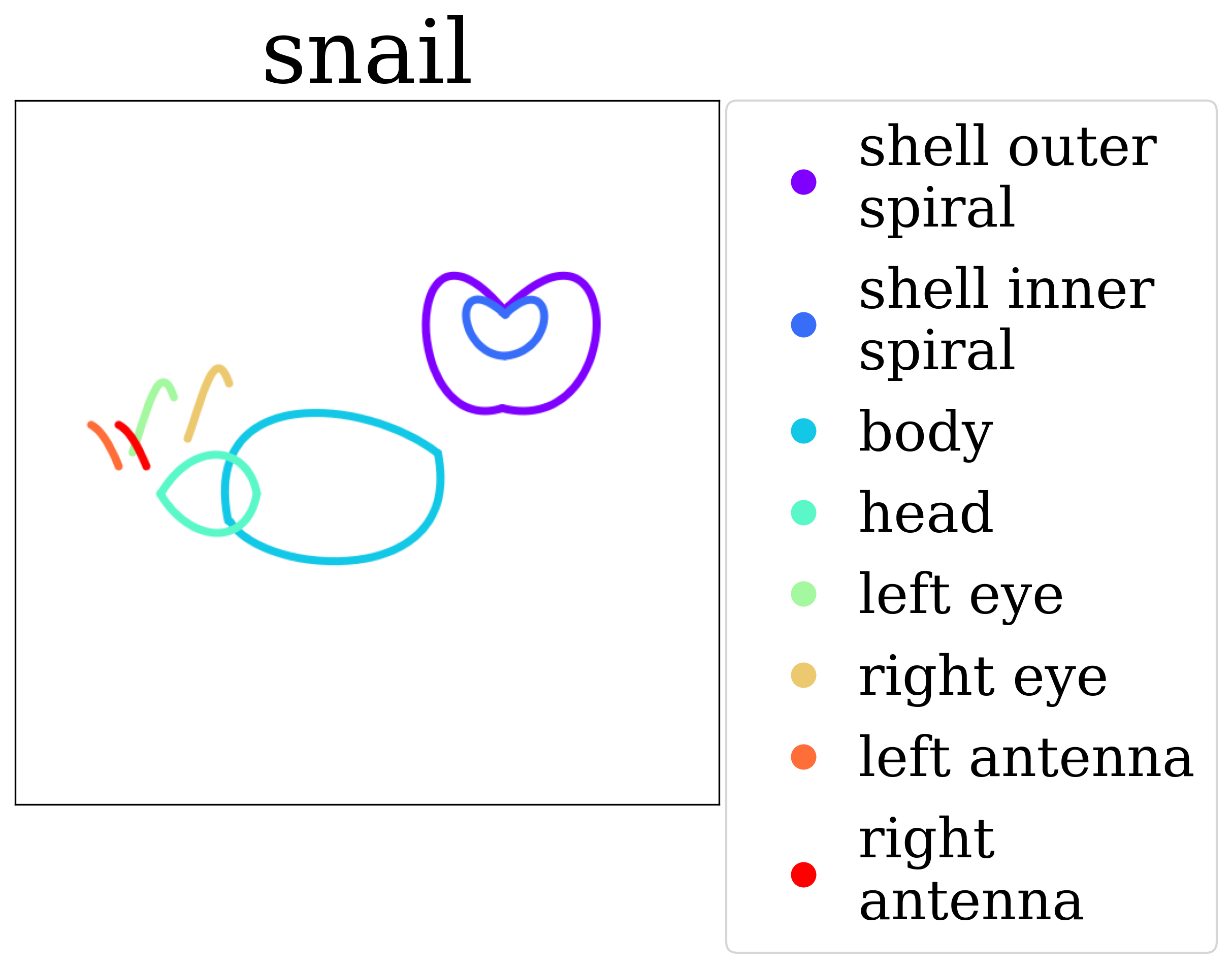} & 
    \includegraphics[width=0.25\linewidth, valign=t]{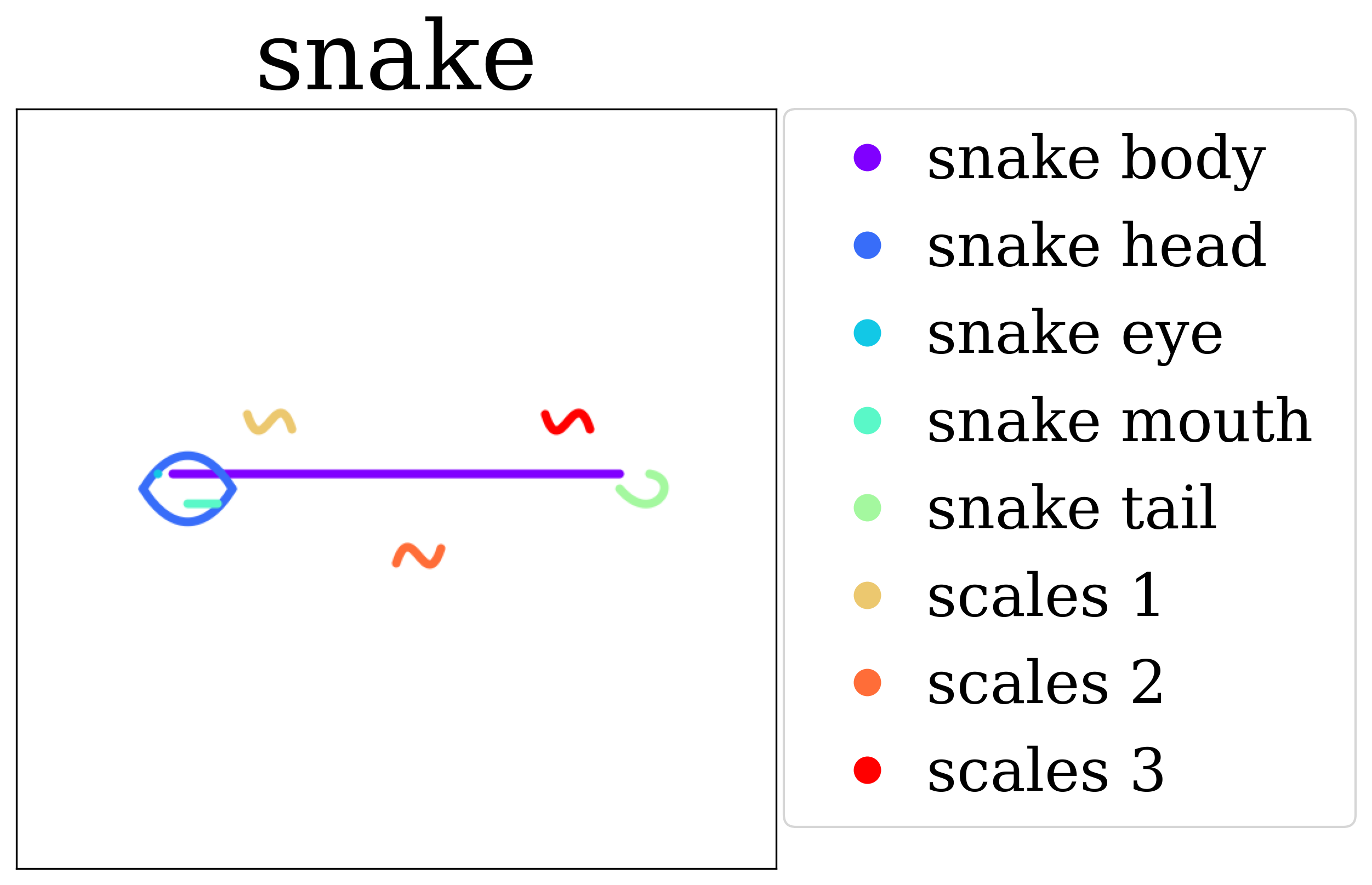} & 
    \includegraphics[width=0.25\linewidth, valign=t]{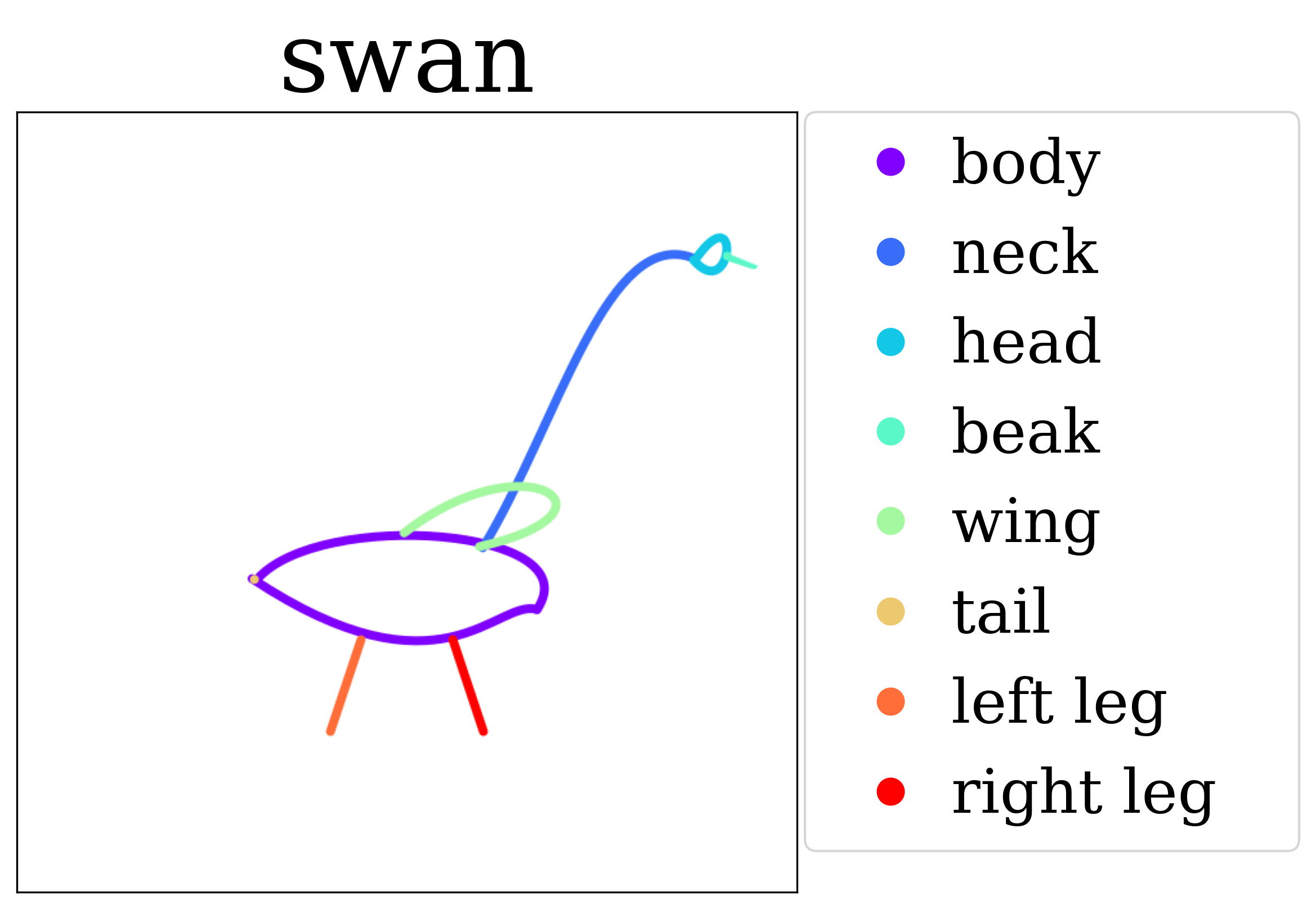} \\
    \includegraphics[width=0.25\linewidth, valign=t]{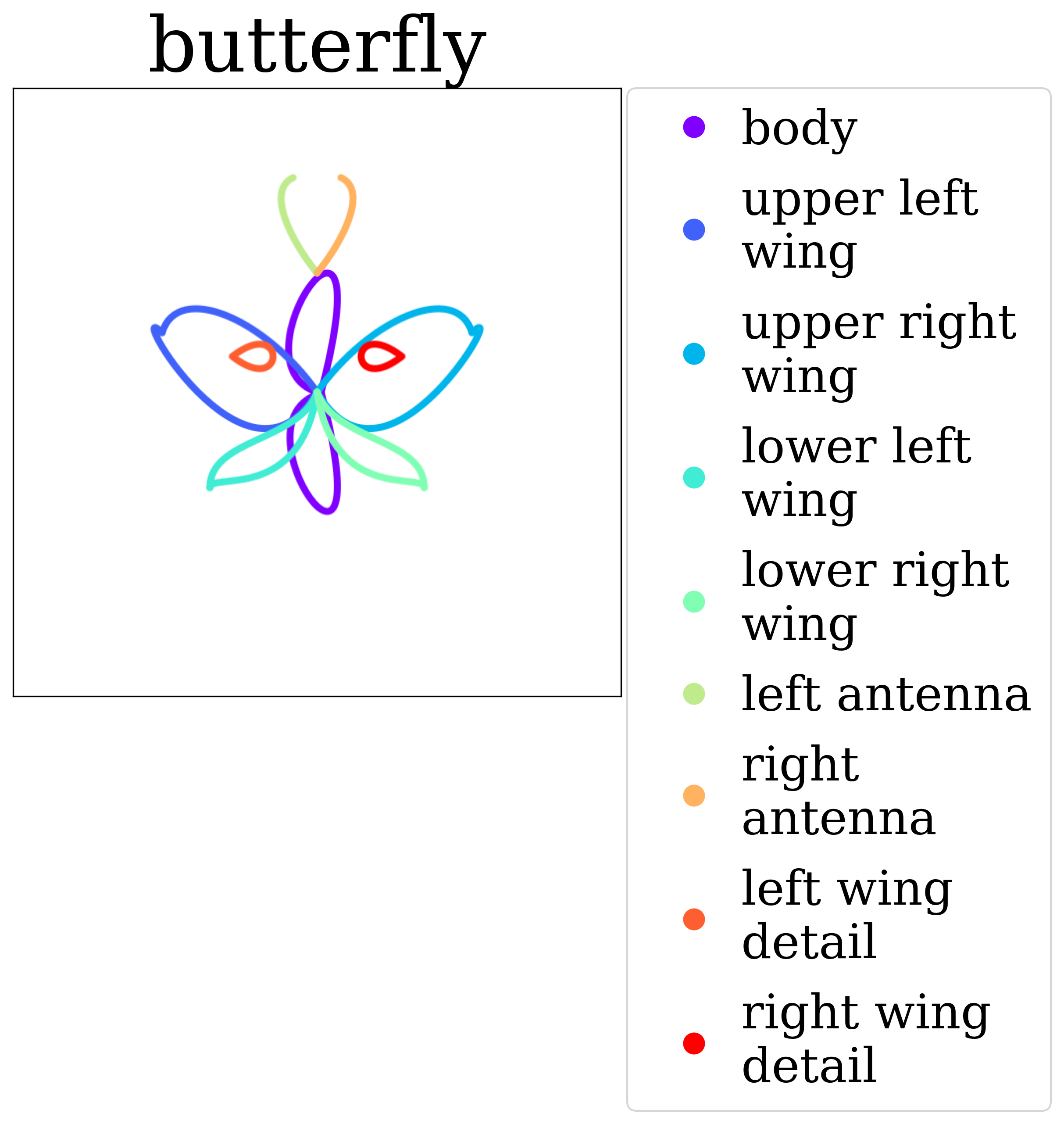} & 
    \includegraphics[width=0.25\linewidth, valign=t]{figs/qualitative/labeled/output_camel_labeled.png} & 
    \includegraphics[width=0.25\linewidth, valign=t]{figs/qualitative/labeled/output_frog_labeled.png} &
    \includegraphics[width=0.25\linewidth, valign=t]{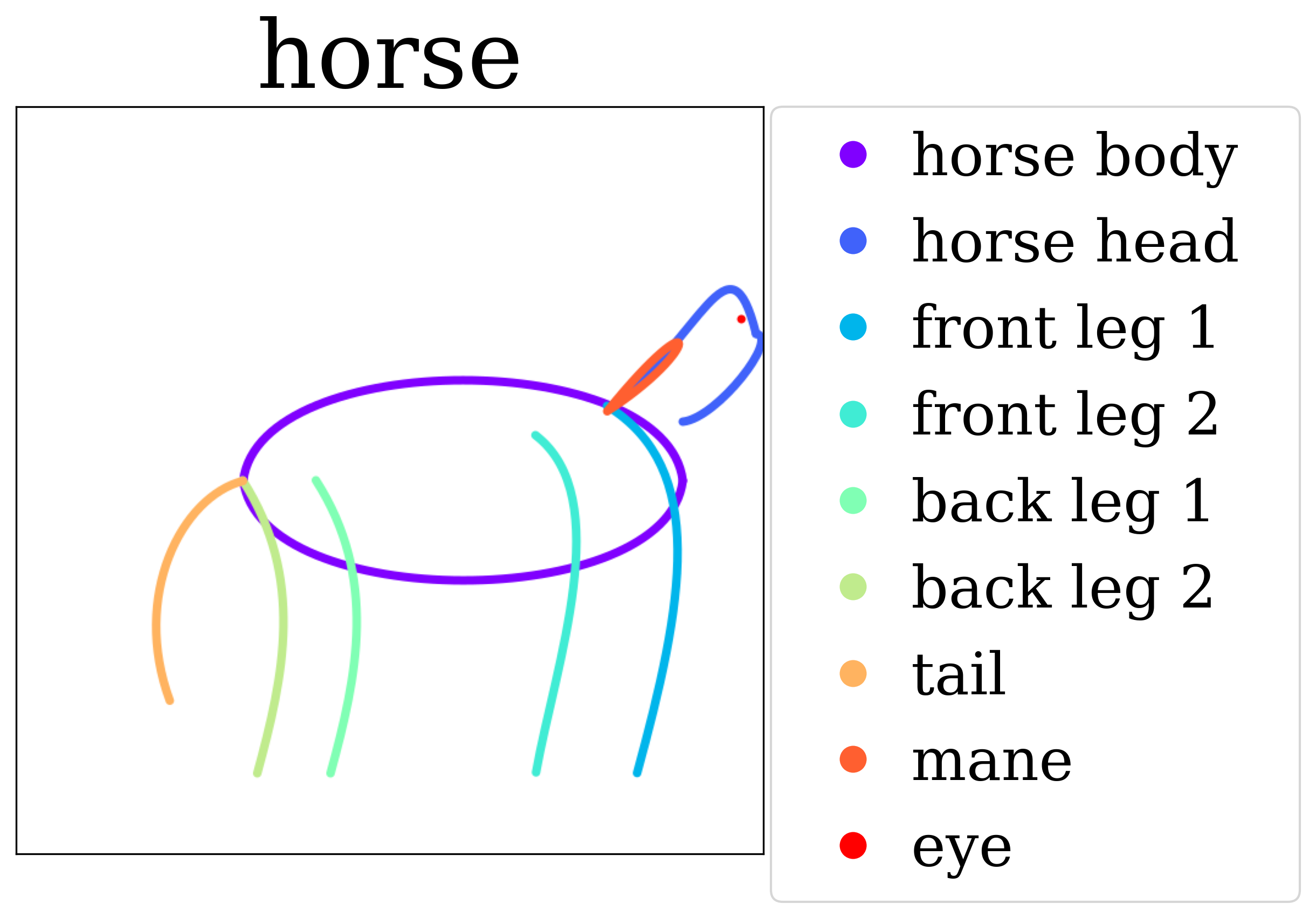} \\
    \includegraphics[width=0.25\linewidth, valign=t]{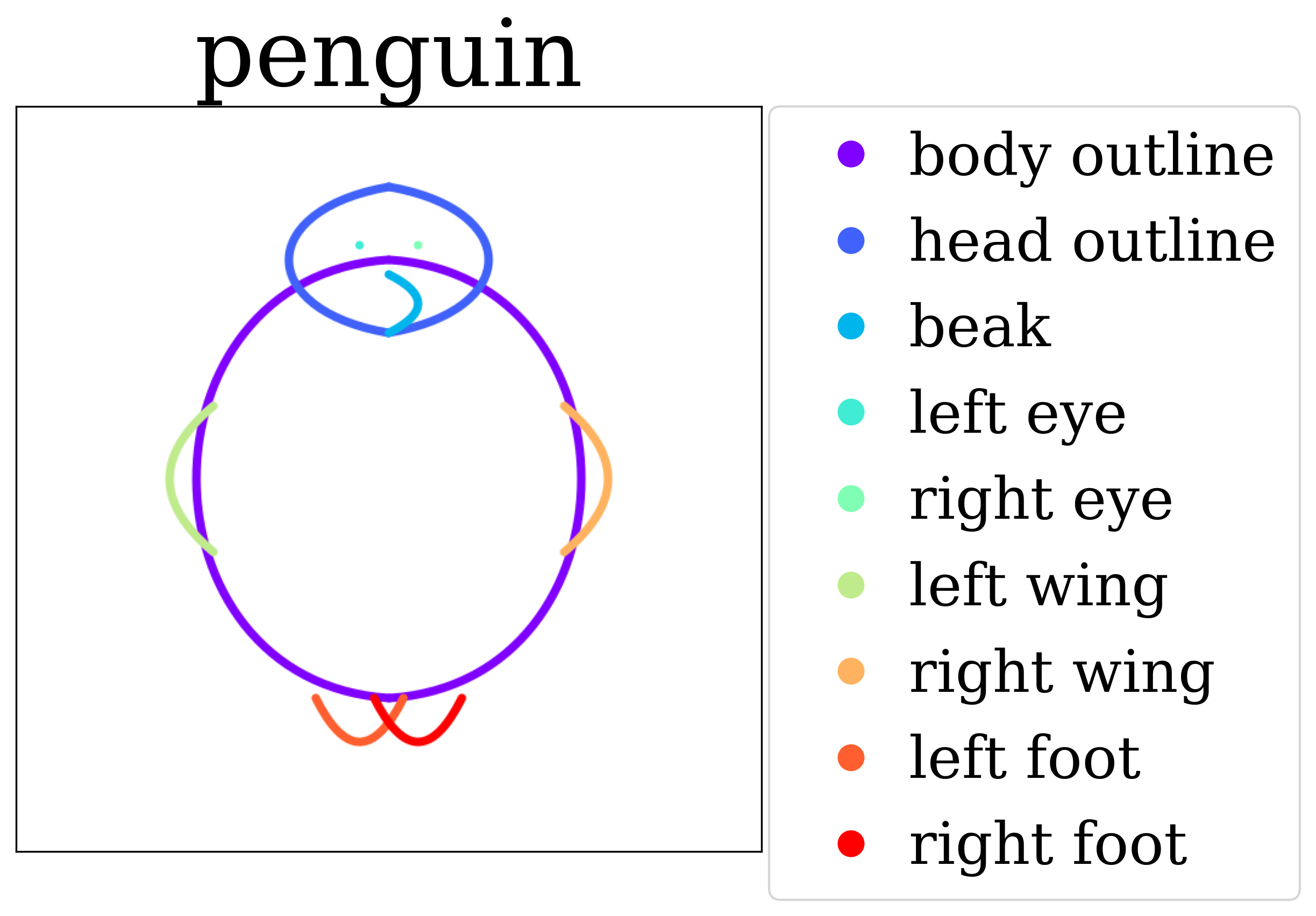} & 
    \includegraphics[width=0.25\linewidth, valign=t]{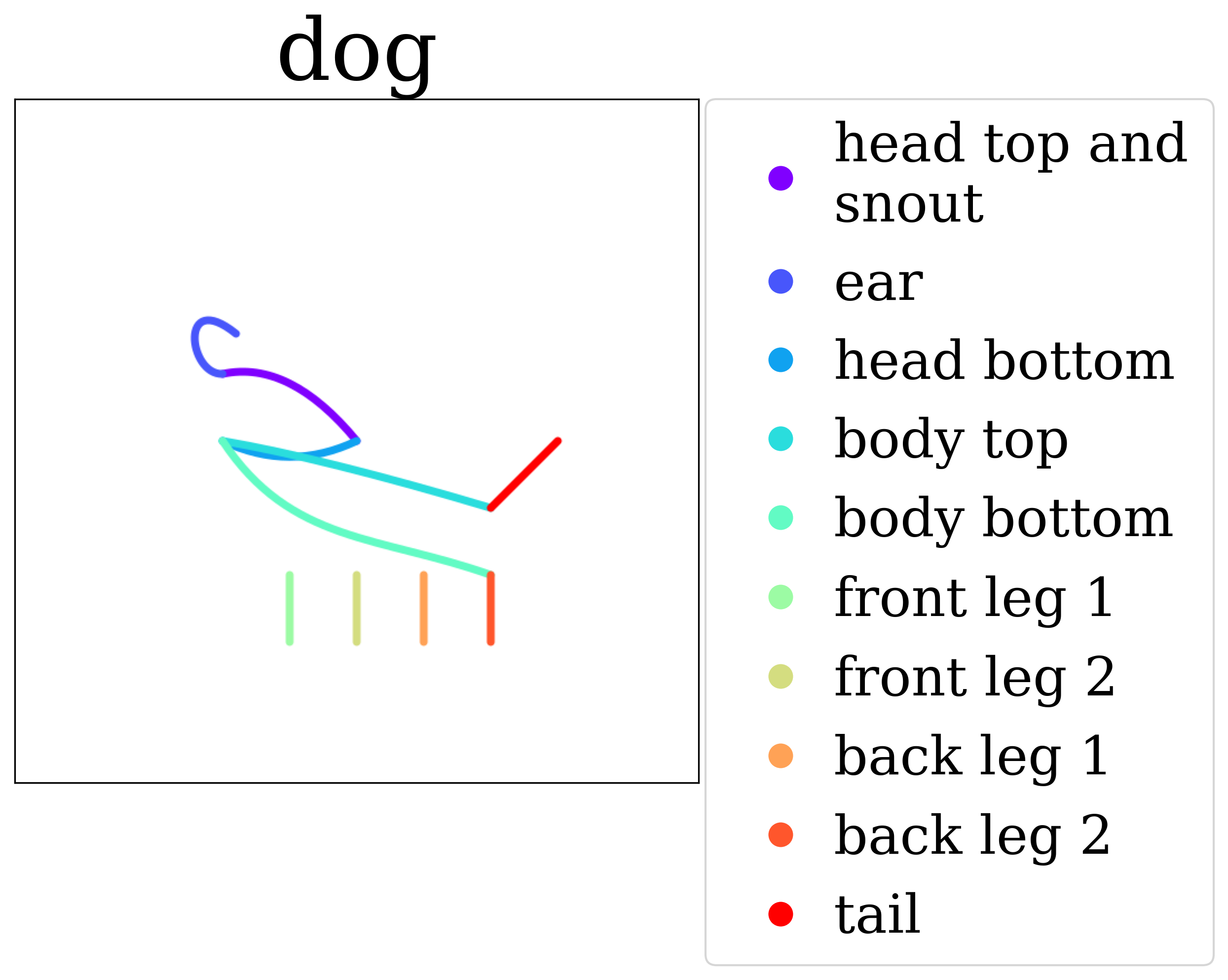} &
    \includegraphics[width=0.25\linewidth, valign=t]{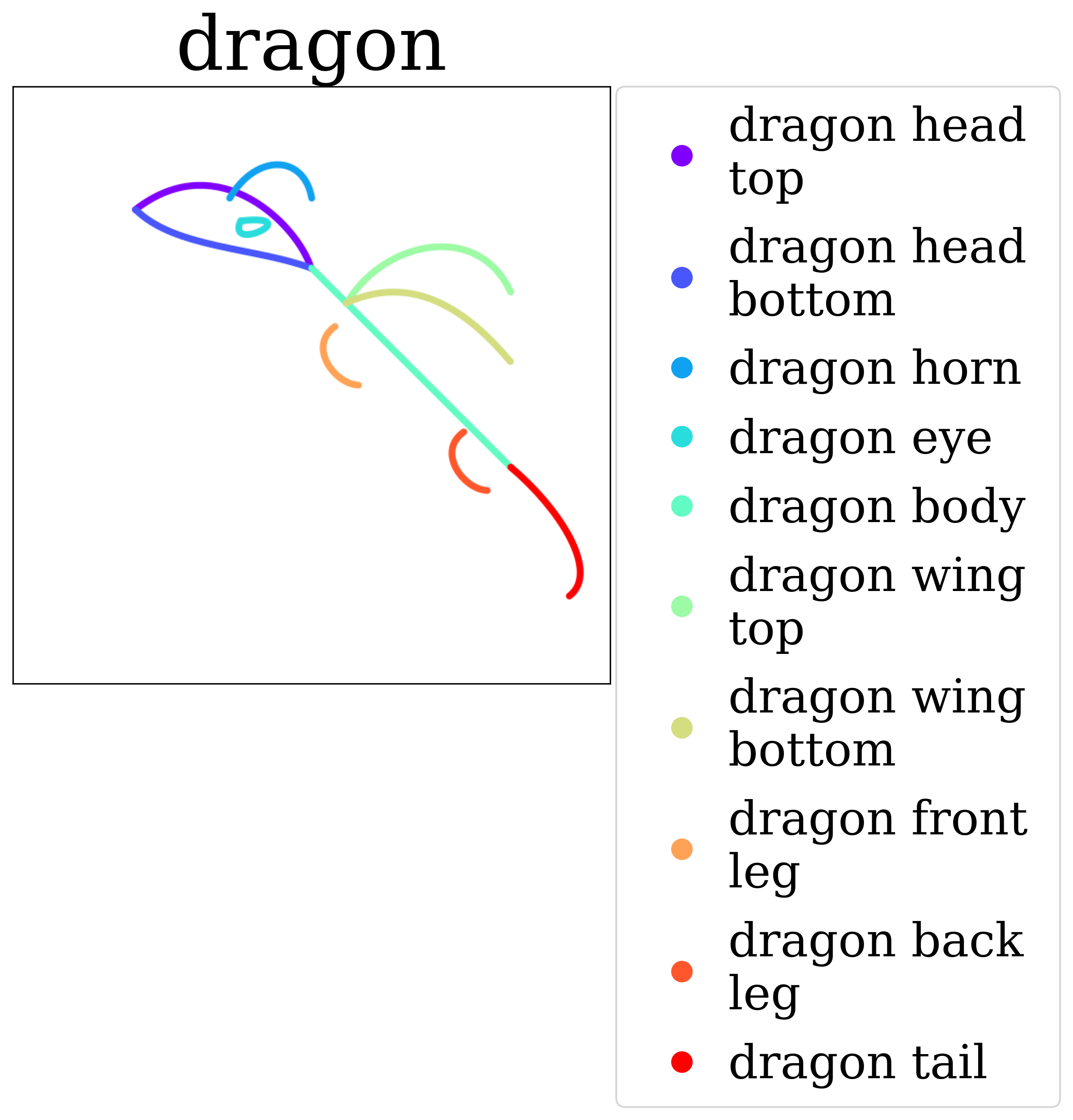} & 
    \includegraphics[width=0.25\linewidth, valign=t]{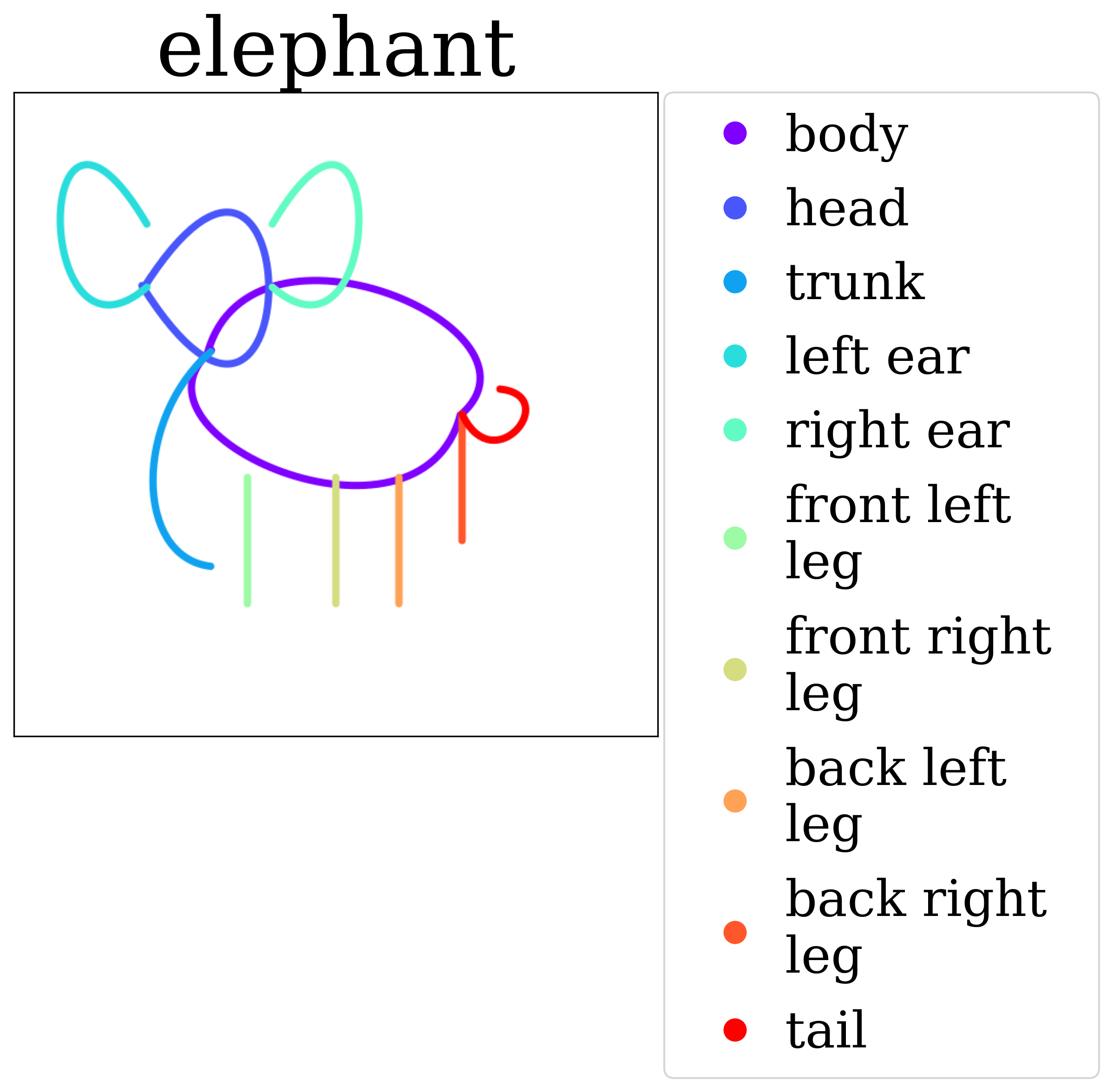} \\
    \includegraphics[width=0.25\linewidth, valign=t]{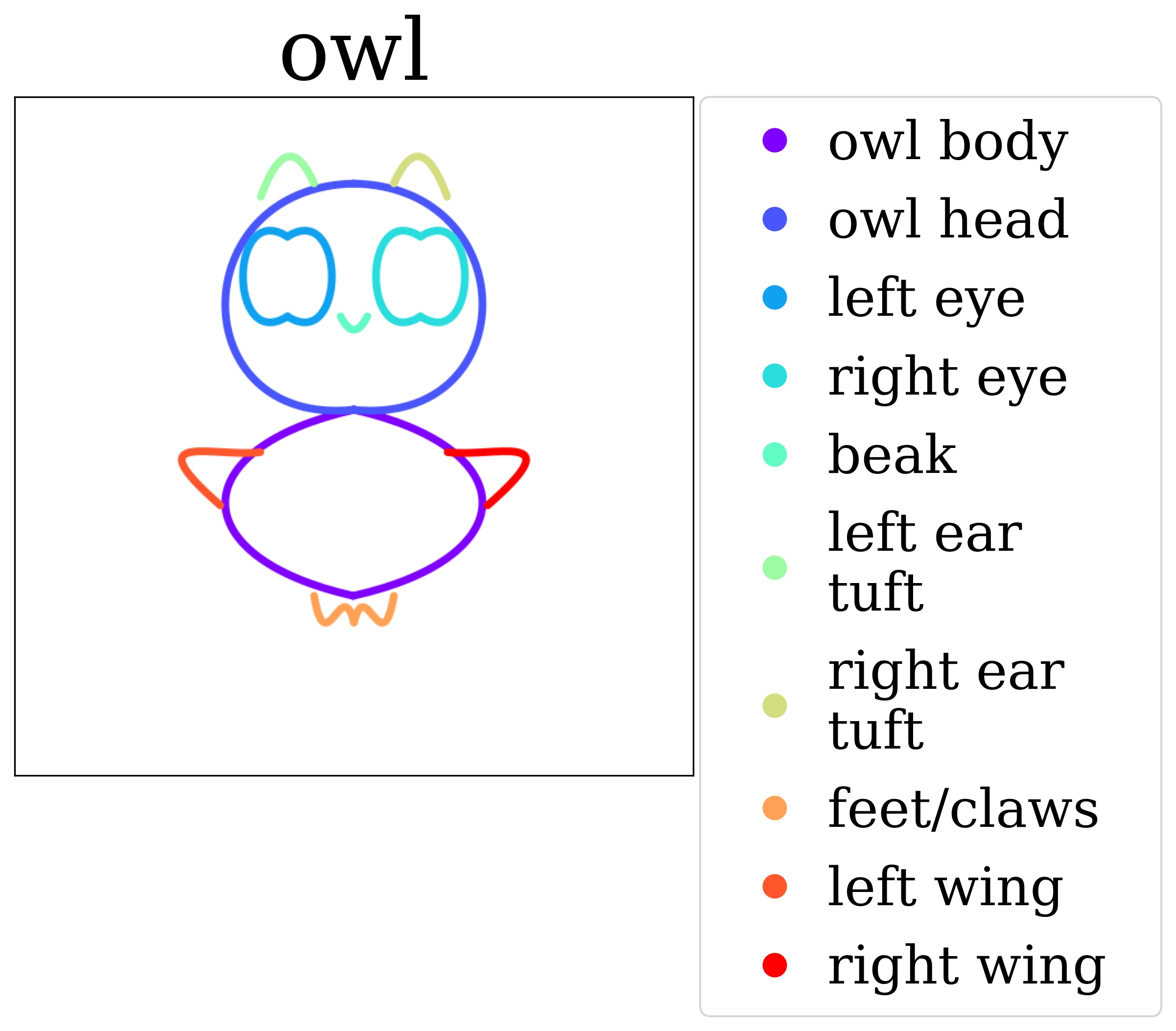} &
    \includegraphics[width=0.25\linewidth, valign=t]{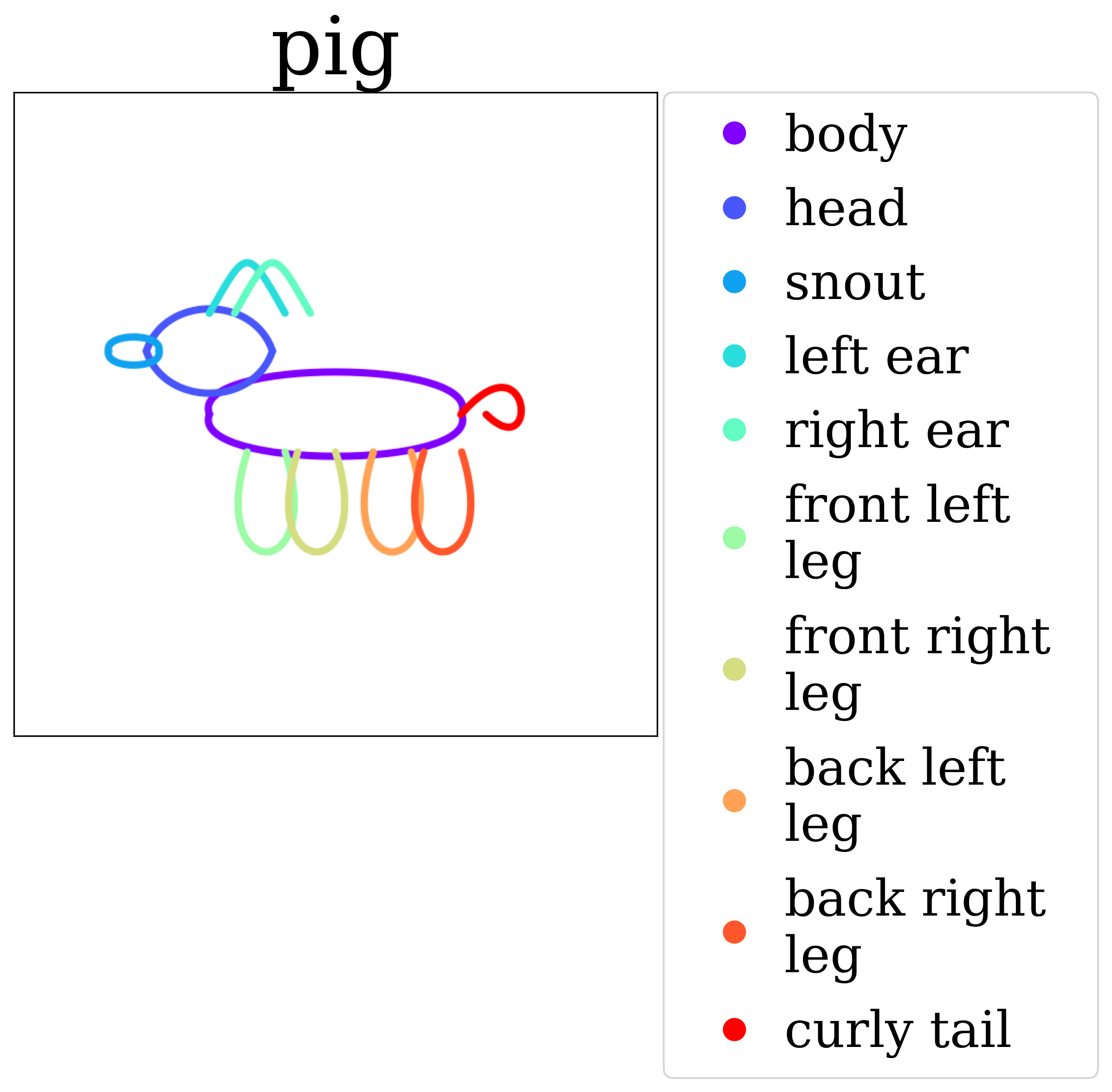} & 
    \includegraphics[width=0.25\linewidth, valign=t]{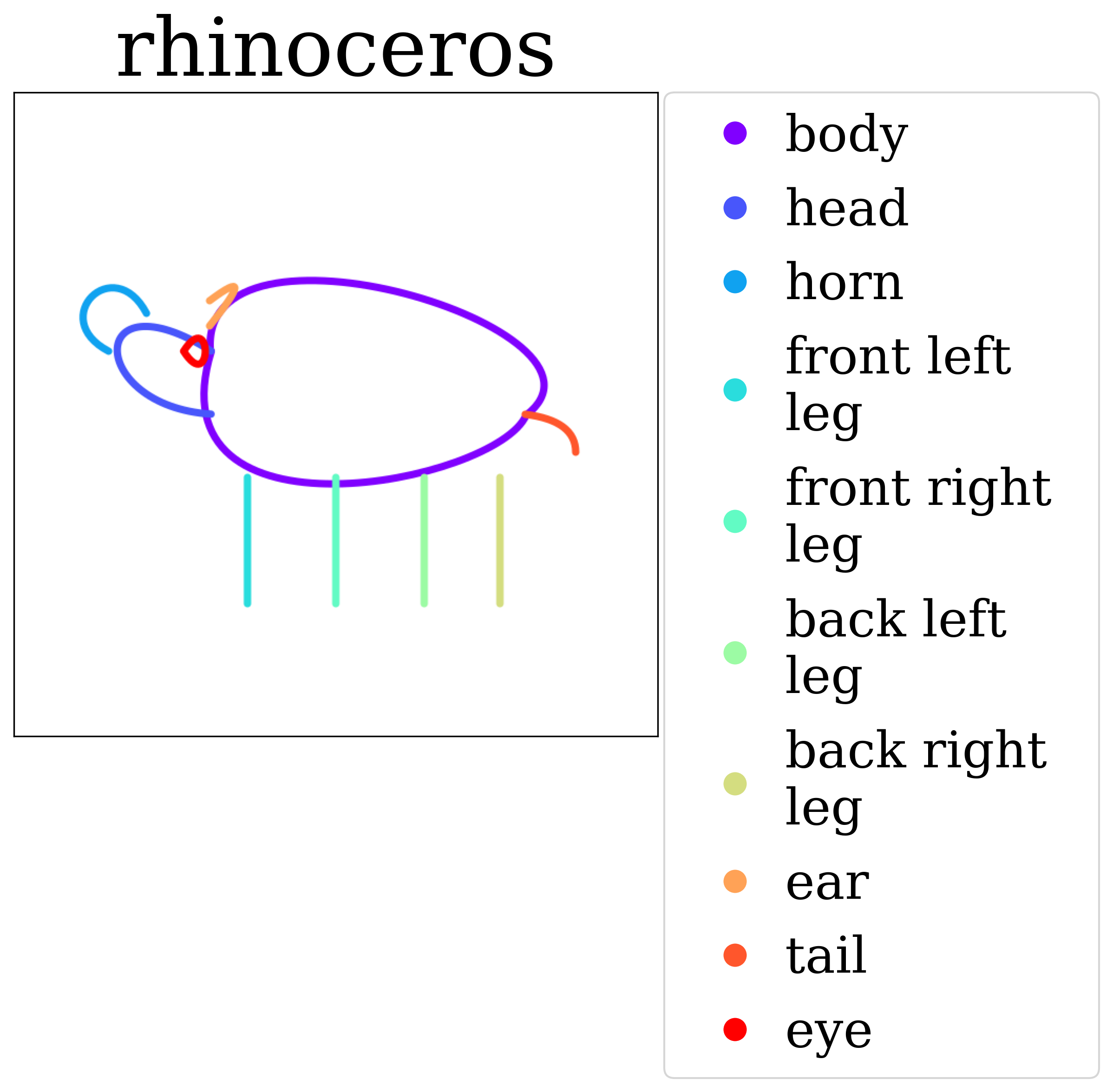} & 
    \includegraphics[width=0.25\linewidth, valign=t]{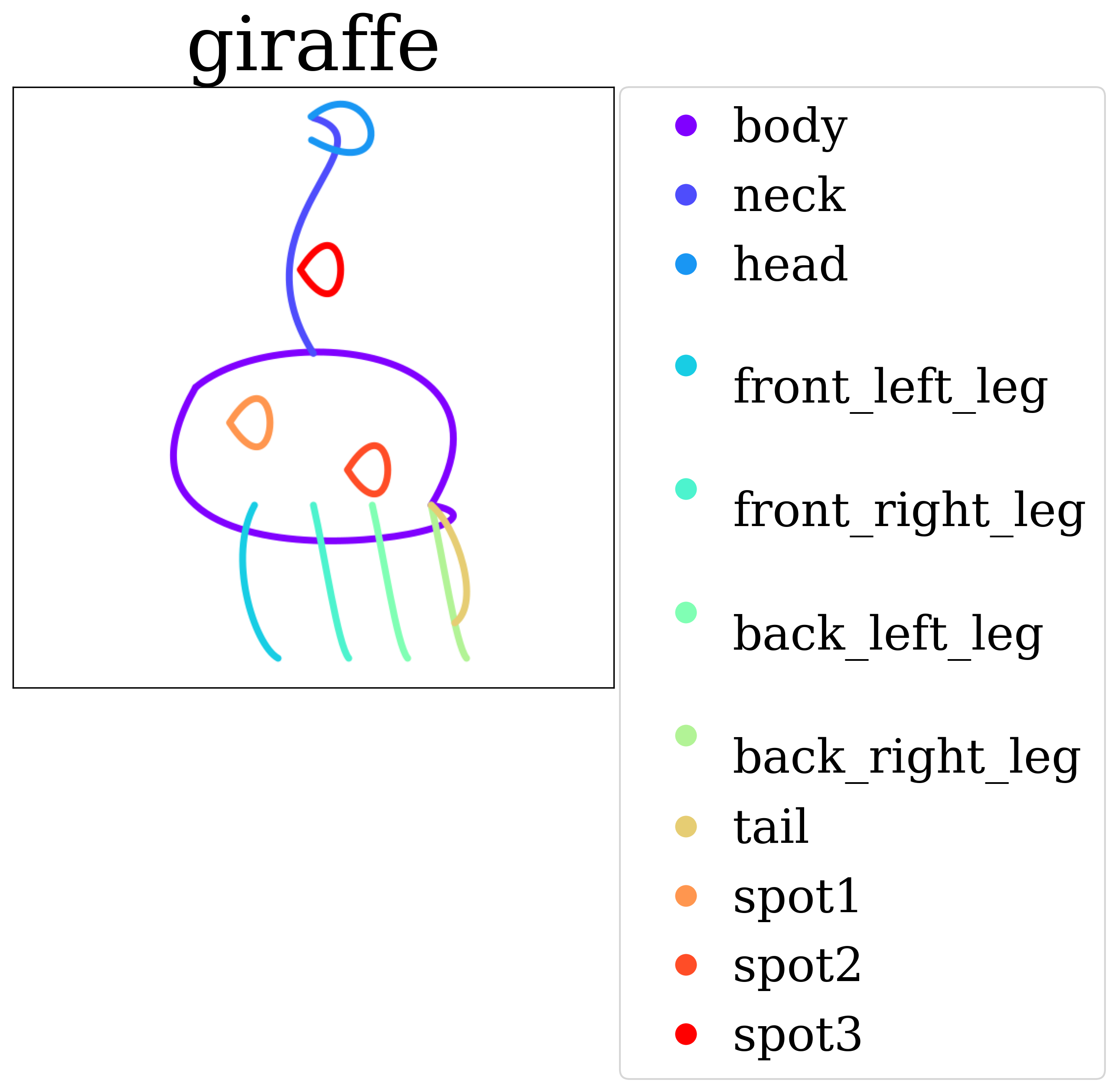} \\ 
    \end{tabular}
    }
    \caption{}
    \label{fig:annotated1}
\end{figure*}

\begin{figure*}[t]
    \centering
    \setlength{\tabcolsep}{0pt}
    {\small
    \begin{tabular}{c c c c}
    \includegraphics[width=0.25\linewidth, valign=t]{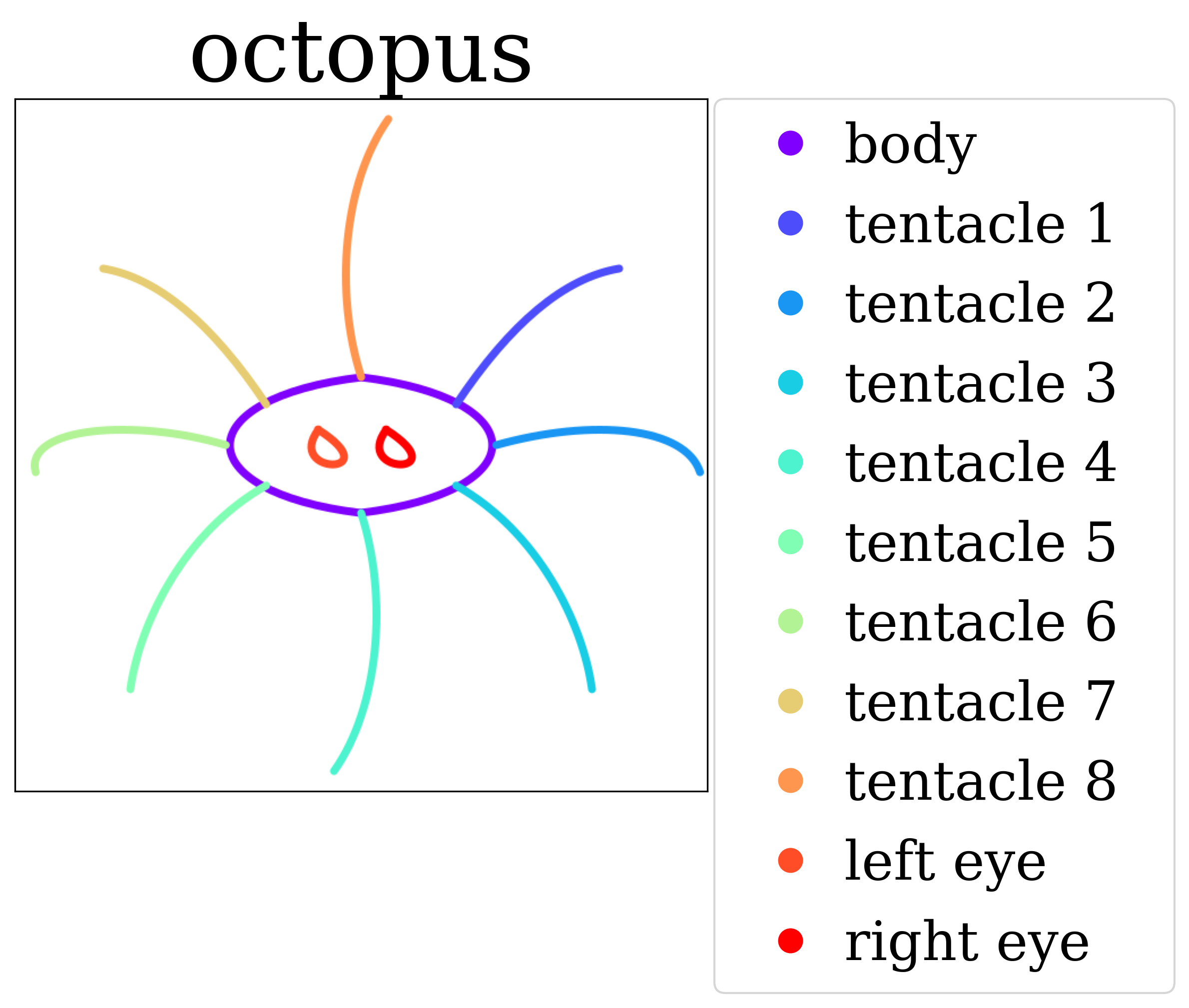} & 
    \includegraphics[width=0.25\linewidth, valign=t]{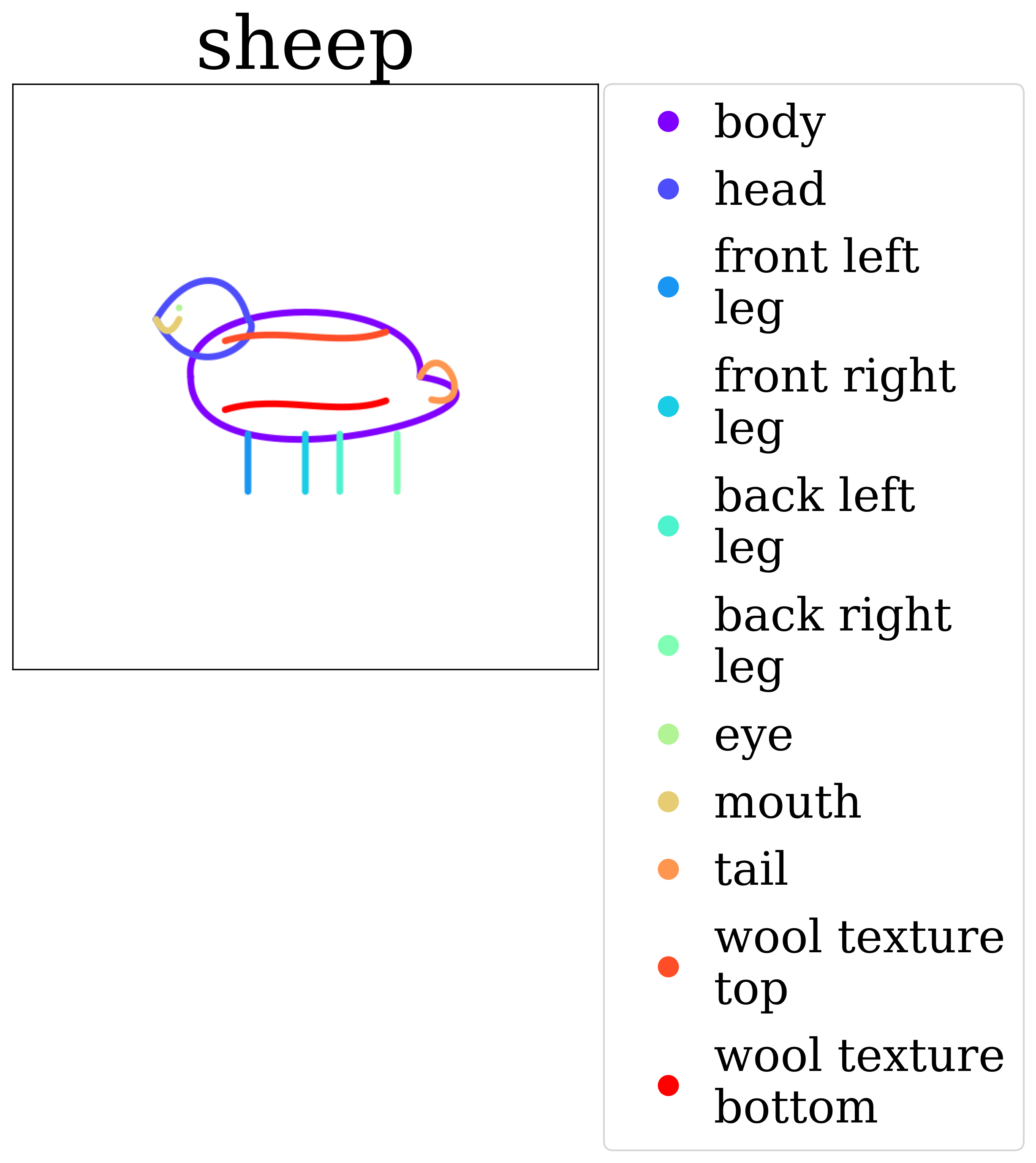} & 
    \includegraphics[width=0.25\linewidth, valign=t]{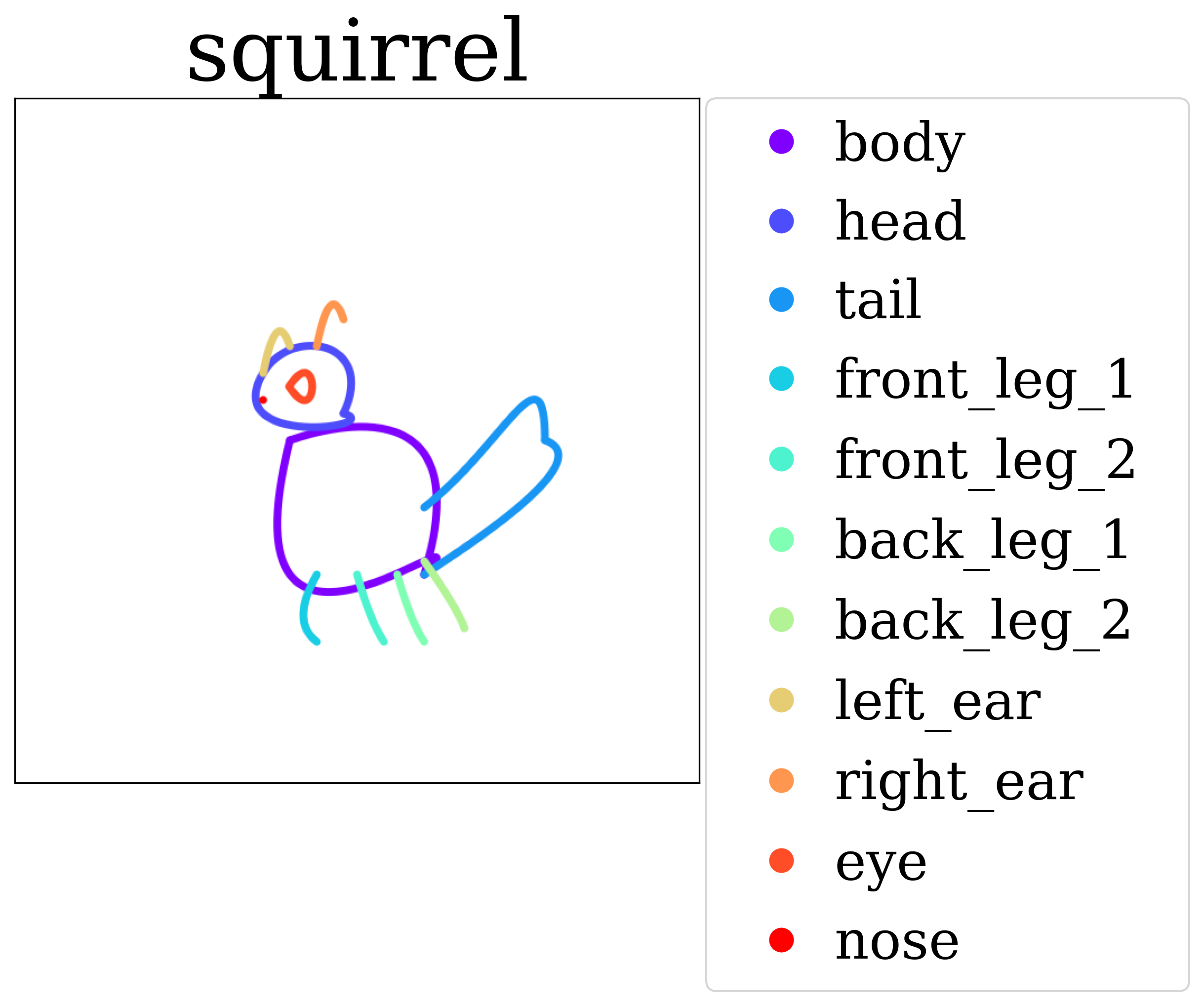} &
    \includegraphics[width=0.25\linewidth, valign=t]{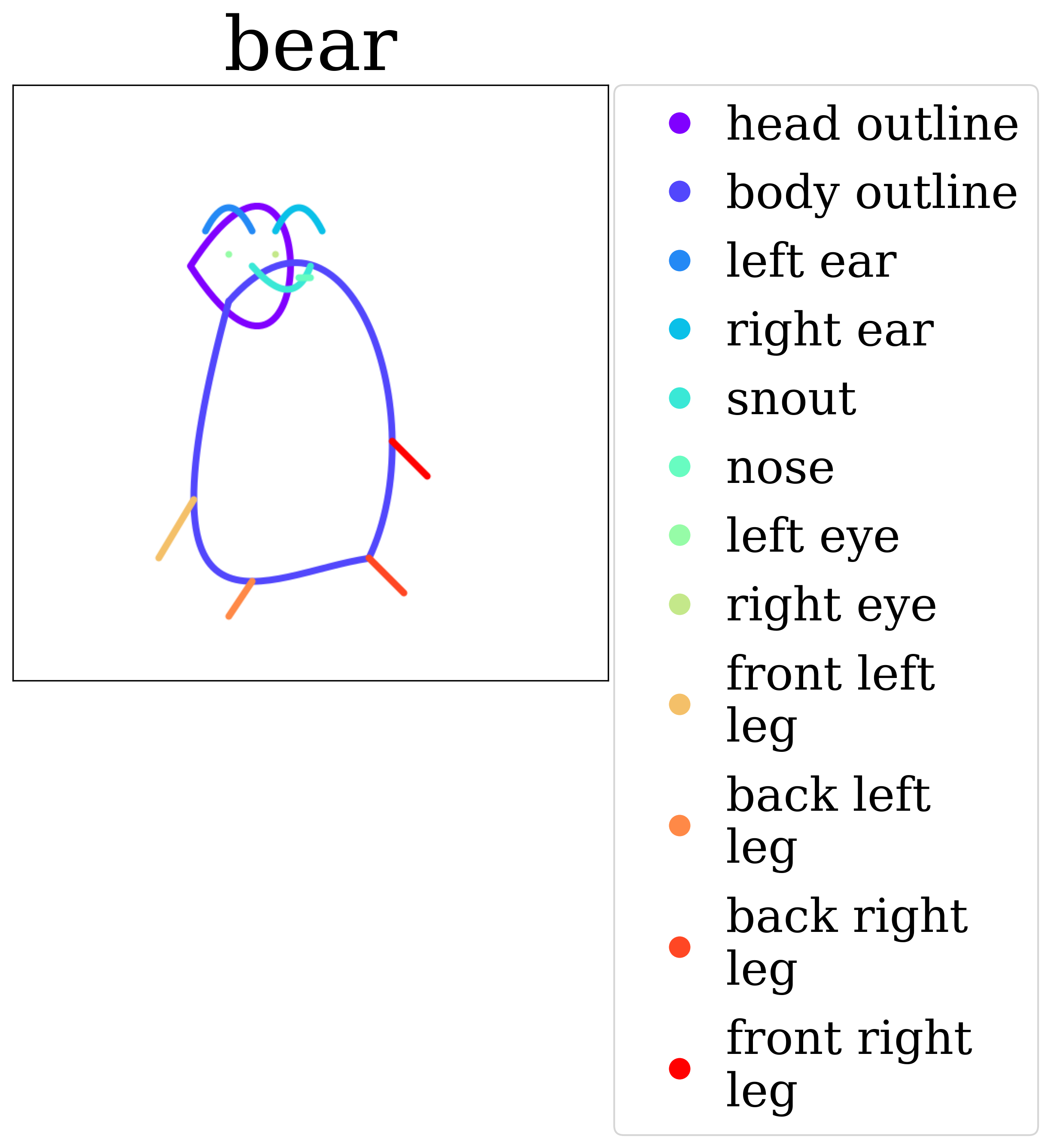} \\
    \includegraphics[width=0.25\linewidth, valign=t]{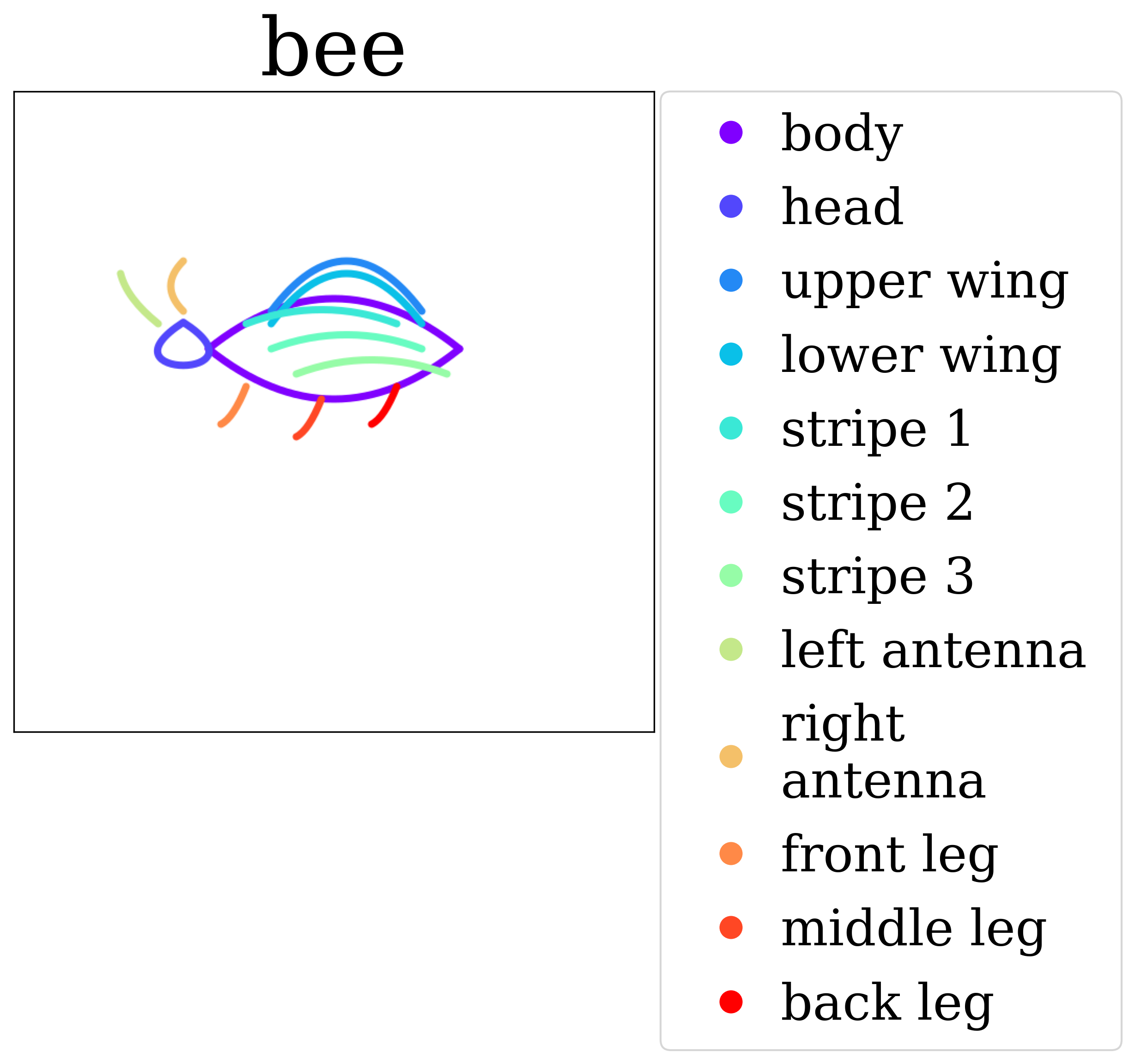} & 
    \includegraphics[width=0.25\linewidth, valign=t]{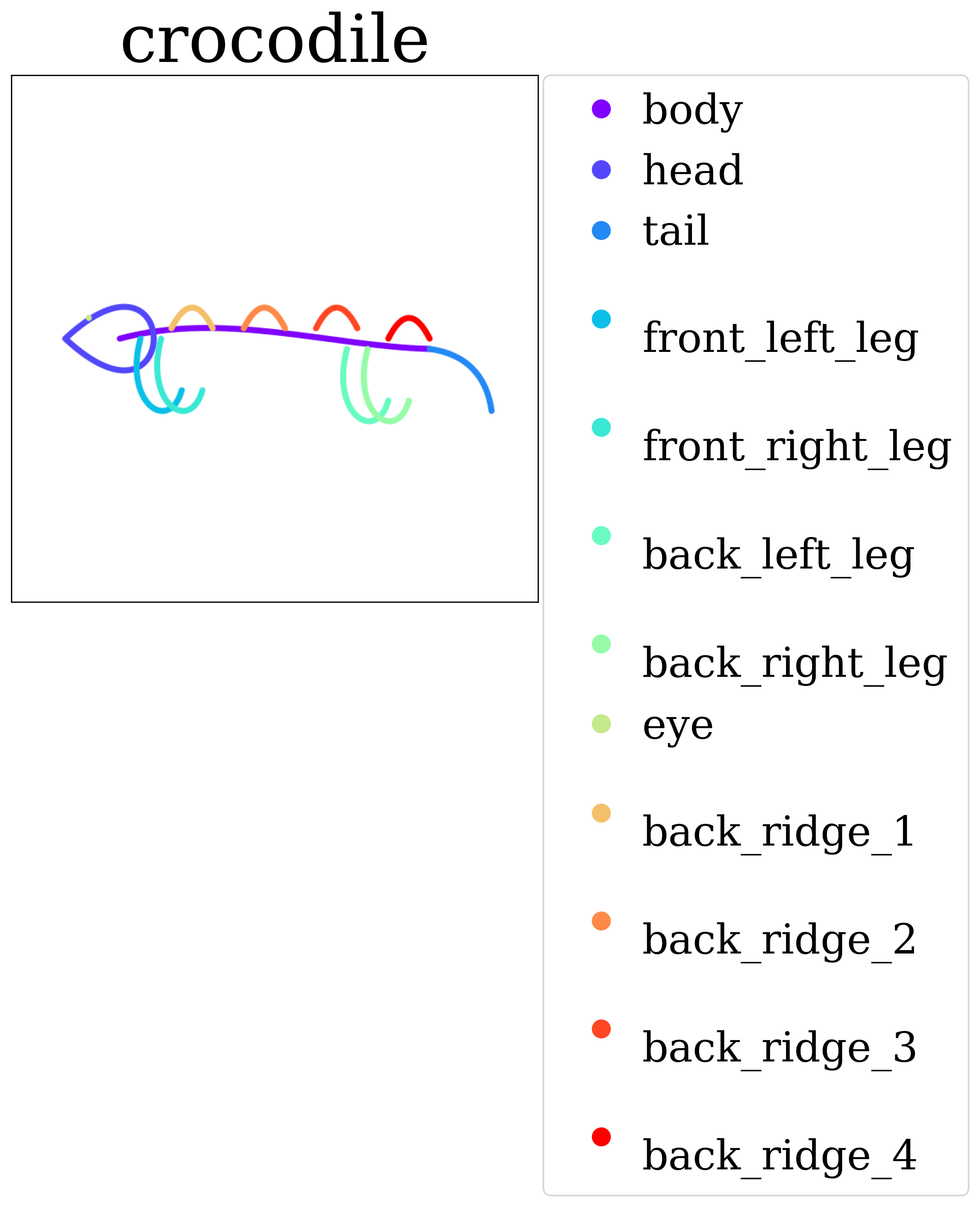} &
    \includegraphics[width=0.25\linewidth, valign=t]{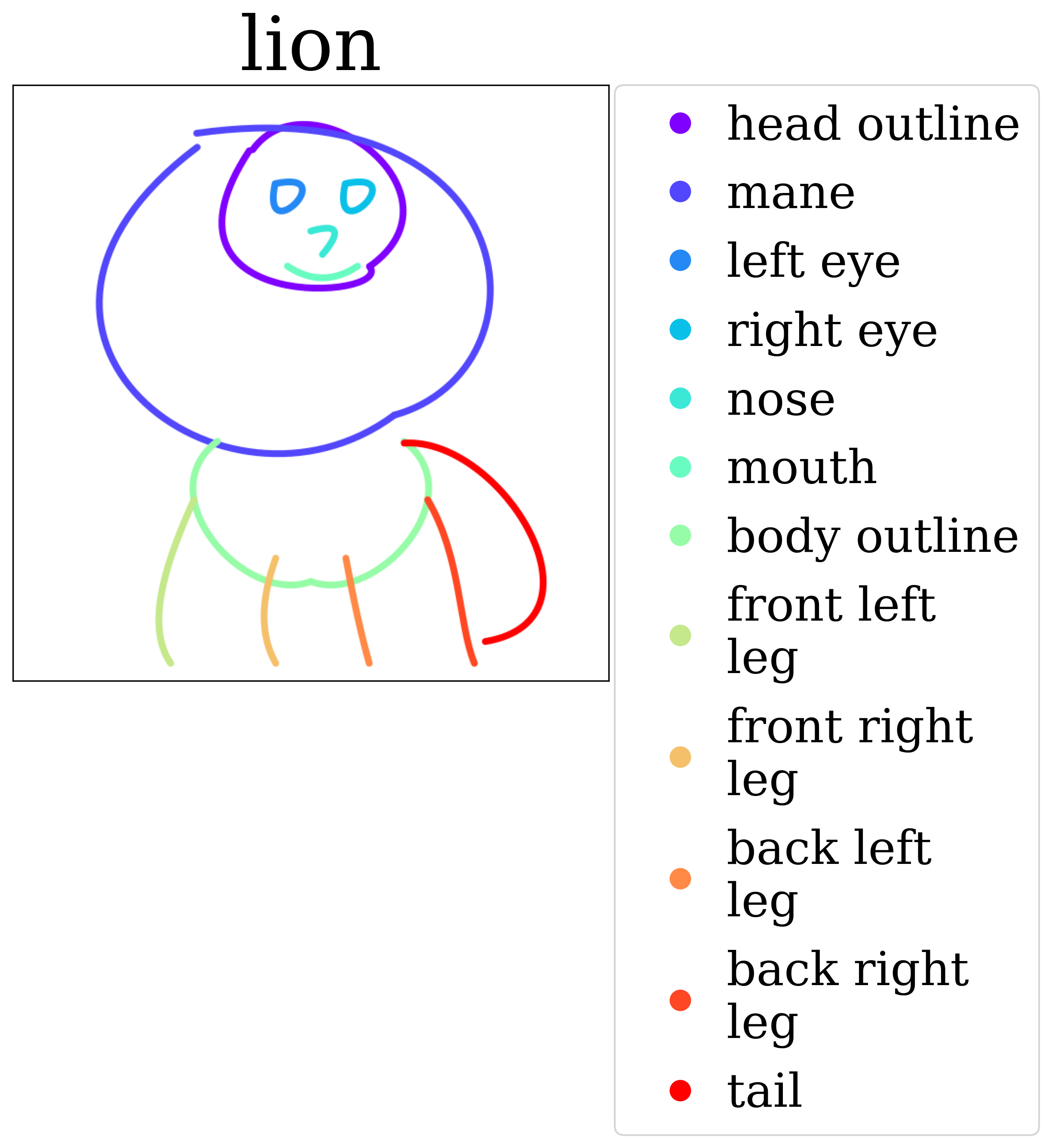} & 
    \includegraphics[width=0.25\linewidth, valign=t]{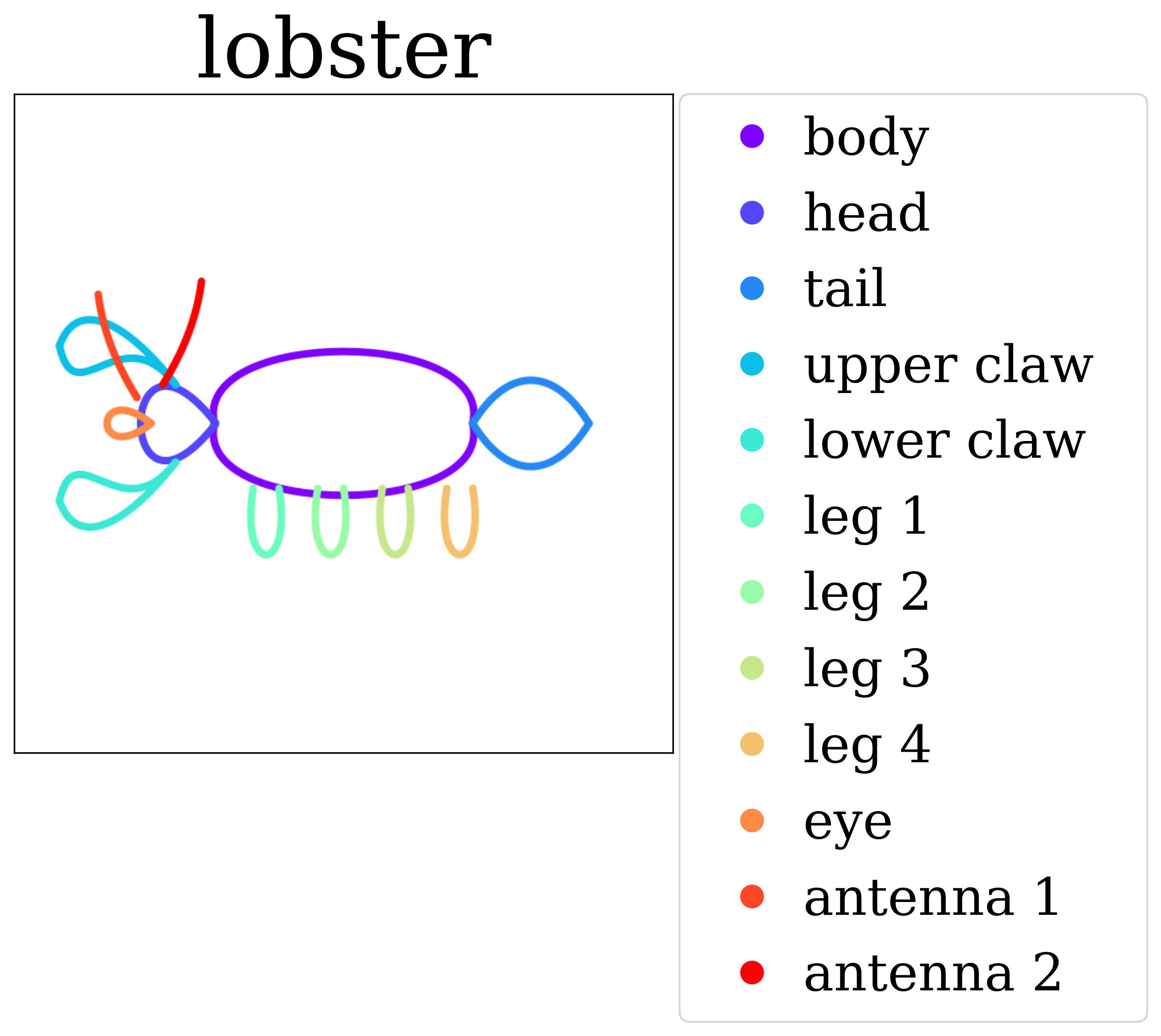} \\
    \includegraphics[width=0.25\linewidth, valign=t]{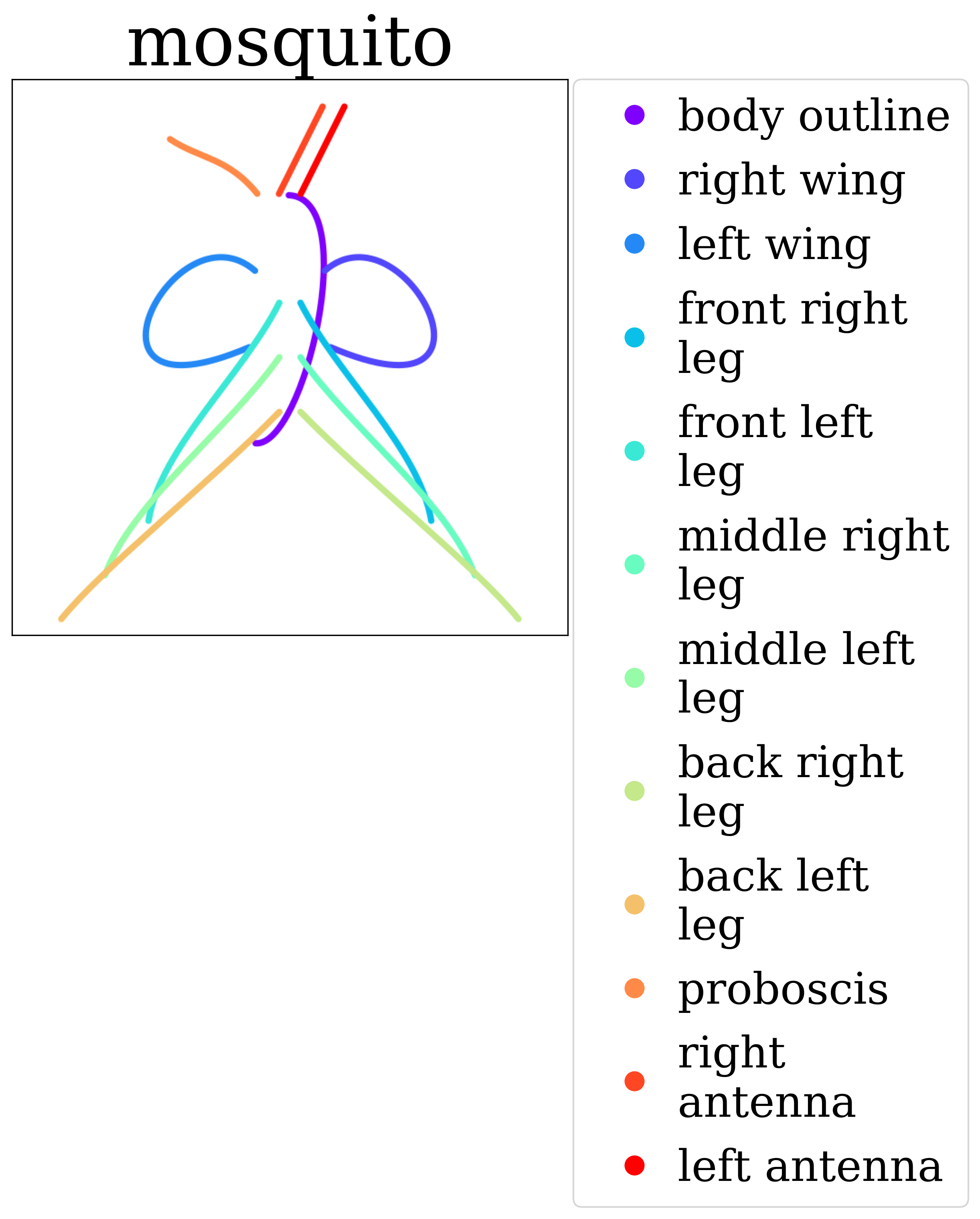} &
    \includegraphics[width=0.25\linewidth, valign=t]{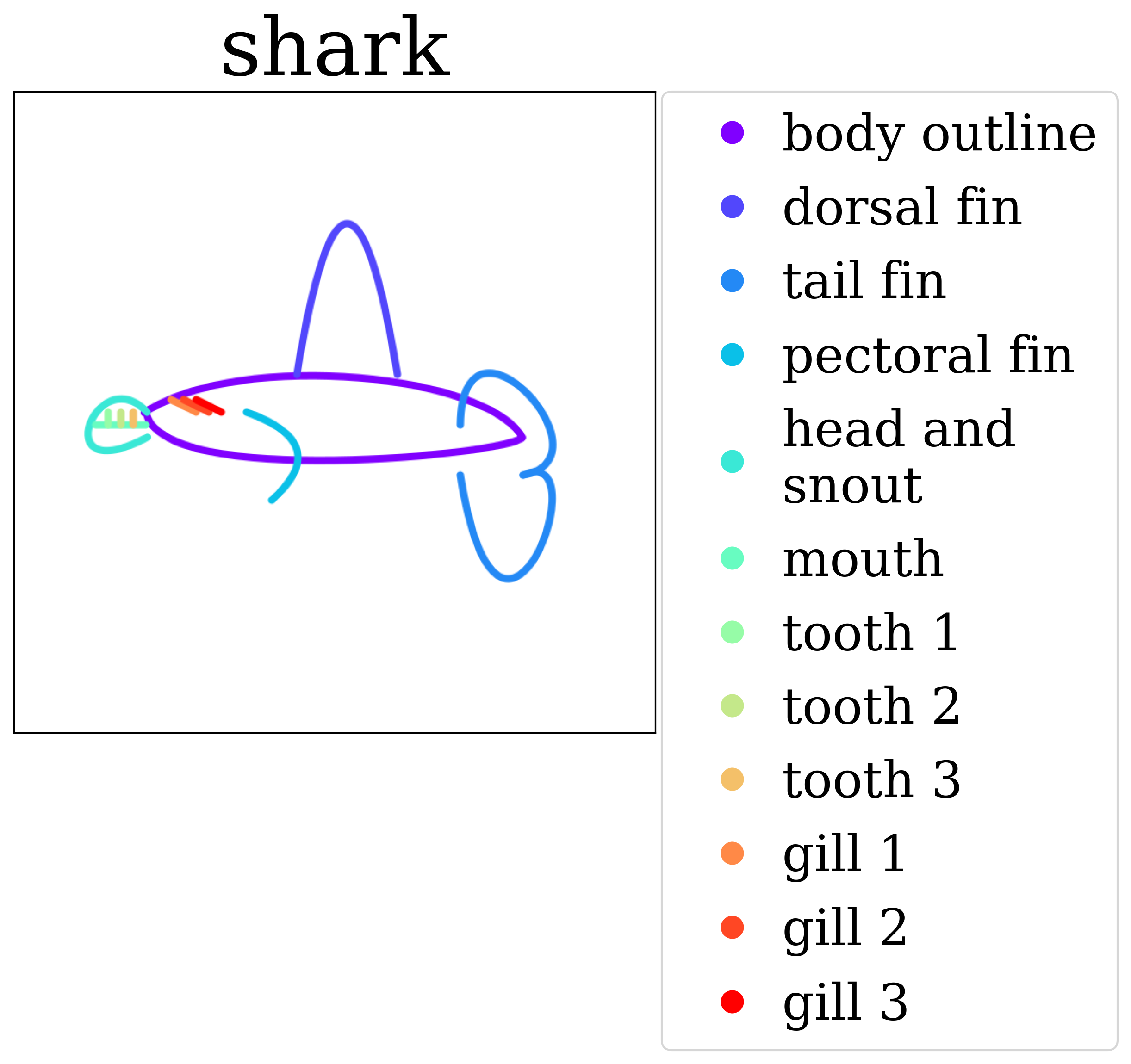} & 
    \includegraphics[width=0.25\linewidth, valign=t]{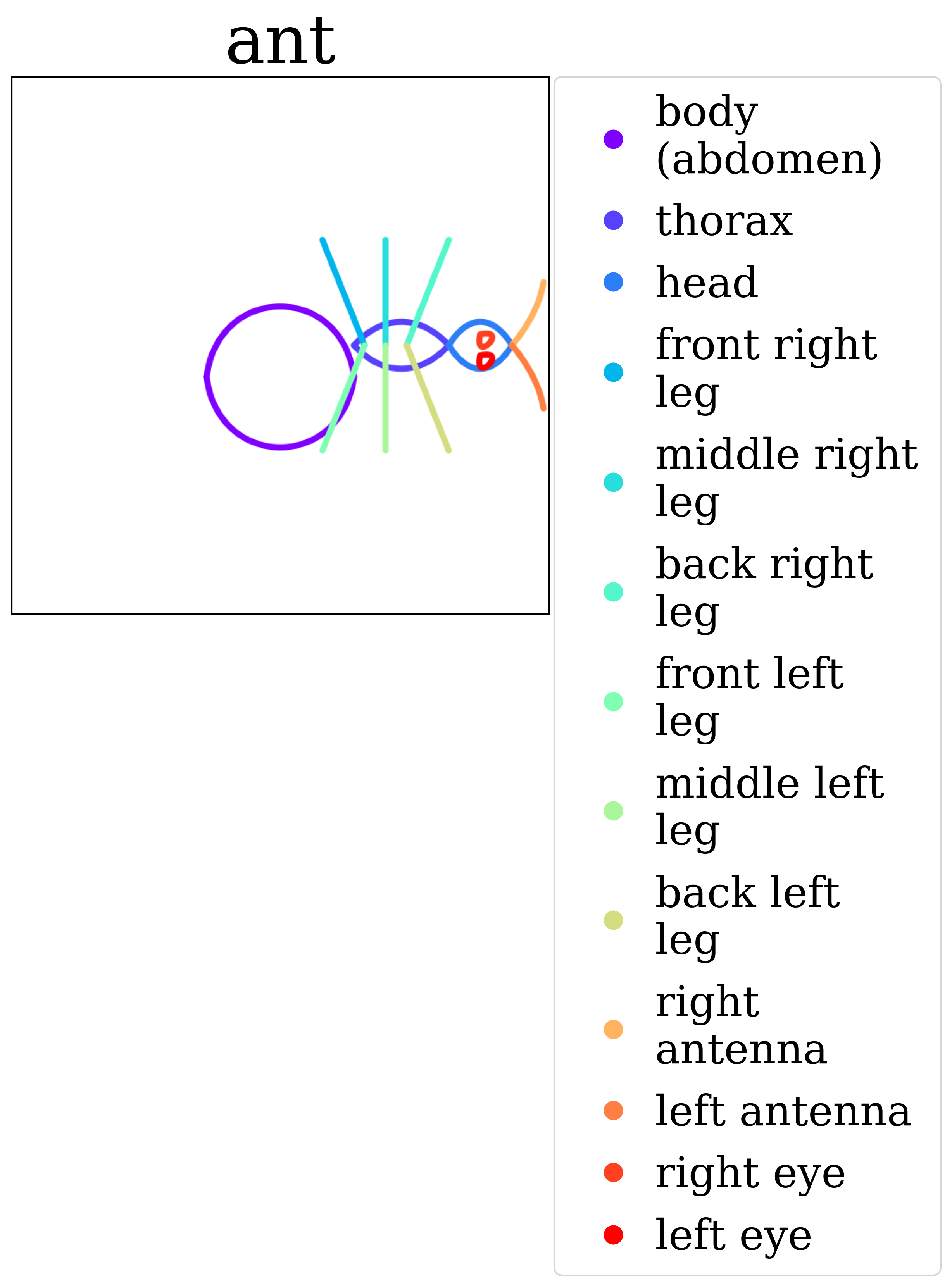} & 
    \includegraphics[width=0.25\linewidth, valign=t]{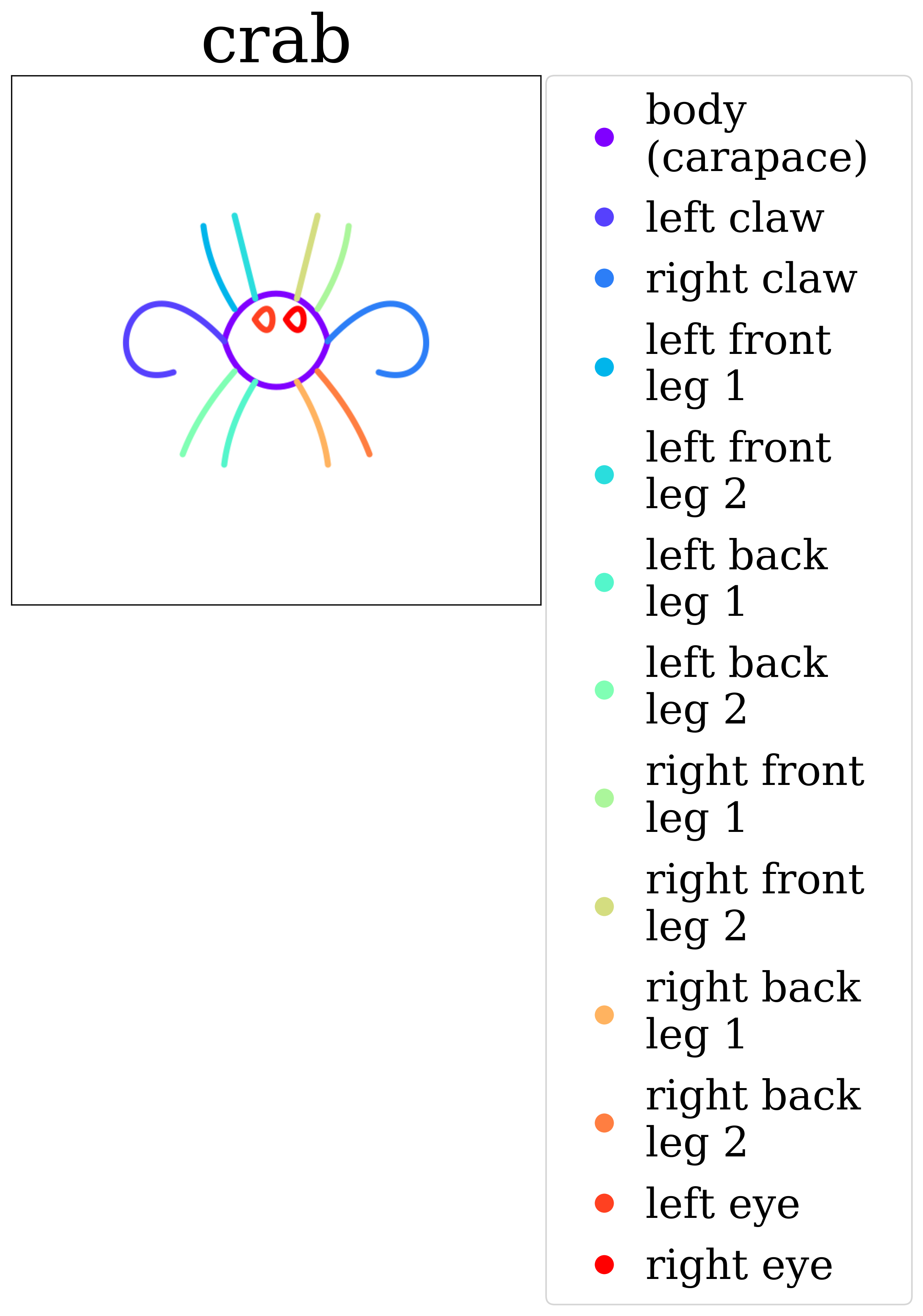} \\ 
    \includegraphics[width=0.25\linewidth, valign=t]{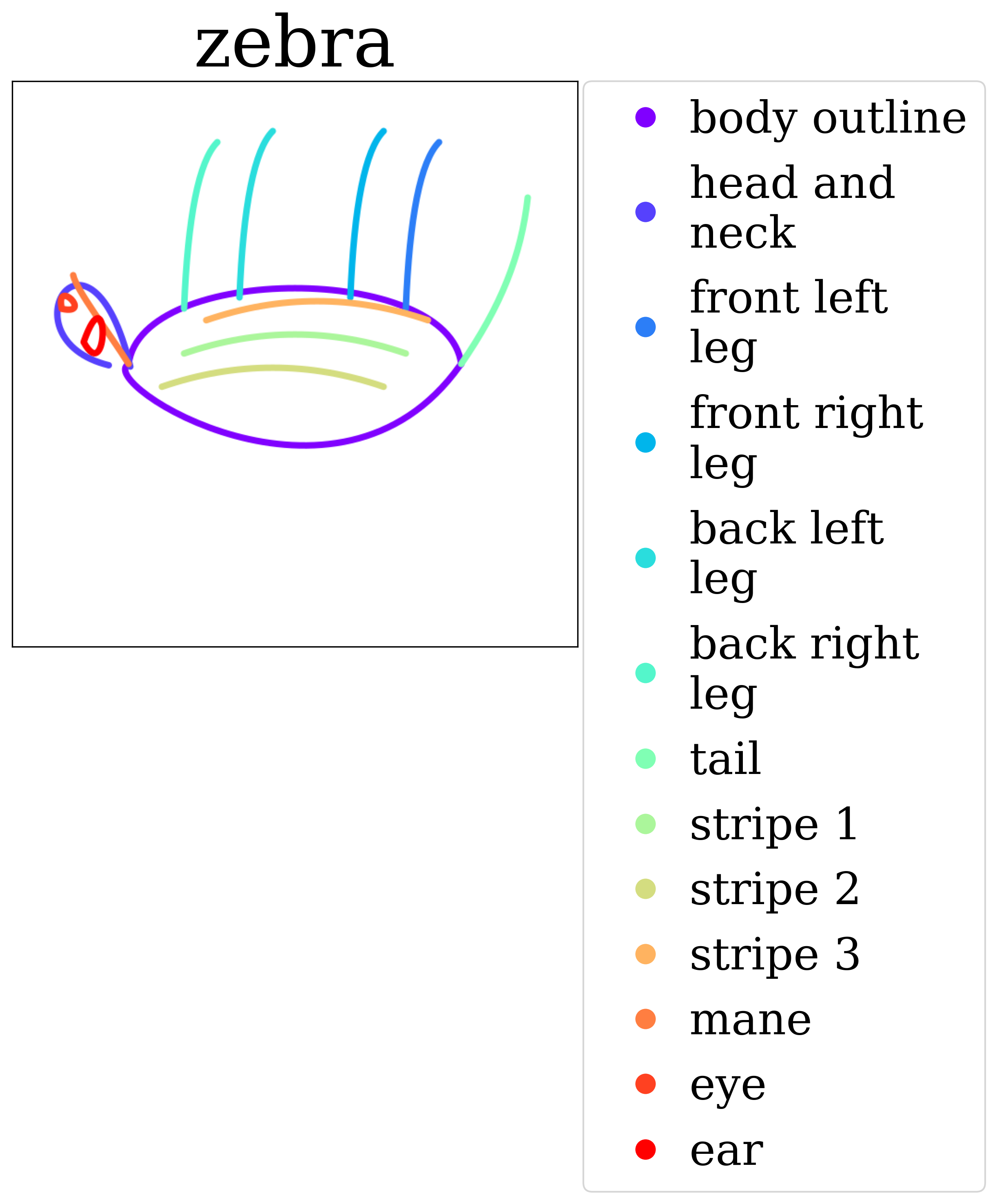} & 
    \includegraphics[width=0.25\linewidth, valign=t]{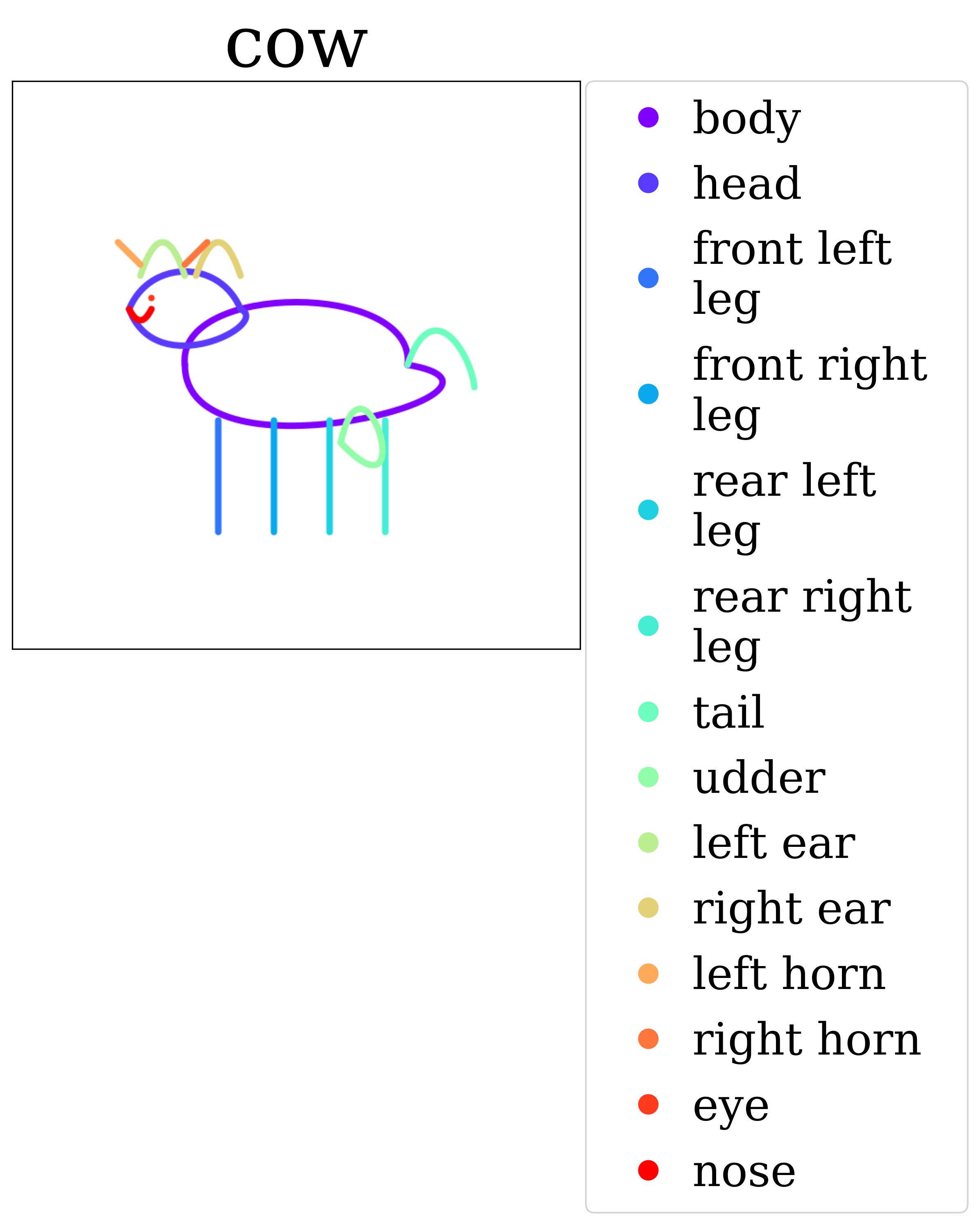} & 
    \includegraphics[width=0.25\linewidth, valign=t]{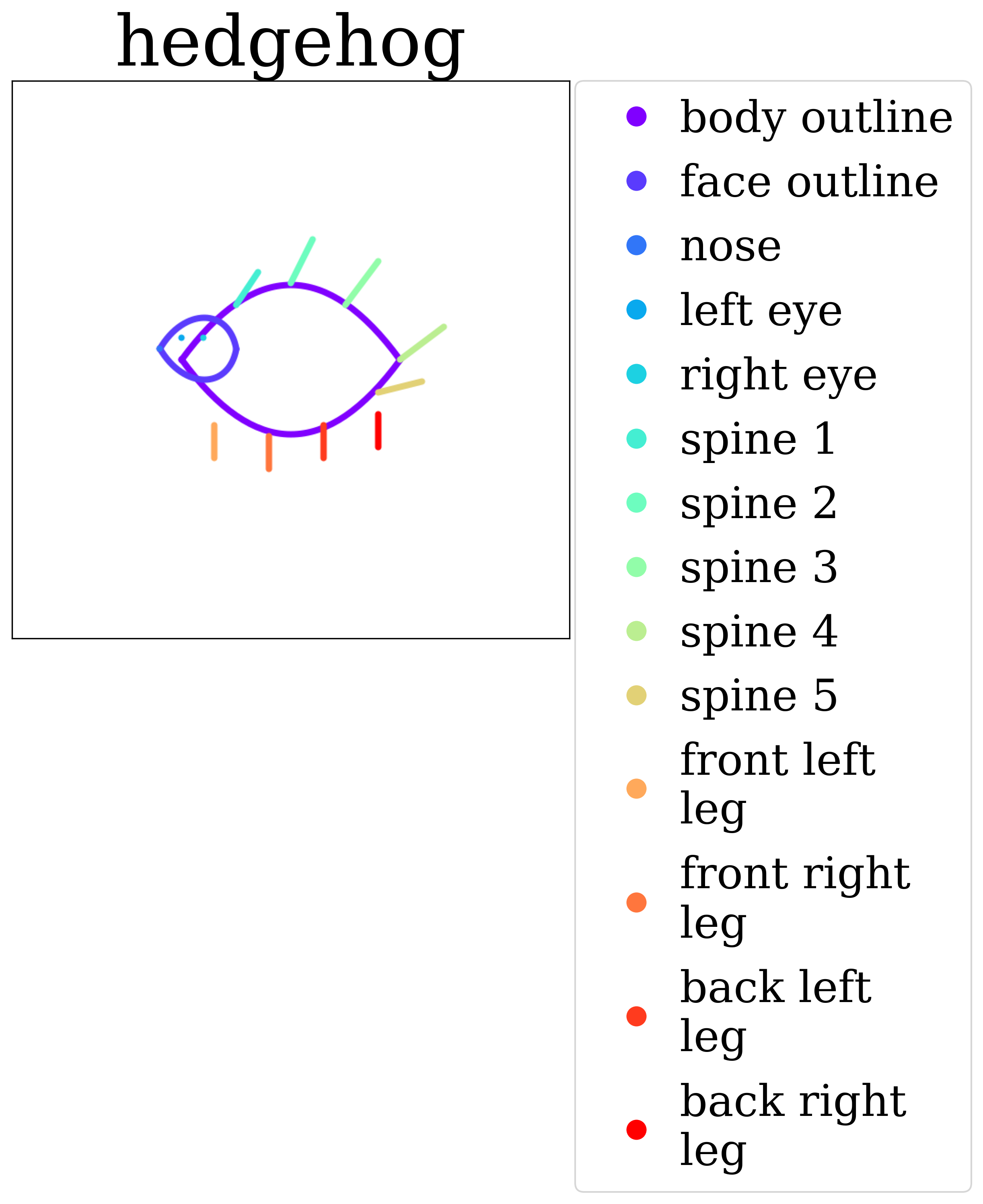} &
    \includegraphics[width=0.25\linewidth, valign=t]{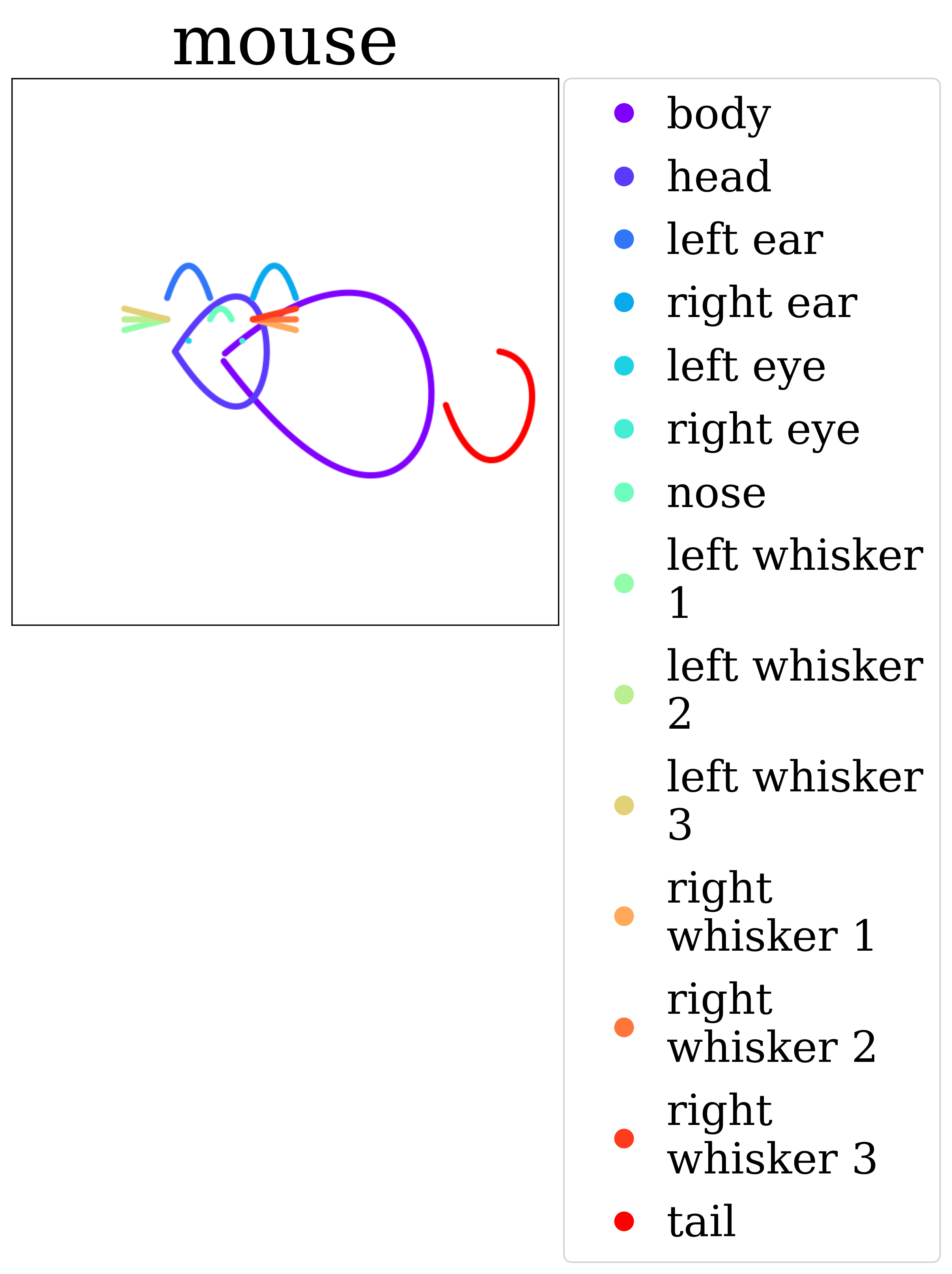} \\

    \end{tabular}
    }
    \caption{}
    \label{fig:annotated2}
\end{figure*}

\begin{figure*}[t]
    \centering
    \setlength{\tabcolsep}{0pt}
    {\small
    \begin{tabular}{c c c c}
        \includegraphics[width=0.25\linewidth, valign=t]{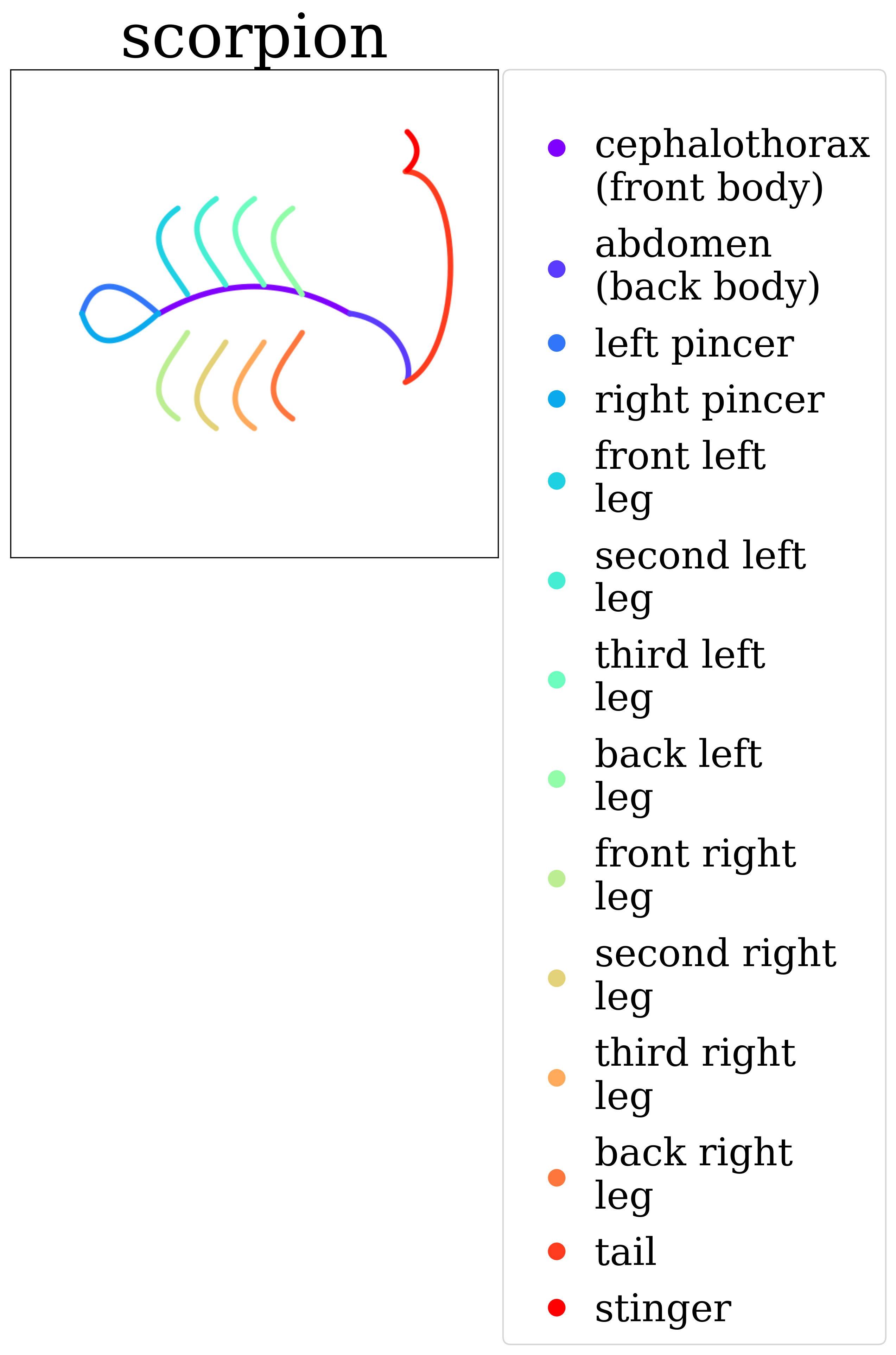} & 
        \includegraphics[width=0.25\linewidth, valign=t]{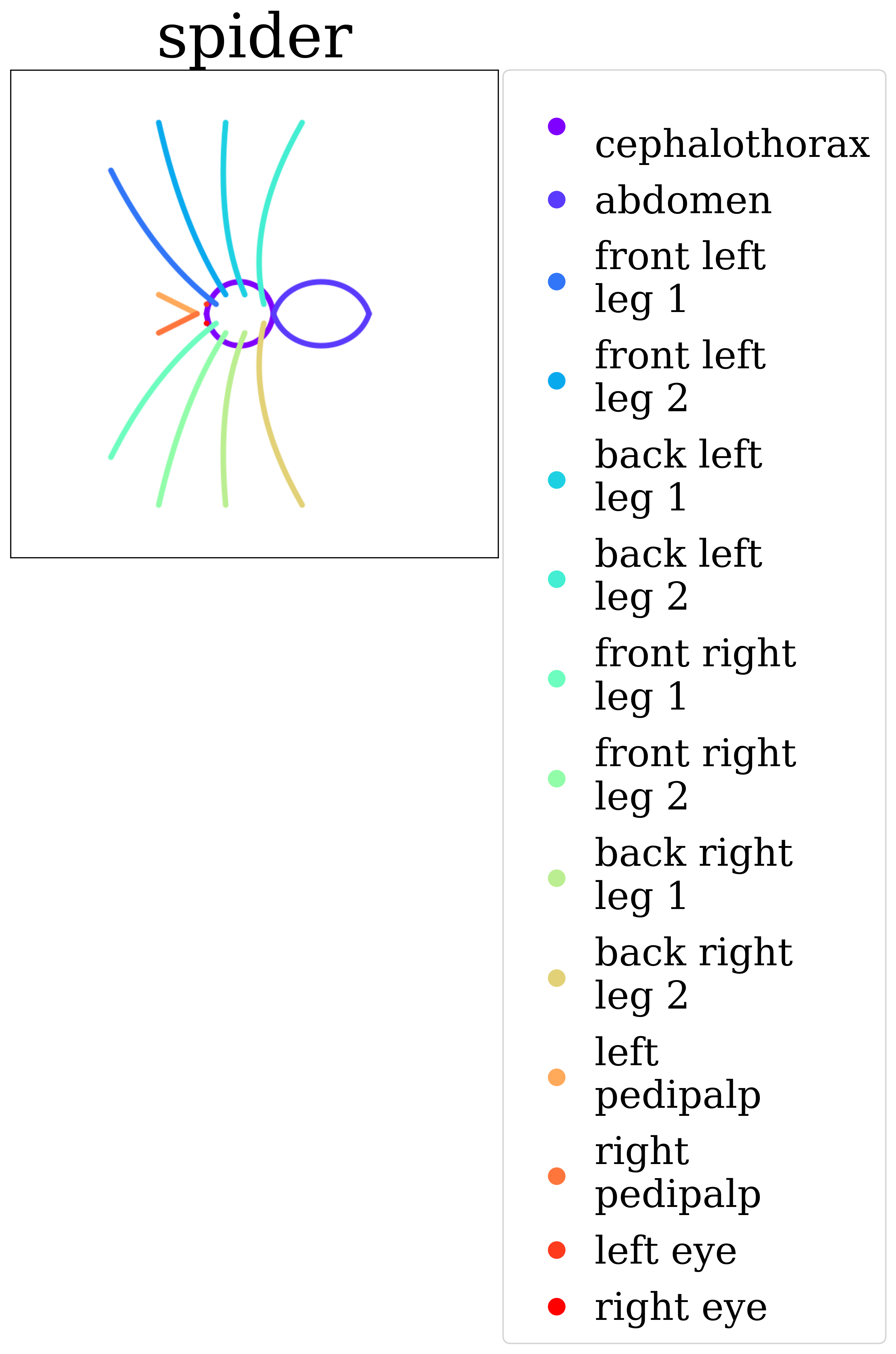} &
        \includegraphics[width=0.25\linewidth, valign=t]{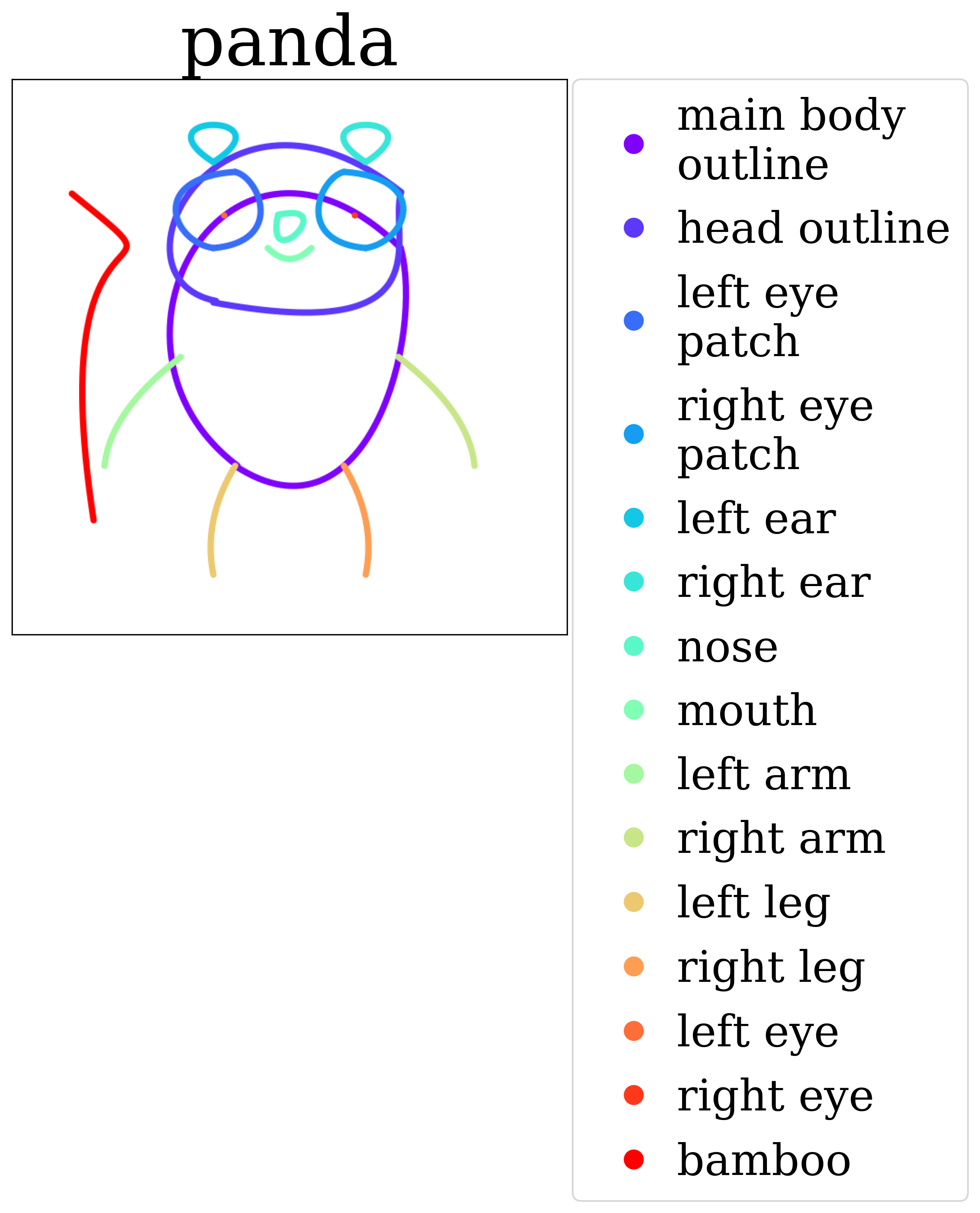} & 
        \includegraphics[width=0.25\linewidth, valign=t]{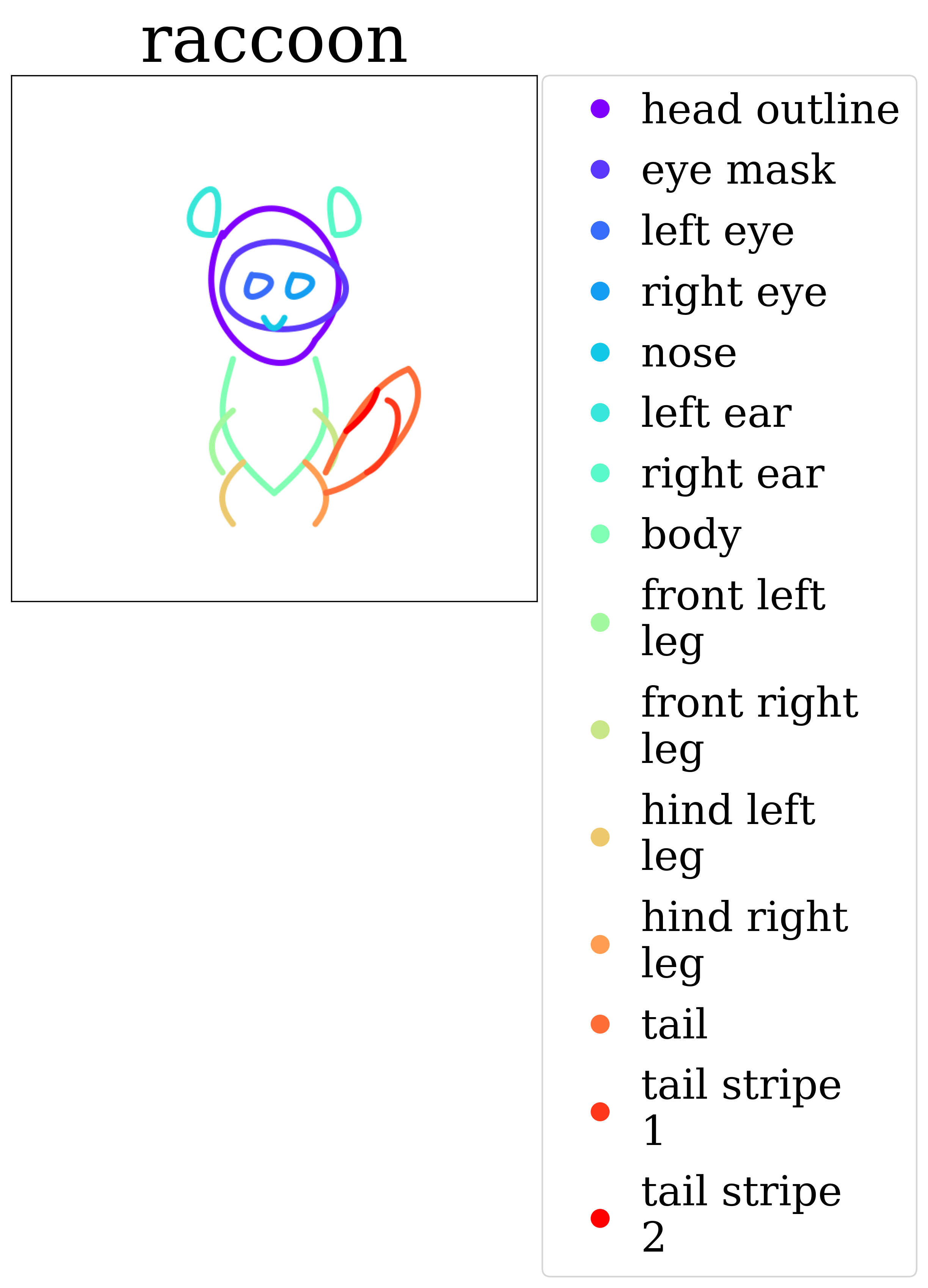} \\ 
        \includegraphics[width=0.25\linewidth, valign=t]{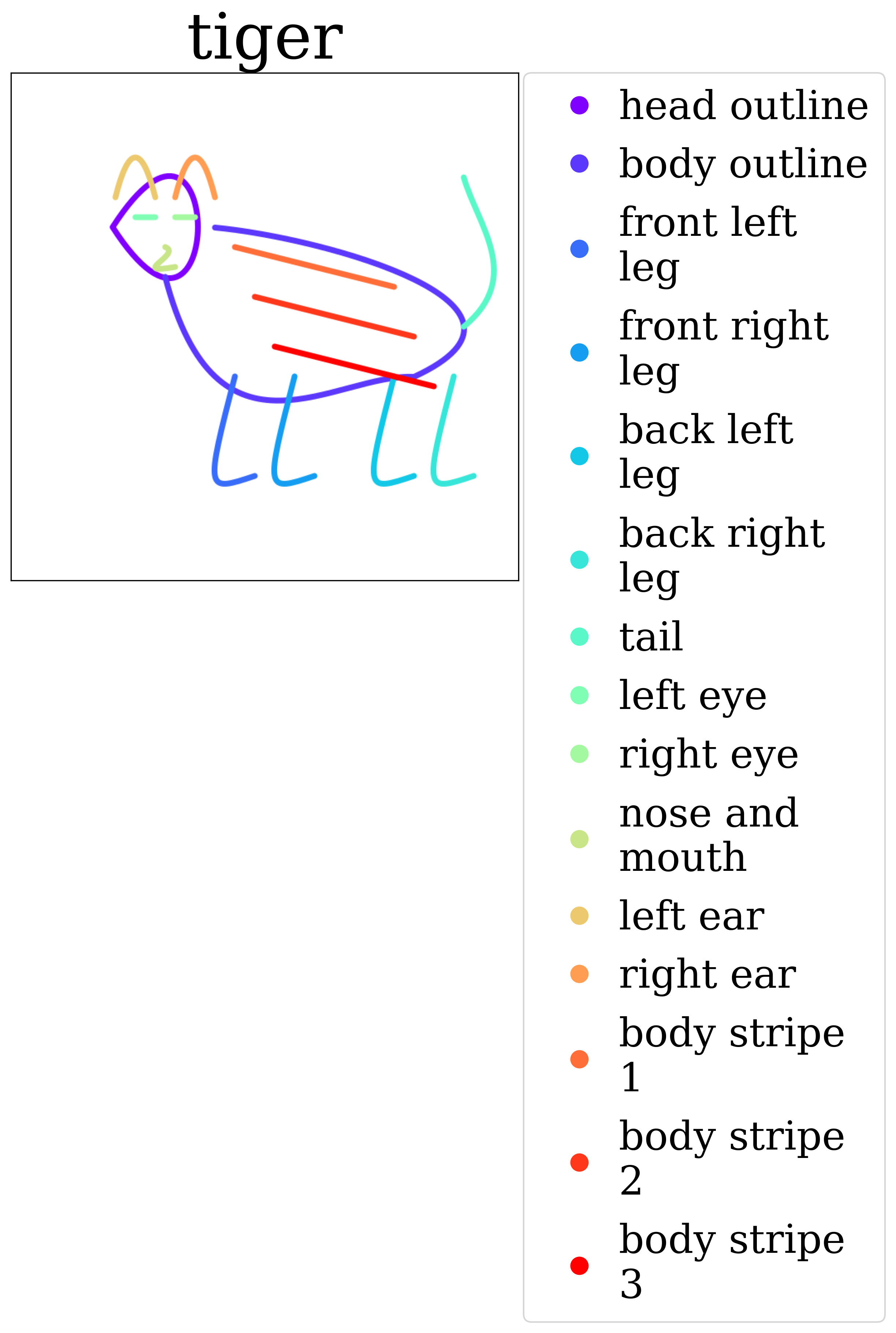} &
        \includegraphics[width=0.25\linewidth, valign=t]{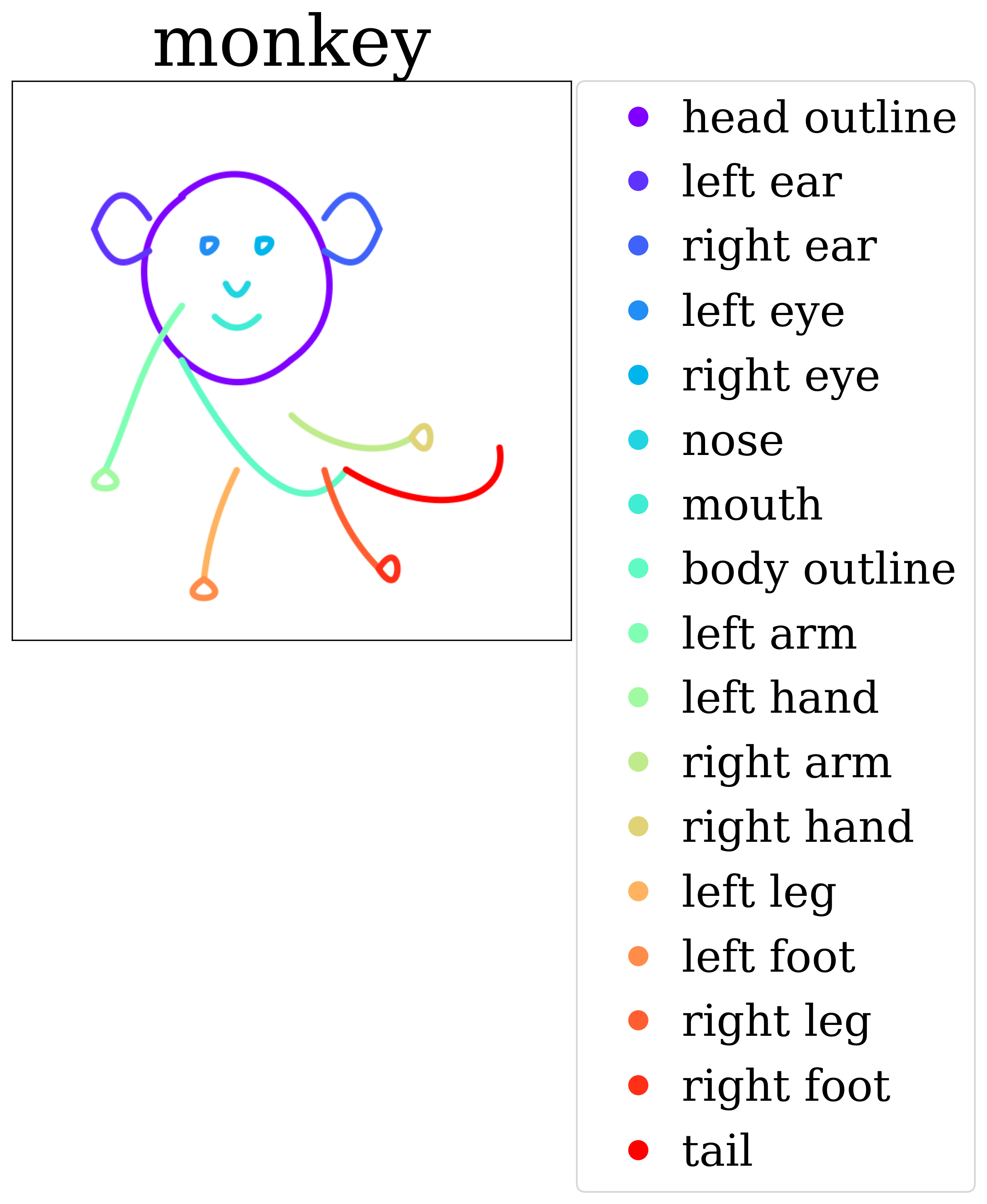} & 
        \includegraphics[width=0.25\linewidth, valign=t]{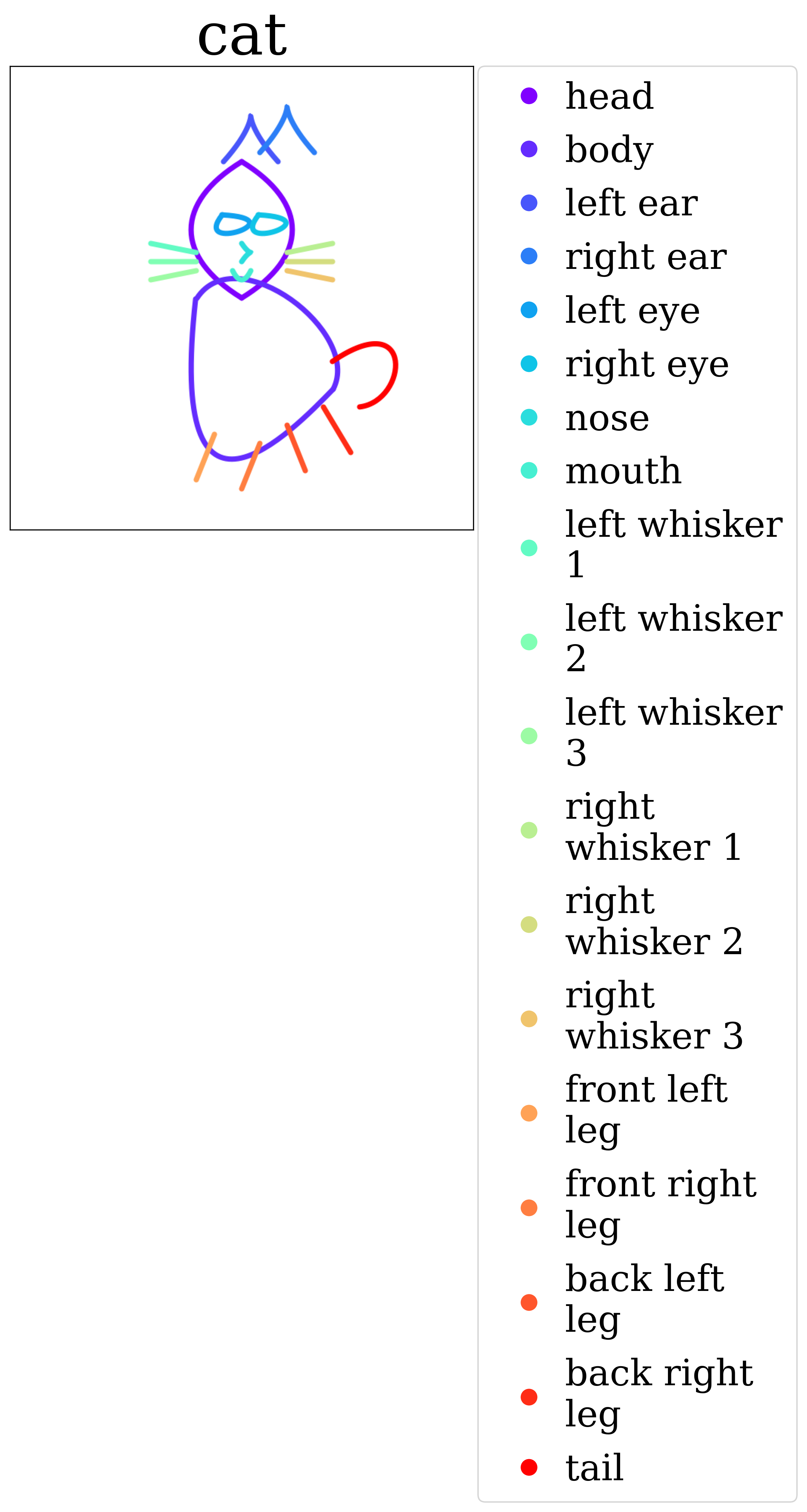} &
        \includegraphics[width=0.25\linewidth, valign=t]{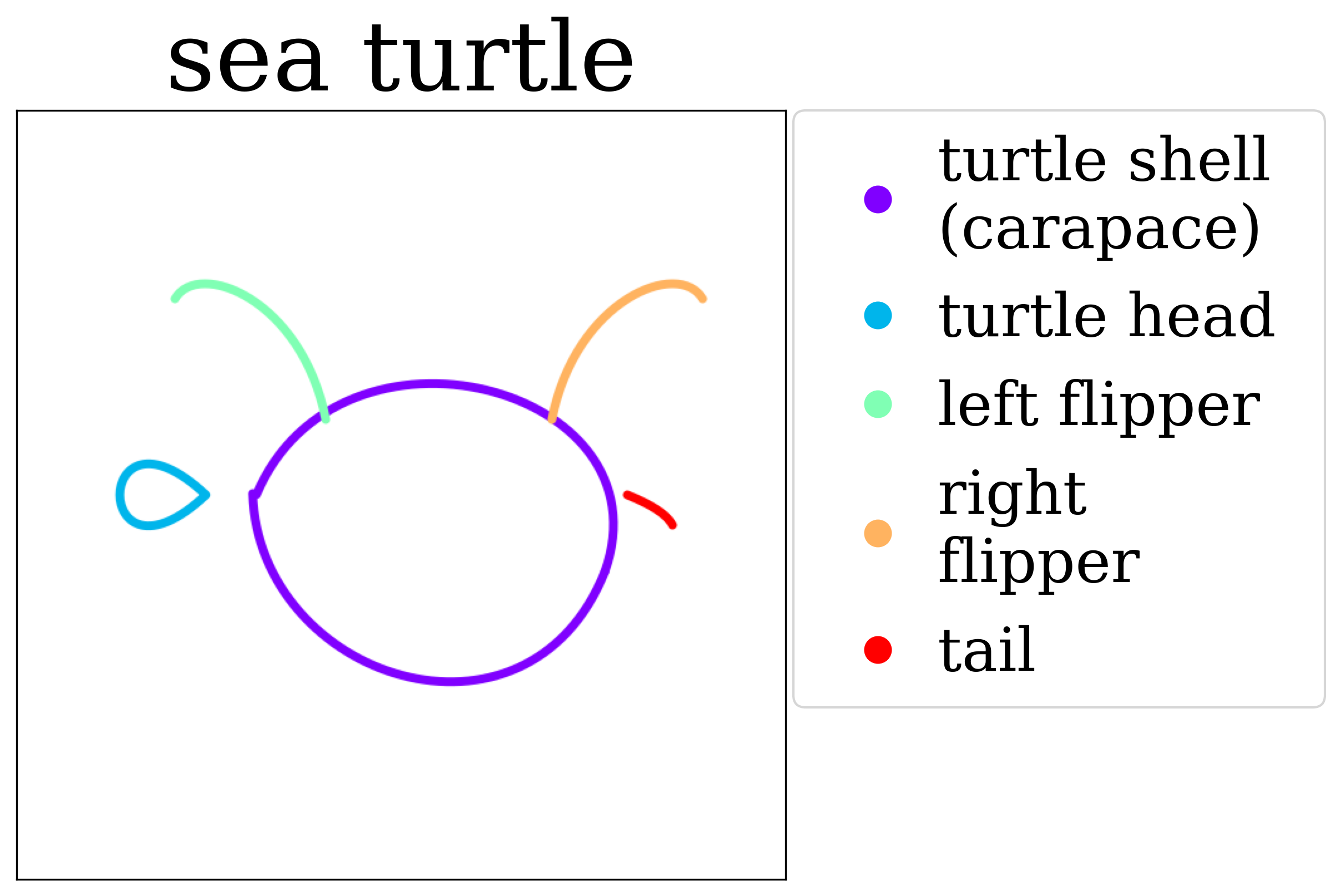}
        
    \end{tabular}
    }
    \caption{}
    \label{fig:annotated3}
\end{figure*}

In Section 5.2 of the main paper, we analyze the sequential nature of our generated sketches. 
In \cref{fig:annotated1,fig:annotated2,fig:annotated3}, we present additional visualizations of annotated sequential sketches of 48 randomly selected animals, with the presented sketches also chosen randomly. As illustrated, due to the extensive prior knowledge of the backbone model, SketchAgent provides meaningful textual annotations for each stroke and sketches in a logical order. Typically, more significant body parts, such as the head and body, are drawn first.
We next provide more details and visualizations of the quantitative analysis shown in Figure 11 of the main paper.
\Cref{fig:stroke-distribution} displays histograms of the number of strokes in QuickDraw sketches (top) and our generated sketches (bottom), as shown in the main paper. Alongside these histograms, we include visualizations of sketches drawn with 1, 4, 7, and 15 strokes. Notably, in the QuickDraw dataset, single-stroke sketches often consist of a single long continuous line, making them recognizable after the first stroke. In contrast, sketches with a larger number of strokes rarely feature long continuous lines. For such cases, the sequential process of adding strokes gradually makes the sketches recognizable after several strokes. \Cref{fig:4-strokes-set,fig:5-strokes-set,fig:6-strokes-set,fig:7-strokes-set} also demonstrates the sequential sketching process for both QuickDraw sketches and those generated by our method, providing a visual context for the trends observed in Figure 11.

\clearpage
\newpage

\subsection{Human-Agent Collaborative Sketching}
\begin{figure}
    \centering
    \includegraphics[width=0.8\linewidth]{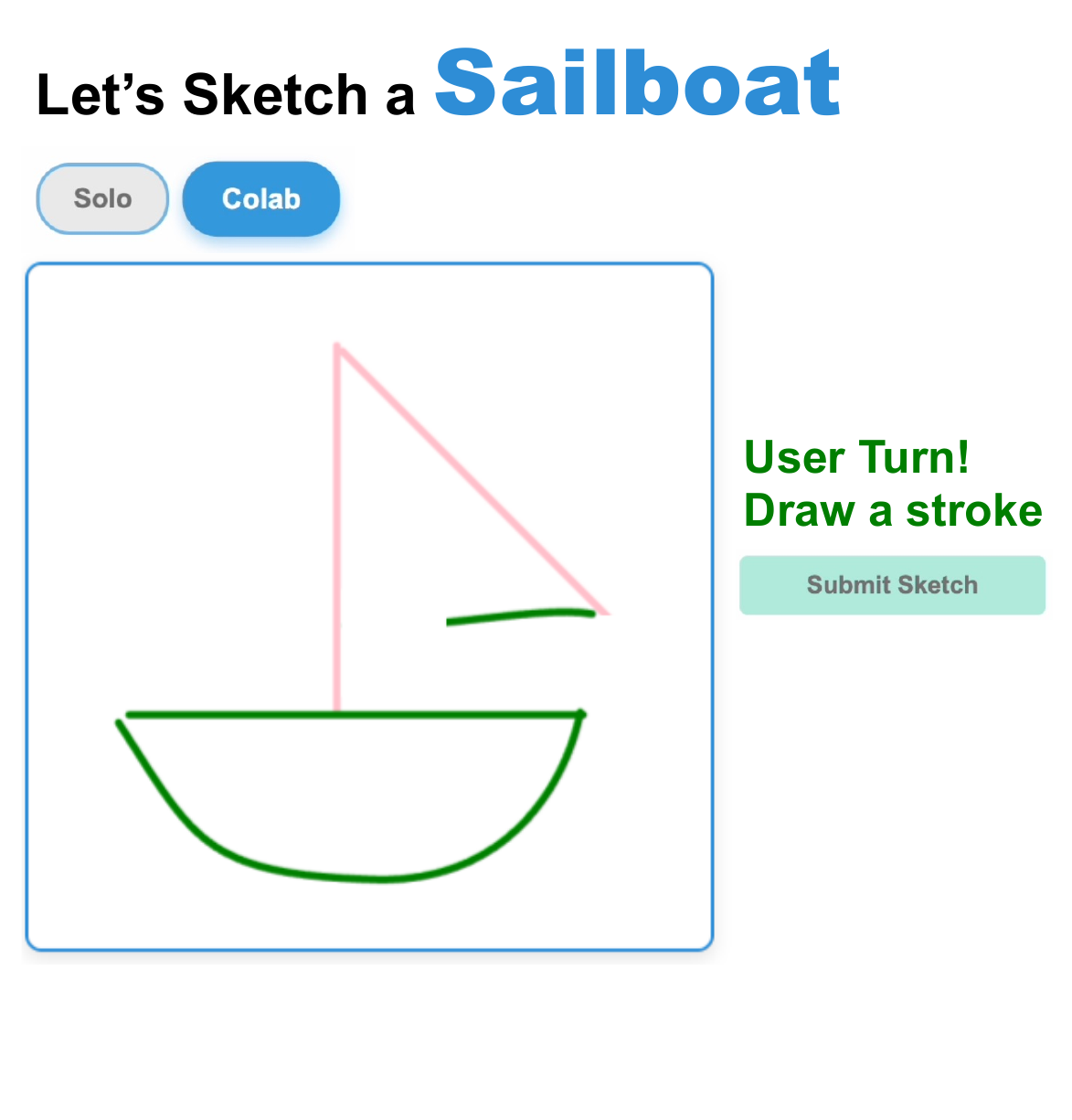}
    \vspace{-1cm}
    \caption{Screenshot of our web interface.}
    \label{fig:interface}
\end{figure}

\begin{figure}
    \centering
    \includegraphics[width=1\linewidth]{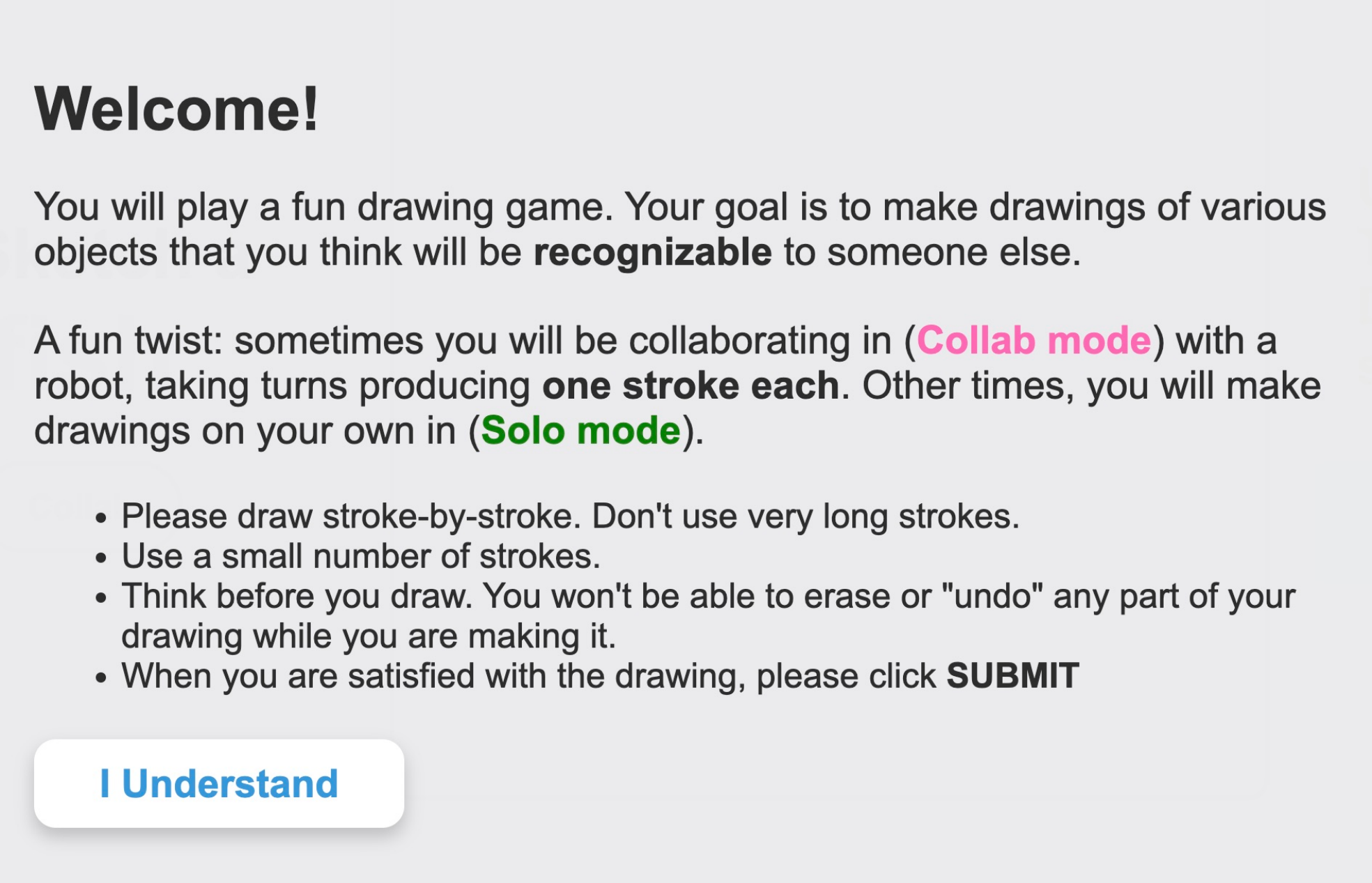}
    \caption{User instructions in the sketching interface.}
    \label{fig:collab-instructions}
\end{figure}

In Section 5.3 of the main paper, we demonstrate that humans and SketchAgent can effectively collaborate to produce meaningful sketches through genuine interaction.
The sketching interface (\cref{fig:interface}) consists of a $400\times400$ plain canvas shared between the user and the agent. It highlights the current concept to be sketched and displays the active sketching mode, which can be either \textit{solo} or \textit{collab}. Additionally, the interface includes a submit button that allows users to finalize the sketch when they consider it complete. In \textit{solo} mode, users independently sketch the given concept using green strokes. In \textit{collab} mode, users and SketchAgent take turns adding strokes, with user strokes displayed in green and agent strokes in pink.
At the beginning of each session, users are provided with general instructions about the experiment and the types of sketches they will be asked to draw (\cref{fig:collab-instructions}). Specifically, they are instructed to create recognizable sketches, stroke by stroke, while minimizing the number of strokes by planning ahead.
Next, users begin by sketching two warm-up concepts in both \ap{solo} and \ap{collab} modes to familiarize themselves with the web environment. Each session includes all eight primary concepts in a randomized order, resulting in a total of 10 sketches per user (including the two warm-up sketches). The concepts are as follows:
\begin{itemize}
    \item \textbf{Warm up concepts}: \textit{jellyfish, house}
    \item \textbf{Text concepts}: \textit{butterfly, fish, rabbit, duck, sailboat, coffee mug, eyeglasses, car}
\end{itemize}

For each concept, participants sketched in both solo and collaboration modes, with the order of these modes randomized to mitigate potential biases. The 30 users are counterbalanced: 15 users produced the first stroke in collaboration with the agent (and all odd-numbered strokes thereafter), while the other 15 users produced the second stroke in collaboration with the agent (and all even-numbered strokes). In total we collected responses from 32 users, however, two users were excluded from the analysis due to incomplete sketching sessions, leaving a total of 30 users. 
In \cref{fig:ex_all_sketches} we present examples of sketches from each mode, focusing on those with high recognition rates across categories. Solo sketches are shown in green, agent-only sketches in pink, and collaborative sketches are depicted with a combination of both colors.
To analyze \ap{collab} and \ap{solo} sketches, we rendered all complete and partial sketches (agent-only and user-only strokes) from SVG to pixel images. We then utilized a CLIP zero-shot classifier, as described in the main paper, to evaluate how effectively each sketch represented the intended concept.
\cref{tab:collab-table} summarizes the results (as shown in the graph in Fig. 12B of the main paper). These results highlight that both users and the agent contributed meaningfully to the final \ap{collab} sketches. Variants of collaborative sketches containing only the user's strokes or only the agent's strokes were found to contain substantially less semantic information about the intended concept compared to the complete collaborative sketches.
Additionally, the average number of strokes per completed sketch was consistent across modes, indicating similar levels of complexity. Specifically, the average stroke counts were as follows: collaborative full sketches: 7.333; solo agent sketches: 7.321; solo user sketches: 7.708. This suggests that collaboration produces sketches with a level of detail comparable to those created independently.

We analyze the classification confusion patterns for collaborative and solo sketches (240 sketches each) in \cref{fig:collab_confusion,fig:solo_user_confusion}, revealing similar trends. For instance, a \ap{coffee cup} was often misclassified as a \ap{teapot}, a \ap{car} was frequently identified as a \ap{turtle}, and a \ap{duck} was misclassified as a \ap{bird}. Additionally, \ap{car} sketches were sometimes mistaken for a \ap{pickup truck}.
In most cases, the misclassifications occur within closely related categories (e.g., \ap{car} to \ap{truck} or \ap{pickup truck}) or among categories sharing similar visual structures (e.g., the rounded dome and four base components of a \ap{car} resembling a \ap{turtle}). This highlights a challenge in emphasizing distinctive features within specific categories, likely stemming from the inherently abstract nature of our sketches.

\begin{figure}
    \centering
    \includegraphics[width=\linewidth]{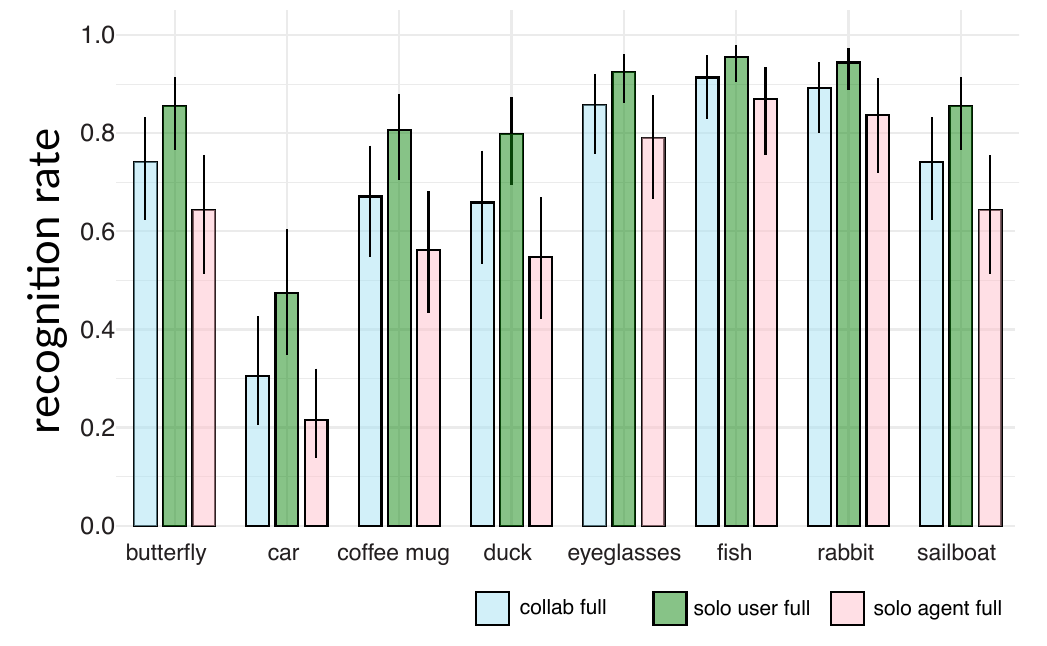}
    \caption{CLIP recognition rate by class for collaborative, solo user, and solo agent full sketches.}
    \label{fig:rate_by_class}
\end{figure}

\begin{figure}
    \centering
    \includegraphics[width=\linewidth]{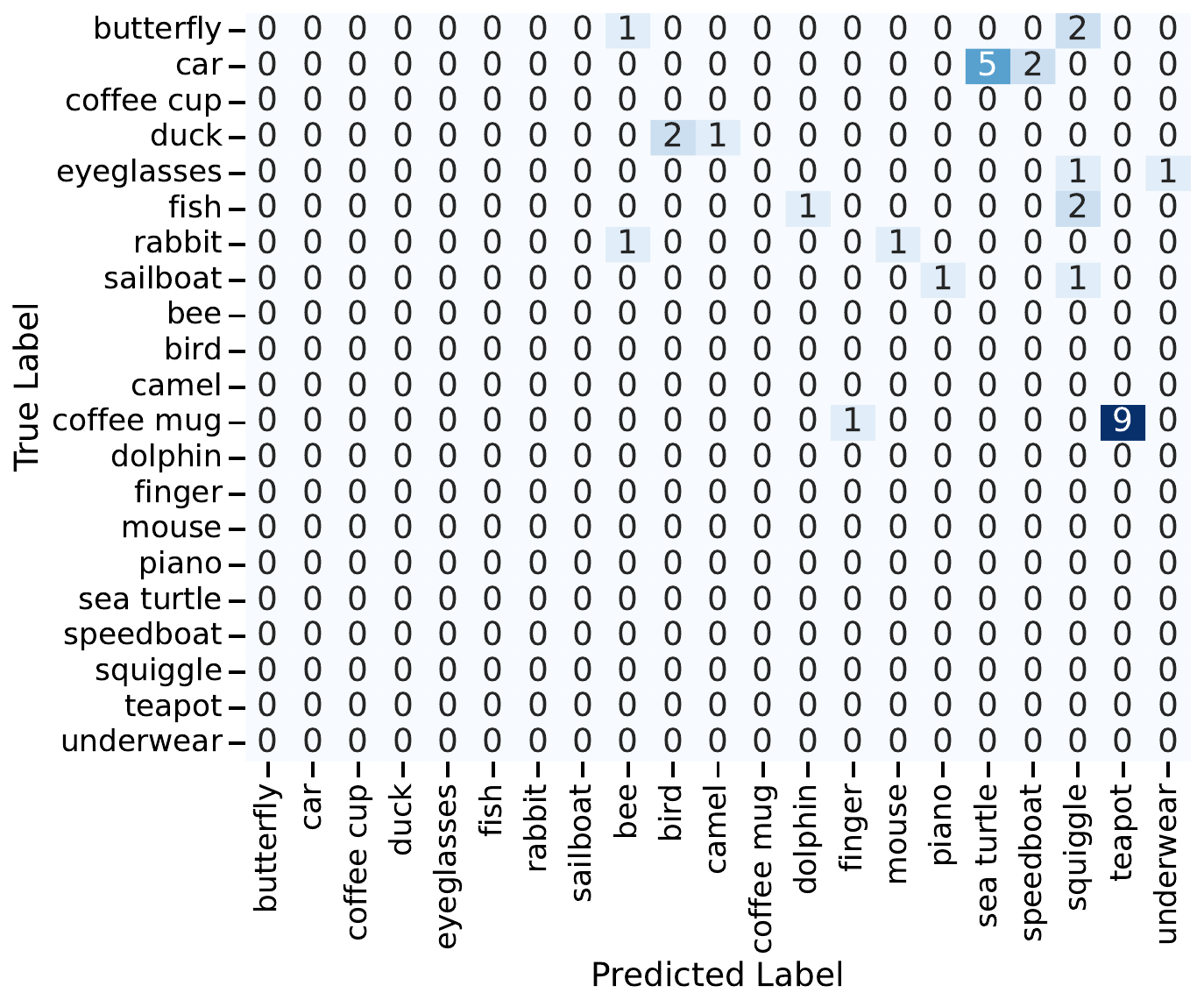}
    \caption{Confusion matrix from CLIP classification with categories from the QuickDraw dataset for 240 collaborative sketches across 8 categories.}
    \label{fig:collab_confusion}
\end{figure}

\begin{figure}
    \centering
    \includegraphics[width=\linewidth]{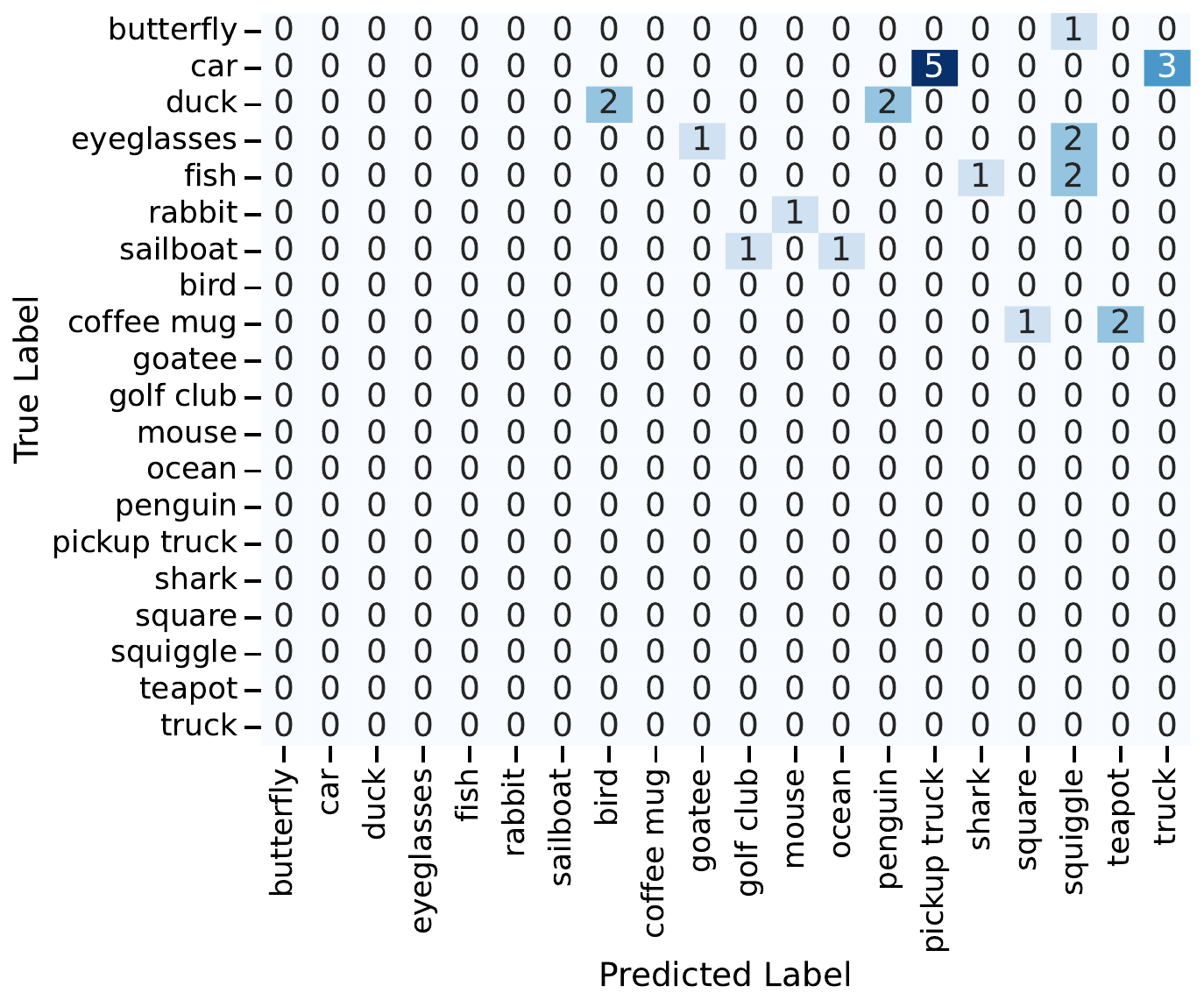}
    \caption{Confusion matrix from CLIP classification with categories from the QuickDraw dataset for 240 solo user sketches across 8 categories.}
    \label{fig:solo_user_confusion}
\end{figure}

\Cref{fig:rate_by_class} presents the recognition rates with $95\%$ confidence interval (CI) error bars for each concept across all three sketching conditions: \ap{collab} (blue), \ap{solo-user} (green), and \ap{solo-agent} (pink). Overall, the recognition rates for collaborative sketches are comparable to those produced by users alone or the agent alone for each unique category.
Notably, sketches of \ap{car} exhibit the lowest recognition rate across all conditions. This is likely due to confusion with semantically similar categories, such as \ap{truck}, \ap{pickup truck}, \ap{airplane}, and \ap{speedboat}, as indicated by the confusion matrices. Similarly, as discussed earlier, \ap{coffee cup} and \ap{duck} are frequently misclassified as related categories with overlapping visual features.

\begin{figure}
    \centering
    \includegraphics[width=\linewidth]{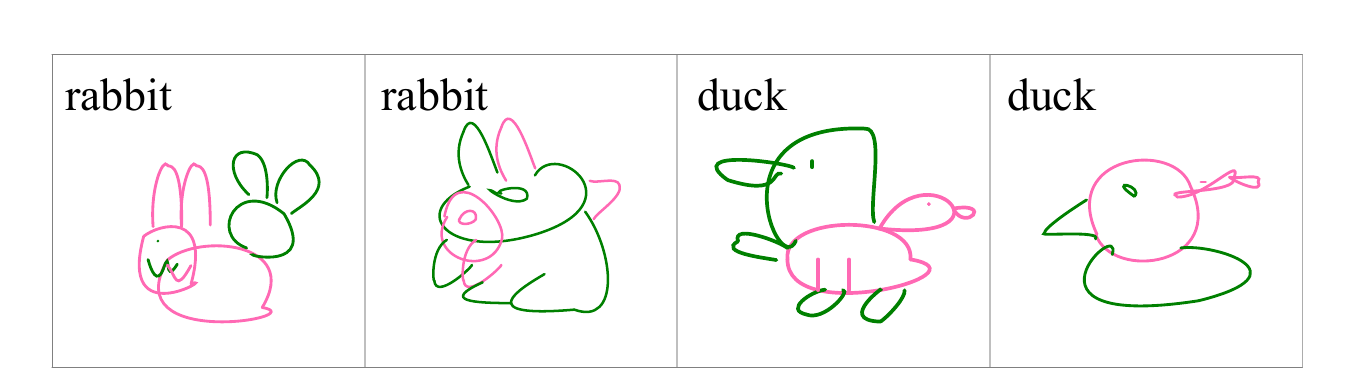}
    \caption{Examples of sketches created in \ap{collab} mode that were correctly classified by CLIP but considered unsuccessful as collaborations due to conflicting agent-user interpretations of sub-components.}
    \label{fig:edge-case-collab}
\end{figure}

\begin{figure}
    \centering
    \includegraphics[trim={3cm 0cm 0cm 0cm},clip,width=0.9\linewidth]{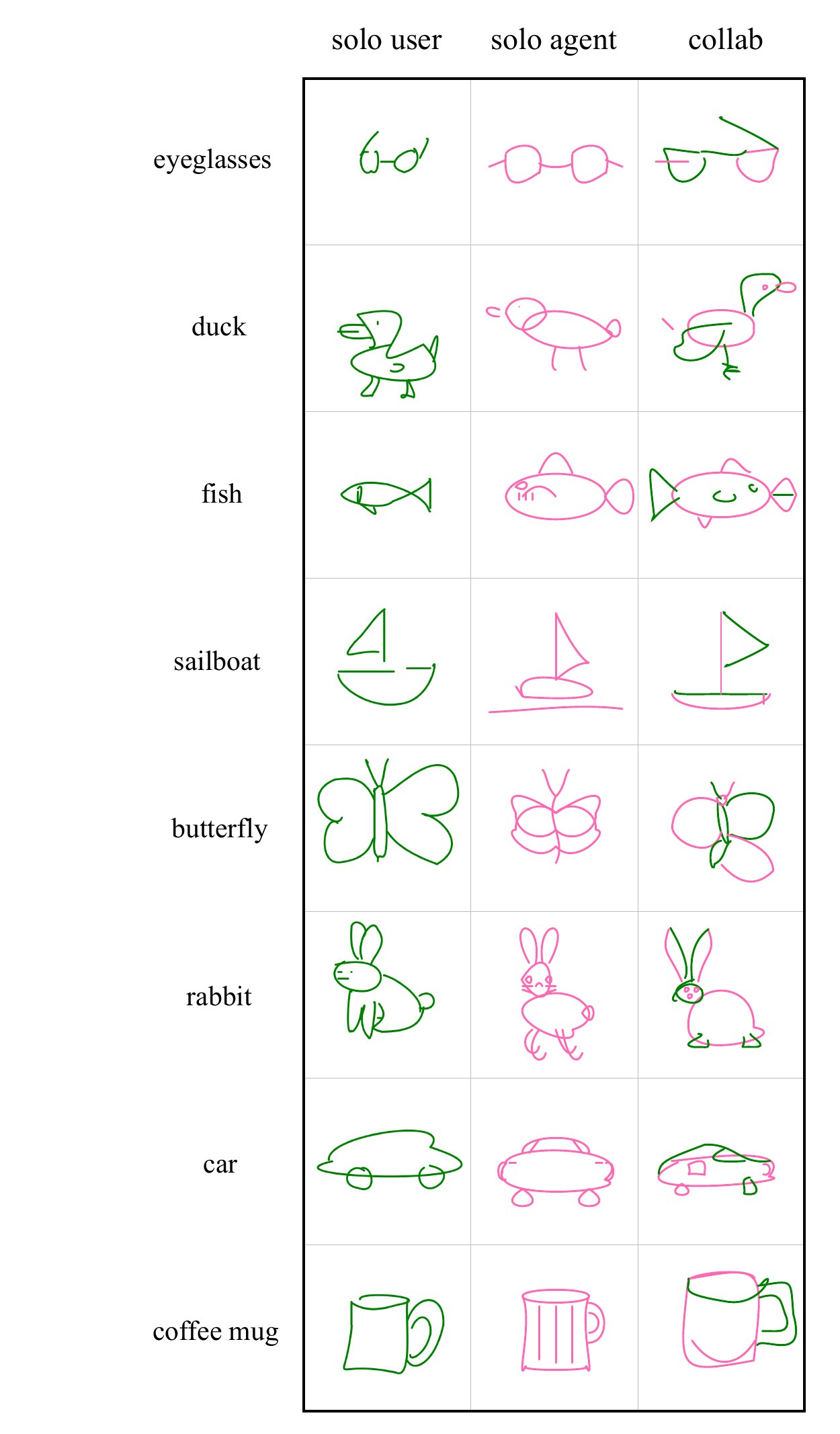}
    \caption{Examples of sketches from our collaborative human study that received high recognition rates. From left to right are sketches drawn in \ap{solo} mode by users, \ap{solo} mode by the agent, and collaboratively by both.}
    \label{fig:ex_all_sketches}
\end{figure}

We observe that in some cases of collaborative sketching (14 out of 240 sketches), the agent-human pair faces challenges in interpreting each other’s intentions and the meanings of strokes. Achieving effective collaboration and communication between different parties \cite{GROSZ1996269} is a challenge that often requires prior planning, social reasoning, and repeated interactions to establish shared intentions and representations. These complex processes continue to be studied across various contexts, including in interactions between humans \cite{mccarthy2021learningcommunicatesharedprocedural,hawkins2021visualresemblancecommunicativecontext,KNOBLICH201159}, between humans and agents \cite{NEURIPS2019_f5b1b89d}, and between agents \cite{Stone_Kaminka_Kraus_Rosenschein_2010}. \cref{fig:edge-case-collab} highlights the few collaborative sketches where the CLIP classification is correct, but the agent and user appear to lack a shared understanding of different stroke groups, resulting in the conflicting creation of duplicate concept components (i.e. two heads).

\begin{table}[t]
\small %
\centering
\label{tab:logistic_regression}
\renewcommand{\arraystretch}{1.2} %
\setlength{\tabcolsep}{1.2 pt} %
\begin{tabular}{lcc}
\hline
\textbf{Variation} & \textbf{Recognition Rate} & \textbf{95\% CI}\\ \hline
Collab full sketch & 0.75 & [0.61, 0.85] \\
Collab agent-only strokes& 0.10 & [0.06, 0.19]\\
Collab user-only strokes & 0.13 & [0.07, 0.23] \\ \hline
\end{tabular}
\caption{Recognition rate and 95\% CI across collaborative full and partial sketches. In collaborative sketches, keeping agent-only strokes or user-only strokes significantly reduces recognizability.}
\label{tab:collab-table}
\end{table}

\subsection{Chat-Based Editing}
In section 5.4 of the main paper we demonstrate chat-based editing using SketchAgent.
Below, we provide more details about the implementation of the experiment we performed.
To enable chat editing, we use the following prompt: 
\textit{\ap{$<$editing instruction$>$. Describe the location of the added concepts first in $<$thinking$>$ tags. Only provide the added strokes. Respond in the same format as before. Be concise.}}, where $<$editing instruction$>$ contains the desired edit such as \textit{\ap{Add glasses to the given cat}}.
The chosen objects per category, as well as the editing prompts, are provided:
\begin{itemize}
    \item \textbf{Animals:} \textit{fish, bird, cat}. Editing instruction: \textit{\ap{Add glasses}, \ap{Add a hat}, \ap{Add a skirt}}. 
    \item \textbf{Outdoor:} \textit{bus, building, boat}. Editing instruction: \textit{\ap{Add a tree to the left of the $<$concept$>$}, \ap{Add a sun on the top right, above the $<$concept$>$}, \ap{Add another smaller $<$concept$>$ to the right of this $<$concept$>$}}.
    \item \textbf{Indoor:} \textit{shelf, nightstand, table}. Editing instruction: \textit{\ap{Add a coffee mug on the top of the $<$concept$>$}, \ap{Add a lamp on the top of the $<$concept$>$}, \ap{Add an indoor plant to the left of the $<$concept$>$}}.
\end{itemize}
The resulting edited sketches are shown in \Cref{fig:editing}.

\begin{figure}[t]
    \centering
    \setlength{\tabcolsep}{2pt}
    {\small
    \begin{tabular}{c c c}
    \begin{tabular}[c]{@{}c@{}}\textcolor{red}{\ap{Tree to the left}} \\ \textcolor{blue}{\ap{Sun on top right}} \\ \textcolor{ForestGreen}{\ap{Smaller $<$concept$>$}} \\\textcolor{ForestGreen}{\ap{to the right}} \end{tabular} & 
        \begin{tabular}[c]{@{}c@{}}\textcolor{red}{\ap{Coffee mug on top}} \\ \textcolor{blue}{\ap{Lamp on top}} \\ \textcolor{ForestGreen}{\ap{Plant to the left}} \end{tabular} &
        \begin{tabular}[c]{@{}c@{}}\textcolor{red}{\ap{Add glasses}} \\ \textcolor{blue}{\ap{Add a hat}} \\ \textcolor{ForestGreen}{\ap{Add a skirt}}\end{tabular} \\ 
        \midrule
        \\
        \ap{Building} & \ap{Nightstand} & \ap{Cat} \\
        \includegraphics[width=0.3\linewidth]{figs/qualitative/editing/output_building.png} &
        \includegraphics[width=0.3\linewidth]{figs/qualitative/editing/output_nightstand.png} &
        \includegraphics[width=0.3\linewidth]{figs/qualitative/editing/output_cat.png}\\
         
        \ap{Boat} & \ap{Shelf} & \ap{Bird} \\
        \includegraphics[width=0.3\linewidth]{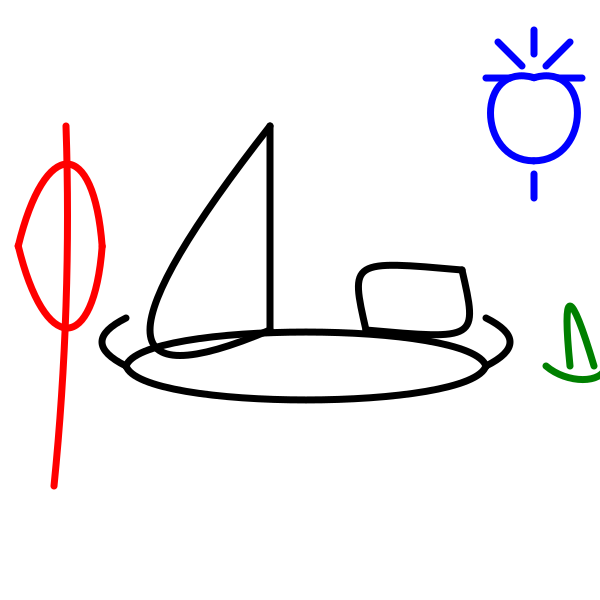} &
        \includegraphics[width=0.3\linewidth]{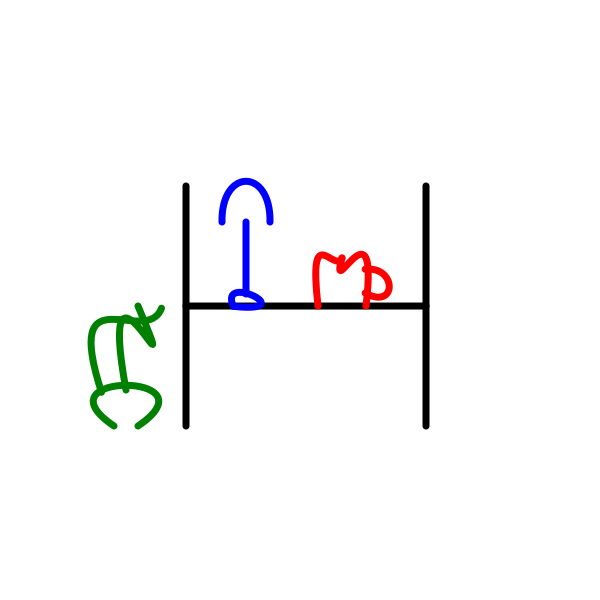} &
        \includegraphics[width=0.3\linewidth]{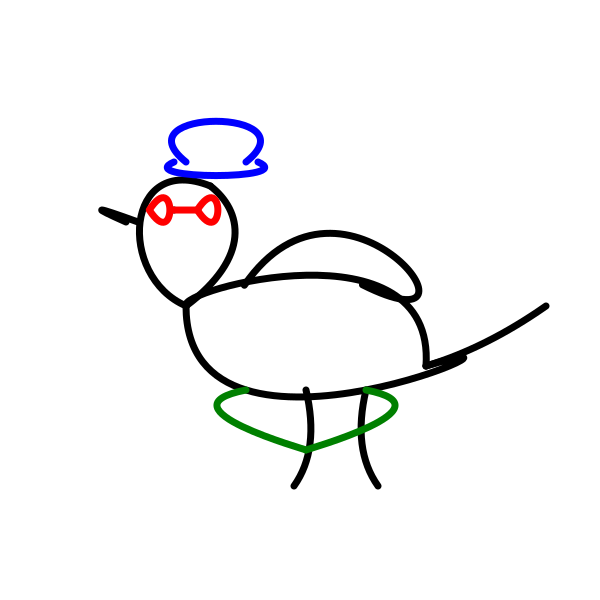}\\
        
        \ap{Bus} & \ap{Table} & \ap{Fish} \\
        \includegraphics[width=0.3\linewidth]{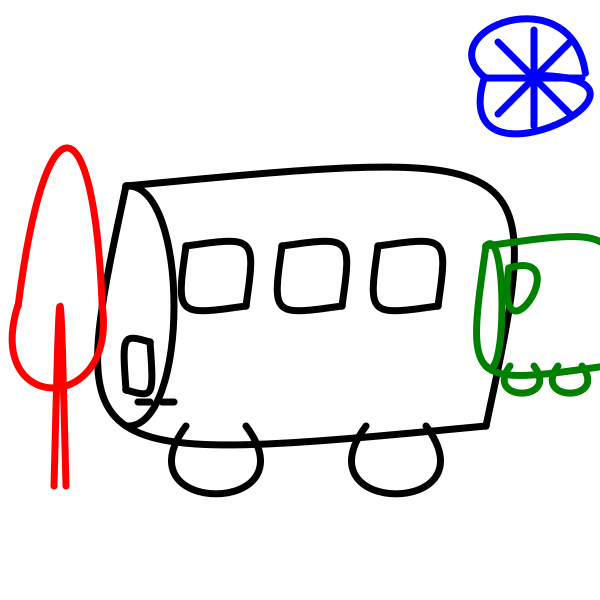} &
        \includegraphics[width=0.3\linewidth]{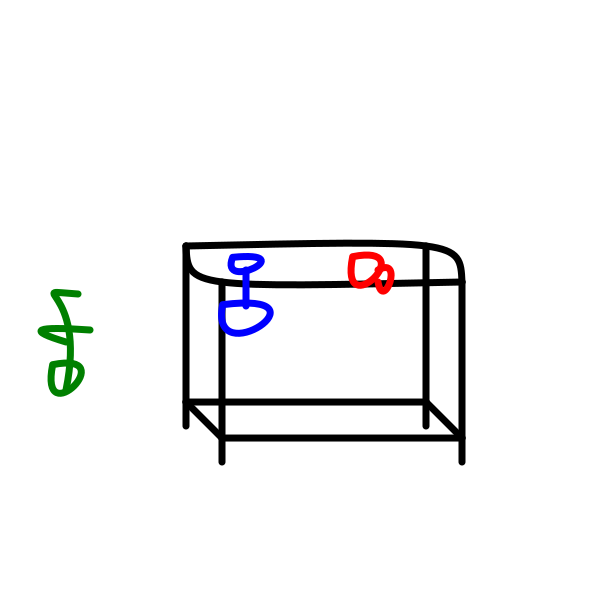} &
        \includegraphics[width=0.3\linewidth]{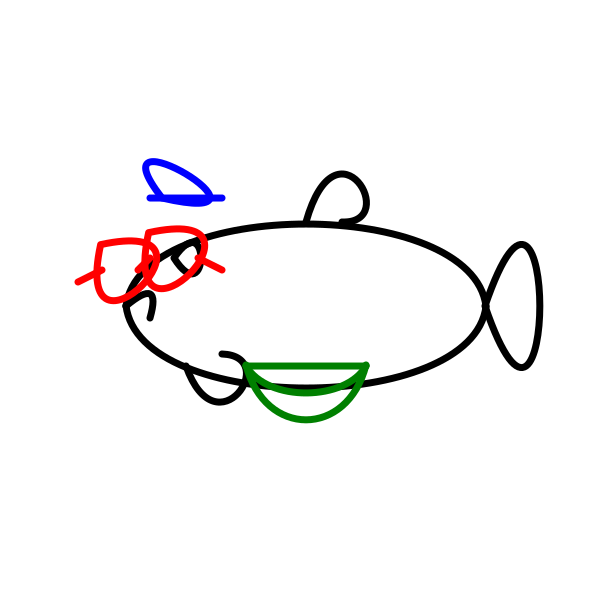}\\

    \end{tabular}
    }
    \caption{Chat-based sketch editing. We iteratively prompt SketchAgent to add objects to sketches through chat dialogues.}
    \label{fig:editing}
\end{figure}

\section{Ablation Study}
In Section 6 of the main paper, we presented an ablation study by systematically removing key components of our method and computing the resulting classification rates. Here, we provide further analyses and discussions on the ablation study.

\begin{figure}[h]
    \centering
    \includegraphics[trim={2cm 1cm 3cm 0.5cm},clip,width=1\linewidth]{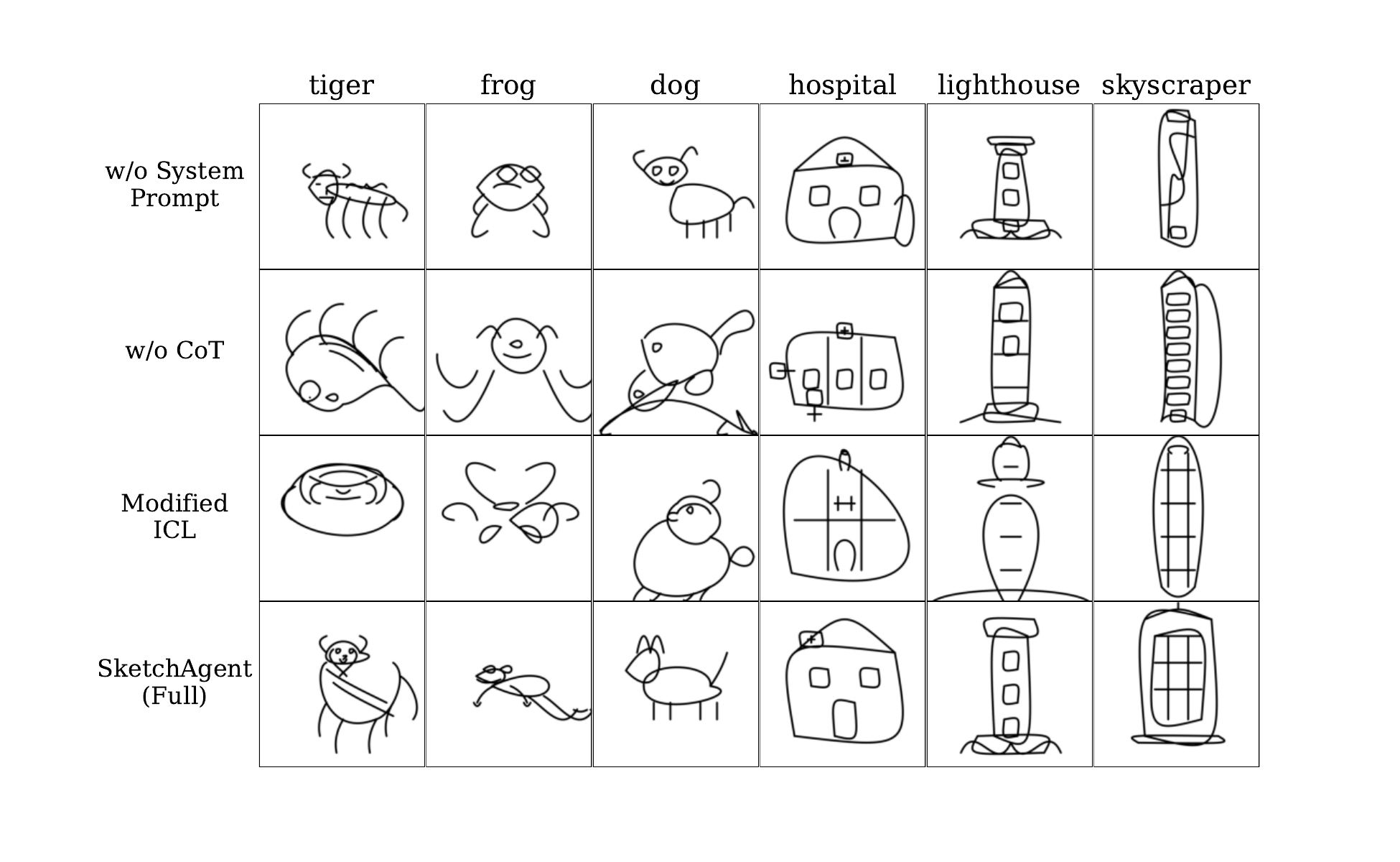}
    \caption{Visualization of sketches produced in different cases of our ablation study.}
    \label{fig:all_ablation_vis}
\end{figure}

Table 2 in the main paper shows the CLIP classification rates for 500 sketches (across 50 categories) per experiment. In \cref{fig:all_ablation_vis} we include a visualization of six sketches from six different concepts, covering both structures and animals. As shown, incorporating chain-of-thought reasoning and our in-context example of a house significantly enhances the quality of the results.

\begin{figure}[h]
    \centering
    \includegraphics[trim={3.6cm 1cm 3cm 0.5cm},clip,width=1\linewidth]{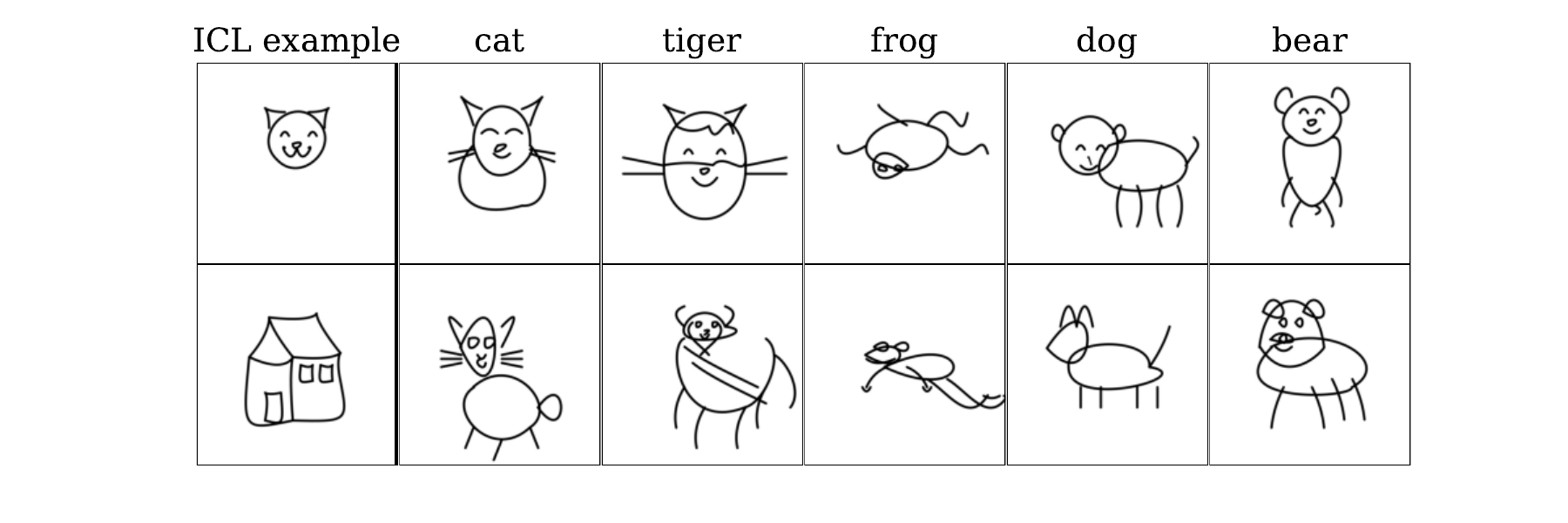}
    \includegraphics[trim={3.6cm 1cm 3cm 0cm},clip,width=1\linewidth]{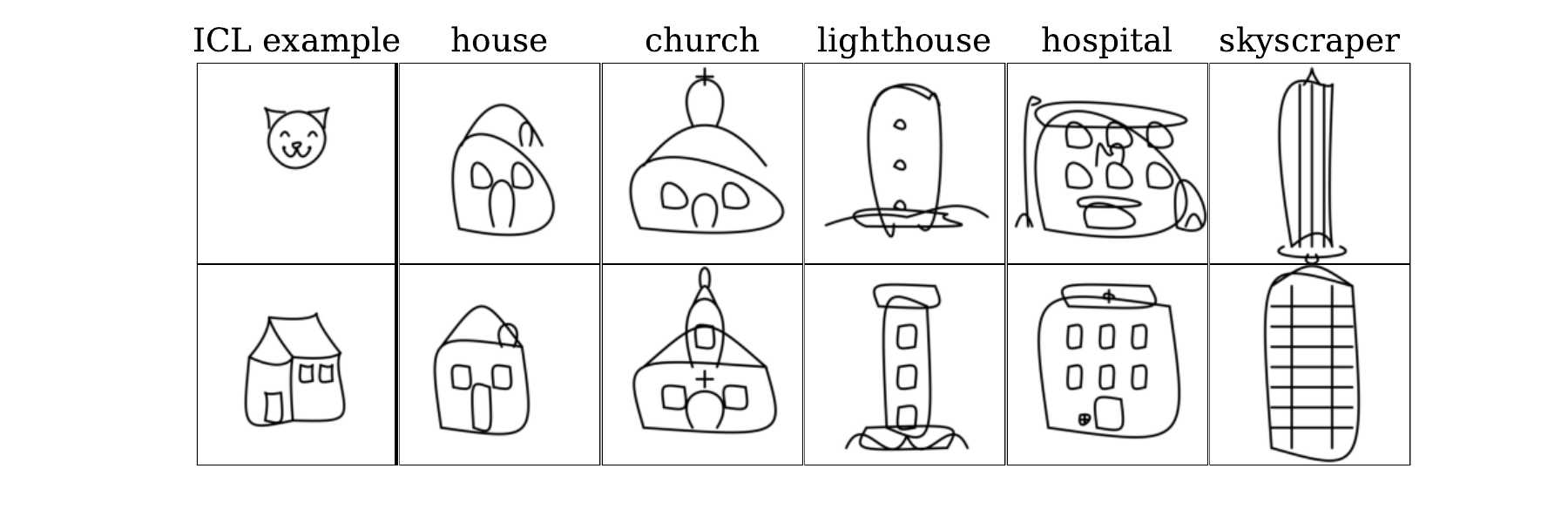}
    \caption{ICL example ablation study. We examine the impact of changing the concept in the ICL example (e.g., from a house to a cat) on the generation of related concepts. The example used in each experiment is shown on the left, with the top figure illustrating the effect on animal concepts and the bottom figure showing the effect on structural concepts.}
    \label{fig:cat_house}
\end{figure}

We find that the examples used in in-context learning (ICL) can influence both the quality and appearance of the generated sketches, suggesting an interesting direction for future research. Here, we analyze the impact of different types of in-context examples.
To investigate whether the theme of the in-context example affects the output (e.g., whether using a house example aids in sketching related concepts like a hospital or if using a cat example helps with sketching other animals), we constructed an alternative sketch of a cat. This sketch used the same number of strokes as the house example to isolate the effect of the theme from complexity. We then applied our method using this alternative example in ICL.
In \cref{fig:cat_house} we illustrate the influence of different ICL examples on related concepts. The example used in each experiment is shown on the left, with the top figure presenting the effect on animal concepts and the bottom figure depicting the effect on structures.
The results indicate that animal sketches are generally more influenced by an animal-based in-context example. For instance, the eyes in the generated sketches tend to resemble the eyes of the cat example more closely, while they vary more when a house example is used. However, there is no definitive conclusion regarding the overall quality or recognizability of these results.
Conversely, for structures (bottom), the use of the cat example seems to result in smoother and more rounded shapes, while sketches generated using the house example generally appear more refined and cohesive.

\begin{figure}[h]
    \centering
    \includegraphics[trim={3cm 1cm 2cm 0.5cm},clip,width=1\linewidth]{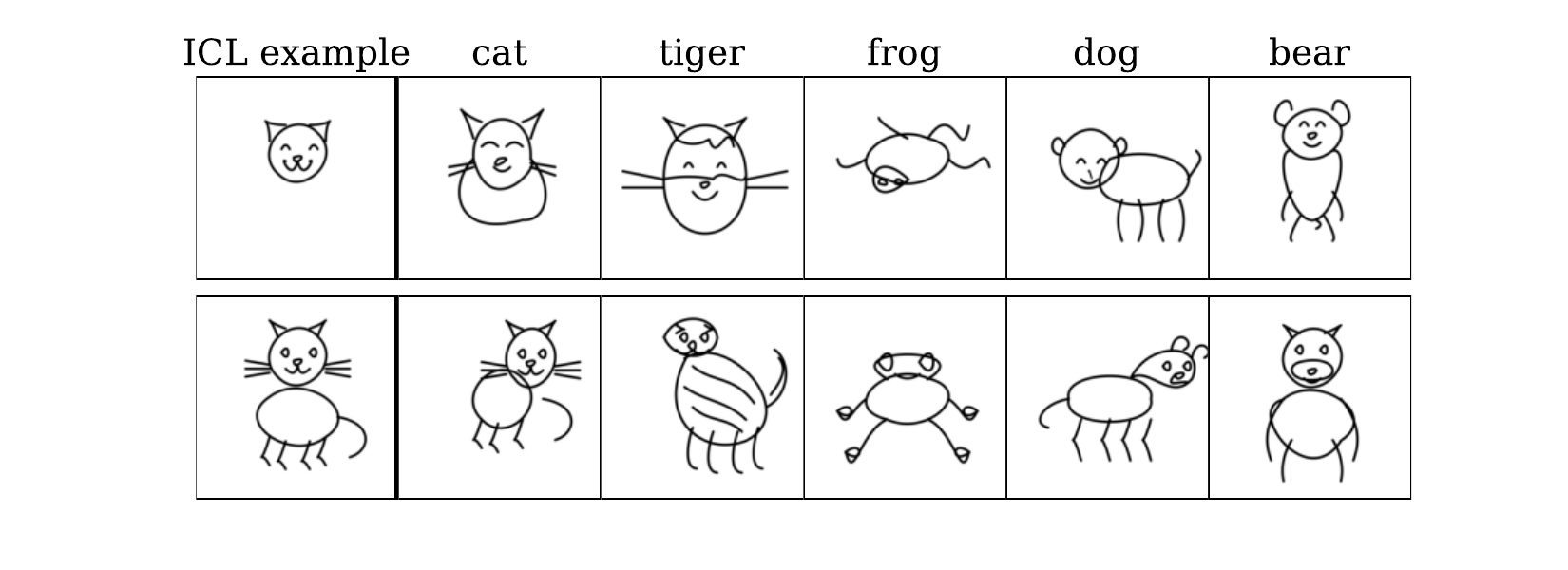}
    \includegraphics[trim={3cm 1cm 2cm 0cm},clip,width=1\linewidth]{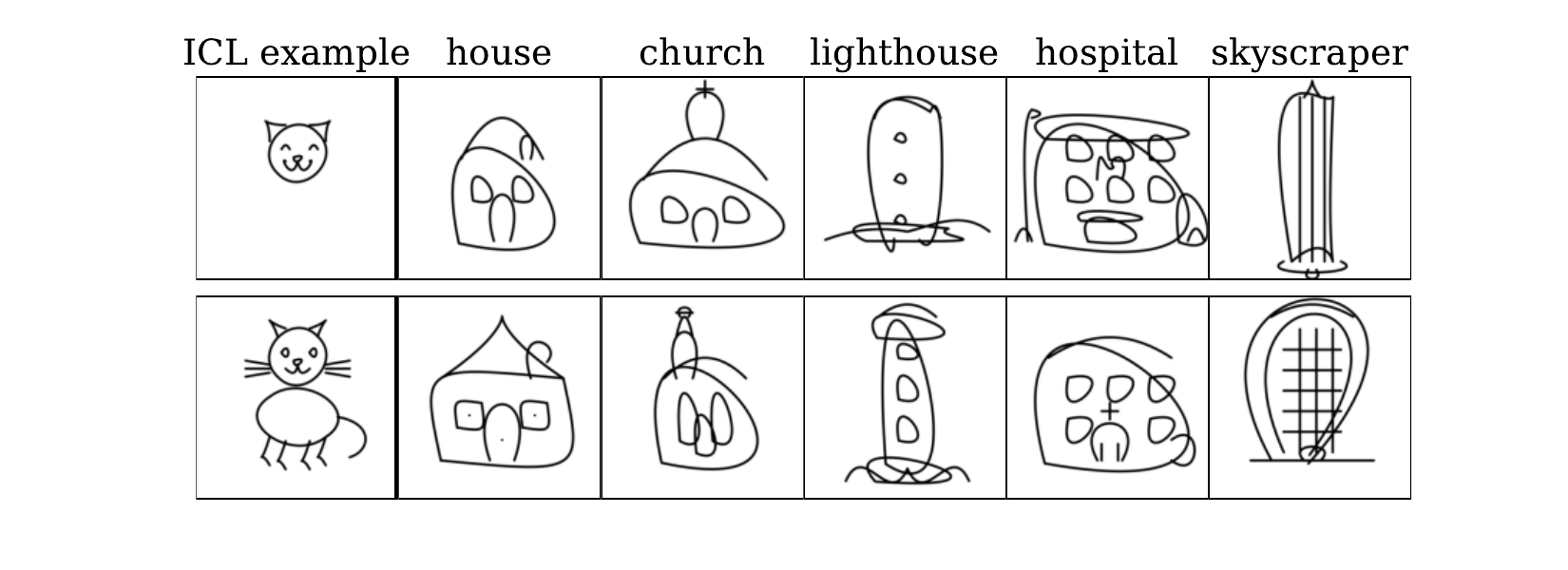}
    \caption{ICL example ablation study. We examine the impact of varying the complexity of the sketch presented in the ICL example while keeping the semantic concept (a cat) constant. The example used in each experiment is shown on the left, with the top figure illustrating the effect on animal concepts and the bottom figure showing the effect on structural concepts.}
    \label{fig:complexity-cat_house}
\end{figure}

We also examine the impact of example complexity, specifically how using a more detailed sketch with additional strokes affects the output. To test this, we enhanced the cat example by adding more details and then applied our method with the new, more complex example. The results are presented in \cref{fig:complexity-cat_house}. As shown, when a more detailed example is used, the generated sketches tend to overfit, closely replicating the original cat sketch. In contrast, using a simpler example leads to greater variation in the output.

\section{Prompts and More Results}
We present the full prompts used in our system, as well as our randomly generated sketches used for the quantitative evaluation presented in Section 5.1 of the main paper, and the full set of sketches made by users and in collaborative mode from our human study.

\begin{figure*}[t]
    \centering
    \small
    \begin{tcolorbox}[colback=gray!5!white,colframe=gray!75!black]
    \small
You are an expert artist specializing in drawing sketches that are visually appealing, expressive, and professional. \\
You will be provided with a blank grid. Your task is to specify where to place strokes on the grid to create a visually appealing sketch of the given textual concept. 
The grid uses numbers (1 to {res}) along the bottom (x axis) and numbers (1 to {res}) along the left edge (y axis) to reference specific locations within the grid. Each cell is uniquely identified by a combination of the corresponding x axis numbers and y axis number (e.g., the bottom-left cell is 'x1y1', the cell to its right is 'x2y1'). 
You can draw on this grid by specifying where to draw strokes. You can draw multiple strokes to depict the whole object, where different strokes compose different parts of the object. \\
To draw a stroke on the grid, you need to specify the following: \\
Starting Point: Specify the starting point by giving the grid location (e.g., 'x1y1' for column 1, row 1). \\
Ending Point: Specify the ending point in the same way (e.g., 'x{res}y{res}' for column {res}, row {res}). \\
Intermediate Points: Specify at least two intermediate points that the stroke should pass through. List these in the order the stroke should follow, using the same grid location format (e.g., 'x6y5', 'x13y10' for points at column 6 row 5 and column 13 row 10). \\
Parameter Values (t): For each point (including the start and end points), specify a t value between 0 and 1 that defines the position along the stroke's path. t=0 for the starting point. t=1 for the ending point. \\
Intermediate points should have t values between 0 and 1 (e.g., "0.3 for x6y5, 0.7 for x13y10").\\
Examples:\\
To draw a smooth curve that starts at x8y6, passes through x6y7 and x6y10, ending at x8y11: \\
Points = ['x8y6', 'x6y7', 'x6y10', 'x8y11'] 
t\_values = [0.00,0.30,0.80,1.00] \\
To close this curve into an ellipse shape, you can add another curve: \\
Points = ['x8y11', 'x11y10', 'x11y7', 'x8y6'] 
t\_values = [0.00,0.30,0.70,1.00] \\
To draw a large circle that starts at x25y44 and ends at x25y44, passing through the cells x32y41, x35y35, x31y29, x25y27, x19y29, x15y35, x18y41: 
Points = ['x25y44', 'x32y41', 'x35y35', 'x31y29', 'x25y27', 'x19y29', 'x15y35', 'x18y41', 'x25y44'] 
t\_values = [0.00, 0.125, 0.25, 0.375, 0.50, 0.625, 0.75, 0.875, 1.00] \\
To draw non-smooth shapes (with corners) like triangles or rectangles, you need to specify the corner points twice with adjacent corresponding t values. 
For example, to draw an upside-down "V" shape that starts at x13y27, ends at x24y27, with a pick (corner) at x18y37: 
Points = ['x13y27', 'x18y37','x18y37', 'x24y27'] 
t\_values = [0.00,0.55,0.5,1.00] \\
To draw a triangle with corners at x10y29, x15y33, and x9y35, start with drawing a "V" shape that starts at x10y29, ends at x9y35, with a pick (corner) at x15y33: \\
Points = ['x10y29', 'x15y33', 'x15y33', 'x9y35'] 
t\_values = [0.00,0.55,0.5,1.00]\\
and then close it with a straight line from x13y27 to x24y27 to form a triangle:\\
Points = ['x13y27', 'x24y27']
t\_values = [0.00,1.00]\\
Note that for a triangle, the start and end points should be different from each other.\\
To draw a rectangle with four corners at x13y27, x24y27, x24y11, x13y11:\\
Points = ['x13y27', 'x24y27', 'x24y27', 'x24y11', 'x24y11', 'x13y11', 'x13y11', 'x13y27']
t\_values = [0.00,0.3,0.25,0.5,0.5,0.75,0.75,1.00]\\
To draw a small square with four corners at x26y25, x29y25, x29y21, x26y21:\\
Points = ['x26y25', 'x29y25', 'x29y25', 'x29y21', 'x29y21', 'x26y21', 'x26y21', 'x26y25']
t\_values = [0.00,0.3,0.25,0.5,0.5,0.75,0.75,1.00]\\
To draw a single dot at x15y31 use:
Points = ['x15y31']
t\_values = [0.00]\\
To draw a straight linear line that starts at x18y31 and ends at x35y14 use:
Points = ['x18y31', 'x35y14']
t\_values = [0.00, 1.00].\\
If you want to draw a big and long stroke, split it into multiple small curves that connect to each other.
These instructions will define a smooth stroke that follows a Bezier curve from the starting point to the ending point, passing through the specified intermediate points.
To draw a visually appealing sketch of the given object or concept, break down complex drawings into manageable steps. Begin with the most important part of the object, then observe your progress and add additional elements as needed. Continuously refine your sketch by starting with a basic structure and gradually adding complexity. Think step-by-step.
\end{tcolorbox}
    \caption{System prompt.}
    \label{fig:system-prompt}
\end{figure*}

\begin{figure*}[h]
    \centering
    \small
    \begin{tcolorbox}[colback=gray!5!white,colframe=gray!75!black]
I provide you with a blank grid. Your goal is to produce a visually appealing sketch of a \{concept\}.\\
Here are a few examples:\\
$<$examples$>$\\
\{gt-sketches\}\\
$<$/examples$>$\\

You need to provide x-y coordinates that construct a recognizable sketch of a {concept}.\\
You will receive feedback on your sketch and you will be able to adjust and fix it. 
Note that you will not have access to any additional resources. Do not copy previous sketches.\\

Think before you provide the x-y coordinates in $<$thinking$>$ tags. \\
First, think through what parts of the {concept} you want to sketch and the sketching order.\\
Then, think about where the parts should be located on the grid.\\
Finally, provide your response in $<$answer$>$ tags, using your analysis.\\

Provide the sketch in the following format with the following fields:\\
$<$formatting$>$\\
$<$concept$>$The concept depicted in the sketch.$<$/concept$>$\\
$<$strokes$>$This element holds a collection of individual stroke elements that define the sketch. \\
Each stroke is uniquely identified by its own tag (e.g., $<$s1$>$, $<$s2$>$, etc.).\\
Within each stroke element, there are three key pieces of information: \\
$<$points$>$A list of x-y coordinates defining the curve. These points define the path the stroke follows.$<$/points$>$\\
$<$t\_values$>$A series of numerical timing values that correspond to the points. These values define the progression of the stroke over time, ranging from 0 to 1, indicating the order or speed at which the stroke is drawn.$<$/t\_values$>$\\
$<$id$>$A short descriptive identifier for the stroke, explaining which part of the sketch it corresponds to.$<$/id$>$\\
$<$/strokes$>$\\
$<$/formatting$>$\\
\end{tcolorbox}
    \caption{User prompt. This prompt contains the specific sketching task as well as details about the expected format.}
    \label{fig:user-prompt}
\end{figure*}

\begin{figure*}[t]
    \centering
    \small
    \begin{tcolorbox}[colback=gray!5!white,colframe=gray!75!black]
$<$example$>$\\
To draw a house, start by drawing the front of the house:\\
$<$concept$>$House$<$/concept$>$\\
$<$strokes$>$\\
    \hspace*{1em}$<$s1$>$\\
    \hspace*{2em}    $<$points$>$'x13y27', 'x24y27', 'x24y27', 'x24y11', 'x24y11', 'x13y11', 'x13y11', 'x13y27'$<$/points$>$\\
    \hspace*{2em}    $<$t\_values$>$0.00,0.3,0.25,0.5,0.5,0.75,0.75,1.00$<$/t\_values$>$\\
    \hspace*{2em}    $<$id$>$house base front rectangle$<$/id$>$\\
    \hspace*{1em}$<$/s1$>$\\
    \hspace*{1em}$<$s2$>$\\
    \hspace*{2em}    $<$points$>$'x13y27', 'x18y37','x18y37', 'x24y27'$<$/points$>$\\
    \hspace*{2em}    $<$t\_values$>$0.00,0.55,0.5,1.00$<$/t\_values$>$\\
    \hspace*{2em}    $<$id$>$roof front triangle$<$/id$>$\\
    \hspace*{1em}$<$/s2$>$\\
$<$/strokes$>$\\

Next we add the house's right section:\\
$<$concept$>$House$<$/concept$>$\\
$<$strokes$>$\\
    \hspace*{1em}$<$s1$>$\\
    \hspace*{2em}    $<$points$>$'x13y27', 'x24y27', 'x24y27', 'x24y11', 'x24y11', 'x13y11', 'x13y11', 'x13y27'$<$/points$>$\\
    \hspace*{2em}    $<$t\_values$>$0.00,0.3,0.25,0.5,0.5,0.75,0.75,1.00$<$/t\_values$>$\\
    \hspace*{2em}    $<$id$>$house base front rectangle$<$/id$>$\\
    \hspace*{1em}$<$/s1$>$\\
    \hspace*{1em}$<$s2$>$\\
    \hspace*{2em}    $<$points$>$'x13y27', 'x18y37','x18y37', 'x24y27'$<$/points$>$\\
    \hspace*{2em}    $<$t\_values$>$0.00,0.55,0.5,1.00$<$/t\_values$>$\\
    \hspace*{2em}    $<$id$>$roof front triangle$<$/id$>$\\
    \hspace*{1em}$<$/s2$>$\\
    \hspace*{1em}$<$s3$>$\\
    \hspace*{2em}    $<$points$>$'x24y27', 'x36y28', 'x36y28', 'x36y21', 'x36y21', 'x36y12', 'x36y12', 'x24y11'$<$/points$>$\\
    \hspace*{2em}    $<$t\_values$>$0.00,0.3,0.25,0.5,0.5,0.75,0.75,1.00$<$/t\_values$>$\\
    \hspace*{2em}    $<$id$>$house base right section$<$/id$>$\\
    \hspace*{1em}$<$/s3$>$\\
    \hspace*{1em}$<$s4$>$\\
    \hspace*{2em}    $<$points$>$'x18y37', 'x30y38', 'x30y38', 'x36y28'$<$/points$>$\\
    \hspace*{2em}    $<$t\_values$>$0.00,0.55,0.5,1.00$<$/t\_values$>$\\
    \hspace*{2em}    $<$id$>$roof right section$<$/id$>$\\
    \hspace*{1em}$<$/s4$>$\\
$<$/strokes$>$\\

Now that we have the general structure of the house, we can add details to it, like windows and a door:\\
$<$concept$>$House$<$/concept$>$\\
$<$strokes$>$\\
    \hspace*{1em}$<$s1$>$\\
    \hspace*{2em}    $<$points$>$'x13y27', 'x24y27', 'x24y27', 'x24y11', 'x24y11', 'x13y11', 'x13y11', 'x13y27'$<$/points$>$\\
    \hspace*{2em}    $<$t\_values$>$0.00,0.3,0.25,0.5,0.5,0.75,0.75,1.00$<$/t\_values$>$\\
    \hspace*{2em}    $<$id$>$house base front rectangle$<$/id$>$\\
    \hspace*{1em}$<$/s1$>$\\
    \hspace*{1em}$<$s2$>$\\
     \hspace*{2em}   $<$points$>$'x13y27', 'x18y37','x18y37', 'x24y27'$<$/points$>$\\
     \hspace*{2em}   $<$t\_values$>$0.00,0.55,0.5,1.00$<$/t\_values$>$\\
     \hspace*{2em}   $<$id$>$roof front triangle$<$/id$>$\\
    \hspace*{1em}$<$/s2$>$

\end{tcolorbox}
    \caption{ICL example. This is the example of a sketch of a house we provide to the model.}
    \label{fig:ICL-prompt}
\end{figure*}

\begin{figure*}[t]
    \centering
    \small
    \begin{tcolorbox}[colback=gray!5!white,colframe=gray!75!black]
    
    \hspace*{1em}$<$s3$>$\\
     \hspace*{2em}   $<$points$>$'x24y27', 'x36y28', 'x36y28', 'x36y21', 'x36y21', 'x36y12', 'x36y12', 'x24y11'$<$/points$>$\\
     \hspace*{2em}   $<$t\_values$>$0.00,0.3,0.25,0.5,0.5,0.75,0.75,1.00$<$/t\_values$>$\\
     \hspace*{2em}   $<$id$>$house base right section$<$/id$>$\\
    \hspace*{1em}$<$/s3$>$\\
    \hspace*{1em}$<$s4$>$\\
      \hspace*{2em}  $<$points$>$'x18y37', 'x30y38', 'x30y38', 'x36y28'$<$/points$>$\\
      \hspace*{2em}  $<$t\_values$>$0.00,0.55,0.5,1.00$<$/t\_values$>$\\
      \hspace*{2em}  $<$id$>$roof right section$<$/id$>$\\
    \hspace*{1em}$<$/s4$>$\\
    \hspace*{1em}$<$s5$>$\\
     \hspace*{2em}   $<$points$>$'x26y25', 'x29y25', 'x29y25', 'x29y21', 'x29y21', 'x26y21', 'x26y21', 'x26y25'$<$/points$>$\\
     \hspace*{2em}   $<$t\_values$>$0.00,0.3,0.25,0.5,0.5,0.75,0.75,1.00$<$/t\_values$>$\\
     \hspace*{2em}   $<$id$>$left window square$<$/id$>$\\
    \hspace*{1em}$<$/s5$>$\\
    \hspace*{1em}$<$s6$>$\\
     \hspace*{2em}   $<$points$>$'x31y25', 'x34y25', 'x34y25', 'x34y21', 'x34y21', 'x31y21', 'x31y21','x31y25'$<$/points$>$\\
     \hspace*{2em}   $<$t\_values$>$0.00,0.3,0.25,0.5,0.5,0.75,0.75,1.00$<$/t\_values$>$\\
     \hspace*{2em}   $<$id$>$right window square$<$/id$>$\\
    \hspace*{1em}$<$/s6$>$\\
    \hspace*{1em}$<$s7$>$\\
     \hspace*{2em}   $<$points$>$'x17y11', 'x17y18', 'x17y18', 'x21y18', 'x21y18', 'x21y11', 'x21y11', 'x17y11'$<$/points$>$\\
     \hspace*{2em}   $<$t\_values$>$0.00,0.3,0.25,0.5,0.5,0.75,0.75,1.00$<$/t\_values$>$\\
     \hspace*{2em}   $<$id$>$front door$<$/id$>$\\
    \hspace*{1em}$<$/s7$>$\\
$<$/strokes$>$\\
and here is the complete example:\\
$<$concept$>$House$<$/concept$>$\\
$<$strokes$>$\\
   \hspace*{1em} $<$s1$>$\\
      \hspace*{2em}  $<$points$>$'x13y27', 'x24y27', 'x24y27', 'x24y11', 'x24y11', 'x13y11', 'x13y11', 'x13y27'$<$/points$>$\\
      \hspace*{2em}  $<$t\_values$>$0.00,0.3,0.25,0.5,0.5,0.75,0.75,1.00$<$/t\_values$>$\\
      \hspace*{2em}  $<$id$>$house base front rectangle$<$/id$>$\\
   \hspace*{1em} $<$/s1$>$\\
  \hspace*{1em}  $<$s2$>$\\
      \hspace*{2em}  $<$points$>$'x24y27', 'x36y28', 'x36y28', 'x36y21', 'x36y21', 'x36y12', 'x36y12', 'x24y11'$<$/points$>$\\
      \hspace*{2em}  $<$t\_values$>$0.00,0.3,0.25,0.5,0.5,0.75,0.75,1.00$<$/t\_values$>$\\
      \hspace*{2em}  $<$id$>$house base right section$<$/id$>$\\
  \hspace*{1em}  $<$/s2$>$\\
\hspace*{1em}    $<$s3$>$\\
      \hspace*{2em}  $<$points$>$'x13y27', 'x18y37','x18y37', 'x24y27'$<$/points$>$\\
      \hspace*{2em}  $<$t\_values$>$0.00,0.55,0.5,1.00$<$/t\_values$>$\\
      \hspace*{2em}  $<$id$>$roof front triangle$<$/id$>$\\
    \hspace*{1em}$<$/s3$>$\\
  \hspace*{1em}  $<$s4$>$\\
      \hspace*{2em}  $<$points$>$'x18y37', 'x30y38', 'x30y38', 'x36y28'$<$/points$>$\\
      \hspace*{2em}  $<$t\_values$>$0.00,0.55,0.5,1.00$<$/t\_values$>$\\
      \hspace*{2em}  $<$id$>$roof right section$<$/id$>$\\
  \hspace*{1em}  $<$/s4$>$\\

\end{tcolorbox}
    \caption{ICL example. This is the example of a sketch of a house we provide to the model.}
    \label{fig:ICL-prompt}
\end{figure*}

\begin{figure*}[t]
    \centering
    \small
    \begin{tcolorbox}[colback=gray!5!white,colframe=gray!75!black]
    
 \hspace*{1em}   $<$s5$>$\\
      \hspace*{2em}  $<$points$>$'x26y25', 'x29y25', 'x29y25', 'x29y21', 'x29y21', 'x26y21', 'x26y21', 'x26y25'$<$/points$>$\\
      \hspace*{2em}  $<$t\_values$>$0.00,0.3,0.25,0.5,0.5,0.75,0.75,1.00$<$/t\_values$>$\\
      \hspace*{2em}  $<$id$>$left window square$<$/id$>$\\
 \hspace*{1em}   $<$/s5$>$\\
\hspace*{1em}    $<$s6$>$\\
      \hspace*{2em}  $<$points$>$'x31y25', 'x34y25', 'x34y25', 'x34y21', 'x34y21', 'x31y21', 'x31y21','x31y25'$<$/points$>$\\
      \hspace*{2em}  $<$t\_values$>$0.00,0.3,0.25,0.5,0.5,0.75,0.75,1.00$<$/t\_values$>$\\
      \hspace*{2em}  $<$id$>$right window square$<$/id$>$\\
\hspace*{1em}    $<$/s6$>$\\
\hspace*{1em}    $<$s7$>$\\
     \hspace*{2em}   $<$points$>$'x17y11', 'x17y18', 'x17y18', 'x21y18', 'x21y18', 'x21y11', 'x21y11', 'x17y11'$<$/points$>$\\
     \hspace*{2em}   $<$t\_values$>$0.00,0.3,0.25,0.5,0.5,0.75,0.75,1.00$<$/t\_values$>$\\
     \hspace*{2em}   $<$id$>$front door$<$/id$>$\\
 \hspace*{1em}   $<$/s7$>$\\
$<$/strokes$>$\\
$<$/example$>$
\end{tcolorbox}
    \caption{ICL example. This is the example of a sketch of a house we provide to the model.}
    \label{fig:ICL-prompt}
\end{figure*}

\begin{figure*}
    \centering
    \includegraphics[trim={0cm 3600pt 0cm 0cm},clip,width=0.5\linewidth]{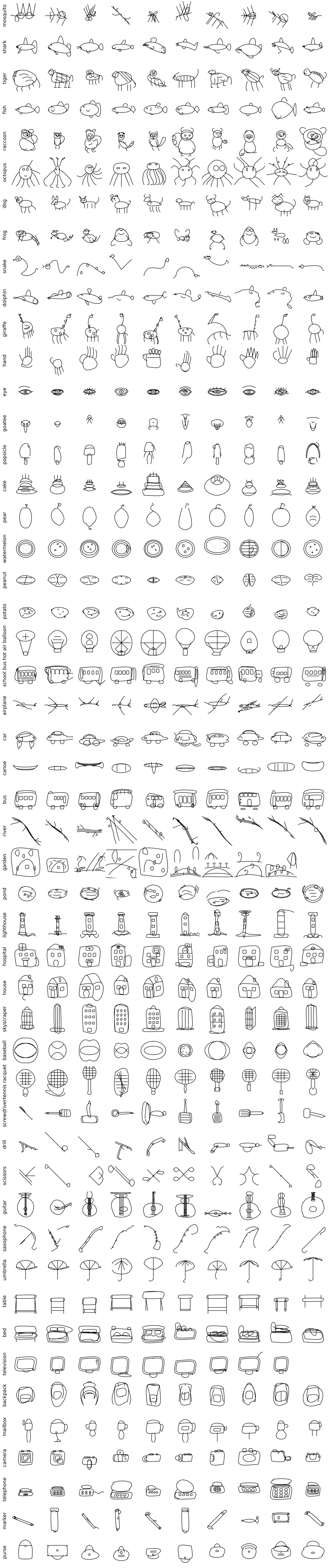}
    \caption{Randomly generated sketches used in the quantitative analysis (ten sketches per category).}
    \label{fig:enter-label}
\end{figure*}

\begin{figure*}
    \centering
    \includegraphics[trim={0cm 0cm 0cm 3600pt},clip,width=0.5\linewidth]{appendix/figs/category-based-quant/ours_all.pdf}
    \caption{Randomly generated sketches used in the quantitative analysis (ten sketches per category).}
    \label{fig:enter-label}
\end{figure*}

\begin{figure*}
    \centering
    \includegraphics[width=1\linewidth]{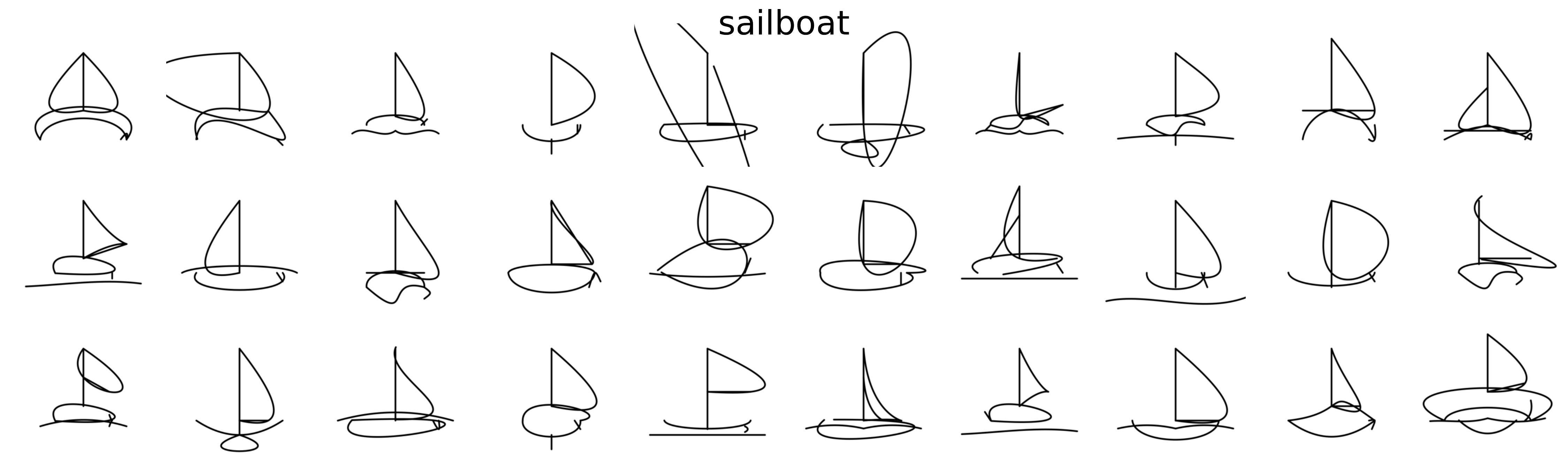}
    \includegraphics[width=1\linewidth]{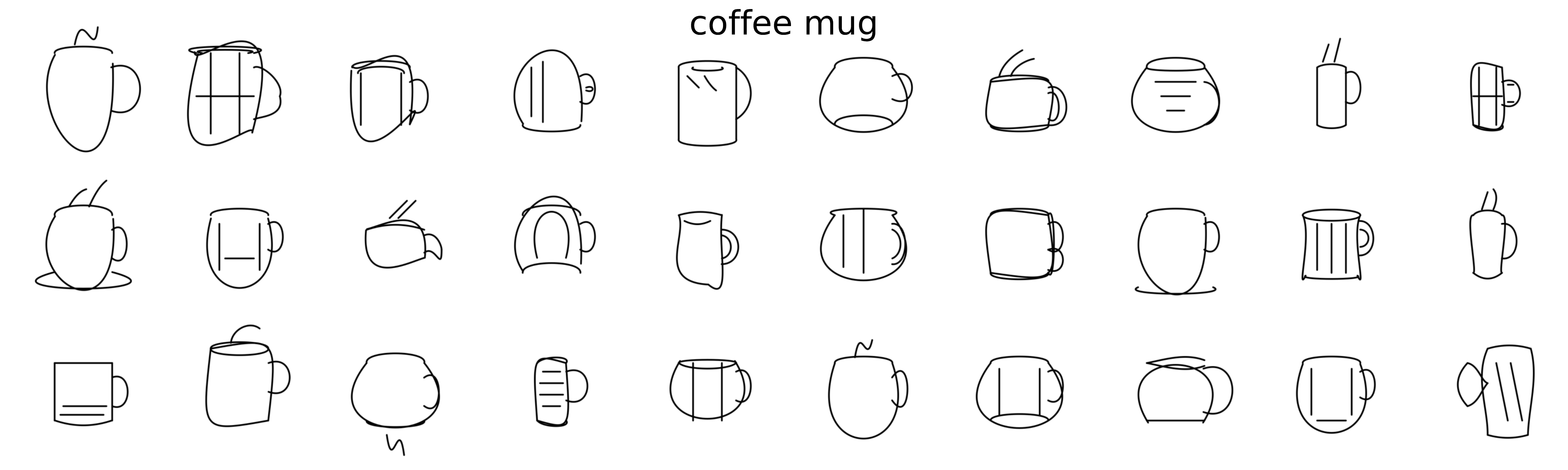}
    \includegraphics[width=1\linewidth]{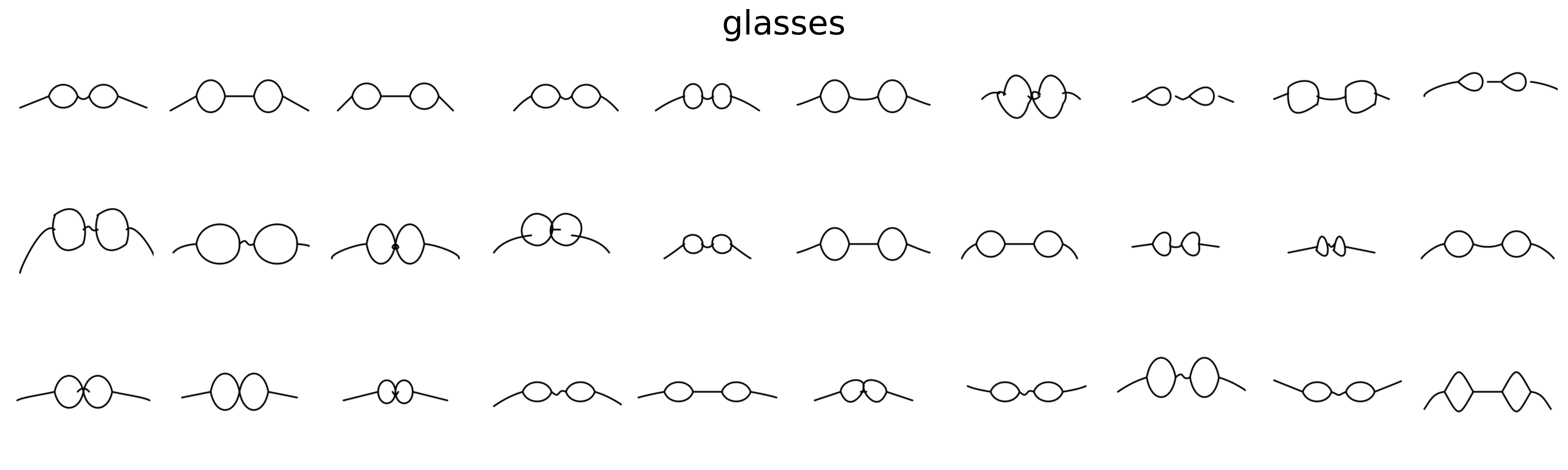}
    \includegraphics[width=1\linewidth]{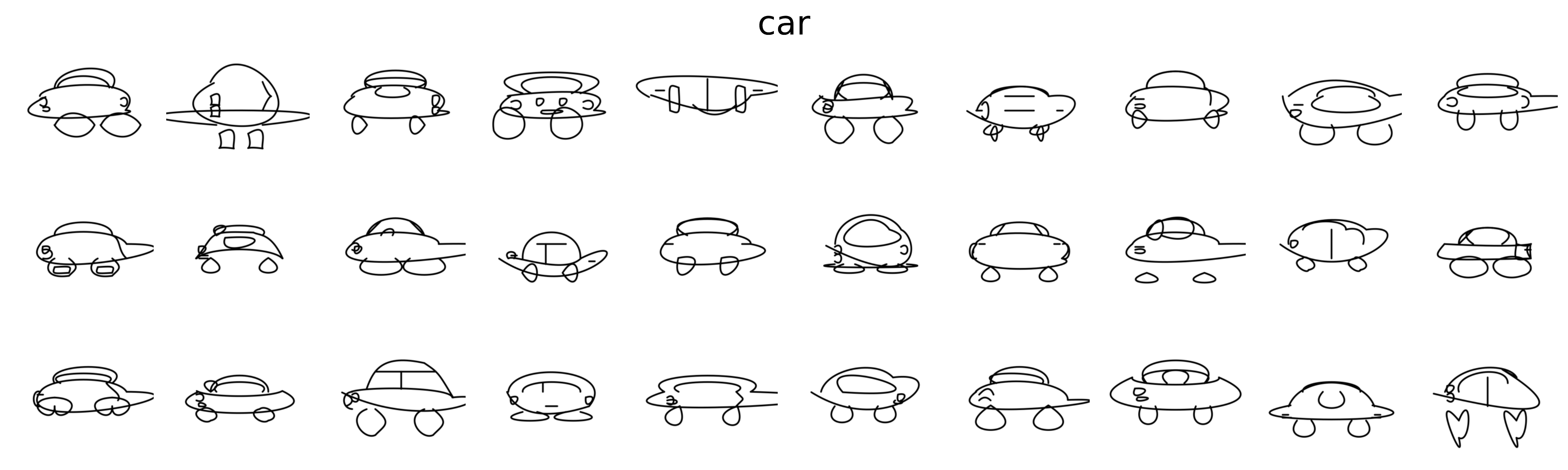}
    \caption{Sketches generated by SketchAgent for the eight categories of our human-agent collaborative study.}
    \label{fig:enter-label}
\end{figure*}

\begin{figure*}
    \centering
    \includegraphics[width=1\linewidth]{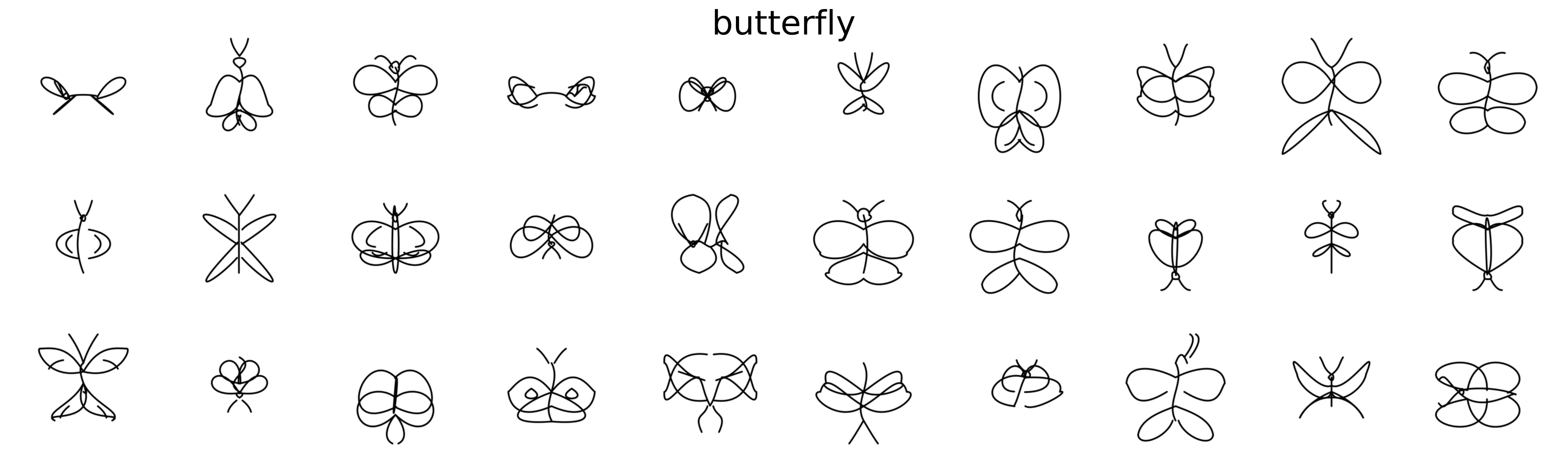}
    \includegraphics[width=1\linewidth]{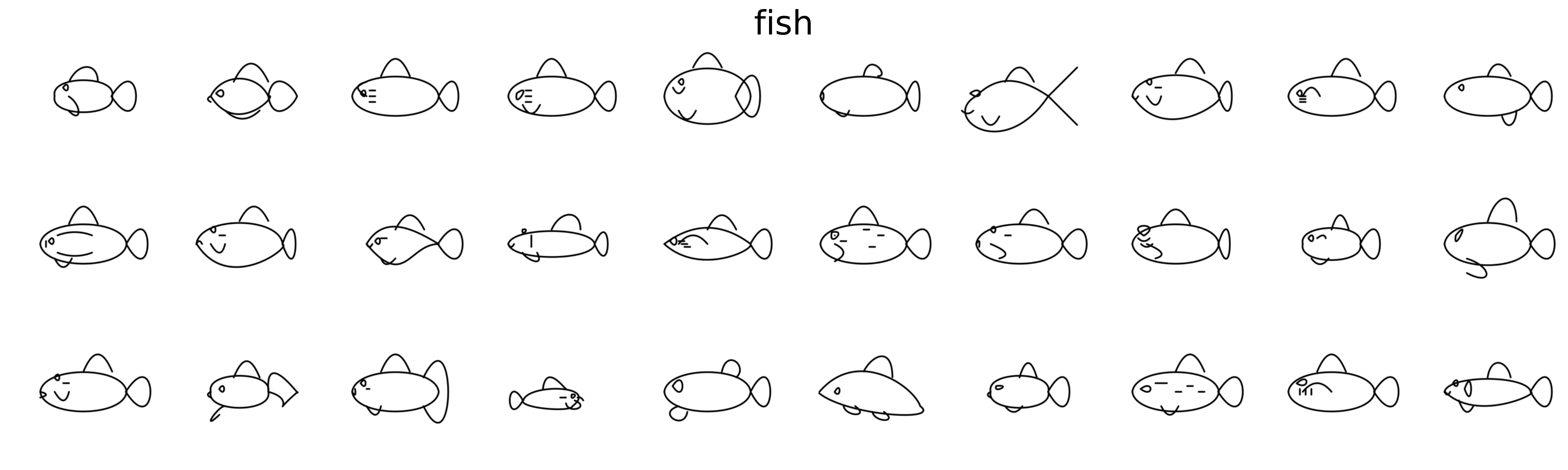}
    \includegraphics[width=1\linewidth]{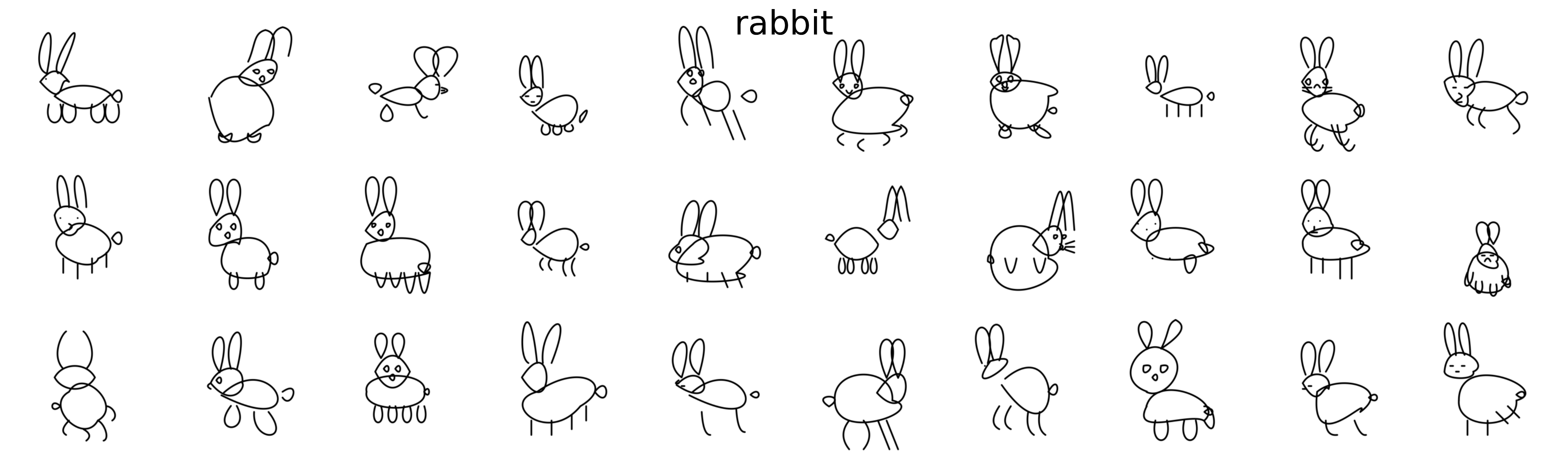}
    \includegraphics[width=1\linewidth]{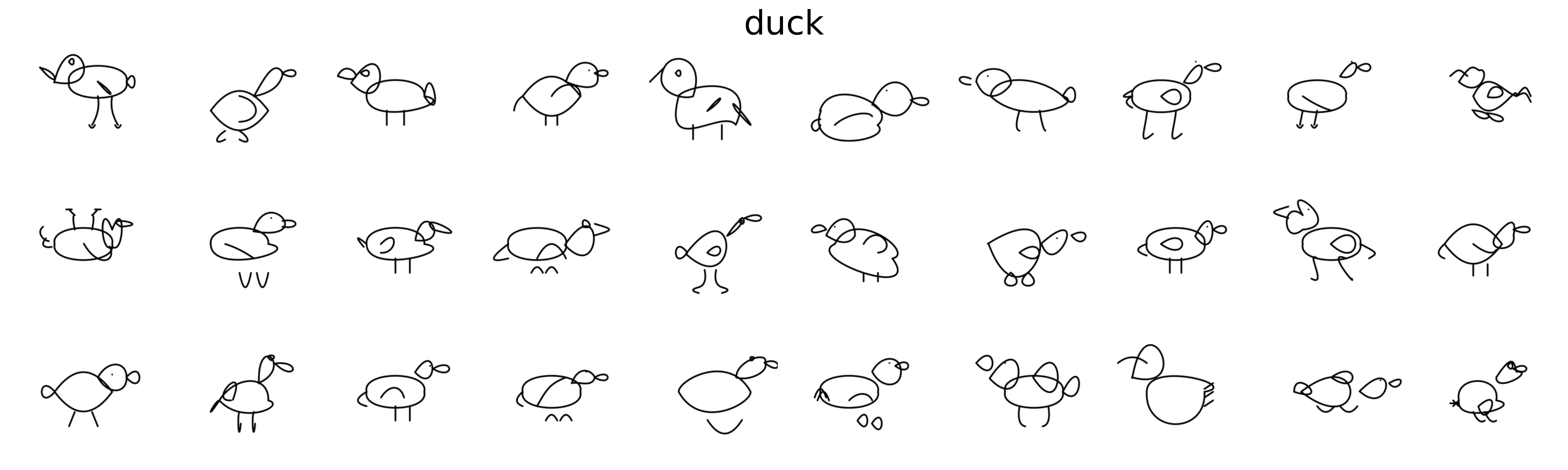}
    \vspace{-0.5cm}
    \caption{Sketches generated by SketchAgent for the eight categories of our human-agent collaborative study.}
    \label{fig:enter-label}
\end{figure*}

\begin{figure*}
    \centering
    \includegraphics[width=1\linewidth]{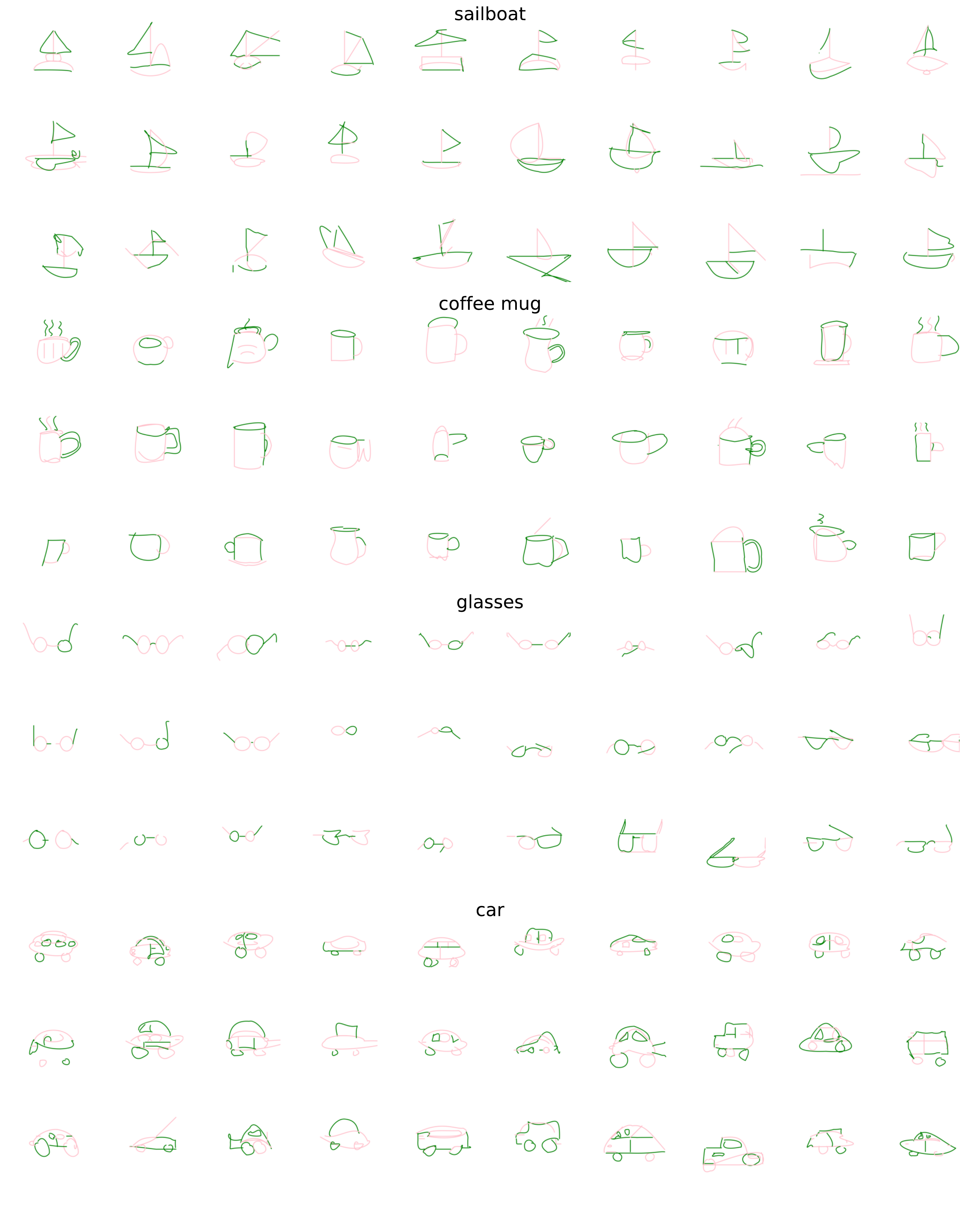}
    \label{fig:collab_all_sketches1}
    \vspace{-0.5cm}
    \caption{Sketches created collaboratively by users and SketchAgent as part of our collaborative human study.}
\end{figure*}

\begin{figure*}
    \centering
    \includegraphics[width=1\linewidth]{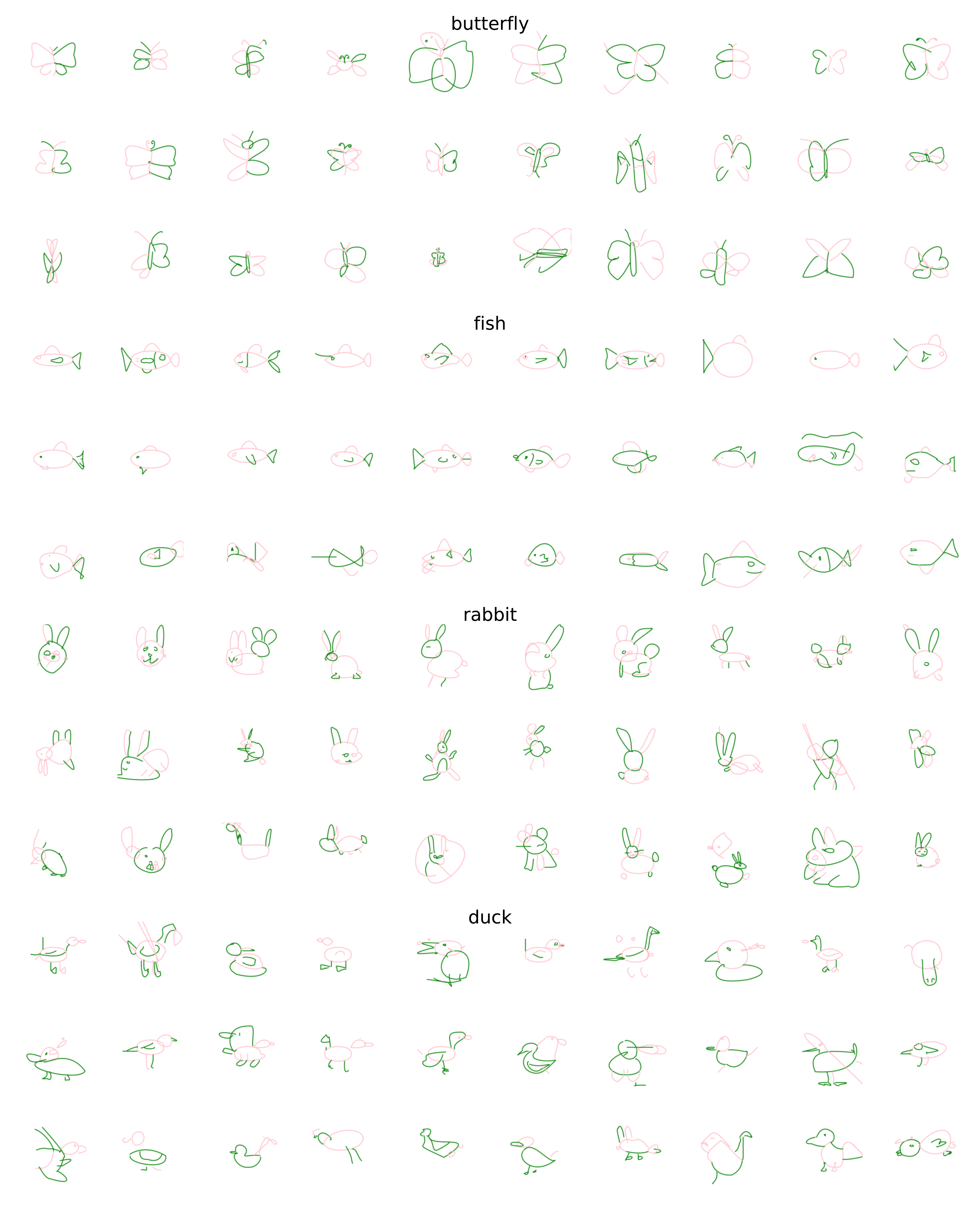}
    \vspace{-0.5cm}
    \caption{Sketches created collaboratively by users and SketchAgent as part of our collaborative human study.}
    \label{fig:collab_all_sketches2}
\end{figure*}

\begin{figure*}
    \centering
    \includegraphics[width=1\linewidth]{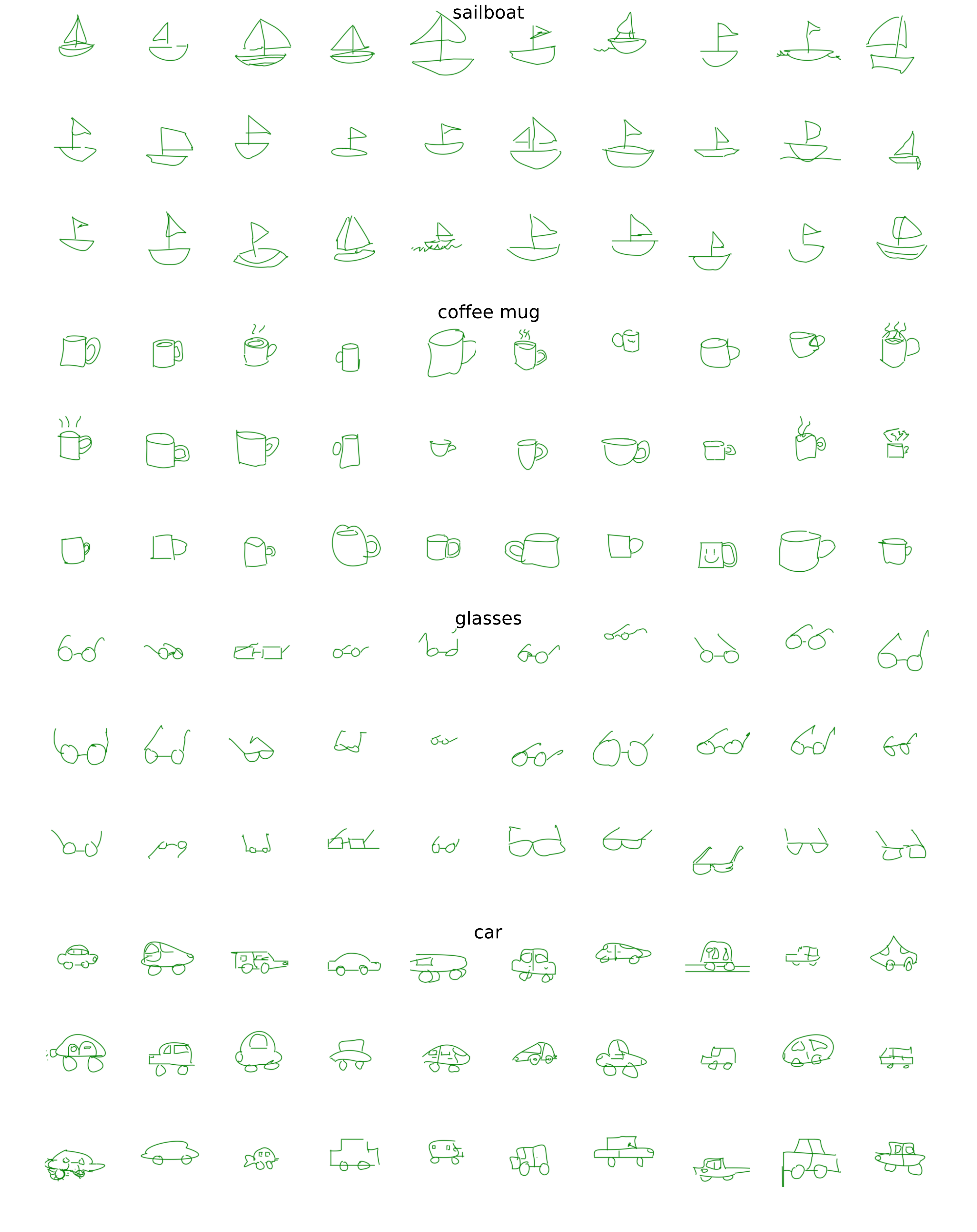}
    \vspace{-0.5cm}
    \caption{Sketches created by users in \ap{solo} mode as part of our collaborative human study.}
    \label{fig:solo_user_sketches1}
\end{figure*}

\begin{figure*}
    \centering
    \includegraphics[width=1\linewidth]{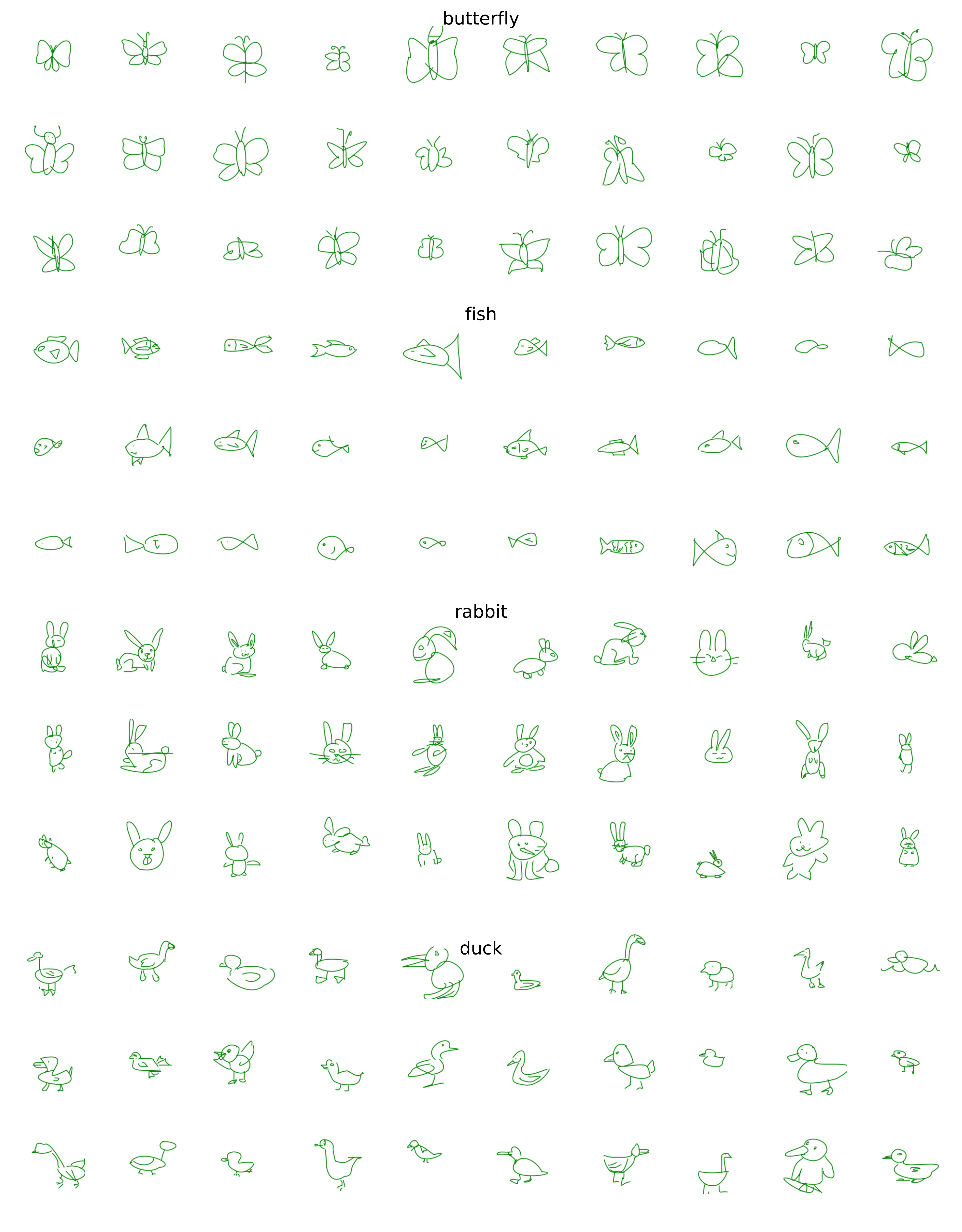}
    \vspace{-0.5cm}
    \caption{Sketches created by users in \ap{solo} mode as part of our collaborative human study.}
    \label{fig:solo_user_sketches2}
\end{figure*}

\end{document}